\documentclass[phd,oneside]{ubcthesis}

\usepackage[unicode=true,
  linktocpage,
  linkbordercolor={0.5 0.5 1},
  citebordercolor={0.5 1 0.5},
  linkcolor=blue]{hyperref}
\usepackage{amsmath, amsfonts, amssymb}
\usepackage{graphicx}
\usepackage{afterpage}
\usepackage{float}
\usepackage{longtable}
\usepackage{pdflscape}
\usepackage[numbers,sort&compress]{natbib}
\usepackage{psfrag}
\usepackage{parskip}
\usepackage{pdfpages}
\usepackage{geometry}
\usepackage{subfigure}
\usepackage{multirow}
\usepackage{booktabs}
\usepackage{makecell} 
\usepackage{threeparttable}
\usepackage{fancybox} 
\usepackage[table]{xcolor}
\usepackage{seqsplit}
\usepackage{bookmark}

\makeatletter
\renewcommand{\l@table}[2]{%
  \@dottedtocline{1}{0em}{2.5em}{Table~#1}{#2}%
}
\renewcommand{\l@figure}[2]{%
  \@dottedtocline{1}{0em}{2.5em}{Figure~#1}{#2}%
}
\makeatother

\newcommand{\emphasizespace}[1]{\emph{#1}\hspace{1em}}

\newcommand{\committeeMember}[5]{#1, #2, #3, #4\\\textit{#5}}

\def \theAuthor {Chenwei Zhang}
\previousdegree{B.Sc., The University of Waterloo, 2018}
\previousdegree{M.Sc., The University of Waterloo, 2021}

\def \theTitle {Applications of Deep Generative Models to DNA Reaction Kinetics and to Cryogenic Electron Microscopy}

\def \theDegree {Doctor of Philosophy} 
\def \theProgramTitle {Computer Science}
\def \theCopyrightyear {2026}
\def \theSubmitdate {March 2026}

\institution{The University Of British Columbia}

\faculty{The Faculty of Graduate and Postdoctoral Studies}
\institutionaddress{Vancouver}

\title{\theTitle}
\author{\theAuthor}
\copyrightyear{\theCopyrightyear}
\submitdate{\theSubmitdate}
\program{\theProgramTitle}

\renewcommand\thepart         {\Roman{part}}
\renewcommand\thechapter      {\arabic{chapter}}
\renewcommand\thesection      {\thechapter.\arabic{section}}
\renewcommand\thesubsection   {\thesection.\arabic{subsection}}
\renewcommand\thesubsubsection{\thesubsection.\arabic{subsubsection}}
\renewcommand\theparagraph    {\thesubsubsection.\arabic{paragraph}}

\floatstyle{ruled}
\newfloat{Program}{htbp}{lop}[chapter]

\begin{document}

\frontmatter



\maketitle                      

\chapter*{}
The following individuals certify that they have read, and recommend to the Faculty of Graduate and Postdoctoral Studies for acceptance, the thesis entitled:


\textbf{\theTitle}

submitted by \textbf{\theAuthor} in partial fulfillment of the requirements for the degree of \textbf{\theDegree} in \textbf{\theProgramTitle}.

\textbf{Examining Committee:}

\committeeMember{Anne Condon}{Professor}{Computer Science}{University of British Columbia} {Co-Supervisor}

\committeeMember{Khanh Dao Duc}{Assistant Professor}{Mathematics}{University of British Columbia} {Co-Supervisor}

\committeeMember{Jiarui Ding}{Assistant Professor}{Computer Science}{University of British Columbia} {Supervisory Committee Member}

\committeeMember{Raymond Ng}{Professor}{Computer Science}{University of British Columbia} {University Examiner}

\committeeMember{Alexandre Bouchard-Côté}{Professor}{Statistics}{University of British Columbia} {University Examiner}

\committeeMember{David Fleet}{Professor}{Computer Science}{University of Toronto} {External Examiner}

\textbf{Additional Supervisory Committee Members:}

\committeeMember{Jörg Gsponer}{Professor}{Michael Smith Laboratories}{University of British Columbia} {Supervisory Committee Member}


\begin{abstract}
This dissertation explores how deep generative models can advance the analysis of challenging biological problems by integrating domain knowledge with cutting-edge deep learning techniques. It focuses on two fundamental areas: DNA reaction kinetics and cryogenic electron microscopy (cryo-EM). 

In the first part, we present ViDa, a biophysics-informed deep learning framework that leverages variational autoencoders (VAEs) and geometric scattering transform to generate biophysically-plausible embeddings of DNA reaction kinetics simulations. These embeddings are further reduced to a two-dimensional Euclidean space to visualize DNA hybridization and toehold-mediated three-way strand displacement reactions. By embedding simulated secondary structures and reaction trajectories into a low-dimensional representation, ViDa preserves structure and clusters trajectory ensembles into reaction pathways, making simulation results more interpretable and revealing new insights into reaction mechanisms. 

In the second part, we address key challenges in cryo-EM density map interpretation and protein structure modeling. We first provide a comprehensive review and benchmarking of state-of-the-art deep learning methods for protein structure modeling (i.e., atomic model building). We propose improved evaluation metrics to assess the performance of these methods and provide guidance for researchers. We then present Struc2mapGAN, a generative adversarial network (GAN) that synthesizes high-fidelity experimental-like cryo-EM density maps from protein structures. We finally present CryoSAMU, a structure-aware multimodal U-Net that enhances intermediate-resolution cryo-EM density maps by integrating density features with structural embeddings from protein large language models through cross-attention mechanisms. 

Overall, these contributions demonstrate the potential of deep generative models to interpret DNA reaction mechanisms and to advance cryo-EM density map analysis and protein structure modeling. 
\end{abstract}

\chapter{Lay Summary}

Artificial intelligence (AI), particularly machine learning (ML) and deep learning (DL), has revolutionized scientific research and practical applications across diverse fields, including biology. In this dissertation, we leverage advanced DL methods to address two challenging biological problems: uncovering the underlying DNA reaction mechanisms and improving cryogenic electron microscopy (cryo-EM) density map analysis to advance protein structure modeling.
In the first part of this dissertation, we introduce a novel DL-based workflow that converts complex DNA reactions into intuitive two-dimensional visual representations. These visualizations make reaction pathways easier to interpret and reveal nuanced insights into the reaction mechanisms.
In the second part, we review and systematically evaluate existing DL methods for protein structure modeling from cryo-EM maps and propose refined metrics to assess their performance. Furthermore, we introduce several novel DL approaches to synthesize realistic cryo-EM maps and enhance the quality of raw experimental maps, thereby facilitating more accurate protein structure modeling. 
\chapter{Preface}

The research presented in this dissertation was developed in collaboration with my co-supervisors, Dr. Anne Condon from the Department of Computer Science and Dr. Khanh Dao Duc from the Department of Mathematics at the University of British Columbia. 
The chapters are based on research papers that are either published or currently unpublished. The use of Generative AI in this dissertation was restricted to employing ChatGPT for rephrasing and grammatical refinement.
\begin{itemize}
    \item Chapters \ref{chap:vida_mlcb} and \ref{chap:vida_2strand} are based on two research papers that were accepted at the Machine Learning in Structural Biology Workshop at the 36th Conference on Neural Information Processing Systems and the 18th Machine Learning in Computational Biology Conference: 
    \begin{enumerate}
        \item Chenwei Zhang, Khanh Dao Duc, and Anne Condon. Visualizing DNA reaction trajectories with deep graph embedding approaches. In NeurIPS 2022 Machine Learning in Structural Biology (MLSB) Workshop, 2022 \cite{vida-nips}.
        \item Chenwei Zhang, Jordan Lovrod, Boyan Beronov, Khanh Dao Duc, and Anne Condon. ViDa: Visualizing DNA hybridization trajectories with biophysics-informed deep graph embeddings. In Machine Learning in Computational Biology, pages 148–162. PMLR, 2024 \cite{vida-mlcb}.
    \end{enumerate}
    - I am the primary contributor in proposing the key ideas, formulating the solutions, collecting the data, developing the model, conducting the experiments, analyzing the results, and writing the papers. Jordan Lovrod contributed to partial data collection and manuscript revision. The other co-authors contributed to providing ideas for improving the model, discussing the results, and revising the papers.

    \item Chaper \ref{chap:vida_3strand} is an extension of Chapter \ref{chap:vida_2strand}. I am the primary contributor in upgrading the model, conducting the experiments and analyzing the results. This chapter has not been published.

    \item Chapter \ref{chap:cryoem_review} is based on a research paper that was accepted at Briefings in Bioinformatics:
    \begin{enumerate}
        \item Chenwei Zhang, Anne Condon, and Khanh Dao Duc. A comprehensive survey and benchmark of deep learning-based methods for atomic model building from cryo-electron microscopy density maps. Briefings in Bioinformatics, 26(4):bbaf322, 2025 \cite{atomicmodelreview_zhang}.
    \end{enumerate}
    - I am the primary contributor in summarizing different approaches, developing the metrics, conducting the experiments, analyzing the results, and writing the paper. The other co-authors contributed to discussing the results and revising the paper.

    \item Chapter \ref{chap:gan} is based on a research paper that was accepted at Bioinformatics Advances:
    \begin{enumerate}
        \item Chenwei Zhang, Anne Condon, and Khanh Dao Duc. Struc2mapGAN: improving synthetic cryogenic electron microscopy density maps with generative adversarial networks, Bioinformatics Advances, vbaf179, 2025 \cite{struc2mapgan}.
    \end{enumerate}
    - I am the primary contributor in proposing the key idea, formulating the solution, collecting the data, developing the model, conducting the experiments, analyzing the results, and writing the paper. The other co-authors contributed to discussing the results and revising the paper.

    \item Chapter \ref{chap:enhancemap} is based on a research paper that was accepted at the 2nd Workshop on Generative AI and Biology for the 42nd International Conference on Machine Learning. 
    \begin{enumerate}    
        \item Chenwei Zhang and Khanh Dao Duc. CryoSAMU: Enhancing 3D cryo-EM density maps of protein structures at intermediate resolution with structure-aware multimodal U-Nets. In ICML 2025 Generative AI and Biology (GenBio) Workshop, 2025 \cite{cryosamu}.
    \end{enumerate}
   - I am the primary contributor in proposing the key idea, formulating the solution, collecting the data, developing the model, conducting the experiments, analyzing the results, and writing the paper. The other co-authors contributed to discussing the results and revising the paper.
\end{itemize}

\tableofcontents                
\listoftables                   
\listoffigures                  
\chapter{Acknowledgements}

My deepest thanks to my supervisor, Anne Condon, whose open-mindedness, patience, rigor, kindness, and humility have shaped me as both a better researcher and a better person. I am also grateful for her support of my conference travels, which opened my horizons and shaped my research path.
I am equally grateful to Khanh Dao Duc, who served as my co-supervisor from my second year onward, for his guidance, support, and collaboration in introducing me to cryo-EM research and helping me connect it with my interests in deep learning. 
I would like to extend my sincere gratitude to my committee members, Jiarui Ding and Jörg Gsponer, for their insightful questions during my proposal and their thoughtful feedback on my thesis. 

Many thanks to Ben Chen, Parson Poon, and Jessica Choi in Computer Science for their assistance with paperwork and travel reimbursements, and to Tony Nguyen in Mathematics for his many timely rescues with remote-server setup.
I would also like to thank Bruce Shepherd for welcoming the Algorithm Lab to summer BBQs in his beautiful backyard. 

I would like to thank my lab mates and collaborators, Jordan Lovrod and Boyan Beronov, for their insightful ideas, thoughtful questions, and stimulating discussions that enriched my research; Aryan Tajmir Riahi, for his collaboration on an interesting cryo-EM project; and Wanxin Li and Geoffrey Woollard, for their steady support throughout my Ph.D.

I am deeply grateful to my wonderful friends, Xin Xin, Wangning Cai, Yuhang Song, Jingxiang Song, and Jiajie Pu, who have accompanied me on this exciting, though at times challenging journey, offering encouragement, laughter, and support.  

Finally, I am profoundly grateful to my partner, Shiyu, whose unwavering strength and encouragement have sustained me through every stage of this journey. Her presence has filled my life with joy, balance, and calm beyond research. 
I also extend my heartfelt gratitude to my parents, Zhenguo and He. Their unconditional love, unwavering support, and unfailing care have been my anchor throughout my life.

\chapter{Dedication}
To my parents. 

\mainmatter

\chapter*{Summary} \label{chap:summary}
\addcontentsline{toc}{part}{Summary} 

A deep generative model (DGM) is a type of machine learning model that learns the underlying probability distribution of data, enabling it to generate new, analogous samples. Beyond their capacity for data generation, DGMs often learn a compact latent representation that captures the essential structure of high-dimensional inputs, effectively performing dimensionality reduction as a by-product. Commonly DGMs include variational autoencoders (VAEs), generative adversarial networks (GANs), Transformer-based models, and diffusion models. 

Over the past few years, DGMs have achieved remarkable success across various domains, including image, video, and natural language processing, and have become increasingly popular in biology. For instance, VAEs have been applied to single-cell RNA sequencing data to capture the high-dimensional variability of gene expression profiles. GANs have transformed biomedical imaging by generating realistic synthetic microscopy images, which are especially useful for training deep neural networks when annotated data is scarce. Transformers have been widely used in protein large language models (pLLMs). More recently, diffusion models have been successfully employed to produce high-fidelity images of cellular and molecular structures. 

In this dissertation, we employ DGMs to solve two challenging biological problems. 
The first challenge is to comprehend the mechanisms that determine DNA reaction kinetics. Molecular programmers would benefit from accurate estimates of the rates of DNA reactions, as they vary dramatically across sequences. Yet, the mechanisms of such effects are not well understood. 
In Chapter \ref{chap:vida_mlcb} of Part \MakeUppercase{\romannumeral 1}, we introduce a novel workflow, called ViDa, for visualizing DNA reaction kinetics. ViDa leverages a VAE integrated with biophysics-informed losses, designed to extract features and reduce the dimensionality for visualizing DNA secondary structure state spaces and reaction trajectories. 
We evaluate ViDa on two well-studied DNA hybridization reactions, as discussed in Chapter \ref{chap:vida_2strand}. 
Furthermore, we extend ViDa to handle more complex DNA reactions. In Chapter \ref{chap:vida_3strand}, we propose an extension called ViDa-3Strand, specifically designed to accommodate DNA toehold-mediated three-way strand displacement reactions. 

The second challenge is to improve the interpretation of cryogenic electron microscopy (cryo-EM) 3D density maps and enhance protein structure modeling.
In Chapter \ref{chap:cryoem_review} of Part \MakeUppercase{\romannumeral 2}, we conduct a comprehensive survey of existing model-building methods, refine current evaluation metrics, and benchmark representative DGMs to offer insights and practical guidance for both cryo-EM practitioners and machine learning researchers in this field.
To further address the scarcity of reliable low-resolution map–model pairs, we introduce in Chapter \ref{chap:gan} a customized GAN model, named Struc2mapGAN, that learns realistic structural features from experimental data to simulate high-fidelity synthetic cryo-EM density maps.
Finally, in Chapter \ref{chap:enhancemap}, we present CryoSAMU, a multimodal U-Net that integrates structural embeddings from a pretrained pLLM with cryo-EM density map features to enhance intermediate-resolution cryo-EM maps for improved protein model building.

In future work, we plan to generalize ViDa to accommodate more complex DNA reactions, such as four-way strand exchange. 
We also aim to enhance Struc2mapGAN to precisely control resolution levels during map generation.
Finally, we plan to investigate advanced pLLMs for structural embeddings and develop new algorithms to integrate multimodal information within CryoSAMU for improved cryo-EM density map enhancement and protein structure modeling.

By leveraging DGMs across these two critical biological challenges, this dissertation advances computational methodologies and deepens biological understanding. Through the integration of VAEs, GANs, and Transformer-based models with domain-specific knowledge, it introduces novel tools for interpreting DNA reaction kinetics and enhancing cryo-EM maps and protein structure modeling. Ultimately, this work contributes to molecular programming and structural biology, paving the way for new discoveries and practical applications in these fields.

\part{Biophysics-informed Variational Autoencoders for Nucleic Acid Reaction Visualization}

\chapter{Introduction}  \label{chap:vida_intro}

Nucleic acids, namely deoxyribonucleic acid (DNA) and ribonucleic acid (RNA), can be engineered beyond their genetic roles as versatile materials for nanotechnology, due to their structural stability and functional diversity. DNA molecules adopt double-helical structures that provides rigidity and stability, enabling the design of motifs such as double crossovers that serve as building blocks for lattices and patterns.
RNA molecules can incorporate functional modules such as aptamers and ribozymes, which can be assembled into nanoparticles to create multifunctional nanosystems for targeted delivery and therapy.
Owing to predictable base-pairing rules, both DNA and RNA sequences can be designed precisely to self-assemble into well-defined nanoscale architectures.

In the past few decades, DNA and RNA nanotechnologies have been developed that are capable of sensing and responding to changes in their environments \cite{dnasensing}, self-assembling into complex structures \cite{dna_assembly}, and simulating computational models such as logic circuits \cite{dnacircuits,dnagate} or artificial neural networks \cite{dnaneuralnets}. The behaviour of these technologies depends on nucleic acid thermodynamics (which can be used to predict properties of nucleic acid systems in equilibrium) and kinetics (which predicts rates of change and folding dynamics). 

Thermodynamics of nucleic acids has been extensively studied, and numerous methods have been developed to predict thermodynamic properties, such as NUPACK \cite{nupack} and ViennaRNA \cite{vienna}.
In contrast, the mechanisms that influence reaction kinetics are less well understood. Predicting reaction kinetics is crucial not only for understanding biological processes, but also for practical implications such as drug design and biotechnology applications. 
Several kinetic simulation tools exist for this purpose, including Multistrand \cite{Schaeffer}, Kinfold \cite{Kinfold}, and oxDNA \cite{oxDNA}. While these simulators produce useful data, their output is often difficult to interpret and summarize. This highlights the need for improved visualization techniques to make simulation results more comprehensible and informative for domain experts. 

Accordingly, Part I of this dissertation focuses on developing powerful visualization tools to facilitate deeper understanding of DNA reaction mechanisms. In particular, we present a deep learning-based approach for analyzing and visualizing DNA reaction kinetics, providing an interactive way to understand key factors such as structural features and kinetic traps that influence reaction rates. Furthermore, we propose evaluation metrics to quantitatively compare visualization results across different methods.

In this introductory chapter, we first provide a comprehensive background of nucleic acid reactions, thermodynamic models of nucleic acids, kinetic models for elementary-step continuous-time Markov chains (CTMCs), the Multistrand DNA reaction simulator, and existing nucleic acid reaction visualization methods. We then provide background on useful deep learning and machine learning-based visualization algorithms, namely variational autoencoders (VAEs) and potential of heat diffusion for affinity-based transition embedding (PHATE). 

Building on this foundation, we introduce our work on adapting VAEs to visualize DNA reaction energy landscapes and trajectories that are output by the Multistrand simulator. 
Specifically, in Chapter \ref{chap:vida_mlcb}, we present \textbf{ViDa}, a novel biophysics-informed semi-supervised variational autoencoder with graph representations for DNA reaction visualization. 
In Chapter \ref{chap:vida_2strand}, we apply ViDa to visualize two-stranded DNA hybridization reactions and quantitatively compare its performance with other state-of-the-art visualization methods.
In Chapter \ref{chap:vida_3strand}, we extend ViDa to \textbf{ViDa-3Strand}, which can accommodate three-stranded reactions. We employ ViDa-3Strand to analyze several well-studied toehold-mediated three-way strand displacement reactions, through which ViDa-3Strand provides nuanced understanding of the reaction kinetics.
In addition, we also discuss useful directions for future work in both Chapters \ref{chap:vida_2strand} and \ref{chap:vida_3strand}.

\section{Nucleic Acid Reactions}

We model nucleic acid interactions (reactions) in a stochastic regime \cite{Schaeffer}. In this system, the reactions operate under fixed experimental conditions, including solution volume $V$, temperature $T$, and concentrations of $Na^{+}$ and $Mg^{2+}$ ions. 
The interacting strands in a nucleic acid reaction are represented by their nucleotide (base) sequences in a multiset $\Psi$. 
A \textit{complex} is a maximal subset of strands in $\Psi$ that are connected through base pairing. 
A \textit{secondary structure} of a complex is a set of all base pairs among the strands within that complex. It can be represented as a graph in which nodes correspond to bases and edges correspond to either base pairs or backbone connections.
A \textit{pseudoknotted} single-stranded secondary structure contains at least one base pair in which exactly one of its nucleotides situates between the nucleotides of another base pair; otherwise, the single-stranded secondary structure is non-pseudoknotted (pseudoknot-free).
A \textit{state} is a multiset of complexes in which each strand $\psi \in \Psi$ is part of exactly one complex. 
A \textit{DNA reaction}, in which DNA complexes fold from one three-dimensional structure into another through the formation or breaking of base pairs, is specified by an initial state (or initial distribution over states) that represents the reactants, and a final state (or final set of states) that represents the products. 
To analyze reaction mechanisms, we are interested in studying \textit{trajectories}, i.e., sequences of secondary structures, from the reactants to the products of a DNA reaction, along with the time to transition from one secondary structure to the next.
Transitions between neighbouring states (secondary structures) correspond to \textit{elementary steps}, i.e., a single base pair forming or breaking.
In this work, we focus on two common types of DNA reactions. 

\begin{figure}[!ht]
  \begin{center}
      \includegraphics[width=\linewidth, trim={0cm 7.2cm 21cm 0cm}, clip]{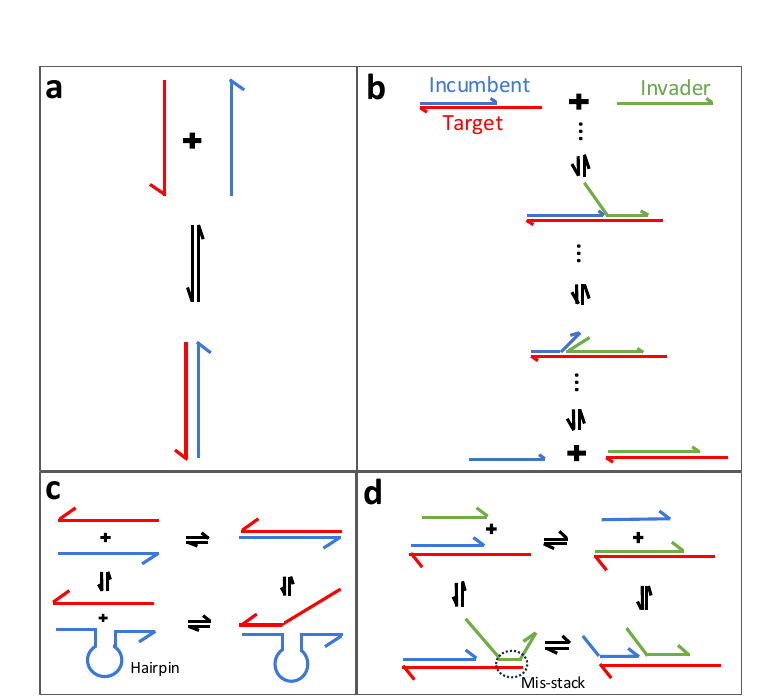}
      \caption{
      \textbf{(a)} Helix association and dissociation reactions. The bottom pair of lines represents the two complementary strands with all bases paired (forming a double helix).
      \textbf{(b)} A toehold-mediated three-way strand displacement reaction. The two figures show coarse-grained representations of the reactions, i.e., it groups many elementary steps into one coarse-grained step.
      \textbf{(c)} A possible hairpin forming during helix association.
      \textbf{(d)} Possible mis-stacked base pairs forming during three-way strand displacement.
      }
      \label{fig:helix_3strand}
  \end{center}
\end{figure}

\paragraph{DNA hybridization and melting}
Figure \ref{fig:helix_3strand}a shows DNA hybridization and melting reactions, also known as DNA helix association and dissociation. In helix association, one unpaired DNA strand hybridizes with its complementary strand to form a fully paired duplex. In the reverse direction, helix dissociation is the reaction whereby the base pairs of a double-stranded DNA duplex break, resulting in two separate strands. As shown in Figure \ref{fig:helix_3strand}c, during helix association, hairpins (self-complementary strand loops) may form in any of the single strands, reducing the hybridization rate by a factor of 2 to 10, depending on the stability of the hairpins \cite{gaohelix}. Helix association and dissociation reactions are important in many cellular processes \cite{dnasensing} and in biotechnological applications such as DNA-based switches \cite{dnaswitch}, motors \cite{dnamotors}, and molecular diagnostics \cite{diagnostics}. For example, a DNA switch can be engineered by coupling DNA hybridization with an external stimulus, such as light or temperature. In this design, the DNA structure can reversibly switch between hybridized and unhybridized states in response to the stimulus.

\paragraph{Toehold-mediated three-way strand displacement}
Another example is toehold-mediated three-way strand displacement, as shown in Figure \ref{fig:helix_3strand}b. An invader strand initially binds to the toehold of a target-incumbent complex, i.e., to a terminal unpaired domain on the partially bound target strand, and then goes on to displace the incumbent strand from this complex via the incremental process of branch migration. Introducing mismatched base pairs can strongly affect the reaction, though the underlying mechanisms are not yet fully understood. Machinek et al. \cite{MachinekThreeway} demonstrated that the reaction rate can be tuned over three orders of magnitude by altering the position of a mismatch introduced by the invader within the displacement domain. In particular, a mismatch near the toehold destabilizes the system once the toehold is bound, forcing the system to spend considerably more time in the toehold-only state. In our study, we observed that stable mis-stacked base pairs formed between the target and invader strands (Figure \ref{fig:helix_3strand}d) can also significantly slow down the overall reaction, as discussed in Chapter \ref{chap:vida_3strand}.
Toehold-mediated three-way strand displacement is a fundamental mechanism for constructing autonomous DNA-based devices, including logit gate \cite{dnagate}, digital circuits \cite{dnacircuits} and oscillators \cite{dnaoscillators}. For example, the presence or absence of the displaced incumbent strand can be treated as a logical signal, with \textit{1} indicating a fluorescent output and \textit{0} indicating no output. To design an \texttt{AND} gate, the target strand is engineered with two toehold regions. Full displacement of the incumbent strand through branch migration occurs only when both input invader stands are present, resulting in the release of the output signal, corresponding to a logical value of \textit{1}.

\section{Thermodynamic Models}

A thermodynamic model assigns each allowed state $x\in\mathcal{X}$ ($\mathcal{X}$ is a set of allowed states) a Gibbs free energy $\Delta G(x)$, measured relative to a reference state. 
At statistical equilibrium, the probability that the state is $x$, at given temperature $T$, is given by the Gibbs-Boltzmann distribution:
\begin{equation}
    \pi(x) = \frac{1}{Z} \, e^{-\beta \cdot \Delta G(x)},
\end{equation}
\begin{equation}
    Z_\beta = \int e^{-\beta \cdot \Delta G(x)} \, dx,
\end{equation}
where $\beta=\frac{1}{k_BT}$, $k_B$ is the Boltzmann constant, and $Z$ is the partition function.
The Boltzmann distribution can be used to determine all thermodynamic equilibrium properties of the system. For example, it can be used to compute the average Gibbs free energy, evaluate the probabilities of specific states, and identify the minimum free energy (MFE) states. It also allows for calculating entropy, heat capacity, and other thermodynamic quantities at equilibrium. 
Thermodynamic models have been developed for predicting DNA and RNA secondary structures \cite{dna_mathews1,dna_mathews2,dna_santalucia1,dna_santalucia2,dna_turner}, as well as for identifying MFE secondary structures \cite{nupack,vienna}.

\section{Kinetic Models for Elementary-Step Continuous-Time Markov Chains}

Many reactions have well-defined end products that can be predicted by thermodynamic models. However, determining whether a system will reach its end state within a reasonable time requires modeling its kinetics. Moreover, kinetic analysis can also uncover poor sequence designs, such as those containing alternate reaction pathways to the same final state or kinetic traps that slow or prevent the desired reaction.
A kinetic model such as Kawasaki model \cite{kawasaki}, Metropolis model \cite{metropolis}, and Arrhenius-type model \cite{lovrod,zolaktafarrhenius}, describes the non-equilibrium dynamics of an elementary-step continuous-time Markov chain (CTMC) through transition rates. 
For any two distinct states $x$ and $x'$, $K(x,x')$ specifies the transition rate from $x$ to $x'$, determined as a function of the free energies of these states in terms of the chosen kinetic model \cite{kawasaki,metropolis,lovrod,zolaktafarrhenius}.
A common property in kinetic models is the \textit{detailed balance} condition: for any adjacent states $x$ and $x'$ (differing by a single base pair), their rates satisfy:
\begin{equation}
    \frac{K(x,x')}{K(x',x)} = \frac{\pi(x')}{\pi(x)} = e^{-\beta(\Delta G(x')-\Delta G(x))}, \quad \forall x,x' \in \mathcal{X}.
\end{equation}
This property guarantees that given sufficient time, the system will reach the same equilibrium state distribution predicted by thermodynamics, i.e., Gibbs-Boltzmann distribution.

A CTMC is defined by an initial probability distribution $\pi_{0}$ over a set of allowed states $\mathcal{X}$ and the \textit{transition rate matrix} $K$: ${\mathcal{X}}^{2} \rightarrow \mathbb{R}$. 
The diagonal entries are given by \cite{markovchains}
\begin{equation}
    K(x,x) = - \sum_{x' \in \mathcal{X}\setminus x} K(x,x').
\end{equation}
The set $I \subset \mathcal{X}$ containing states with non-zero initial probability is called the initial set of states. The \textit{transition probability matrix} $P: \mathcal{X}^{2} \rightarrow [0, 1]$ is the normalized rate matrix
\begin{equation}
    P(x,x') = \frac{K(x,x')}{-K(x,x)}.
\end{equation}
The \textit{transition matrix} (or \textit{propagator}) is
\begin{equation}
    Q_t = e^{\,tK}, \quad t \in \mathbb{R}_{\ge 0},
\end{equation}
which describes the transient dynamics of a CTMC.
A stochastic process $\{X(t)\}_{t\in\mathbb{R}_{\ge 0}}$ follows this CTMC if
\begin{equation}
    \mathbb{P}\!\left( X(t_0)=x_0, \ldots, X(t_n)=x_n \right)
= \pi_0(x_0)\, \prod_{m\in[0,n-1]} Q_{\,t_{m+1}-t_m}(x_m, x_{m+1}),
\end{equation}
for any $n\in\mathbb{N}_0$, $t_0<\ldots<t_n\in\mathbb{R}_{\ge0}$, and $x_0,\ldots,x_n\in\mathcal{X}$. This implies that trajectory probabilities can be arbitrarily decomposed into segment probabilities in terms of Markovianity, where the probability of each segment depends solely on its spatial endpoints $(x_m,x_{m+1})$, temporal endpoints $(t_m,t_{m+1})$, and the transition rate matrix.

For a fixed final state $F\subset\mathcal{X}$, the \textit{mean first passage time} (MFPT), $\tau_F(x): \mathcal{X}\rightarrow\mathbb{R}_{\ge0}$, represents the expected time to reach $F$ for the first time from each state. It satisfies the system:
\begin{equation}
    - \tau_F(x) \cdot K(x, x) = 1 + \sum_{x' \in \mathcal{X} \setminus x} \tau_F(x') \cdot K(x, x'),
\quad \forall x \in \mathcal{X} \setminus \{F\}.
\end{equation}

For sufficiently small CTMCs where all states can reach $F$, this system can be solved numerically \cite{markovchains}.
The MFPT from the initial states $I$ to the final state $F$ is then obtained by taking the expectation of $\tau_F$, $\mathbb{E}_{x\sim\pi_0}[\tau_F(x)]$, over the initial state distribution:
\begin{equation}
    \text{MFPT}_{\pi_0\rightarrow F} = \sum_{x\in\mathcal{X}} \pi_0(x) \tau_F(x).
\end{equation}

When the state space of a CTMC is large, direct matrix-based computation of the MFPT is infeasible, therefore, Monte Carlo estimation, via the \textit{stochastic simulation algorithm} (SSA) \cite{Gillespie} is applied. However, SSA is prohibitively inefficient for rare event simulation, especially in systems with multiple metastable regions. To overcome this, many path sampling algorithms \cite{WE,umbrella,ffs} are employed to focus on important states or transitions of interest. 

In this work, we focus on the elementary-step CTMC model for simulating multi-stranded nucleic acid reactions \cite{Schaeffer}, with further details provided in Section \ref{sec:multistrand}. In this model, states represent secondary structures and transitions represent elementary base-pair changes. The number of possible states, i.e., secondary structures, in the elementary-step CTMC can grow exponentially with the total strand length.

\section{The Multistrand Simulator} \label{sec:multistrand}

In our work, we use the \textit{Multistrand} simulator to generate DNA reaction trajectories.
Multistrand is an elementary-step CTMC model designed to simulate thermodynamic and kinetic process for various DNA or RNA-strand interactions ignoring formation of pseudoknotted structures. As the name suggests, Multistrand is able to handle systems involving several distinct strands. 
Multistrand allows for Watson-Crick base pairs (A-T and G-C in DNA and A-U and G-C in RNA), as well as for optional wobble pairs (G-T in DNA and G-U in RNA).
Because the secondary structure state space is known to scale exponentially in the length of the strands, the simulator uses a Gillespie sampling approach, i.e., SSA \cite{Gillespie}, rather than representing the entire state space of secondary structures explicitly. The rates between adjacent states are determined by a kinetic model, which is chosen in a way that detailed balance is satisfied, and that the equilibrium state distribution is in line with thermodynamic predictions made by both NUPACK \cite{nupack} and ViennaRNA \cite{vienna}.

\begin{figure}[!ht]
\centering
\begin{Sbox}
\begin{minipage}{\textwidth}
\small
\begin{verbatim}
----------------------------------------------------------------------------------
                                                        |   t[us]   | dG[kcal/mol]
----------------------------------------------------------------------------------
    AGATCAGTGCGTCTGTACTAGCACA+TGTGCTAGTACAGACGCACTGATCT
[1] ....(.((((..........)))).+.((((..........)))).).... | 0.0000000 |    -1.737    
[1] ...((.((((..........)))).+.((((..........)))).))... | 0.1153000 |    -2.137    
[1] ...(((((((..........)))).+.((((..........)))))))... | 0.1802000 |    -1.787    
[1] ....((((((..........)))).+.((((..........)))))).... | 0.2768000 |    -1.387    
[1] ....((((((...(.....))))).+.((((..........)))))).... | 0.5369000 |    +1.392        
...

[1] ((((((((((..........))...+...((..((....)))))))))))) | 17.760000 |    -1.310
[1] (((((((.((..........))...+...((..((....)))).))))))) | 17.800000 |    -1.829
[1] ((((((((((..........))...+...((..((....)))))))))))) | 17.860000 |    -1.310
[1] (((((((.((..........))...+...((..((....)))).))))))) | 17.940000 |    -1.829
[1] .((((((.((..........))...+...((..((....)))).)))))). | 18.140000 |    -1.961
...

[1] .(((((((((((((((((((((...+...))))))))))))))))))))). | 71.570000 |   -27.496    
[1] .((((((((((((((((((((((..+..)))))))))))))))))))))). | 71.620000 |   -26.401    
[1] .((((((((((((((((((((((.(+).)))))))))))))))))))))). | 71.810000 |   -26.201    
[1] (((((((((((((((((((((((.(+).))))))))))))))))))))))) | 71.900000 |   -26.069    
[1] (((((((((((((((((((((((((+))))))))))))))))))))))))) | 71.910000 |   -30.522    
\end{verbatim}
\end{minipage}
\end{Sbox}
\fbox{\TheSbox}
\caption{Sample output data produced using the \textit{first step mode} of Multistrand.}
\label{fig:multistrand}
\end{figure}

The Multistrand simulator offers several simulation modes to suit different user needs. In this study, we use the \textit{first step mode} (FSM) for all reaction simulations.
FSM divides each reaction trajectory into two stages: the initial binding step and the subsequent folding trajectory. 
The initial (bimolecular) step simulates a ``join'' move, where two molecules $A$ and $B$ interact and form a single base pair. The initial secondary structure states of $A$ and $B$ are chosen by Boltzmann sampling from their respective state spaces. The rate of this step is calculated based on all possible join moves from the initial state of $A$ and $B$. The subsequent (unimolecular) folding step is simulated until one of two stop conditions is met: (i) non-reactive: the molecules dissociate into one of the $A+B$ states, or (ii) reactive: the molecules react to form the product state. The rate of this step is calculated based on the first passage time, i.e., the time taken after the ``join'' move to reach the final product state.
These unimolecular and bimolecular rates can then be used to obtain overall rate for a reaction, as detailed in \cite{Schaeffer,kinda}.
In our study, we collect both reactive and non-reactive trajectory samples.
FSM is well suited for simulations at low concentrations because the rate of the initial bimolecular step is proportional to the simulated concentration and is therefore potentially much slower than the unimolecular step. 

As depicted in Figure \ref{fig:multistrand}, which shows a reactive simulation of a helix association reaction, the initial [\texttt{\seqsplit{5$'$-....(.((((..........)))).-3$'$+5$'$-.((((..........)))).)....}}-3$'$] and final [\texttt{\seqsplit{5$'$-(((((((((((((((((((((((((-3$'$+5$'$-)))))))))))))))))))))))))-3$'$}}] structures are represented in ``dot-parenthesis'' (dp) notation (see Supplementary \ref{dp_notation} for details). In this notation, matching parentheses indicate base pairs, dots represent unpaired bases, and the ``+" sign separates two complementary strands. The second and last columns in the figure list the reaction simulation time (in terms of sampled trajectory time, as opposed to the wall-clock time) and the corresponding free energy of the state, respectively. 
For DNA hybridization reactions, the Multistrand simulation can produce hundreds of thousands of elementary steps within a single trajectory. In contrast, for toehold-mediated three-way strand displacement reactions that are considerably more complex, it can produce hundreds of millions of elementary steps within a single trajectory. In addition, the number of elementary steps can vary substantially depending not only on strand length but also on sequence composition, even among strands of equal length.

\section{DNA Reaction Visualization}

Given a set of trajectories generated by the Multistrand simulator, how can we provide a visualization that will help synthetic biologists and molecular programmers understand key aspects of reaction kinetics, such as different types of reaction pathways and kinetic traps along those pathways during the unimolecular step? 
For complex reactions like three-stranded strand displacement, however, Multistrand can produce hundreds of millions of elementary steps within a single trajectory. This requires a way to compress these steps into biologically meaningful representations for visualization. The challenge is further increased by the need to showcase hundreds of such trajectories for analysis.

\begin{figure}[!ht]
  \centering
  \includegraphics[width=\linewidth,trim={0cm 17cm 0cm 0cm}, clip]
  {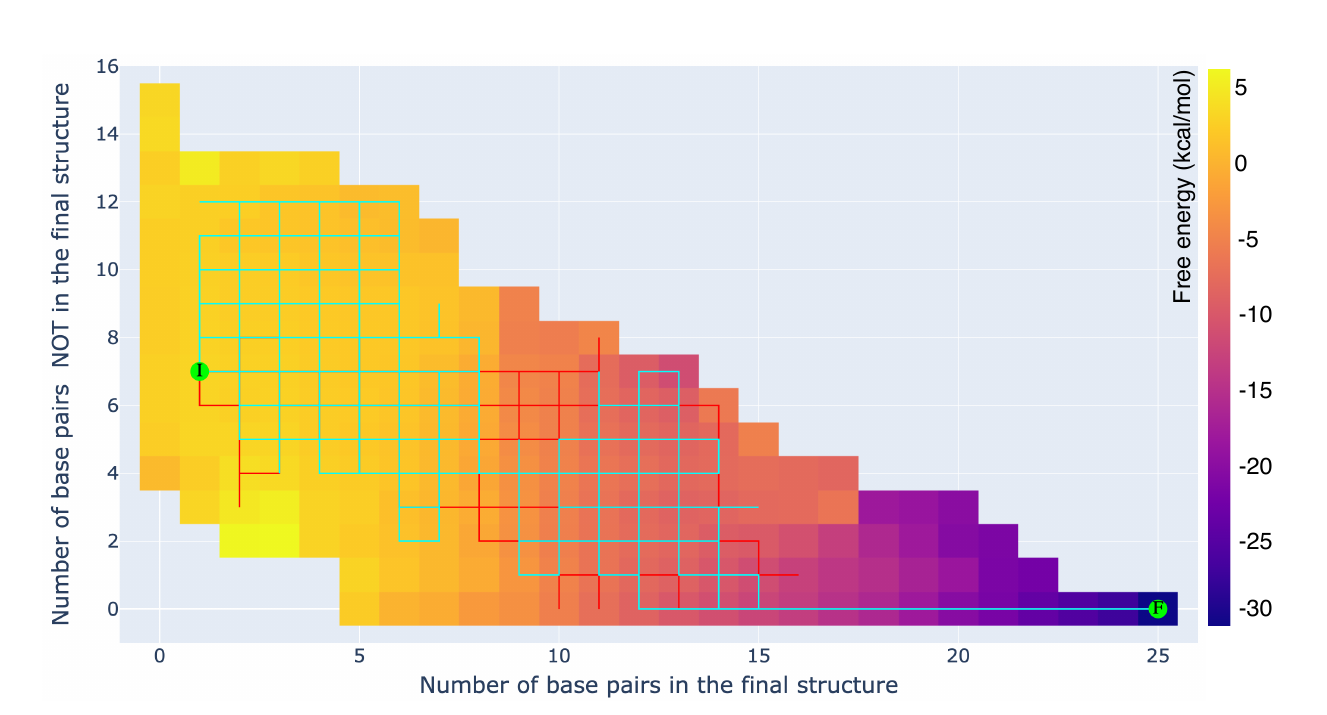}
  \caption{Coarse-grained visualization of a DNA hybridization reaction Gao-P4T4 (see Chapter \ref{chap:vida_2strand} for details). Each grid cell is an ensemble of secondary structures. A structure is in cell $(x,y)$ exactly $x$ of its base pairs contribute to the desired helix, and exactly $y$ of its base pairs are {\em not} part of the desired helix.
  The cyan and red traces represent two distinct trajectory samples, corresponding to the same samples shown in Figure \ref{fig:gao_p4t4_vida}a.
  The initial state (denoted by the green circle marked $I$) corresponds to the grid cell (1,7) and the final state (denoted by the green circle marked $F$) corresponds to the grid cell (25,0), since the strands have 25 bases each. 
  In this coarse-grained representation, both traces share the same initial state coordinates.
  }
  \label{fig:gao_p4t4_cg}
\end{figure}

There are two different but related directions with respect to showcasing nucleic acid reactions based on their sampled trajectories. One is visualizing energy landscapes. The other is visualizing trajectories through those landscapes. 
One of the most basic visualizations just uses ``dot-parenthesis" notation to represent a secondary structure, and a sequence of such strings to represent a trajectory (as shown in Figure \ref{fig:multistrand}). This method does not situate the trajectory in the overall energy landscape. Visualizations of energy landscapes use mappings of the high-dimensional state space to 2-D or 3-D space. One example, for DNA hybridization, is shown in Figure \ref{fig:gao_p4t4_cg}: the cell $(x,y)$ represents the set of secondary structures in the sampled trajectories, where $x$ indicates the number of base pairs in the final structure (fully paired DNA helix), and $y$ indicates the number of bases pairs not in the final structure. This visualization is able to involve the full energy landscape, however, with this scheme, structurally dissimilar secondary structures may be mapped to the same macrostate (cell), making it difficult to interpret each macrostate and trajectories through them. Flamm et al. \cite{barriertrees,Badelt} use barrier trees to visualize landscapes. This approach is effective in presenting the landscape and its overall shape and characteristics, but does not provide a way to visualize trajectories through such landscapes. Castro et al. \cite{GSAE} use deep graph embedding techniques to uncover the energy landscape of RNA secondary structure. However, this approach does not address how to visualize trajectories through those landscapes and it cannot visualize DNA reactions. Kinefold generates a movie showing reaction kinetics but is limited to displaying one trajectory, and does not show the overall energy landscape \cite{kinefold}. 
Overall, it is crucial to design and implement better visualization tools that can visualize both reaction energy landscapes in a biologically meaningful way, as well as reaction trajectories that can reveal the reaction kinetics.

\section{Background on Deep Generative Models -- VAE}

Deep generative models provide a powerful approach to learn meaningful representations from complex, high-dimensional data. 
Specifically, variational autoencoders (VAEs) \cite{VAE} are well suited for learning latent representations that capture essential features of the data, for example, thermodynamics and kinetic properties in the context of DNA reactions. In combination with a manifold learning methods, PHATE \cite{PHATE} that excels at preserving both local and global structure in biological datasets, VAEs allow for mapping high-dimensional reaction trajectories into interpretable two-dimensional spaces for visualization.

\subsection{Variational Autoencoder}

An autoencoder \cite{AE} consists of an encoder and a decoder, connected through a bottleneck layer. High-dimensional input data are passed through the encoder, which maps them to a low-dimensional output latent representation. This latent vector passes though the bottleneck layer and is then reconstructed into high-dimensional output by the decoder. The network is trained by comparing the reconstructed outputs with the original inputs and minimizing the reconstruction loss via gradient descent, with network weights updated through backpropagation.

\begin{figure}[!ht]
  \centering
  \includegraphics[width=.85\linewidth,trim={0cm 22.5cm 12cm 0cm}, clip]
  {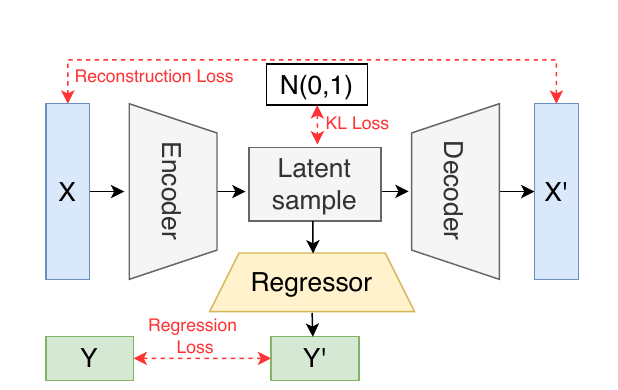}
  \caption{The architecture of a semi-supervised VAE model, along with its associated loss terms. Here, $X'$ is the reconstruction of input $X$ produced by VAE, $Y$ is an additional input associated with $X$, and $Y'$ is the predicted value by the regressor network.}
  \label{fig:semivae}
\end{figure}

A traditional autoencoder focuses solely on minimizing reconstruction loss, regardless of the regularity of its latent space, which often leads to severe overfitting. To address this, a variational autoencoder (VAE) \cite{VAE} encodes each input as a probability distribution over the latent space instead of a single point. In this setting, the total loss of the VAE network is composed of two parts: the reconstruction loss which compares the difference between output and input data and the latent space loss (KL loss) which regularizes the latent space by enforcing the encoder-resulting distribution to be close to a standard normal distribution.
By default, a VAE is an unsupervised network that does not require labeled input. 
In some applications, it is useful to concatenate the VAE with an auxiliary regression network, so that the VAE becomes a semi-supervised model \cite{GSAE}, as shown in Figure \ref{fig:semivae}. Consequently, in addition to reconstruction and latent penalties, a regression penalty is introduced to the overall network.
In our work, we take this approach, augmenting the latent space of the VAE with biophysically meaningful features, by providing additional constraints that regularize the latent representation and embedding it with richer biophysical information, as discussed in Chapter \ref{chap:vida_mlcb}.

\subsection{Geometric Scattering Transform}
DNA secondary structures can be viewed as graphs, as described earlier. This viewpoint enables the use of graph signal processing techniques, such as the geometric scattering transform \cite{scatteringtransform,zou2020graph,gama2018diffusion}.
The geometric scattering transform generalizes the traditional scattering transform that captures features from Euclidean signals (e.g., image or audio) to graph data. 
Let $G=(V,E,W)$ be a weighted graph with adjacency matrix $A\in\mathbb{R}^{n\times n} \quad (n=|V|)$, diagonal matrix of graph vertex degrees $D$, and transition operator $T=AD^{-1}$. A scalar graph signal is any $f\in\mathbb{R}^{n}$ assigning a real value to each vertex. The geometric scattering builds a multiscale, permutation-invariant representation of $(G,f)$ using diffusion wavelets \cite{diffusionwavelets}. 
These wavelets are constructed using the lazy random-walk diffusion operator
\begin{equation}
    P=\frac{1}{2}(I+T) = \frac{1}{2}(I+AD^{-1}),
\end{equation}
and powers $P^{t}$ describe $t$-step diffusion transition probabilities between graph nodes. 
The operator $P^{t}$ acts as a low-pass filter that averages $f$ over neighborhoods whose size is controlled by $t$. Therefore, the filtered signal $P^{t}f$ only preserves low frequencies over the graph. 
Similarly, the residuals $(I-P^{t})$ form high-pass filters controlled by $t$. 
Following diffusion wavelets, band-pass graph wavelets at dyadic scales $2^{j}$ are defined by differences of diffusions
\begin{equation}
    \Psi_{j} = P^{2^{j-1}} - P^{2^{j}} = P^{2^{j-1}}\bigl(I - P^{2^{j-1}}\bigr), \qquad j=1,\dots,J,
\end{equation}
where $J$ defines the largest scale of interest, subject to $2^{J}\ \lesssim t$. 
Collecting low-pass and band-pass outputs yield wavelet coefficients $\mathcal{W}f=\{P^{t}f, \Psi_jf \}^{\log_2t}_{j=1}$.
To guarantee the stability and permutation invariance to local graph deformations, absolute value nonlinearities are applied to the wavelet coefficients, which are then aggregated using global statistical moments (e.g., mean or $q$-moments).
As a result, the geometric scattering transform provides a multiscale, permutation-invariant Euclidean representation of the graph $G$, which can be used directly within various machine learning frameworks, including neural networks.

\subsection{Geometric Scattering Autoencoder}

Castro et al. developed a deep graph embedding framework, called the geometric scattering autoencoder (GSAE) that combines the geometric scattering transform with VAEs, to study energy landscapes of RNA secondary structures \cite{GSAE}. GSAE has three major parts: an untrained geometric scattering transform, a trained variational autoencoder (VAE) and a trained auxiliary regression network, where the latter two networks together form a semi-supervised VAE. The geometric scattering transform first extracts continuous high-dimensional features, called scattering coefficients, from the discrete input graph, and these are then embedded into low-dimensional representations through the semi-supervised VAE. This embedding approximately retains important biophysical information, such as free energy, that can be used for further study. However, this approach is currently limited to single-stranded secondary structures, whereas many nucleic acid reactions of interest are typically multi-stranded, and it does not address the visualization of trajectories through such energy landscapes. Our work is build upon GSAE by augmenting it with domain-specific features and extensively extending it to support multi-stranded DNA secondary structures, as well as to visualize reaction trajectories across the energy landscape.

\subsection{Potential of Heat Diffusion for Affinity-Based Transition Embedding}

Potential of heat diffusion for affinity-based transition embedding (PHATE) is a nonlinear, unsupervised dimensionality reduction method designed to embed high-dimensional data into a low-dimensional space, typically two-dimensional Euclidean space \cite{PHATE}. PHATE captures both local and global nonlinear structures by combining principles from manifold learning, diffusion geometry, and multidimensional scaling, demonstrating consistent visualization results across varying datasets, such as single-cell RNA sequencing and gut microbiome data.

Given a set of high-dimensional data points, PHATE first computes pairwise distances and converts them into affinities rather than directly relying on Euclidean distances. This affinity representation alleviates the curse of dimensionality and better preserves local neighborhood relationships. To capture global structure, PHATE performs a diffusion process by constructing a Markov transition matrix and propagating affinities through a $t$-step random walk. This diffusion process is modeled analogously to the heat equation, and the optimal diffusion step $t$ is determined based on the von Neumann entropy \cite{VNE}. While diffusion distances are effective for capturing manifold geometry, they tend to lose sensitivity for far-apart points. To address this, PHATE introduces the potential distance to measure the far apart distances between log-transformed transition probabilities from the powered diffusion operator. This transformation enhances the separation of distant points while retaining meaningful local relationships. Finally, PHATE applies metric multidimensional scaling (MDS) \cite{MDS} to the potential distance matrix, yielding a low-dimensional (usually 2D) embedding.
In our framework, we employ PHATE after obtaining the VAE embeddings, mapping these learned latent representations into two-dimensional coordinates for subsequent visualization.

\section{Summary of Contributions}
The contributions of Part I of this dissertation are as follows:
\begin{enumerate}
    \item We propose ViDa, a novel biophysics‑informed deep learning framework that leverages variational autoencoders and geometric scattering transforms to generate biophysically plausible embeddings for visualizing DNA reaction state spaces and trajectories. See Chapter \ref{chap:vida_mlcb}.
    \item We apply ViDa to DNA hybridization reactions and conduct both visual and quantitative comparisons against state‑of‑the‑art visualization methods. We introduce new evaluation metrics to measure the local structure preservation of embedded energy landscapes and smoothness of embedded trajectories. Across these metrics, ViDa reveals superior performance compared to existing methods. See Chapter \ref{chap:vida_2strand}.

    \item We extend ViDa to ViDa‑3Strand and apply it to toehold-mediated three‑way strand displacement reactions. The visualization results demonstrate that ViDa‑3Strand is capable of interpreting underlying reaction mechanisms. See Chapter \ref{chap:vida_3strand}.
\end{enumerate}

\chapter{ViDa: Visualizing DNA Reaction Trajectories with Biophysics-Informed Deep Graph Embeddings} \label{chap:vida_mlcb}

This chapter is a modified version of two papers by Chenwei Zhang, et al., published in NeurIPS 2022, Machine Learning in Structural Biology (MLSB) Workshop (\url{https://arxiv.org/abs/2311.03409}) \cite{vida-nips} and in Machine Learning in Computational Biology (MLCB) 2023 Proceedings (\url{https://proceedings.mlr.press/v240/zhang24a/zhang24a.pdf}) \cite{vida-mlcb}.

\section{Motivations and Contributions}

Given the limitations of existing visualizations for DNA reaction kinetics that we discussed in the previous chapter, we aim to develop an interactive visualization tool that makes it possible for domain specialists to quickly gain intuition about how secondary structures influence reaction folding pathways and to identify kinetic traps. To this end, we introduce a new visualization method called \emph{\textbf{ViDa}}, which uses a semi-supervised variational autoencoder with graph embeddings. 
The assessment of ViDa is motivated by two key questions: (\romannumeral1) Do ViDa's energy landscape embeddings, influenced by biophysics-informed loss terms during model training, preserve local structure? 
That is, are secondary structures with similar base pairs and energies clustered together?
(\romannumeral2) More importantly, do trajectories laid out on the energy landscapes progress smoothly and reveal meaningful underlying reaction mechanisms, such as alternative reaction pathways and kinetic traps? 
In this chapter, we present ViDa as a visualization tool for DNA reaction trajectories. It embeds DNA secondary structures emitted by elementary-step reaction simulators in a 2D landscape, using a semi-supervised VAE embedding that leverages domain knowledge to determine custom training loss terms. 

Our main contributions in this chapter are:
\begin{enumerate}
    \item We develop a new workflow for dimensionality reduction and visualization for both DNA secondary structure state space and reaction trajectories. Additionally, ViDa supports interactive exploration of the state space and trajectories.
    \item We augment ViDa with three biophysically informed features of the DNA reaction domain during training, leading to significant quality improvements over the state-of-the-art methods in visualizing DNA reactions.
\end{enumerate}

In this chapter, we introduce the design of the ViDa framework, the loss functions, and the implementation. In Chapters \ref{chap:vida_2strand} and \ref{chap:vida_3strand}, we will introduce ViDa's applications in two-stranded (DNA hybridization) and three-stranded (toehold-mediated three-way strand displacement) reactions, respectively.

\section{The ViDa Model}

\begin{figure}[!ht]
  \centering
  \includegraphics[width=\linewidth,trim={0cm 30cm 0cm 15cm} ,clip]
  {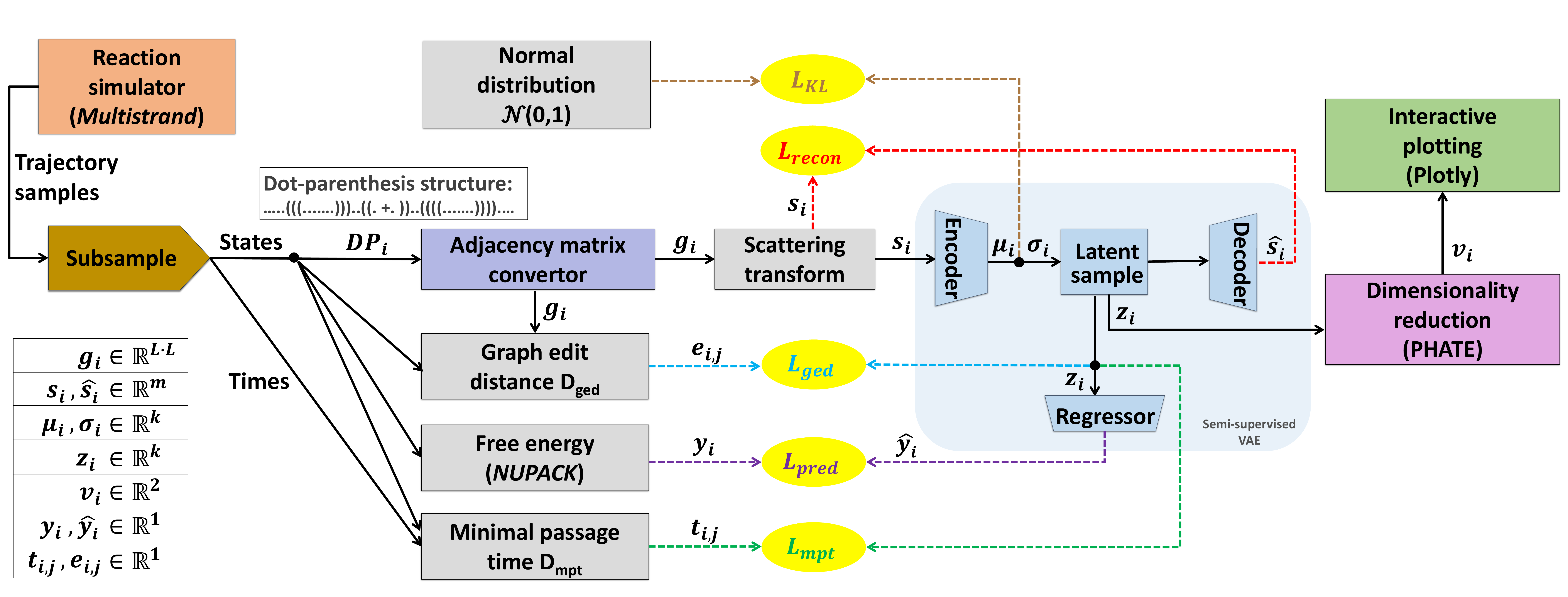}
  \caption{The ViDa framework consists of several major parts: the Multistrand reaction simulator, a converter from dp notation to adjacency matrices, an untrained geometric scattering transform, a trained semi-supervised VAE, the nonlinear DR technique PHATE, and post-processing components for interactive plotting and/or clustering. $DP_i$ is a sampled secondary structure, $g_i$ its graph adjacency representation and $s_i$ the corresponding vector of scattering coefficients, whereas 
  $\mu_i$ and $\sigma_i$ are the mean and standard deviation of the multivariate latent distribution,
  $z_i$ is a latent sample with dimension of $k=25$,
  $\hat{s}_i$ is the reconstructed scattering transform,
  $y_i$ and $\hat{y}_i$ are the simulated and regressed free energy values,
  $t_{i,j}$ and $e_{i,j}$ are the minimum passage time distance and graph edit distance between the secondary structures $i,j$,
  and $v_i$ is a 2D embedding of $z_i$.
  The training loss is composed of five terms: $L_{recon}$, $L_{KL}$ (unsupervised), $L_{pred}$, $L_{mpt}$, and $L_{ged}$ (supervised).}
  \label{fig:ViDa}
\end{figure}

The ViDa framework pipeline is illustrated in Figure \ref{fig:ViDa}. 
An input set of secondary structures (state), represented using dp notation, their corresponding energies, as well as transition times between consecutively occupied states, were extracted from simulated Multistrand trajectories. 
Each state was converted to a graph adjacency matrix, with a node per nucleotide and two types of edges: strand backbones as determined by the primary structure, and complementary base pairs in the secondary structure. The resulting set of graphs ${G=\{g_1,g_2,...,g_n\}}$ was then passed through
a geometric scattering transform, which converts graph signals ${g_i \in \mathbb{R}^{L\times L}}$ into scattering coefficient vectors ${s_i \in \mathbb{R}^{m}}$,
where $L$ is the sum of the lengths of the single-stranded sequences, $n$ is the total number of simulated states, and usually ${m > L^2}$.

Out of these coefficient vectors, 90\% were randomly assigned to the training set for the supervised VAE model, and the remaining 10\% were assigned to the testing set. 
The encoder network was comprised of two fully connected layers, followed by batch norm layers and RELU activations, and the decoder was chosen to be mirror symmetric.
In order to guide the training and to regularize the embedding space, the latent samples produced by the encoder, ${z_i \in \mathbb{R}^d}$ with ${d \ll m}$, were additionally processed by a regressor network for predicting the free energy.

The overall VAE loss was augmented with three domain-specific predictors: (1) the free energy, evaluated at each sampled $z_i$; (2) the ``minimum passage time'' distance $D_{mpt}$ (detailed in the next section); and (3) the graph edit distance $D_{ged}$ for pairs ${\left(z_i,z_j\right)}$.
After training, all scattering coefficient vectors ${s_i \in \mathbb{R}^{m}}$, including those from both the training and test sets, were fed into the trained VAE model to generate latent vectors ${Z_G=\{z_1,z_2,...,z_n\}}$. Finally, the dimensionality reduction algorithm, PHATE \cite{PHATE} was applied to these latent vectors, producing the 2D embedding $V_G=\{v_1,v_2,..,v_n\}, v_i \in \mathbb{R}^2$ for visualization and clustering.

\section{Domain-Specific Losses} \label{vidaloss}
The total training loss for the ViDa model is made up of five terms:
\begin{equation}
    L_{tot} = \alpha L_{recon} + \beta L_{KL} + \gamma L_{pred} + \delta L_{mpt} + \epsilon L_{ged},
\end{equation}
where $\alpha$, $\beta$, $\gamma$, $\delta$, and $\epsilon$ are hyperparameters which can be tuned later.
The reconstruction loss $L_{recon}$ and latent loss $L_{KL}$ constitute the original VAE model, and we include
three domain-specific regression terms, namely the free energy loss $L_{pred}$ for the auxiliary regression network, the minimum passage time distance loss $L_{mpt}$, and the graph edit distance loss $L_{ged}$. The free energy loss is calculated using free energy $y_i$ assigned by the Multistrand simulator: 
\begin{equation}
    L_{pred}= \sum_{i}\left(\hat{y}_i - y_i\right)^2.
\end{equation}

We define the minimum passage time distance loss as
\begin{equation}
    \begin{split}
    L_{mpt}= \sum_{i,j}w_{i,j}\cdot{\left(||z_i - z_j|| - t_{i,j}\right)^2}.
    \end{split}
\end{equation}
Here, $t_{i,j}$ is the estimated minimum passage time from state (node) $i$ to state $j$, computed from the simulated trajectories. Concretely, $t_{i,j}$ was obtained as the shortest path length between nodes $i$ and $j$ on a weighted directed graph constructed in a preprocessing stage from the sampled Multistrand trajectories. In this graph, a directed edge $i \rightarrow j$ exists if the transition from state $i$ to state $j$ was observed in the dataset of transitions. The edge weight was then chosen to represent the expected holding time of the source state $i$, estimated as the empirical average of its sampled outgoing transition times.
$w_{i,j}=p_ip_j$ is an importance weight, based on the empirical probabilities $p_i$ and $p_j$ of states $i$ and $j$.

Analogously, we define the graph edit distance loss as
\begin{equation}
    \begin{split}
    L_{ged}= \sum_{i,j}{\left(||z_i - z_j|| - e_{i,j}\right)^2} \,,
    \end{split}
\end{equation}
where $e_{i,j}$ is the graph edit distance between states $i$ and $j$.

The $k$ nearest neighbours of each state were identified under two complementary distance measures to balance structural and temporal similarity while reducing computational cost. 
For the minimum passage time distance, the $k$ neighbours of state $i$ were chosen as the $k$ states $j$ with the smallest precomputed passage times $t_{i,j}$, calculated using Dijkstra's algorithm \cite{dijkstra} as implemented in the NetworkX Library \cite{networkx}. 
For the graph edit distance, secondary structures were represented as adjacency matrices during preprocessing, allowing the distance to be computed as the $\ell_1$ (Manhattan) distance between flattened adjacency matrices. These flattened matrices were then indexed using the Annoy Library \cite{annoy}, enabling approximate nearest neighbour search to avoid exhaustive pairwise comparisons. For each state $i$, its $k$ nearest neighbours $j$ and the corresponding distances $e_{i,j}$ were efficiently retrieved in this way.
In this work, we set $k=100$ for all experiments. 

The free energy term, $L_{pred}$, guides the network to embed states with similar energies close to each other while repelling states with very different energies. The minimum passage time term, $L_{mpt}$, and the graph edit distance term, $L_{ged}$, encourage that states with similar kinetic properties and geometric structures are positioned close to each other. Together, these three auxiliary terms help preserve local structure.

\section{Training}
The ViDa model is intended to be trained separately for each DNA reaction. For our experiments,
the bottleneck dimension of the VAE was set to $d=25$ and training was performed using PyTorch’s Adam optimizer.
The maximum epoch size was set to $100$, with an early stopping applied ($\texttt{patience}=10$) to prevent overfitting. The batch size was set to $256$.

For DNA hybridization reactions (detailed in Chapter \ref{chap:vida_2strand}), the VAE loss hyperparameters were set to $\alpha=1$, $\beta=0.0001$, and $\gamma=0.3$ for both Gao-P4T4 \cite{gaohelix} and Hata-39 \cite{hata} cases. For Gao-P4T4, $\delta=0.0004$ and $\epsilon=0.00004$, whereas for Hata-P39, $\delta=0.0001$ and $\epsilon=0.0001$.
The choice of $\alpha$, $\beta$, and $\gamma$ follows a similar hyperparameter strategy as reported in the GSAE framework \cite{GSAE}, which provided stable training in our setting. For the additional terms $\delta$ and $\epsilon$, we empirically explored larger values but observed unstable optimization, and therefore adopted smaller ones.
The initial learning rate was set to $0.0001$ and then dynamically adjusted by the ReduceLROnPlateau scheduler with default parameters, except for reducing ${\texttt{patience}}$ from $10$ to $5$ to trigger learning rate adjustments earlier, allowing faster convergence. 

For DNA three-stranded displacement reactions (detailed in Chapter \ref{chap:vida_3strand}), the VAE loss hyperparameters were set to $\alpha=1$, $\beta=0.0001$, $\gamma=0.3$, $\delta=0.0004$, and $\epsilon=0.00001$ for all Machinek \cite{MachinekThreeway} cases. The hyperparameter strategy here is similar to that described above.
The initial learning rate was set to $0.00005$ and then dynamically adjusted by the ReduceLROnPlateau scheduler with default parameters, except for ${\texttt{patience}=5}$. 

Since manual calibration of hyperparameters is tedious and time-consuming, to further facilitate hyperparameter tuning while training the VAE network, we utilized Optuna \cite{optuna}, an automatic hyperparameter optimization software framework that employs an advanced Bayesian optimization, Tree-structured Parzen Estimator (TPE) \cite{tpe}.
Specifically, we fixed three hyperparameters: $\alpha=1$, $\beta=0.0001$, and $\gamma=0.3$ since these values proved robust across different reaction types, including both two-stranded and three-stranded reactions, based on our tests. The remaining three hyperparameters were automatically tuned using Optuna within the following searching spaces: learning rate from $0.000001$ to $0.0001$, $\delta$ from $0.0001$ to
$0.001$, and $\epsilon$ from $0.00001$ to
$0.0001$. 
This automated hyperparameter tuning procedure has been deployed in our current ViDa version.

The ViDa model was trained on an Apple M1 Pro with a 10-core CPU, 14-core GPU, 16-core Neral Engine, and 32 GB RAM. Training typically required between $30$ minutes to $1$ hour, depending on the dataset size.
For PHATE, the number of landmarks was set to $2000$, the decay rate to $40$ and the number of nearest neighbours to $5$. 
The ViDa model can be assessed at \url{https://github.com/chenwei-zhang/ViDa}.

\section{Interactive Plotting and Supporting Measurements}
\begin{figure}[!ht]
  \centering
  \includegraphics[width=\linewidth,trim={0cm 17.8cm 0cm 0cm},clip]
  {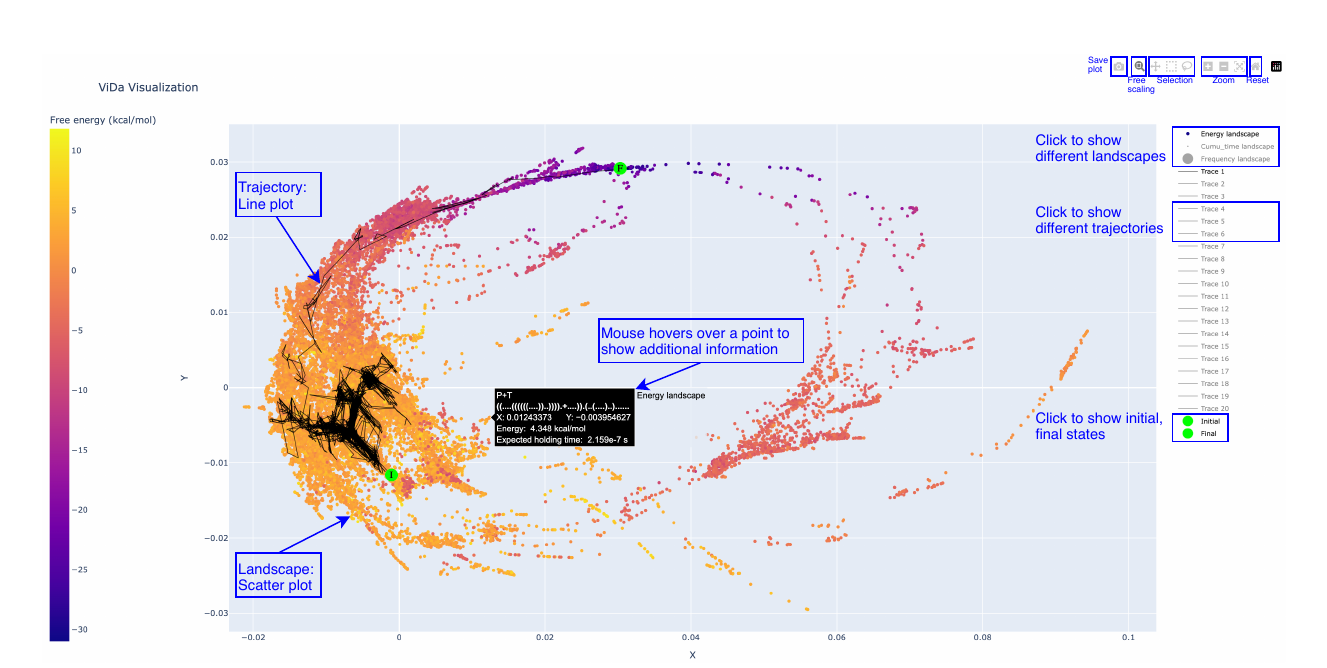}
  \caption{An instructional example showing how to use the designed interactive plotting tool.}
  \label{fig:vida_interactive}
\end{figure}

ViDa provides interactive functionality through the Plotly library \cite{plotly}. Figure \ref{fig:vida_interactive} illustrates an interactive plot generated by ViDa as an example for a DNA hybridization reaction. For ease of use, the interactive plots are exported in HTML format. An example plot from this work can be downloaded at: \url{https://drive.google.com/file/d/12rSpwWLM49bEEhgHQfOJpeKUsSh9Yr4Z/view?usp=sharing}.

In addition to visualization, we provide supporting code to compute complementary statistical measures that enhance the interpretation of reaction trajectories, such as the average reaction time across all simulated trajectories, the proportion of reactive versus non-reactive trajectory samples of a given reaction, and more. 
We also introduce new metrics to quantitatively assess structure preservation of embedded energy landscapes, including the average energy difference and the average graph edit distance difference, as well as trajectory smoothness measured by distortion, as detailed in the next chapter.
We further apply an unsupervised clustering method, Density-Based Spatial Clustering of Applications with Noise (DBSCAN), to identify kinetic traps, as detailed in Supplementary Section \ref{DBSCAN}.
\chapter{ViDa Application I: Visualizing DNA Hybridization} \label{chap:vida_2strand}

This chapter is a modified version of a paper by Chenwei Zhang, et al., published in Machine Learning in Computational Biology (MLCB) 2023 Proceedings (\url{https://proceedings.mlr.press/v240/zhang24a/zhang24a.pdf}) \cite{vida-mlcb}.

\section{Motivations and Contributions}

In this chapter, we employ ViDa to visualize the state space and trajectories of DNA hybridization reactions. Note that while our focus here is on double-stranded complexes,ViDa is also suitable for single-stranded structures such as hairpins.

Our main contributions are:
\begin{enumerate}
    \item We evaluate ViDa on two well-studied DNA reactions with different mechanisms.
    \item We design new evaluation metrics to quantitatively assess the local structure preservation of embedded energy landscapes and the smoothness of trajectories visualized by ViDa.
    \item Our visualization results suggest that ViDa can provide new insights into the underlying mechanisms.
\end{enumerate}

\section{Datasets and Simulations}

\subsection{Datasets}

We present and assess ViDa's visualizations of two DNA hybridization reactions, wherein two unbound complementary strands bind and fold into a double-stranded helix. The first reaction, which we denote by Gao-P4T4, is from Gao et al.'s experimental study \cite{gaohelix}, and the second, which we denote by Hata-39, is from Hata et al.'s experimental study \cite{hata}. The sequences for the two reactions are shown in Table \ref{tab:samples_2strand}, along with some key possible secondary structure motifs for each.

\begin{table}[!ht]
\caption{Sequences of reactions Gao-P4T4 \cite{gaohelix} and Hata-39 \cite{hata}, and examples of key sequence-dependent secondary structure motifs that affect their reactive pathways and reaction rate.}
\vspace{1em}
\label{tab:samples_2strand}
\resizebox{\textwidth}{!}{%
\centering
\begin{tabular}{ll}
\toprule
\multicolumn{2}{l}{\textbf{Gao-P4T4} (25 bases per strand)} \\
& \quad\quad\quad\quad Probe Strand (5' $\rightarrow$ 3')\quad\quad\quad\quad\quad  \quad Target Strand (5' $\rightarrow$ 3') \\
sequences: & \texttt{5$'$-AGATCAGTGCGTCTGTACTAGCACA-3$'$ \quad 5$'$-TGTGCTAGTACAGACGCACTGATCT-3$'$} \\
possible hairpins (size 3): & \texttt{5$'$-....(((.....)))..........-3$'$ \quad 5$'$-..........(((.....)))....-3$'$} \\
possible hairpins (size 4): & \texttt{5$'$-......((((..........)))).-3$'$ \quad 5$'$-.((((..........))))......-3$'$} \\
hybridized structure: & \texttt{5$'$-(((((((((((((((((((((((((-3$'$ + 5$'$-)))))))))))))))))))))))))-3$'$} \\
\midrule
\multicolumn{2}{l}{\textbf{Hata-39} (23 bases per strand)} \\
& \quad\quad\quad\quad Probe Strand (5' $\rightarrow$ 3') \quad\quad\quad\quad Target Strand (5' $\rightarrow$ 3') \\
sequences: & \texttt{5$'$-CCATCAGGAATGACACACACAAA-3$'$ \quad 5$'$-TTTGTGTGTGTCATTCCTGATGG-3$'$}  \\
possible hairpin (size 3): & \texttt{5$'$-.(((.....)))...........-3$'$ \quad 5$'$-.......................-3$'$} \\
possible mis-stack (size 7): & \texttt{5$'$-..............(((((((..-3$'$ + 5$'$-....)))))))............-3$'$} \\
hybridized structure: & \texttt{5$'$-(((((((((((((((((((((((-3$'$ + 5$'$-)))))))))))))))))))))))-3$'$} \\
\bottomrule
\end{tabular}
}
\end{table}

\subsection{Multistrand Simulations}
All simulations for both Gao-P4T4 and Hata-39 were performed using Multistrand’s \textit{first step mode}. In each simulation, the initial state was Boltzmann-sampled from the set of all structures containing exactly one inter-strand base pair. Simulations were terminated either when the two strands unbound or when all bases paired with their intended complements. The simulation conditions, consistent with the associated experimental setup, are listed in Table \ref{tab:multistrand_condition}, and the kinetic model rate parameters are provided in Supplementary Table \ref{tab:multistrand_rates}.

\subsubsection{Gao-P4T4 Simulation}
The strands in Gao-P4T4, which involve 25 bases each, were designed such that 4-stem hairpins could form \cite{gaohelix}. The experimental hybridization measurements from this study are currently best understood with the follow-up analyses by Schreck et al. \cite{schreck}. They argue that the 4-stem hairpins slow hybridization by destabilizing partially formed duplexes, in addition to occluding potential binding sites and impeding the ``zippering'' of strands.

For this case, we generated 5000 trajectory samples, comprising 224 reactive pathway samples and 4776 non-reactive pathway samples, using the Metropolis rate model. All 40789 unique states identified during simulation were included in the embedding.

\subsubsection{Hata-39 Simulation}
The strands in Hata-39, which involve 23 bases each, were designed with the intention of making mis-nucleation and hairpin formation unlikely \cite{hata}. Hata-39 is currently best understood with the follow-up analyses by Lovrod et al. \cite{lovrod}. They show that the Hata-39 sequence gives rise to important secondary structures that are not common among hybridization reactions, and not generally considered in hybridization models. More specifically, it is possible for these strands to form stable stacks (3+ consecutive desired inter-strand base pairs), stable mis-stacks (3+ consecutive \textit{undesired} inter-strand base pairs), and hairpins (of size 3+) \textit{simultaneously}, leading to a diverse set of reactive pathways. The analysis involves a definition of eight \textit{structural types} of secondary structures, which we use in this subsection to color the states in each plot and argue about the quality of our embedding. 

For this case, trajectories were collected until 50 reactive pathway samples were generated, resulting in a total of 3145 trajectory samples, including 3095 non-reactive ones. The Arrhenius rate model was used for the simulations, and all 56702 unique states identified during simulation were included in the embedding. \footnote{Note that the Hata-39 simulation data were directly taken from Lovrod et al. \cite{lovrod}.}

\begin{table}[!t]
\caption{Multistrand simulation conditions for Gao-P4T4 and Hata-39.}
\vspace{0.5em}
\label{tab:multistrand_condition}
\resizebox{\textwidth}{!}{%
\small
\centering
\begin{tabular}{l|ccccc}
\toprule
  Reaction & Temp. (\textdegree{}C)  & $Na^{+}$ Conc. (M) & $Mg^{2+}$ Conc. (M) & Probe Conc. (M) & Target Conc. (M) \\
\midrule
Gao-P4T4  & 20 &  0.5  & 0 & $1\times10^{-6}$ & $1\times10^{-6}$   \\
\midrule
Hata-39  & 25 & 0.195 & 0 & $5\times10^{-8}$ & $5\times10^{-8}$   \\
\bottomrule
\end{tabular}
}
\end{table}

\section{Experiments and Results}

\subsection{Case Study 1: Gao-P4T4}

\subsubsection{ViDa Preserves Local Structure in Energy Landscapes} \label{finding1}

\begin{figure}[!hp]
    \centering
    \includegraphics[width=.84\linewidth, trim={0cm 3cm 0cm 0cm} ,clip] {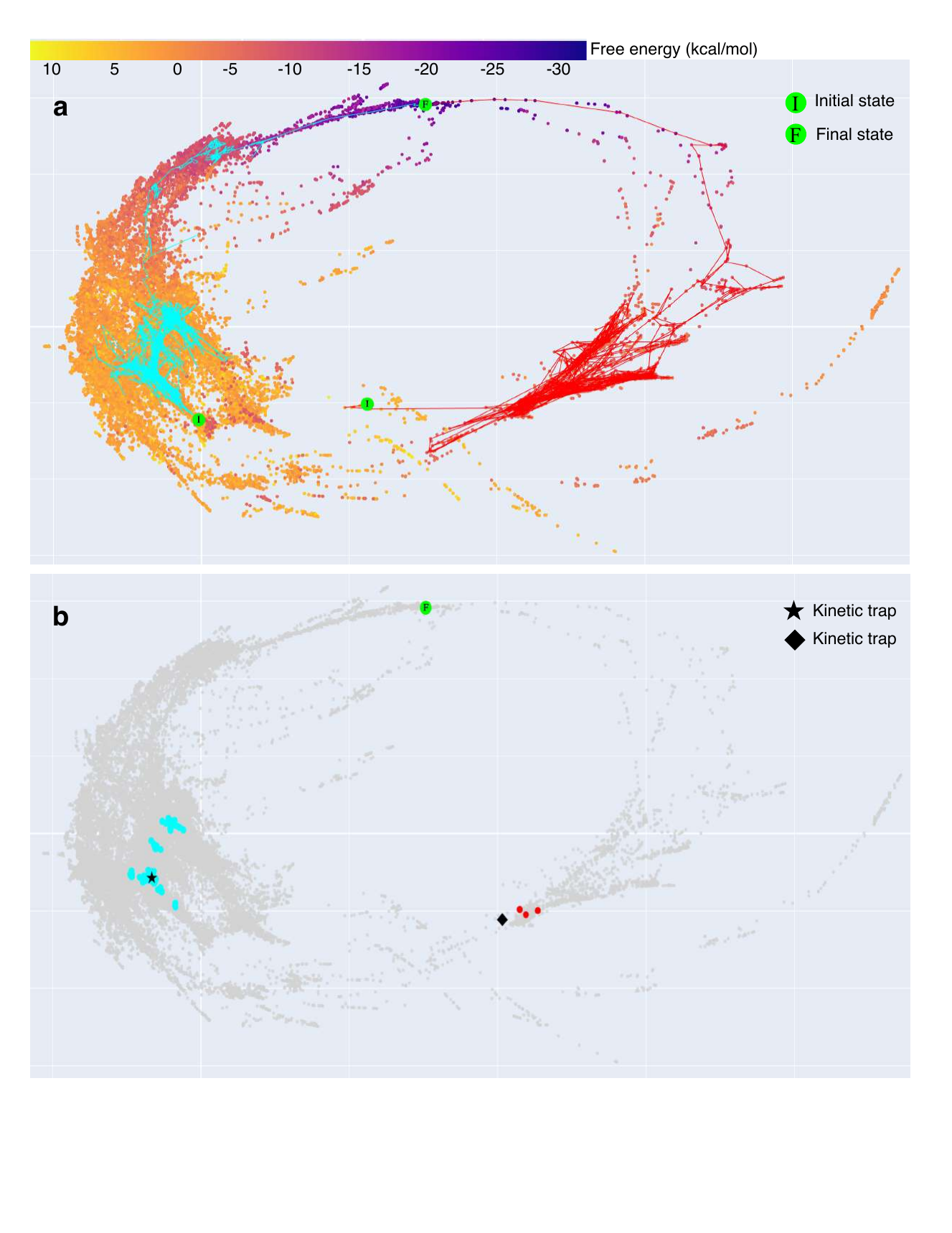}
    \caption{\small
    ViDa embedding results for Gao-P4T4. Each point represents a secondary structure state. The green circle marked $I$ ($F$) denotes the initial (final) state.
    \textbf{(a)} 2D embedding of secondary structure states. The color of each state refers to its free energy. The red and cyan traces represent two different trajectory samples. 
    \textbf{(b)} Results of density-based spatial clustering of applications with noise (DBSCAN) \cite{DBSCAN} on the 2D embedded states. DBSCAN identifies two clusters (cyan and red). The star and diamond symbols indicate the minimum free energy state within each cluster. States that are clustered as noise or filtered out are shown in light gray.
    DBSCAN was applied with parameters $\texttt{eps}=0.0034$ and $\texttt{min\_samples}=4$ (see Supplementary Section \ref{DBSCAN}).
    }
    \label{fig:gao_p4t4_vida}
\end{figure}

The secondary structure embedding for Gao-P4T4 is shown in Figure \ref{fig:gao_p4t4_vida}a. The free energy, which is superimposed on the embedding plot, follows a high-to-low trend from the initial state to the final hybridized state. 
By manually hovering over the points in the interactive plot, we find that neighbouring structures often only differ by a few base pairs, qualitatively suggesting that ViDa preserves local structure.

To quantitatively assess the local structure preservation, we evaluate two aspects: (i) whether neighboring states (secondary structures) in the embedding have similar energies; (ii) whether neighboring states in the embedding have similar base-pairing patterns. Energy similarity is measured by the average energy difference between each structure with its neighbouring structures, while structural similarity is measured by the graph edit distance (GED) difference to these neighbours. 
Accordingly, we introduce two metrics: (i) the average energy difference between each structure and its $K$ nearest neighbours in the 2D embedding, and (ii) the average GED difference under the same criterion.
Because the neighborhood size $K$ can be varied, small $K$ values emphasize fine-grained local neighborhoods, whereas larger $K$ values extend the neighborhood and assess progressively broader local relationships.

\begin{figure}[!hp]
    \centering
    \includegraphics[width=.83\linewidth, trim={0cm 4cm 0cm 0cm} ,clip] {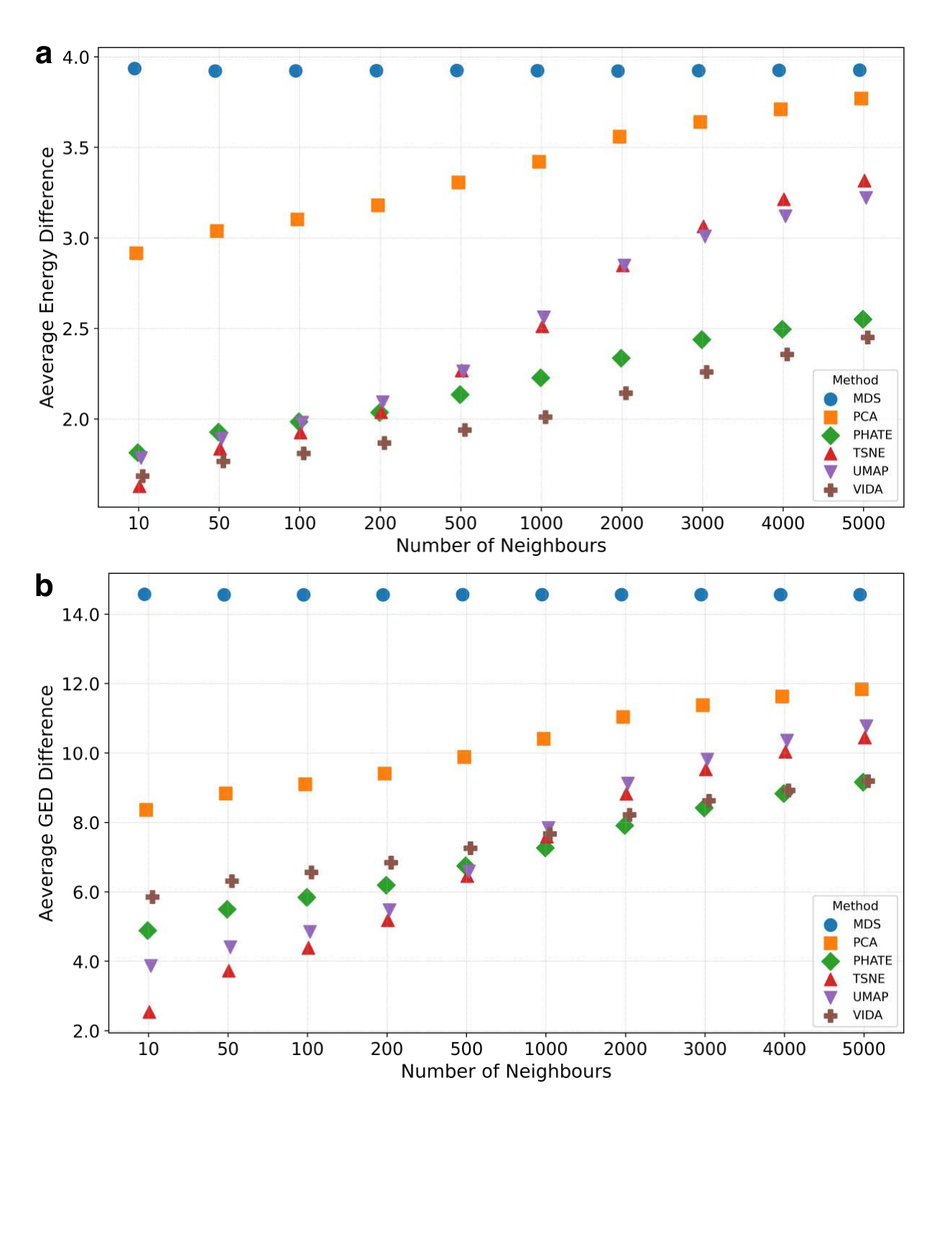}
    \caption{\small
    Scatter plot comparison of local structure preservation across different embedding methods, including ViDa, PCA, PHATE, UMAP, t-SNE, and MDS. The figure illustrates the average difference in \textbf{(a)} energy and \textbf{(b)} graph edit distance (GED) between original state and their K nearest neighbours in the embedding space, for $K=\{10, 50, 100, 200, 500, 1000, 2000, 3000, 4000, 5000\}$. The corresponding numerical values are provided in Supplementary Table \ref{tab:local_preserve}.
    }
    \label{fig:gao_p4t4_localprerserve}
\end{figure}

We compare these metrics across varying neighborhood sizes $K$ for ViDa and several general-purpose dimensionality reduction (DR) methods including PCA, PHATE, UMAP, t-SNE, and multidimensional scaling (MDS) \footnote{The corresponding visualization plots are provided in Supplementary Section \ref{fig:novida}.}. 
The results are shown in Figure \ref{fig:gao_p4t4_localprerserve}. 
For the average energy difference, ViDa consistently achieves the lowest values nearly all $K$, except at $K=10$ where it ranks second lowest. This indicates that ViDa effectively captures energy similarity robustly across different neighborhood sizes. 
For the average GED difference, t-SNE and UMAP maintain relatively small differences when $K < 200$, but their performance degrades sharply as $K$ increases. This behavior reflects their strong bias toward preserving only very local neighborhoods. These patterns are consistent with their visualizations in Supplementary Figures \ref{fig:novida}b and c.
In contrast, ViDa and PHATE maintain low average GED differences across all $K$, demonstrating stable preservation of local structure even when the neighborhood size is expanded.

\subsubsection{ViDa Provides A More Nuanced Understanding of Reaction Mechanisms} \label{finding3}

The embedding (Figure \ref{fig:gao_p4t4_vida}a) separates states into two main branches. The left branch (cyan trace side) corresponds to the reactive pathway in which the helix begins forming at the \texttt{5$'$} end of the probe strand. In other words, this branch contains most structures of the form \texttt{\seqsplit{5$'$-{((([$\ldots$]}-3$'$+5$'$-{[$\ldots$])))}-3$'$}}, where the \texttt{\seqsplit{[$\ldots$]}}s together comprise a legal dp sub-structure. These structures often coincide with the breaking of stable 4-stem hairpins with large loops formed in the pre-collision stage (see Table \ref{tab:samples_2strand} and Supplementary Figure \ref{fig:hairpins}c), and is therefore a slow reactive pathway. In contrast, the right branch (red trace side) corresponds to the reactive pathway in which inter-strand base pairs form near the \texttt{3$'$} end of the probe strand (structures of the form \texttt{\seqsplit{5$'$-{[$\ldots$](((}-3$'$+5$'$-{)))[$\ldots$]}-3$'$}}), and small 3-stem hairpin often form in both strands (see Table \ref{tab:samples_2strand} and Supplementary Figure \ref{fig:hairpins}d). Statistical analysis shows that approximately 3\% of reactive reactions (6 out of 224) proceed via this right-branch pathway.

Laying out trajectories on the embedding, we find two dense regions (see Figure \ref{fig:gao_p4t4_vida}a). We hypothesize the presence of kinetic traps within these regions. To delve deeper into these regions of interest, we first excluded less significant states with exceedingly short cumulative time (we set a cumulative time threshold of $5\times10^{-4} s$ for filtering) and then employed DBSCAN to cluster the post-filtered states (see Figure \ref{fig:gao_p4t4_vida}b). We obtained two clusters, that also each locate around states with minimum free energy (MFE). 
Upon investigating these traps, as detailed in Table \ref{tab:gao_2traps}, we first found that for the kinetic trap in the cyan cluster, annotated as the star symbol, its corresponding secondary structure is \texttt{\seqsplit{5$'$-.((((.((((..........)))).-3$'$ + 5$'$-.((((..........)))).)))).-3$'$}}. The 4-stem hairpins that are extremely stable and hard to break have the same structures as the design of Gao-P4T4 (see Table \ref{tab:samples_2strand}). The cumulative reaction time for this state containing two 4-stem hairpins is notably long at $1.36\times10^{-1}$ s, thereby decelerating the overall reaction in a manner consistent with the computational analysis by Schreck et al. \cite{schreck}. 
The second kinetic trap located in the red cluster, annotated as the diamond symbol, has a secondary structure of \texttt{\seqsplit{5$'$-....(((.....))).((((((((.-3$'$ + 5$'$-.)))))))).(((.....)))....-3$'$}}, with two 3-stem hairpins at both strands (see Supplementary Figure \ref{fig:hairpins}d). These hairpins are also stable, resulting in a relatively long cumulative reaction time of $3.24\times10^{-3}$ s. Notably, these 3-stem hairpins form at completely different locations from the 4-stem hairpins, indicating that they are not intermediate states in 4-stem hairpin formation. Instead, they suggest an alternative reaction mechanism.
Moreover, to avoid bias arising from the choice of cumulative time threshold, which could lead to different DBSCAN clustering results, we applied a smaller threshold of $4\times10^{-5} s$. The resulting cluster plots are shown in Supplementary Figure \ref{fig:dbscan_SI}. This analysis yields five clusters with MFE states annotated using distinct symbols, as detailed in Supplementary Table \ref{tab:gao_5traps}. We observed that the two states with the longest and second-longest cumulative reaction times correspond exactly to the two major traps identified earlier. The remaining three states exhibit short cumulative reaction times and can be regarded as minor traps. This finding corroborates our previous analysis and further reinforce the evidence for the two major kinetic traps.
In summary, our visualization highlights the kinetic trap created by the designed 4-stem hairpins in Gao-P4T4 reaction, which is stable enough to significantly slow down hybridization. Additionally, we identified a second major kinetic trap in Gao-P4T4, with 3-stem hairpins on both strands, which exacerbates the slowness of the reaction process.

\begin{table}[!t]
\caption{Kinetic traps with their associated dot-parenthesis notations, cumulative times, and free energies, filtered using a cumulative time threshold of $5\times10^{-4} s$.}
\vspace{0.5em}
\label{tab:gao_2traps}
\resizebox{\textwidth}{!}{%
\small
\centering
\begin{tabular}{c|ccc}
\toprule
  Kinetic Trap & Secondary Structure with DP-Notation (5' $\rightarrow$ 3')  & Cuml. Time (s) & $\Delta G$ (kcal/mol) \\
\midrule
$\bigstar$  & \texttt{5$'$-.((((.((((..........)))).-3$'$ + 5$'$-.((((..........)))).)))).-3$'$} &  $1.36\times10^{-1}$  &  $-5.87$  \\
\midrule
$\blacklozenge$  & \texttt{5$'$-....(((.....))).((((((((.-3$'$ + 5$'$-.)))))))).(((.....)))....-3$'$}  & $3.24\times10^{-3}$ &  $-10.6$  \\
\bottomrule
\end{tabular}
}
\end{table}

\subsubsection{Domain-Specific Features Improve Trajectory Smoothness} \label{finding2}

Laying out the trajectories on the embedding, all trajectories proceed nicely along the branches (Figure \ref{fig:gao_p4t4_vida}a). Additionally, for all trajectory plots, we did not observe large jumps occurring along the traces, confirming that nearby secondary structures on simulated trajectories tend to be placed nearby in the embeddings. 
In order to quantify this smoothness property, we use a custom metric for distortion/stretch. We define the \emph{average distortion} of an embedding as the frequency-weighted mean Euclidean distance between the images of secondary structure pairs that occur consecutively in the trajectory dataset, normalized by the embedding diameter of all states. In Table \ref{tab:smoothness} we compare the average distortion achieved by ViDa and by general-purpose DR methods including PCA, PHATE, UMAP, t-SNE, and MDS, and find that ViDa achieves a significantly lower average distortion than all other considered methods. 
On the one hand, these general-purpose DR methods do not take into account any domain knowledge beyond the secondary structure data itself, and thereby their visualizations and smoothness are relatively poor. The comparisons suggest the importance of incorporating domain-specific knowledge when training neural networks to make a biophysically-plausible visualization tool, such as our custom loss terms (Section \ref{vidaloss}) that penalize the distortion of local structure. 
On the other hand, we also compared an MDS embedding which only leverages the biophysics-based distance measure of minimum passage time (see Supplementary Figure \ref{fig:novida}e). However, trajectories are densely concentrated across the embedding, making it infeasible to distinguish different folding pathways.
These results emphasize the significance of integrating deep graph embeddings and distance loss metrics for achieving superior results.
In combination with the visualizations, they demonstrate that ViDa can embed the reaction trajectories while preserving some continuity in time.

\begin{table}[!ht]
    \caption{Comparison of average distortion for different embedding methods for Gao-P4T4. Lower values indicate better trajectory smoothness.}
    \label{tab:smoothness}
    \vspace{0.5em}
    \small
    \centering
    \begin{tabular}{l|cccccc}
        \toprule
        \cmidrule{1-7}
        Metric  & \textbf{ViDa (ours)}  & PCA & PHATE  & UMAP & t-SNE  & MDS \\
        \midrule
        \emph{Avg. distortion  $\downarrow$} & $\mathbf{0.003}$  &  $0.219$  &  $0.009$ &  $0.245$  &  $0.260$  & $0.537$  \\
        \bottomrule
    \end{tabular}
\end{table}

\subsubsection{Comparison with State-Of-The-Art Coarse-Grained Visualizations for Hybridization} \label{finding4}

In Figure \ref{fig:gao_p4t4_cg}, we show a coarse-grained representation of Gao-P4T4, similar to visualizations in \cite{MachinekThreeway,PE}. Each secondary structure is mapped to a single macrostate based on (1) the number of base pairs that correspond exactly to base pairs in the desired helix and (2) the number of base pairs that do \textit{not} contribute to the desired helix, for instance base pairs involved in hairpins or mis-stacks. Each macrostate is therefore an ensemble of secondary structures. These sorts of coarse-grained visualizations are easily adjustable, do not require training, and have the capacity to represent all possible secondary structure states and trajectories. However, with this scheme, structurally dissimilar secondary structures may be mapped to the same macrostate, making it difficult to interpret each macrostate and trajectories through them, and to distinguish between different reaction mechanisms. 
In contrast, ViDa's fine-grained embedding overcomes this limitation. ViDa's plots show distinct reaction trajectories, enabling users to interpret reaction mechanisms more straightforwardly and accurately.

\subsubsection{Ablation Study of ViDa}

\begin{figure}[!hp]
    \centering
    \includegraphics[width=\linewidth,trim={0cm 15cm 0cm 0cm}, clip] {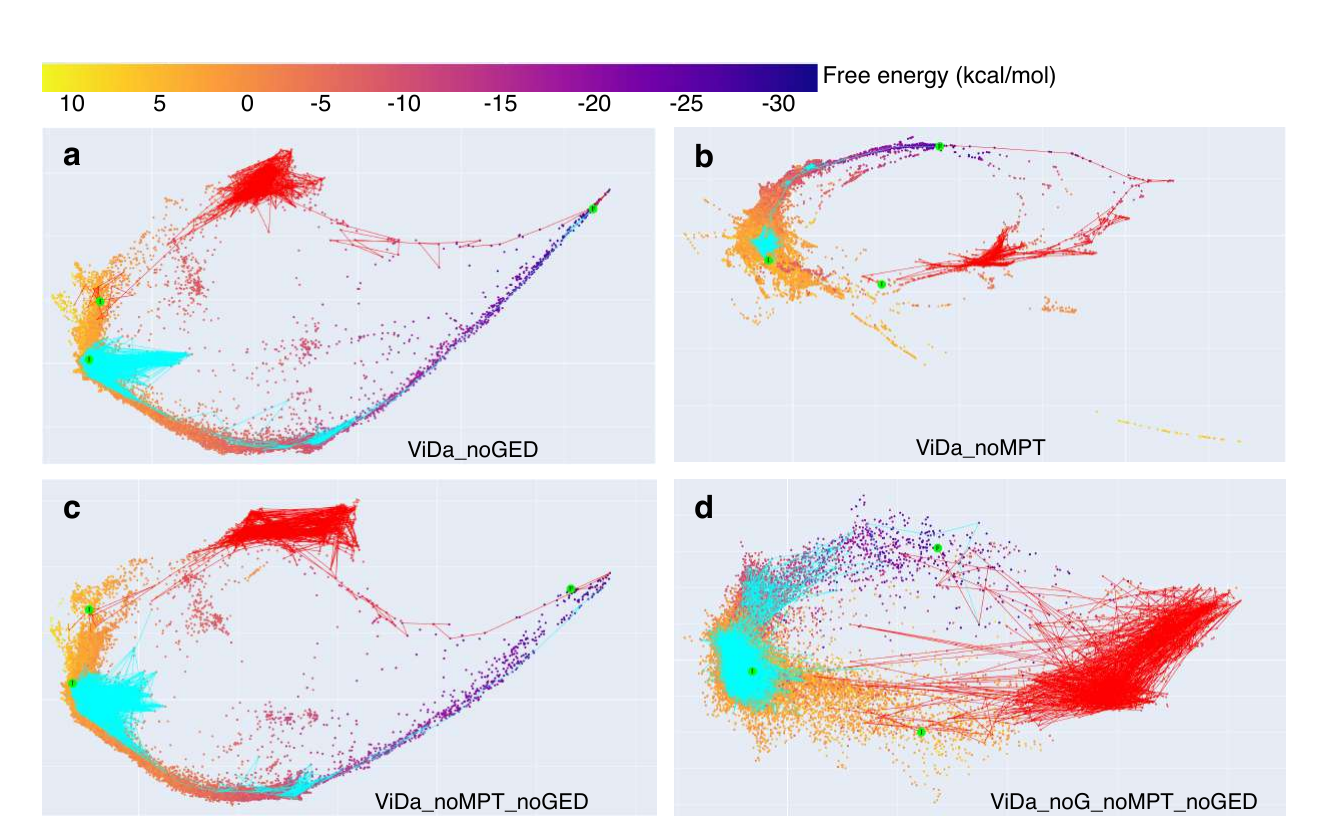}
    \caption{
    Ablated ViDa's visualization plots: \textbf{(a)} ViDa\_noGED, \textbf{(b)} ViDa\_noMPT, \textbf{(c)} ViDa\_noMPT\_noGED, and \textbf{(d)} ViDa\_noG\_noMPT\_noGED.
    Trajectories laid out on the embedding for Gao-P4T4. Each point represents a secondary structure state. The color of each point represents the value of free energy. 
    The cyan and red traces represent two distinct trajectory samples, corresponding to the same samples shown in Figure \ref{fig:gao_p4t4_vida}a.
    The initial and final states are indicated by the green circles marked $I$ and $F$, respectively.
    }
    \label{fig:vida_ablation_viz}
\end{figure}

We incorporate three domain-specific features, including energy (G), minimum passage time (MPT), and graph edit distance (GED), to encourage ViDa to capture richer domain knowledge.  
To evaluate the impact of these factors on ViDa's performance, we conduct four ablation studies by training ViDa under the following conditions: (i) without GED (ViDa\_noGED), (ii) without MPT (ViDa\_noMPT), (iii) without both GED and MPT (ViDa\_noMPT\_noGED), and (iv) without G, GED and MPT (ViDa\_noG\_noMPT\_noGED). Visualizations of these ablation variants for the Gao-P4T4 reaction are shown in Figure \ref{fig:vida_ablation_viz}.

\begin{figure}[!hp]
    \centering
    \includegraphics[width=.9\linewidth, trim={0cm 3cm 0cm 0cm} ,clip] {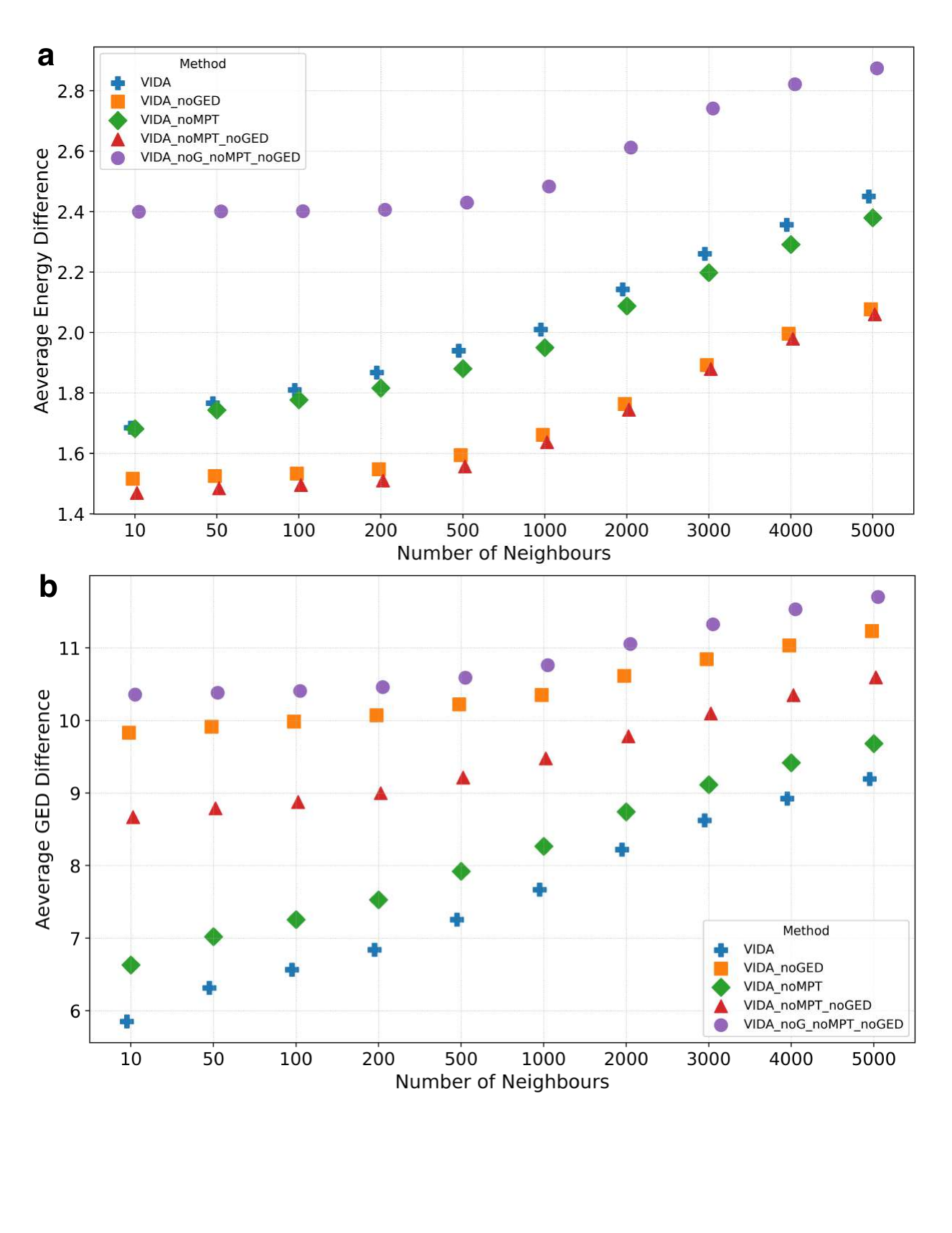}
    \caption{
    Scatter plot comparison of local structure preservation across different ablated ViDa methods, including ViDa\_noGED, ViDa\_noMPT, ViDa\_noMPT\_noGED, and ViDa\_noG\_noMPT\_noGED. The figure illustrates the average difference in \textbf{(a)} energy and \textbf{(b)} graph edit distance (GED) between original state and their K nearest neighbours in the embedding space for various $K$. The corresponding numerical values are provided in Supplementary Table \ref{tab:vida_ablation_local_preserve}.
    }
    \label{fig:vida_knn_ablation}
\end{figure}

Removing the GED term (Figure \ref{fig:vida_ablation_viz}a) weakens local structure preservation, resulting in a higher average GED difference (see Figure \ref{fig:vida_knn_ablation}b). and a much larger distortion (increasing from 0.003 to 0.024, see Table \ref{tab:smoothness_ablation}). 
Removing the MPT term (Figure \ref{fig:vida_ablation_viz}b) causes a slight increase in both average GED difference and distortion (from 0.003 to 0.007). 
When both GED and MPT terms are removed (Figure \ref{fig:vida_ablation_viz}c), the embedding is constrained only by the energy term. This leads to the lowest average energy difference, as expected. However, it comes at the cost of worse local GED preservation and higher distortion (0.025) in trajectory smoothness.
Finally, removing all three domain-specific terms collapses both local structure preservation and trajectory smoothness, leading to the worst average energy and GED differences and a significantly deterioration in distortion (increased to 0.033). This is also revealed in its visual embedding in Figure \ref{fig:vida_ablation_viz}d.
Together, these results confirm that the three domain-specific terms complement on another, and that their combination enhances both local structure preservation and trajectory smoothness in ViDa.

\begin{table}[!ht]
    \caption{Comparison of average distortion for different ablated ViDa methods for Gao-P4T4. Lower values indicate better trajectory smoothness.} \label{tab:smoothness_ablation}
    \vspace{0.5em}
    \small
    \centering
    \resizebox{\textwidth}{!}{%
    \begin{tabular}{l|ccccc}
    \toprule
    \cmidrule{1-6}
    Metric  & \textbf{ViDa}  & ViDa\_noGED & ViDa\_noMPT  & ViDa\_noMPT\_noGED & ViDa\_noG\_noMPT\_noGED  \\
    \midrule
    \emph{Avg. distortion  $\downarrow$} & $\mathbf{0.003}$  &  $0.024$  &  $0.007$ &  $0.025$  &  $0.033$   \\
    \bottomrule
    \end{tabular}
    }
\end{table}

\subsection{Case Study 2: Hata-39} \label{finding5}

\subsubsection{ViDa Embedding Is Compatible with Structural Types}

\begin{figure}[!hp]
    \centering
    \includegraphics[width=.86\linewidth, trim={0cm 2cm 0cm 0cm} ,clip]{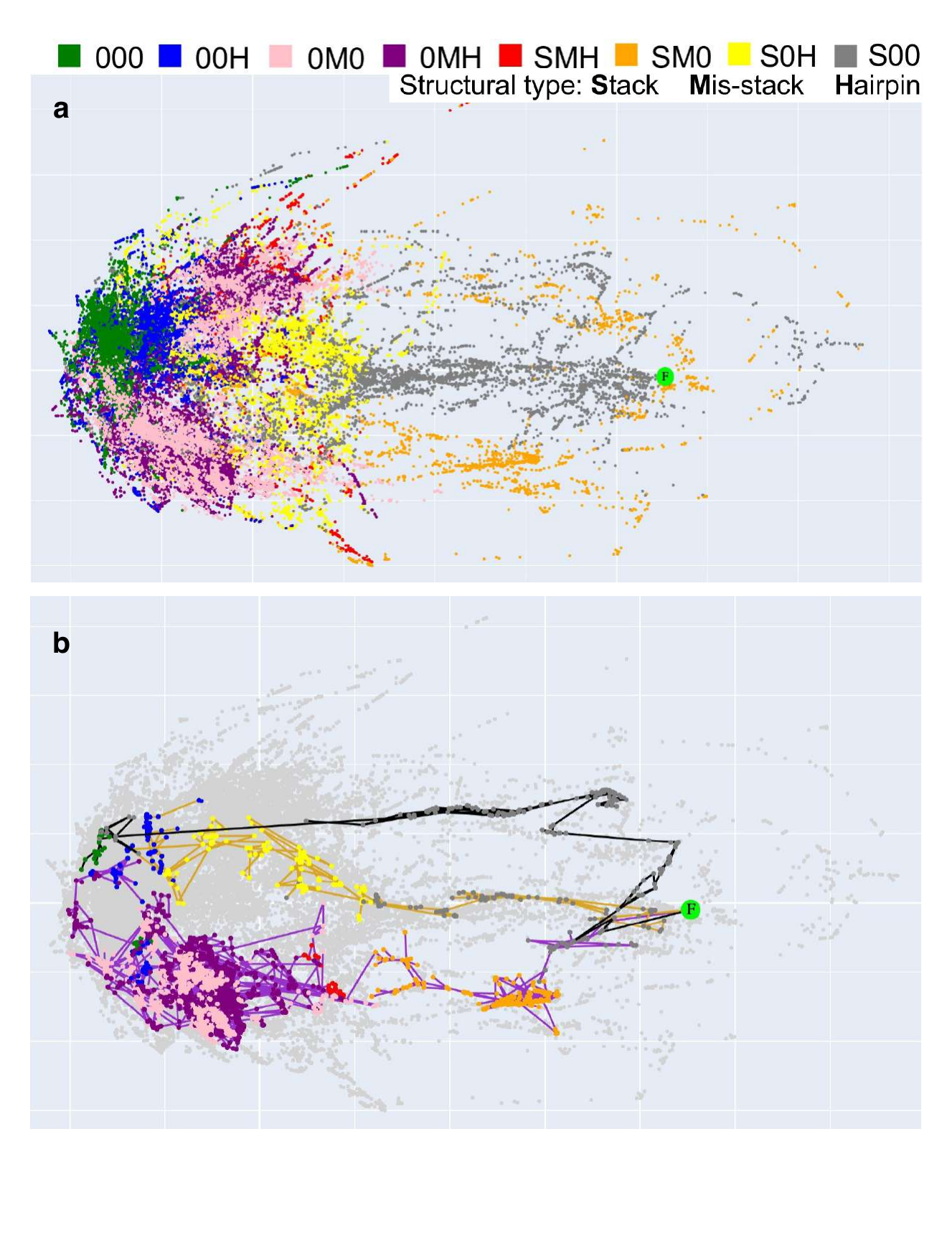}
    \caption{\small
    ViDa embedding results for Hata-39. Each point represents a secondary structure state. The green circle marked $F$ denotes the final (hybridized) state. 
    \textbf{(a)} 2D embedding of secondary structure states. The color of each state refers to its structural type \cite{lovrod}.  For instance, \textit{SM0} denotes the type of secondary structure states with at least one stack, at least one mis-stack, and no hairpins. There are eight structural types in total. 
    \textbf{(b)} Three reactive trajectories laid over the embedding. States that do not lie on one of the three trajectories are shown in light gray.}
    \label{fig:hata39_vida}
\end{figure}

The secondary structure embedding for Hata-39 is shown in Figure \ref{fig:hata39_vida}a. To assess the quality of the embedding and establish compatibility with previous work, we color each state according to its structural type, which is determined by whether there is at least one correctly hybridized stack (\textbf{S}) or not (\textbf{0}), at least one mis-stack (\textbf{M}) or not (\textbf{0}), and at least one hairpin (\textbf{H}) or not (\textbf{0}) \cite{lovrod}. Schematic representations for stacks, mis-stacks, and hairpins are given in Supplementary Figure \ref{fig:hairpins}. 
Although training ViDa does not receive these structural labels as input, states with the same, or similar, types tend to be close together in the embedding, implying that ViDa captures local structure. For instance, states of type \textit{0M0} (pink) are nearby states of type \textit{0MH} (purple), and indeed these structures are closely related because they contain a similar mis-stack, which dominates the hairpins in this reaction (see Table \ref{tab:samples_2strand}). Moreover, states with very different structures are far apart in the embedding. For instance, the states of type \textit{00H} (blue) are generally far from states of type \textit{SM0} (orange), which is reasonable since they do not share any significant structural motifs stacks, mis-stacks, or hairpins.

\subsubsection{ViDa Can Distinguish Between Different Reaction Mechanisms} \label{finding6}

The structural labels can also highlight the reaction mechanisms captured by the embedding. Figure \ref{fig:hata39_vida}b shows three trajectory samples, which are representative of three distinct reaction mechanisms, laid over the embedded secondary structures. The black trajectory is an example of a direct hybridization reaction mechanism, which is extremely fast, but only accounts for $\sim$10\% of the sampled reactive trajectories. The orange and purple trajectories illustrate slower, more complex reactive pathways that, in the case of this reaction, are much more common. Similar to the dominant mechanism described for Gao-P4T4, the orange trajectory includes the formation of a 3-stem hairpin, such as 
\texttt{\seqsplit{5$'$-{.(((.....))).((((((((..}-3$'$+5$'$-{..))))))))..((.....))..}-3$'$}}, 
making this pathway slower than the direct pathway. The purple trajectory, although it also involves the formation of a 3-stem hairpin, is qualitatively distinct from the other two trajectories in its formation of a stable mis-stack, e.g., 
\texttt{\seqsplit{5$'$-.(((.....)))..(((((((..-3$'$+5$'$-....)))))))............-3$'$}}. 
These three reaction mechanisms, originally found and illustrated by Lovrod et al. \cite{lovrod}, are also distinguished by ViDa, suggesting that our embeddings are biophysically meaningful.

\section{Conclusion and Discussion}

In this chapter, we apply ViDa on two well-studied DNA hybridization reactions. We show how ViDa can visually cluster trajectory ensembles into reaction mechanisms, therefore making simulation results more interpretable. ViDa also supports interactive exploration of the landscape and trajectories (details not included).

In the context of multi-stranded reactions, an important direction for improving our method is the partitioning of secondary structure microstates into clusters corresponding to different strand-level complexes, i.e., into macrostates defined by the subset of available strands which are actually bound into a complex.
For our simple example of DNA hybridization, states without inter-strand base pairs (dissociated states) should ideally be separated from those with inter-strand base pairs (associated states). For reactions involving three strands, such as three-way strand displacement, there should be five distinct groups (one group without inter-strand base pairs, three groups with a single dissociated strand each, and one group with the three-way complex). In Chapter \ref{chap:vida_3strand}, we extend ViDa to accommodate three-way strand displacement reactions, and the resulting visualizations show partial separation of different groups.

Another interesting direction is to extend the approach of overlaying embeddings with structural types, as done for the Hata-39 reaction, to gain more insight into reaction mechanisms.
In addition, it would be also useful if trajectory samples could be classified automatically according to their time (e.g., fast) and probability (e.g., rare) to understand the contribution of individual energy basins to the overall kinetics.
\chapter{ViDa Application II: Visualizing DNA Toehold-Mediated Three-Way Strand Displacement} \label{chap:vida_3strand}

\section{Motivations and Contributions}

In the previous chapter, we introduced ViDa to visualize DNA hybridization trajectories, focusing solely on single- or two-stranded DNA reactions. However, numerous complex reactions involve three or more strands, presenting an opportunity to expand the framework's applicability.
DNA strand displacement reactions have been widely used in designing advanced nanotechnological devices \cite{nanotech1,nanotech2}. Kinetic control of such reactions is essential in autonomous molecular machinery and molecular computation \cite{MachinekThreeway,Zhang_natchem}. It is well established that introducing mismatched base pairs can strongly affect kinetic rates \cite{mismatch}. However, the mechanisms of such effects are not fully understood.
In this chapter, we present \emph{\textbf{ViDa-3Strand}}, an extension of ViDa tailored to accommodate toehold-mediated three-way strand displacement reactions.

Our main contributions are:
\begin{itemize}
    \item We develop ViDa-3Strand for visualizing both DNA secondary structure state space and reaction trajectories of three-way strand displacement. 
    \item We apply ViDa-3Strand to several experimentally studied reaction systems from Machinek et al. \cite{MachinekThreeway}, providing better visual insights into the underlying reaction mechanisms. 
\end{itemize}

\section{Datasets and Simulations}

\subsection{Datasets}

We analyze four toehold-mediated three-way strand displacement reactions from Machinek et al. \cite{MachinekThreeway}, as listed in Table \ref{tab:3strand_data}. 
Each reaction involves a target, an incumbent, and an invader strand. 
The target strand has 27 bases and the invader has 24 bases in each reaction. The incumbent strand that interacts with the reporter has 33 bases while the incumbent that does not interact with the reporter has 16 bases.
The reporter is a DNA duplex carrying a fluorophore–quencher pair that produces a fluorescence signal upon displacement, thereby enabling indirect monitoring of reaction kinetics. In reporter-based systems, displacement of the incumbent strand leads to the release of a fluorophore-labeled strand from the reporter duplex. In contrast, in reporter-free systems, the incumbent strand itself is fluorophore-labeled. Using a reporter avoids the need to label primary reactants directly, minimizing potential perturbations of their interactions.

Both the perfect-toehold8 and proximal-toehold8 reactions include 8-base toeholds in the invader (highlighted in blue). In proximal-toehold8, the invader contains a mismatched base ``C'' near the toehold (highlighted in red). \footnote{Note that both perfect-toehold8 and proximal-toehold8 reaction simulation datasets in this work were obtained from our coworker Lovrod.}
Similarly, both the perfect-toehold7-reporter and proximal-toehold7-reporter reactions have 7-base toeholds in the invader, with the invader of proximal-toehold7-reporter containing a mismatched base ``C'' near the toehold.

\begin{table}[!hp]
\centering
\renewcommand{\arraystretch}{1.1}
\caption{A series of sequences for mismatch-free and mismatched three-way strand displacement reactions \cite{MachinekThreeway}. Toehold bases in the invader are shown in blue and mismatched bases are shown in red. The bases in the incumbent
that interact with the reporter are shown in italics.}
\label{tab:3strand_data}
\vspace{1em}
\resizebox{\textwidth}{!}{
\begin{tabular}{llp{10.5cm}}
\toprule
Reaction & Species & Sequence (5' $\rightarrow$ 3') \\
\midrule
\midrule
\multicolumn{3}{l}{\textbf{Mismatch-Free Reactions}} \\
& Target & CCC TCC ACA TCA ACC TCA AAC TCA CC \\
\makecell[l]{perfect-\\toehold8}  & Incumbent & GGT GAG TTT GAG GTT G \\
& Invader & GGT GAG TTT GAG GTT G\textcolor{blue}{AT GTG GAG} \\
\midrule
& Target & CCC TCC ACA TTC AAC CTC AAA CTC ACC \\
\makecell[l]{perfect-toe\\hold7-reporter}  & Incumbent & \textit{TGG TGT TTG TGG GTG TGG TGA G}TT TGA GGT TGA \\
& Invader & GGT GAG TTT GAG GTT GA\textcolor{blue}{A TGT GGA} \\
\midrule
\midrule
\multicolumn{3}{l}{\textbf{Mismatched Reactions}} \\
& Target & CCC TCC ACA TCA ACC TCA AAC TCA CC \\
\makecell[l]{proximal-\\toehold8} & Incumbent & GGT GAG TTT GAG GTT G \\
& Invader & GGT GAG TTT GAG GTT \textcolor{red}{C}\textcolor{blue}{AT GTG GAG} \\
\midrule  
& Target & CCC TCC ACA TTC AAC CTC AAA CTC ACC \\
\makecell[l]{proximal-toe\\hold7-reporter} & Incumbent & \textit{TGG TGT TTG TGG GTG TGG TGA G}TT TGA GGT TGA \\
& Invader & GGT GAG TTT GAG GTT \textcolor{red}{C}A\textcolor{blue}{A TGT GGA} \\
\bottomrule
\end{tabular}
}
\end{table}

\subsection{Multistrand Simulations}
All simulations for the four reactions were performed using Multistrand’s \textit{first step mode} with the Arrhenius rate method. In each simulation, the initial state was Boltzmann-sampled from the set of all structures containing exactly one inter-strand base pair, between the invader strand and target-incumbent complex.
Simulations were terminated when the invader strand dissociated or when it fully displaced the incumbent strand and the incumbent subsequently dissociated. The simulation conditions, consistent with the associated experimental setup, are listed in Table \ref{tab:multistrand_condition_3strand}, and the rate parameters are provided in Supplementary Table \ref{tab:multistrand_rates}.
Since Multistrand simulates each elementary step during reacting, a single trajectory for complex processes such as three-stranded displacement may involve hundreds of millions of intermediate secondary structures. To substantially reduce trajectory size, we sampled one secondary structure at fixed time intervals, specifically, every $1\times10^{-7}$ simulating seconds. To ensure a sufficiently large set of unique secondary structures that forms a relatively large state space while avoiding excessive computational cost, we targeted approximately 70000 to 140000 unique secondary structure states. The simulation results are as follows:
\begin{itemize}
    \item perfect-toehold8: 1000 trajectories were generated, containing 107187 unique secondary structures. Among them, 177 trajectories were successful, i.e., the incumbent strand is completely displaced and dissociated from the target.
    \item proximal-toehold8: 1000 trajectories were generated, containing 69962 unique secondary structures. Among them, only 1 trajectory was successful.
    \item perfect-toehold7-reporter: 400 trajectories were generated, containing 142090 unique secondary structures. Among them, 27 trajectories were successful.
    \item proximal-toehold7-reporter: 2000 trajectories were generated, containing 131679 uni\-que secondary structures. Among them, all trajectories failed, i.e., the invader strand detaches from the target.
\end{itemize}

\begin{table}[!t]
\caption{Multistrand simulation conditions for Machinek's reactions.}
\vspace{0.5em}
\label{tab:multistrand_condition_3strand}
\resizebox{\textwidth}{!}{%
\small
\centering
\begin{tabular}{l|ccccc}
\toprule
Reaction & Temp. (\textdegree{}C)  & $Na^{+}$ Conc. (M) & $Mg^{2+}$ Conc. (M) & Invader Conc. (M) & Target/Incumbent Conc. (M) \\
\midrule
perfect-toehold8  & 23 &  0.05  & 0.01 & $1\times10^{-8}$ & $1\times10^{-8}$   \\
\midrule
perfect-toehold7-reporter  & 23 & 0.05 & 0.01 & $5\times10^{-9}$ & $5\times10^{-9}$   \\
\midrule
proximal-toehold8  & 23 & 0.05 & 0.01 & 1$\times10^{-8}$ & $1\times10^{-8}$   \\
\midrule
proximal-toehold7-reporter  & 23 & 0.05 & 0.01 & $1\times10^{-8}$ & $1\times10^{-8}$   \\
\bottomrule
\end{tabular}
}
\end{table}

\subsection{Kinetic Hypotheses by Machinek et al.}

The reaction rate of strand displacement is known to depend both on the toehold strength (i.e., on its length and sequence), and on potential positions of a mismatch introduced by the invader within the displacement domain \cite{MachinekThreeway,Zhang_natchem}. Yet, understanding the underlying reaction mechanism remains a challenge. Machinek et al. \cite{MachinekThreeway} visualized the reaction’s free-energy landscapes and used a coarse-grained molecular dynamics simulator known as OxDNA \cite{oxDNA} to investigate the effects of mismatches on strand displacements. They suggest that a mismatch close to the toehold significantly destabilizes the system, increasing the likelihood of detachment in the invader-target pairing and consequently reducing the overall reaction rate.

\section{Experiments and Results}

\subsection{Mismatch-Free Reactions}

\subsubsection{ViDa-3Strand Preserves Local Structure}

\begin{figure}[!hp]
  \begin{center}
      \includegraphics[width=0.83\linewidth, trim={0cm 1.5cm 0cm 0cm}, clip]{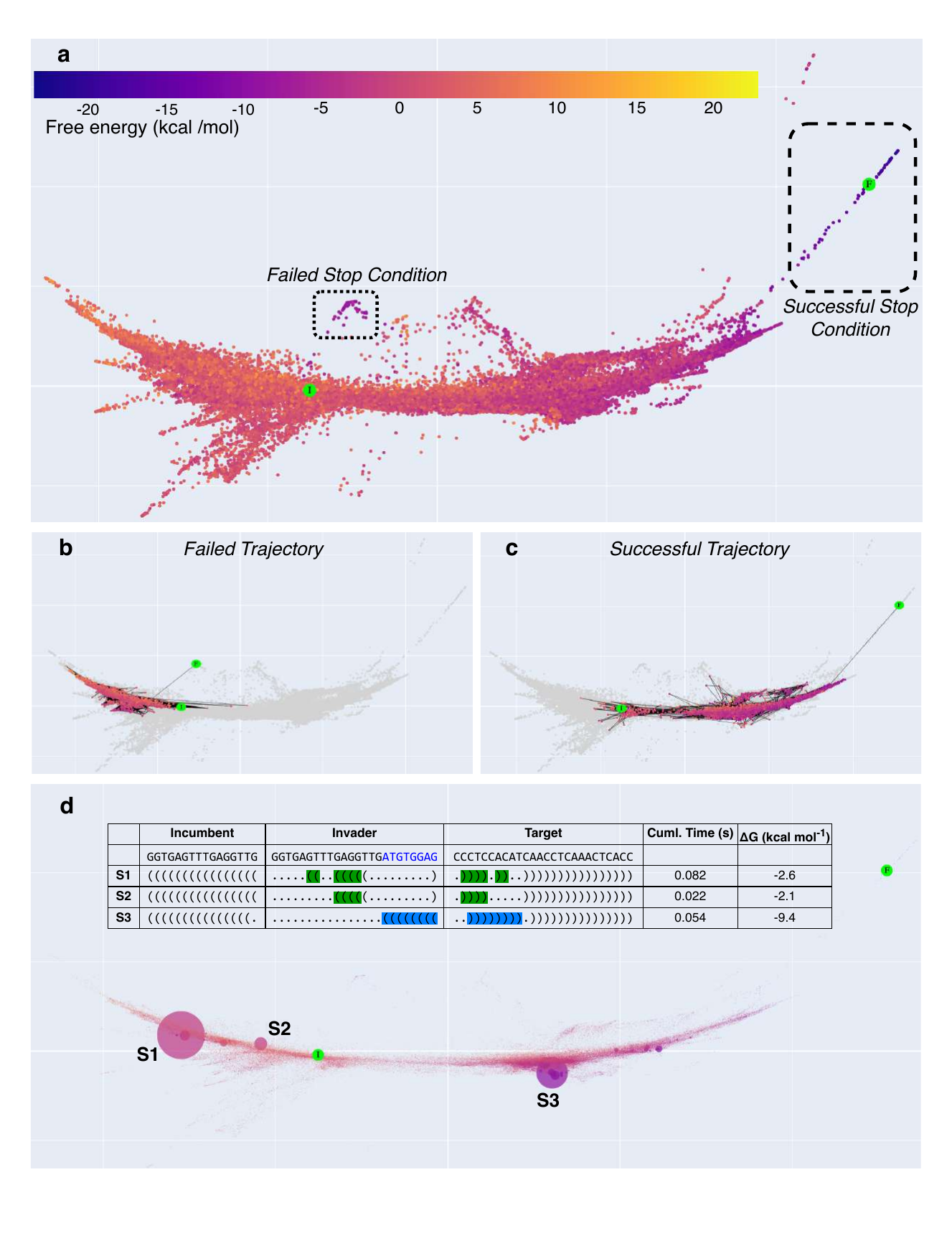}
      \caption{\small 
      \textbf{(a)} ViDa-3Strand-generated secondary structure state space with states laid out on the energy landscape for the perfect-toehold8 reaction. Each point refers to a state with the color representing the value of free energy. The green blobs annotated by ``I'' and ``F'' indicate specific initial and successful final states, respectively.
      \textbf{(b)} and \textbf{(c)} show representative failed and successful trajectories, respectively.
      \textbf{(d)} The diameter of each point is proportional to its cumulative time, with larger points indicating longer time. The table shows corresponding dot-parenthesis notation, cumulative time, and free energy of the selected states. Mis-stacked base pairs and fully paired toeholds are highlighted in green and blue, respectively.
      }       
      \label{fig:perfect_toehold8}
  \end{center}
\end{figure}

\begin{figure}[!hp]
  \begin{center}
      \includegraphics[width=0.83\linewidth, trim={0cm 1cm 0cm 0cm}, clip]{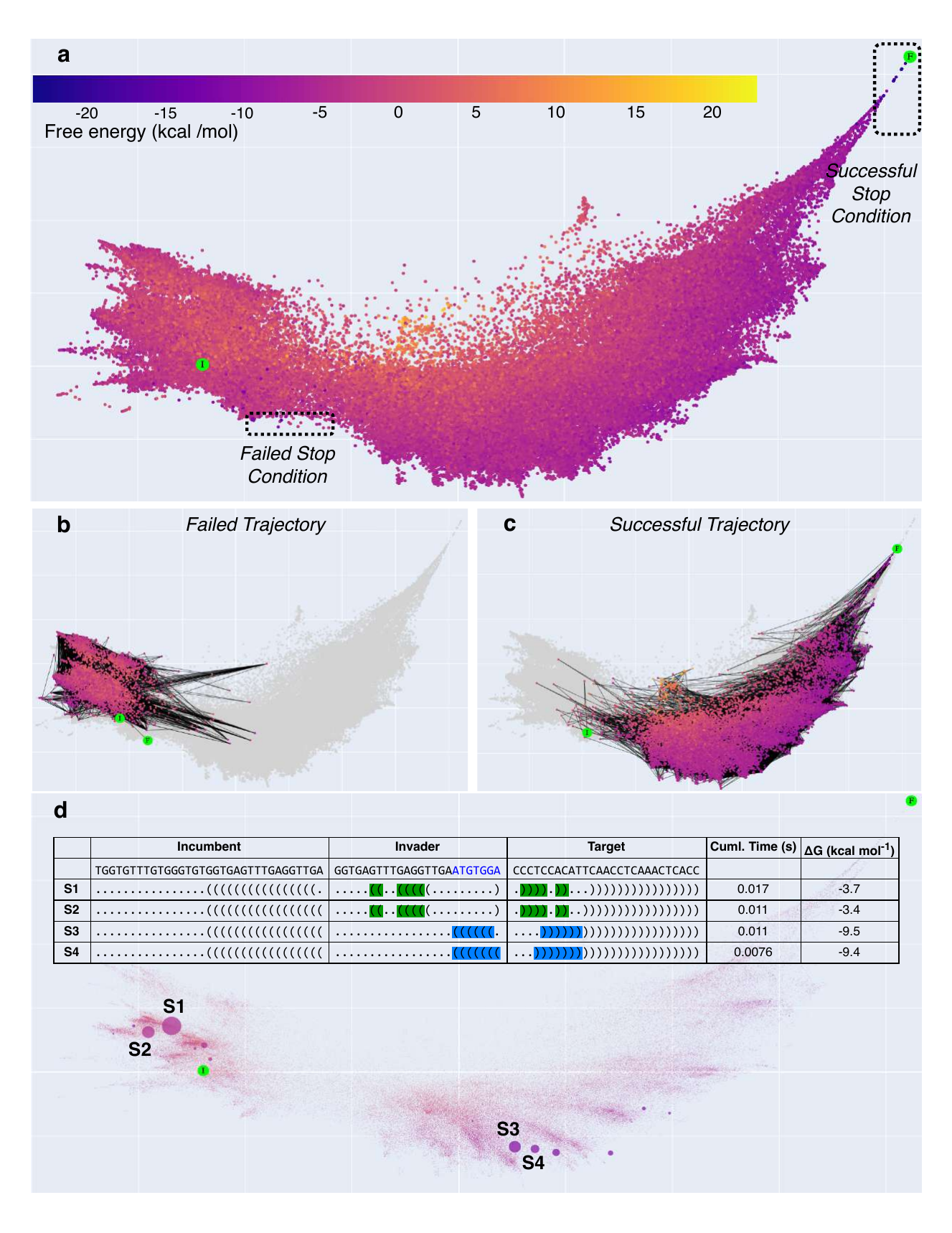}
      \caption{\small 
      \textbf{(a)} ViDa-3Strand-generated secondary structure state space with states laid out on the energy landscape for the perfect-toehold7-reporter reaction. Each point refers to a state with the color representing the value of free energy. The green blobs annotated by ``I'' and ``F'' indicate specific initial and successful final states, respectively.
      \textbf{(b)} and \textbf{(c)} show representative failed and successful trajectories, respectively.
      \textbf{(d)} The diameter of each point is proportional to its cumulative time, with larger points indicating longer time. The table shows corresponding dot-parenthesis notation, cumulative time, and free energy of the selected states. Mis-stacked base pairs and fully paired toeholds are highlighted in green and blue, respectively.
      }
      \label{fig:perfect_toehold7_reporter}
  \end{center}
\end{figure}

We first analyzed the mismatch-free reaction, perfect-toehold8. The state space embedding generated by ViDa-3Strand is shown in Figure \ref{fig:perfect_toehold8}a. The free energy distribution across the embedding follows a high-to-low trend from the initial state to the successful final state. Interestingly, we observed that the successful final states cluster together, enclosed by the black box (labeled ``Successful Stop Condition'') in the right corner of the plot. This suggests that the secondary structures with identical chemical configurations (in this case, the incumbent strand detached from the target and the invader strand paired with the target) are gathered close by, demonstrating that Via-3Strand also preserves local structure. A similar observation is evident for the ``Failed Stop Condition'', where the failed final states also cluster together within the corresponding black box in the plot. Moreover, we investigated the neighbouring states in this embedding plot and found that they differ by only a few base pairs, providing additional evidence of local structure preservation.
To test generalizability, we further applied ViDa-3Strand to the perfect-toehold7-reporter reaction, as shown in Figure \ref{fig:perfect_toehold7_reporter}a. We found that the free energy also follows a high-to-low trend, with the successful final states clustering together and the failed final states forming a separate cluster, indicating preservation local structure.
More visualization results for both perfect-toehold8 and perfect-toehold7-reporter reactions are shown in Supplementary Sections \ref{sup:per_t8} and \ref{sup:per_t7_reporter}.

\subsubsection{ViDa-3Strand Reveals Nuanced Understanding of Strand Displacement}

We laid out a failed and a successful trajectory, each with a relatively long reaction time, on the state space embedding in Figure \ref{fig:perfect_toehold8}b and c, respectively. Interestingly, they exhibit distinct reaction paths within the embedding space: the successful trajectory proceeds towards the right side, moving from the initial state to the successful final one, whereas the failed trajectory proceeds towards the left side and loops back before reaching to the failed final state.
We also observed dense regions along both successful and failed trajectories, which we hypothesized are kinetic traps that occur during the reaction. To further investigate, we computed the cumulative reaction time for each state and visualized each state as a point in the embedding, with the diameter of each point proportional to its corresponding cumulative time, where larger points indicate longer reaction time, as shown in Figure \ref{fig:perfect_toehold8}d. Notably, we noticed several relatively larger points in both successful and failed regions. The top three states with the longest cumulative times are listed in the table of Figure \ref{fig:perfect_toehold8}d. We found that the states (annotated as S1 and S2) contain mis-stacked base pairs, including
 \( \begin{array}{c}
\texttt{5$'$-GAGG-3$'$} \\
\texttt{3$'$-CTCC-5$'$}
\end{array} \ \text{and} \ \begin{array}{c}
\texttt{5$'$-GT-3$'$} \\
\texttt{3$'$-CA-5$'$}
\end{array} \) 
between the invader and target strands, highlighted in green. These stable mis-stacked base pairs inhibit proper pairing between the toeholds of the invader and target strands, thereby trapping the reaction process. To escape these mis-stacked configurations, the invader usually need to detach from the target, leading to reaction failure. Note that in some cases (see Supplementary Figures \ref{fig:per_t8_succ1} and \ref{fig:per_t8_succ2}), the mis-stacked invader proceeds toward the final state without detaching from the target, but it still requires a prolonged time to escape from the trap. We therefore attribute these two states as kinetic traps that were not reported by previous studies.
For state S3 along the successful region, all eight toeholds of the invader are fully paired with the target. This state requires a relatively long reaction time to initiate the branch migration process. We also attribute this state as a kinetic trap.
We further applied ViDa-3Strand to the perfect-toehold7-reporter reaction. As expected, the successful and failed trajectories reveal distinct reaction paths. We also observed the same mis-stacked base pairs between the invader and target strands (see structures S1 and S2 in the failed trajectory region in Figure \ref{fig:perfect_toehold7_reporter}c and the table in Figure \ref{fig:perfect_toehold7_reporter}d). These mis-stacked configurations block the toehold pairing process, thereby increasing reaction time.

\subsection{Mismatched Reactions}

\begin{figure}[!hp]
  \begin{center}
      \includegraphics[width=0.95\linewidth, trim={0cm 4cm 0cm 0cm}, clip]{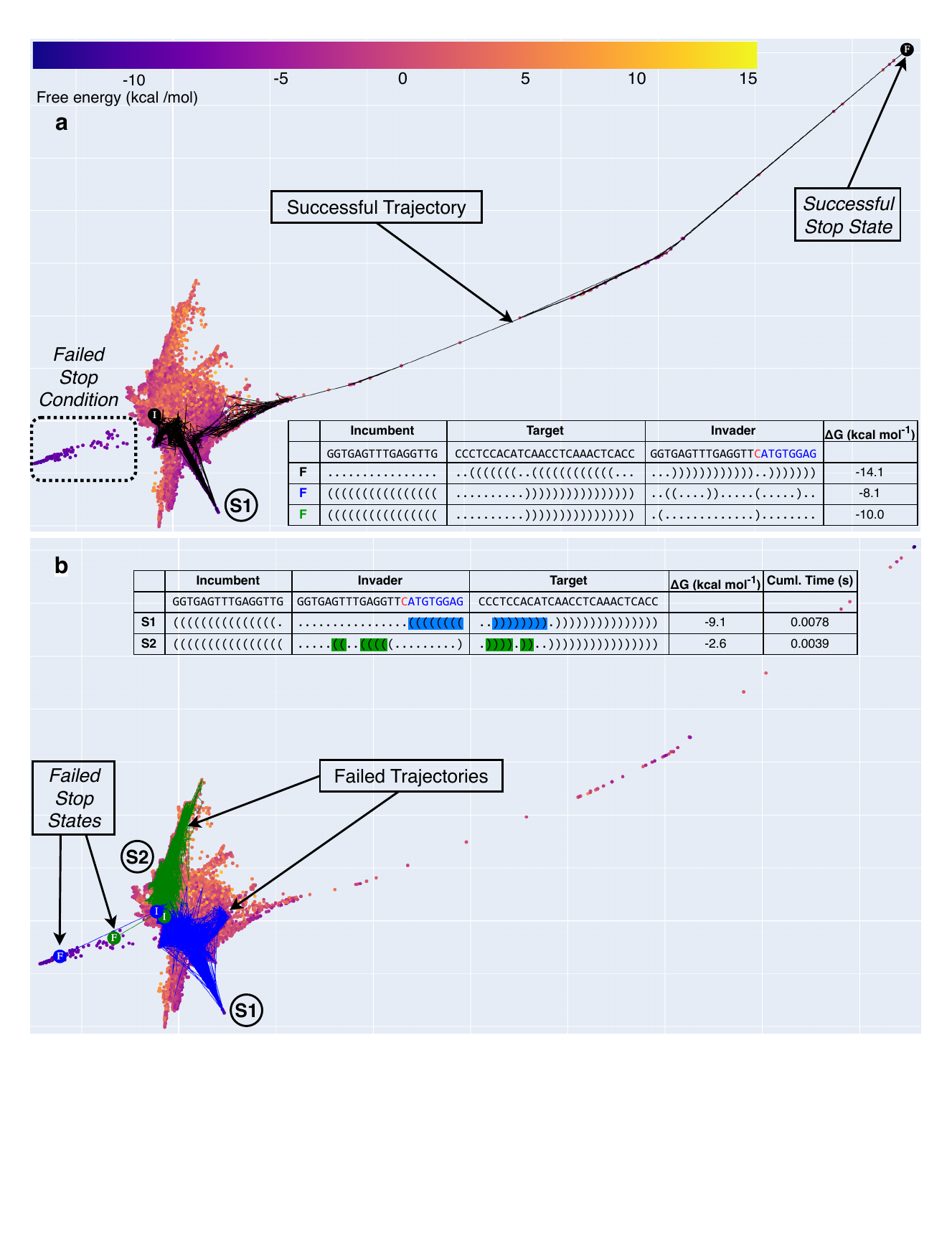}
      \caption{\small
      ViDa-3Strand-generated secondary structure state space with states laid out on the energy landscape for the proximal-toehold8 reaction. Each point refers to a state with the color representing the value of free energy.
      The tables in \textbf{(a)} and \textbf{(b)} show corresponding dot-parenthesis notation, cumulative time, and free energy of the selected states. Mis-stacked base pairs and fully paired toeholds are highlighted in green and blue, respectively.
      \textbf{(a)} A representative successful trajectory is laid out on the landscape, colored in black.
      \textbf{(b)} Two representative failed trajectories with different reaction mechanisms are laid out on the landscape, colored in blue and green. 
      }
      \label{fig:proximal_toehold8}
  \end{center}
\end{figure}

\subsubsection{ViDa-3Strand Captures Both Thermodynamics and Kinetics Properties}
Most simulations failed in the mismatched reactions, including proximal-toehold8 and proximal-toehold7-reporter. This is because mismatched bases can destabilize the system, thereby forcing detachment between the invader and target strands \cite{MachinekThreeway}. As a result, the simulation terminates when using Multistrand's ``first step'' mode.

\begin{figure}[!hp]
  \begin{center}
      \includegraphics[width=\linewidth, trim={0cm 15.5cm 0cm 0cm}, clip]{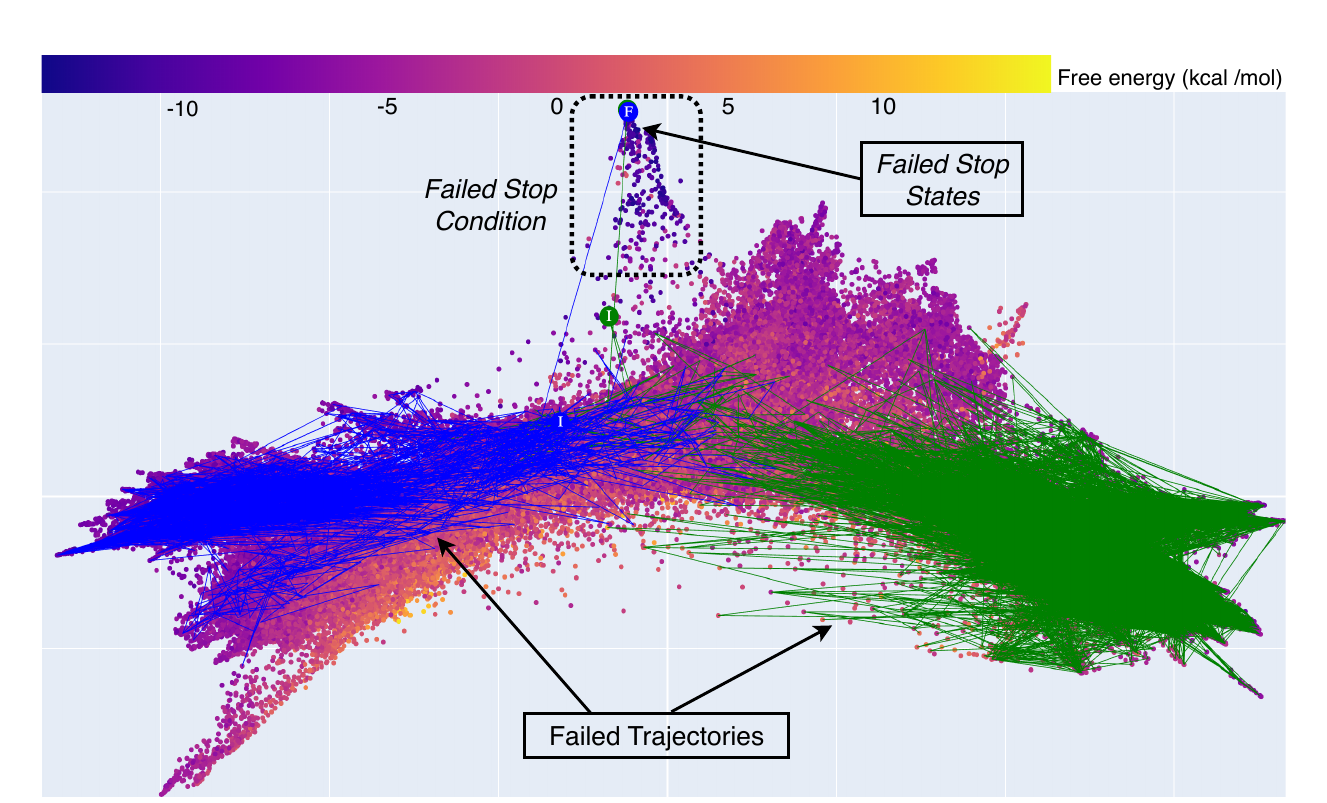}
      \caption{
      ViDa-3Strand-generated secondary structure state space with states laid out on the energy landscape for the proximal-toehold7-reporter reaction. Each point refers to a state with the color representing the value of free energy.
      Two representative failed trajectories with different reaction mechanisms are laid out on the landscape, colored in blue and green. 
      }
      \label{fig:proximal_toehold7_reporter}
  \end{center}
\end{figure}

Figure \ref{fig:proximal_toehold8} shows the secondary structure state space overlaid with both successful and failed trajectories. In particular, Figure \ref{fig:proximal_toehold8}a clearly illustrates the solely successful reaction pathway captured and distinguished by ViDa-3Strand, demonstrating its ability to capture the underlying reaction pathway. The trajectory starts from the initial state (annotated with a black bulb) and proceeds to the intermediate state S1, where the toeholds are full paired. However, because the mismatched base ``C'' (highlighted in red in the table) lines adjacent to the toehold, the trajectory either requires a very long time to bypass this mismatch and initiate branch migration, or results in unbinding of the toeholds and subsequent reaction failure, as shown by the blue trajectory in Figure \ref{fig:proximal_toehold8}b. 
Interestingly, we also observed another distinct failed reaction pathway, shown by the green trajectory. In this scenario, mis-stacked bases pairs form between the invader and target strands, therefore inhibiting proper toehold binding. Notably, this mis-stacked configuration follows the same pattern as the mis-stacked base pairs observed in the perfect-toehold8 reaction, providing further evidence that such configurations
 \( \begin{array}{c}
\texttt{5$'$-GAGG-3$'$} \\
\texttt{3$'$-CTCC-5$'$}
\end{array} \ \text{and} \ \begin{array}{c}
\texttt{5$'$-GT-3$'$} \\
\texttt{3$'$-CA-5$'$}
\end{array} \) 
introduce kinetic traps.
We noticed that a few rare failed trajectories in which the invader initially binds to the toeholds of the target and attempts to bypass the mismatch but fails. It then detaches from the toeholds, forms mis-stacked configurations, and ultimately detaches completely from the target, terminating the reaction (see Supplementary Figure \ref{fig:prox_t8_fail1}).

Similarly, we visualized the secondary state space and reaction trajectories for the proximal-toehold7-reporter reaction, in which no successful reaction was simulated. As expected, two distinct failed reaction pathways emerged, as shown by the blue and green trajectories in Figure \ref{fig:proximal_toehold7_reporter}. The green trajectory displays the same pattern of mis-stacked base pairs, whereas the blue trajectory shows toeholds that initially fully pair and subsequently detach. We also observed that the failed stop states cluster together in both visualizations, further suggesting the preservation of local structure. More visualization results for both proximal-toehold8 and proximal-toehold7-reporter reactions are shown in Supplementary Sections \ref{sup:prox_t8} and \ref{sup:prox_t7_reporter}.

\section{Discussion and Future Work}

In this chapter, we extend ViDa to ViDa-3Stand that is tailored to accommodate three-stranded reactions. We apply ViDa-3Strand on four well-studied toehold-mediated three-way strand displacement reactions. Our visualization results demonstrate ViDa-3Strand's ability to preserve local structure and reveal nuanced insights into the mechanisms of strand displacement reactions. 

However, several limitations remain.
For instance, each of Machinek's reaction can involve hundreds of millions of intermediate secondary structures for a single trajectory. Considering that we need at least a few hundred such trajectories for training our neural network, this requires larger GPU memory size and increases training cost. Moreover, displaying all these secondary structures in 2D space may not be the most effective way.
To mitigate this issue, we have employed Multistrand's ``first step'' mode to generate trajectories and sampled secondary structures at fixed time intervals to shorten each trajectory. However, this sampling strategy may overlook important intermediate structures. To address this, it would be valuable to first obtain a coarse-grained representation of the state space before applying ViDa.

It would also be valuable to upgrade ViDa-3Strand to accommodate more complex reactions, such as four-way strand exchange \cite{fourway}. These reactions contain longer sequences, resulting in larger graph representations of secondary structures compared to current reactions, and consequently, increased memory consumption and computational cost. To address this, it would be advantageous to adopt sparse metrics or adjacency lists instead of dense matrices for representing graphs, which can significantly reduce memory usage. Additionally, strategies such as merging similar nodes or pruning unimportant edges could be applied to further minimize graph size and improve computational efficiency.

In addition, for more complex reactions, the lengths of the trajectories can be substantial, resulting in an extremely expansive secondary structure state space. We noticed that most of these secondary structures are repeated. For instance, four-strand reactions can generate hundreds of millions intermediate secondary structures for a single trajectory, while only hundred of thousands structures are unique. Considering that we need at least a few hundred such trajectories for training our neural network, this creates a significant burden on the memory storage. Therefore, to improve the efficiency of storing trajectory data, we plan to use \textit{Multigrain} \cite{multigrain}, a package designed for post-processing simulations of interacting nucleic acid strands from Multistrand. This package provides an efficient means of storing trajectory data through indexed representation and data compression.

\part{Deep Generative Models for Single-Particle Cryo-EM}

\chapter{Introduction} \label{chap:cryoemintro}

Single-particle cryogenic electron microscopy (cryo-EM) is a powerful technique for structural determination of biological macromolecules such as proteins and nucleic acids, which has been widely used in structure-based drug discovery \cite{dd1,dd2}. Unlike X-ray crystallography, cryo-EM can accommodate large structures, complexes and membrane proteins that are challenging to crystallize. In addition, cryo-EM is capable of capturing heterogeneous conformations of biomolecules, leading to more accurate structural characterizations \cite{diiorio2023novel}.
Its impact on structural biology was highlighted by the 2017 Nobel Prize in Chemistry, awarded to Dubochet, Henderson, and Frank for their pioneering contributions \cite{2017nobel}.

Over the past decade, cryo-EM has achieved remarkable progress, routinely yielding density maps at near-atomic resolutions and facilitating discoveries across structural biology and drug development \cite{henderson2015overview,de2021cryo,cryoemworkflow,dd1,dd2}. Despite this success, several challenges remain. Cryo-EM density maps (or cryo-EM maps for simplicity) often contain noise, artifacts, and resolution heterogeneity, complicating accurate interpretation. Moreover, automated construction of protein atomic models from cryo-EM maps remains challenging, especially at intermediate and low resolutions. These limitations underscore the need for improved computational methods that can enhance density maps and construct more accurate atomic models.

Deep generative models have emerged as promising techniques to address these challenges. Architectures such as generative adversarial networks (GANs) \cite{gan,emgan}, deep graph neural networks (GNNs) \cite{GNN,modelangelonature}, deep U-Net-based models \cite{unet,deeptracer}, transformers \cite{transformer,cryoten}, and diffusion models \cite{diffusion,DiffModeler} have shown great potential for learning powerful representations of 3D cryo-EM density maps. In parallel, protein large language models (pLLMs) such as ESM series \cite{esm1,esm2,esm3,esmif}, capture sequential and residue-level structural priors that can serve as complementary features for enhancing cryo-EM maps and guiding protein model construction \cite{modelangelonature}. Together, these approaches open new opportunities to integrate data-driven generative models with biologically-informed priors to advance cryo-EM analysis.

Accordingly, Part II of this dissertation focuses on developing novel deep learning tools to enhance cryo-EM density maps and to facilitate more accurate protein atomic model building. Specifically, we first present a systematic survey and comprehensive benchmarking of state-of-the-art deep learning methods for protein structure modeling, using our improved evaluation metrics. In addition, we introduce methods that employ GANs to synthesize high-fidelity, experimental-like cryo-EM density maps from protein structures, as well as multimodal U-Nets that integrate structural information from pLLMs via cross-attention mechanisms \cite{transformer} to enhance intermediate-resolution cryo-EM maps.

In this introductory chapter, we first provide background on cryo-EM density maps, the computational pipeline for cryo-EM analysis, and the challenges associated with automated cryo-EM map enhancement and protein model construction. We then introduce the foundations of deep generative models, including GANs and pLLMs, which serve as the backbone and auxiliary networks in our proposed approaches.

Building on this foundation, in Chapter \ref{chap:cryoem_review}, we present a comprehensive survey of deep learning (DL)-based model-building methods to complement existing reviews, highlighting recent DL approaches have demonstrated notable advancement. We categorize the various DL architectures employed and explicitly discuss their pros and cons. Moreover, we propose improved evaluation metrics and benchmark several state-of-the-art approaches for protein model building from cryo-EM density maps. Furthermore, we offer valuable insights and guidance for computer scientists, computational biologists and cryo-EM practitioners in the field.
In Chapter \ref{chap:gan}, we present \textbf{Struc2mapGAN}, a novel method that utilizes generative adversarial networks to synthesize high-resolution, experimental-like cryo-EM density maps from protein structures.
In Chapter \ref{chap:enhancemap}, we present \textbf{CryoSAMU}, a novel method designed to enhance 3D intermediate-resolution cryo-EM density maps of protein structures using structure-aware multimodal U-Nets that integrate density map features and structure features derived from protein large language models. 
In addition, we also discuss useful future directions in Chapters \ref{chap:cryoem_review}, \ref{chap:gan}, and \ref{chap:enhancemap}.

\section{Cryogenic Electron Microscopy}

Cryogenic electron microscopy (cryo-EM) has emerged as a revolutionary technique in structural biology over the past decade. Earlier cryo-EM density maps were often limited to low resolution, restricting their widespread use. This changed dramatically during the ``resolution revolution'' of the 2010s, driven by advances in electron microscopy hardware, improvements in image processing algorithms, and increased computational resources \cite{de2021cryo}. These developments enabled cryo-EM to routinely achieve near-atomic resolution structures.

In contrast to X-ray crystallography that requires crystalization of the target protein, single-particle cryo-EM allows direct visualization of purified biomolecules in solution. This makes cryo-EM particularly valuable for studying proteins and complexes that are challenging or infeasible to crystallize, such as large macromolecular assemblies, membrane proteins, and structurally flexible complexes \cite{cheng2015single,nogales2015cryo}. The exponential growth in deposited cryo-EM density maps and corresponding atomic models has led to cryo-EM becoming a powerful complement to X-ray crystallography and nuclear magnetic resonance (NMR), and in many cases, serving as the preferred technique for structure determination. 
An exponentially growing number of cryo-EM density maps and their associated protein structures have been deposited in the Electron Microscopy Data Bank (EMDB) \cite{EMDB} and the Protein Data Bank (RCSB PDB) \cite{PDB}. As of August 2025,  a total of 48984 cryo-EM maps and 28822 corresponding structures has been released.

Beyond static structural determination, cryo-EM can also capture dynamic molecular processes. By resolving multiple conformations from a single specimen, it reveals motions and conformational changes of molecules, providing crucial insights into structural heterogeneity \cite{heterogeneity,CryoDRGN}. Overall, single-particle cryo-EM has rapidly developed into an indispensable technique in structural biology, enabling high-resolution characterization of biomolecular structures and dynamics that were previously inaccessible.

\section{Cryo-EM Data Processing Pipeline}

\begin{figure}[!ht]
  \begin{center}
      \includegraphics[width=\linewidth, trim={0cm 0cm 0cm 0cm}, clip]{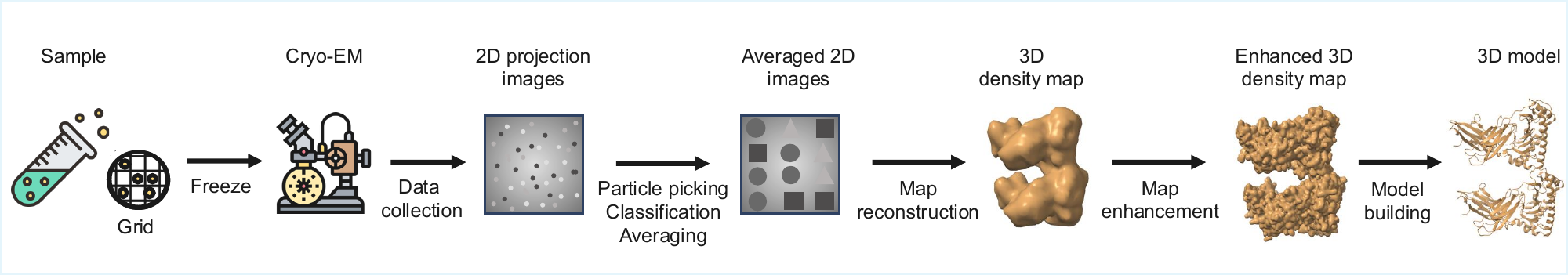}
      \caption{The cryo-EM data processing workflow. Some icons were obtained from \texttt{Flaticon} \url{https://www.flaticon.com}.}
      \label{fig:cryoEM-workflow}
  \end{center}
\end{figure}

Single-particle cryo-EM involves a multi-step computational workflow that transforms raw electron microscope images into a 3D structure. As illustrated in Figure \ref{fig:cryoEM-workflow}, a typical pipeline spans from sample preparation and imaging to map reconstruction, enhancement (sharpening), and model building. 
The key computational stages include image alignment and CTF correction, particle picking, 2D classification, 3D reconstruction, density map enhancement, and model building.
We briefly review these steps next, along with additional background on methods for generating synthetic cryo-EM density maps from protein structures. In particular, we specialize on map enhancement, atomic model building, and simulation methods, as they are covered in Chapters \ref{chap:cryoem_review}, \ref{chap:gan}, and \ref{chap:enhancemap} of this dissertation.

\subsection{Alignment and CTF Correction}
Raw cryo-EM movies are aligned to correct beam-induced motion, and the microscope's contrast transfer function (CTF) is estimated and corrected to restore signal contrast \cite{CTF}. These steps enhance the signal-to-noise ratio (SNR) in particle images, which is critical given the low contrast of vitrified biological samples.

\subsection{Particle Picking}
After electron microscope projection and imaging, hundreds of thousands micrographs containing weakly visible protein projections are obtained, which need to be further processed.
Early manual picking (done by eye) is slow and biased, while traditional automated methods, such as template matching \cite{findem,templatematching}, often produce many false positives and missed rare particle views. Recent machine learning–based tools such as Topaz \cite{Topaz} and crYOLO \cite{crYOLO} learn from small training sets and achieve much higher accuracy, reducing the need for tedious post-picking curation.

\subsection{2D Classification}
Picked particles are compared against each other, and similar particles are clustered into groups to produce class averages with enhanced details. This step filters out blurred or nonsensical particles, provides clean views of the protein, and enables rapid assessment of data quality and conformational heterogeneity. Commonly used tools for 2D classification include RELION \cite{relionpostprocess} and cryoSPARC \cite{cryoSPARC}.

\subsection{3D Reconstruction}
Reconstructing 3D density maps from 2D projections requires determining their unknown orientations, typically using projection-matching algorithms against low-resolution reference models \cite{projection-matching}. These orientations and reconstructions are refined iteratively until convergence, yielding progressively accurate maps. To avoid model bias and estimate resolution, modern workflows leverage a ``gold standard'' approach, splitting data into two halves and comparing their independent reconstructions through Fourier Shell Correlations (FSCs) \cite{goldstandard}. With recent advances in deep learning, new approaches have been introduced to facilitate 3D reconstruction and address challenges of conformational and compositional heterogeneity in protein complexes \cite{CryoDRGN,CryoPoseNet}.

\subsection{Map Enhancement} \label{sec:mapsharpening}
While cryo-EM 3D maps form the foundation for molecular structure determination, raw maps are often unsuitable for direct use because they usually lack sufficient contrast. This loss of detail can result from molecular motion and heterogeneity, imaging artifacts, and incoherent averaging of image data \cite{eminherent}. To improve interpretability, raw 3D reconstructed maps are typically sharpened by boosting high-frequency signals that are dampened during reconstruction.

Traditional map enhancement or sharpening approaches, including Phenix Autosharpen \cite{phenixautosharpen} and RELION postprocessing \cite{relionpostprocess}, are based on global B-factor correction. This technique enhances the amplitude of high-frequency Fourier components in raw cryo-EM maps. However, global B-factor-based methods encounter difficulties with maps exhibiting heterogeneous local resolutions, often leading to over- or under-sharpening in specific regions. Despite local B-factor-based sharpening algorithms, such as LocalDeblur \cite{LocScale}, LocScale \cite{LocScale}, and LocSpiral \cite{LocSpiral}, have been developed to alleviate this limitation, these methods still suffer from poor accuracy in estimating the local resolution of maps, which is crucial for precise local B-factor sharpening.

DeepEMhancer \cite{deepemhancer} is a pioneering DL-based fully automatic method that leverages a 3D U-Net model to mimic local sharpening effects and enhance map features. Subsequently, CryoFEM \cite{cryofem} that employs convolutional neural networks (CNNs) and EM-GAN \cite{emgan} that utilizes generative adversarial networks (GANs) have been introduced to further enhance cryo-EM maps. Most recently, with the emergence of transformers, CryoTEN \cite{cryoten} that adopts 3D UNETR++ transformer \cite{unetr++} and EMReady \cite{emready} that adopts a Swin transformer architecture \cite{swintransformer}, both have shown superior performance in enhancing map quality for accurate protein structure modeling.

\subsection{Model Building} \label{sec:modelbuilding}
Cryo-EM density maps typically form large 3D grids consisting of voxels with varying intensity levels, usually ranging from approximately $100^3$ to $500^3$ voxels, which are further processed to build 3D atomic models (structures).
Yet, automated construction of atomic models from cryo-EM density maps is a challenging process due to the presence of artifacts and noise in the map, varying and inconsistent resolution across the entire map caused by molecular flexibility or radiation damage, and the lack of homologous or predicted structures \cite{phenix,modelangelonature,embuild}. 

Conventional methods for protein model building involving optimization-driven approaches progressively refine an atomic model by minimizing an energy function that reflects physical forces, including electrostatic interactions, steric clashes, and bond lengths \cite{phenix,rosetta}. However, these algorithms are computationally intensive and usually require substantial manual input to construct precise models.

In this context, early data-driven methods that utilize classical machine learning (ML) techniques such as K-Nearest Neighbours (KNN) \cite{RENNSH}, K-means clustering \cite{Pathwalking}, and Support Vector Machine (SVM) \cite{SSELearner}, have been proposed, but are limited to secondary structures or simplified backbone models, and not suited for large complex structures.  

To address these limitations, a variety of DL-based approaches have been introduced in recent years. Early efforts primarily focused on identifying secondary structure elements (SSEs) such as $\alpha$-helix and $\beta$-sheets from density maps with resolution ranging from intermediate to low \cite{Emap2sec,Emap2sec+,EMNUSS,CNN-classifier}. With the increasing availability of high-resolution density maps, recent DL techniques predict atom types and coordinates, from backbone structures that consist of a repeated sequence of nitrogen (N), alpha-carbon (C$_\alpha$), and beta-carbon (C$_\beta$) atoms to full-atom  models that include both backbone (main-chain) and side-chain atoms \cite{deeptracer,embuild,modelangelo,Deepmainmast,DiffModeler,DEMO-EM}. 
Concurrently, the recent revolution in protein structure prediction driven by AlphaFold and related methods leveraging AI \cite{alphafold2,alphafold1,rosettafold,rosettafoldallatom,esm2,omegafold,rgn2,itassermtd}, has led to the integration of these sequence-to-structure approaches for atomic model building from density maps. We can categorize these methods into the following two types. \footnote{A comprehensive survey of these approaches is provided in Chapter \ref{chap:cryoem_review}.}

\begin{figure}[!hp]
\centering
\includegraphics[width=\linewidth, trim={1cm 1.5cm 1cm 1cm}, clip]{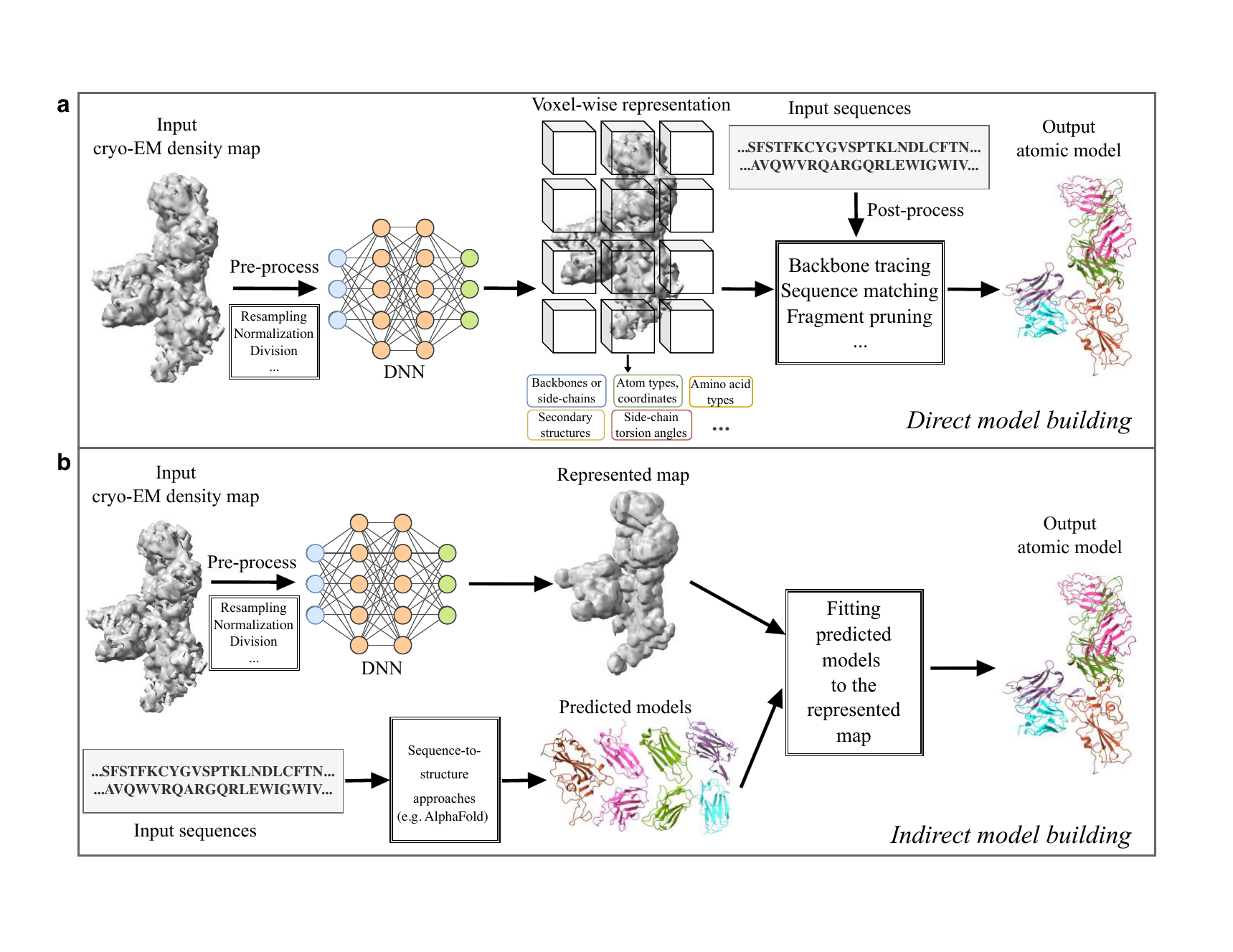}
\caption{DL-based approaches (\textbf{a}) direct model building and (\textbf{b}) indirect model building. Note that the illustration of indirect model-building approaches represents an overview of some of the approaches. The illustrated protein complex structure is the SARS-CoV-2 spike in complex with antibodies B1-182.1 and A19-61.1 (PDB-ID: 7TBF; EMDB-ID: 25797; Resolution: 3.1 {\AA}) \cite{7TBF}. 
}
\label{fig:modelbuildingworkflow}
\end{figure}

\emphasizespace{Direct model building}
Some methods (\cite{Emap2sec,Emap2sec+,CNN-classifier,AAnchor,Cascaded-CNN,A2-Net,DeepMM,Structure-Generator,Haruspex,EMNUSS,deeptracer,deeptracer2.0,modelangelonature,Cryo2Struct}) leverage deep neural networks (DNNs) to predict voxel-wise representations from the map directly, where each voxel is annotated with its associated backbone, secondary structure element (SSE), amino acid type, C$_\alpha$ atom, and/or side-chain torsion angle. The predicted C$_\alpha$ atoms are then connected into chains by employing Traveling Salesman Problem (TSP) solvers \cite{tsp1,tsp2}, Vehicle Routing Problem (VRP) solvers \cite{vrp1,vrp2}, or Monte Carlo Tree Search (MCTS) \cite{MCTS}.
We illustrate this workflow in Figure \ref{fig:modelbuildingworkflow}a.

\emphasizespace{Indirect model building}
Some methods \cite{embuild,fff,DiffModeler} use DNNs to learn a representation that strengthens the backbone features of the original map, yielding DL-represented backbone maps. 
Independently, and in contrast with the direct-model-building methods, models of chains, domains, or complexes are obtained using sequence-to-structure approaches such as AlphaFold \cite{alphafold2}.
These predicted models are then aligned to the DL-represented maps, using rigid-body fitting \cite{rigid1,rigid2}, partial fitting \cite{empot}, semi-flexible fitting \cite{embuild} and/or flexible fitting \cite{flexible1,flexible2,flexible3} algorithms, and are assembled to obtain a final complete model. We illustrate this workflow in Figure \ref{fig:modelbuildingworkflow}b.

\section{Simulations of Cryo-EM Density Maps} \label{sec:simulation_cryoem}

After outlining the computational workflow of cryo-EM, it is important to emphasize that simulated density maps also play a crucial role in structural biology. Such simulations provide a controlled way to generate cryo-EM-like density maps directly from atomic models (e.g., PDB structures). These synthetic maps are widely used for various applications, such as guiding the particle-picking process before reconstruction of experimental maps, assisting rigid-body fitting with use of simulated maps, and severing as ground-truth targets for neural network training.

Simulation-based methods for generating density maps from their associated PDB structures are based on the convolution of atom points with resolution-lowering point spread functions such as Gaussian, triangular, or hard-sphere \cite{spread1, spread2}. 
Given a PDB structure containing $M$ atoms, a general Gaussian simulation formula for producing a density value $\rho$ at a grid point $\mathbf{x}$ is expressed as:
\begin{equation} \label{eq:simulation}
    \rho(\mathbf{x}) = \sum_{i=1}^{M} \theta Z_i e^{-k|\mathbf{x}-\mathbf{r}_i|^2},
\end{equation}
where $Z_i$ represents the atomic number and $\mathbf{r}_i$ is the position vector of the i-th heavy atom, $\theta$ is a scaling factor and $k$ is defined in terms of the resolution \cite{emready,embuild}. 
Such methods including \emph{e2pdb2mrc} in EMAN2 \cite{eman2} (originally called \emph{pdb2mrc} in EMAN1 \cite{eman1}), \emph{molmap} in UCSF ChimeraX \cite{chimerax}, \emph{StructureBlurrer} in TEMPy2 \cite{tempy2}, and \emph{pdb2vol} in Situs \cite{situs}, generate 3D density maps using a Gaussian point spread function, where the real-space dimension corresponds to the desired resolution, which varies based on the specific resolution convention of each package.

\section{Current Challenges in Automation}

Upon running the aforementioned pipeline, practitioners overall face significant challenges and bottlenecks that limit the ability to produce a structure from a 3D cryo-EM density map. The most significant obstacles relate to map resolution quality, as maps at worse than $\sim$ 3.5-4 {\AA} resolution typically prevent accurate side-chain assignment, while conformational heterogeneity can further complicate backbone tracing and local fitting \cite{resrange,heterogeneity}. With advances in high-throughput cryo-EM that allow for generating large collections of data, the need for automated model building methods has become more pressing, with an increasing fraction of deposited maps remain unsolved or partially interpreted \cite{wwpdb2024emdb,EMDB}. 

In this regard, the variety of recent DL-based methods introduced provide a new direction for improving automated model building (see Section \ref{sec:modelbuilding}). However, their advantages and limitations have not been systematically compared. As a result, practitioners often lack clear guidance on which tools are most suitable for their specific applications. Additionally, current evaluation metrics for model building quality are limited and often inadequate. 
To address these challenges, in Chapter \ref{chap:cryoem_review}, we survey existing model building methods, refine current metrics, and benchmark several cutting-edge approaches. We also provide practical guidance for practitioners and outline future directions for applying machine learning to cryo-EM and structural biology.

Besides, reliable map-model pairs at lower resolutions remain scarce \cite{wriggers2015numerical}, resulting in insufficient training data for optimizing model building tools. In this context, simulating accurate synthetic cryo-EM maps is crucial, as it expands the pool of ground-truth datasets for training \cite{embuild,fff,modelangelo,Deepmainmast,emready,emgan,cryofem,cryoten}.
However, existing simulation methods often fail to capture complex structural features, leading to notable discrepancies between simulated and experimental maps (see Section \ref{sec:simulation_cryoem}). To address this, in Chapter \ref{chap:gan}, we introduce a generative adversarial network that learns realistic characteristics from experimental data to improve the simulation of cryo-EM density maps. 

In addition, raw cryo-EM density maps are often unsuitable for model building because they typically suffer from low contrast. However, existing DL-based methods \cite{deepemhancer,emready,cryoten,cryofem,emgan} are generally optimized for high-resolution maps and tend to neglect complementary sources such as structural context (see Section \ref{sec:mapsharpening}). To overcome these limitations, in Chapter \ref{chap:enhancemap}, we propose a multimodal neural network that integrates structural information with map features to enhance, for specifically intermediate-resolution cryo-EM map enhancement.

\section{Background on Deep Generative Models -- GAN and pLLM}

We leverage generative adversarial networks (GANs) and protein large language models (pLLMs) for our designed methods outlined in Chapter \ref{chap:gan} and Chapter \ref{chap:enhancemap}, respectively. This section provides detailed background information on both GANs and pLLMs.

\subsection{Generative Adversarial Network}
\begin{figure}[!ht]
\centering
\includegraphics[width=\linewidth, trim={0cm 21cm 1cm 0cm}, clip]{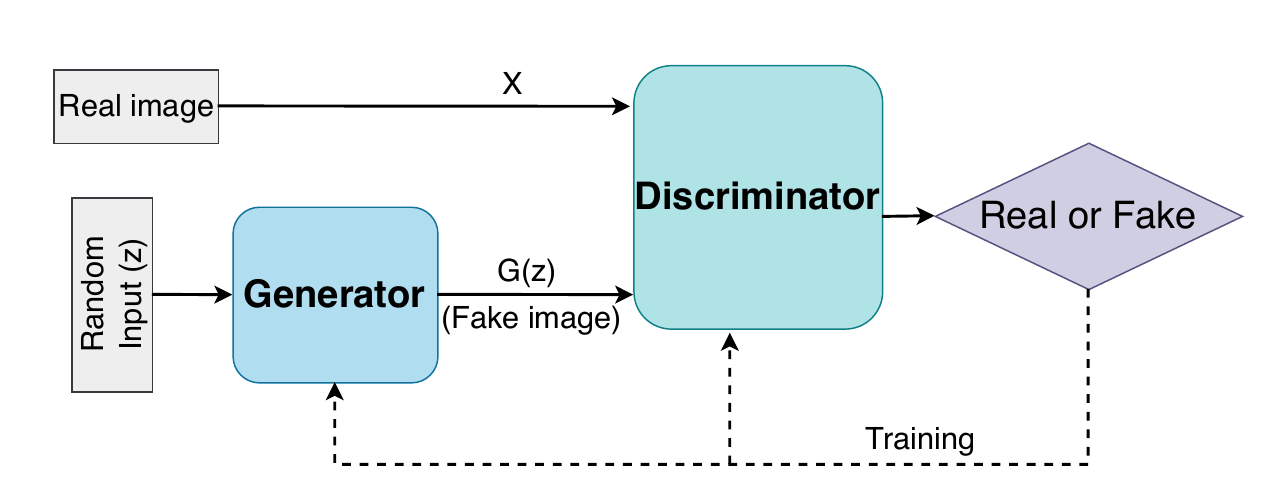}
\caption{The architecture of the GAN model.}
\label{fig:gan_general}
\end{figure}

Generative Adversarial Network (GAN) is one of the most prevalent generative models widely employed in image generation, super-resolution, and 3D object generation \cite{gan}. The GAN architecture comprises two main models, a generator \emph{$G$} and a discriminator \emph{$D$} trained together, as shown in Figure \ref{fig:gan_general}.
\emph{$G$} generates a batch of images (we call them fake images), and these images are fed along real images (ground-truth reference images) into \emph{$D$} to be classified as real or fake. During training, \emph{$G$} generates fake images to fool \emph{$D$}, while \emph{$D$} is updating its parameters to discriminate the fake ones. 
Mathematically speaking, the two models \emph{$G$} and \emph{$D$} compete in a two-player minimax game with the objective function $L(G,D)$: 
\begin{equation}
    \min_G \max_D V(D, G) = \mathbb{E}_{\mathbf{x}} [\log D(\mathbf{x})] + \mathbb{E}_{\mathbf{z}} [\log (1 - D(G(\mathbf{z})))].
\end{equation}
The generator minimizes the $\log (1 - D(G(\mathbf{z})))$ term predicted by \emph{$D$} for fake images. Conversely, the discriminator maximizes the log probability of real images, $\log D(\mathbf{x})$ and the log probability of correctly identifying fake images, $\log (1 - D(G(\mathbf{z})))$.
Over the past few years, the GAN architecture has been adapted for various purposes and has demonstrated superior performance in multiple domains \cite{wgan,cyclegan,cryogan,pix2pix,emgan}.

\subsection{Protein Large Language Model}
The advancement of protein large language models (pLLMs) has enabled unprecedented insights into protein structure, function, and evolution \cite{esm1,esm2,esm3,esmif,proteinbert,prostt5,proteinmpnn}. In analogy to human texts, protein sequences are treated as ``biological texts'' and input into pLLMs to capture contextual information inherent in the sequences. Notable examples of such models include the ESM family \cite{esm1,esm2,esm3}, which are pretrained on vast datasets of protein sequences using the masked language modeling strategy, allowing them to develop rich representations that encapsulate evolutionary information. 

Addressing the inverse problem of predicting protein sequences from given structures, ESM-IF1 \cite{esmif} has been developed. Trained on 12 million protein structures derived from AlphaFold2 \cite{alphafold2}, ESM-IF1 predicts protein sequences from backbone atom coordinates (N, C$_\alpha$, and C atoms). By recovering sequences from structures, ESM-IF1 captures both sequence preferences and structural context, including local geometry and secondary structure features. These traits make ESM-IF1 a compelling choice for generating structure-aware embeddings that complement the map-only modality. 


\section{Summary of Contributions}
The contributions of Part II of this dissertation are as follows:
\begin{enumerate}
    \item To provide an up-to-date overview of recent progress in automated model building, we conduct a comprehensive survey of DL-based methods, distinguishing between direct approaches that rely solely on density maps and indirect ones that integrate sequence-to-structure predictions from AlphaFold \cite{alphafold2}. Among these two classes, we also cover the underlying architectures and discuss the benefits and limitations of different methods. In addition, to address limitations of existing evaluation metrics for atomic model-building methods from cryo-EM density maps, we introduce a set of newly designed metrics by considering the alignment quality of predicted models. Using these improved metrics, we conduct a thorough benchmark of cutting-edge atomic model-building methods against traditional physics-based approaches using 50 cryo-EM density maps at varying resolutions. See Chapter \ref{chap:cryoem_review} for more details.
    \item To generate high-quality, experimental-like cryo-EM density maps from given PDB (atomic) structures, we develop struc2mapGAN, a novel generative adversial network (GAN) \cite{gan} that uses a U-Net++ architecture \cite{unet++} as the generator, with an additional L1 loss term and further processing of raw experimental maps to enhance learning efficiency. Struc2mapGAN shows superior overall performance against simulation-based methods across various evaluation metrics and demonstrates practical suitability for generating large-scale maps. See Chapter \ref{chap:gan} for more details.
    \item To enhance intermediate-resolution cryo-EM density maps, we develop CryoSAMU, the first structure-aware multimodal network that combines 3D map features with corresponding structural features through cross-attention mechanisms. Notably, Cr\-yoSAMU achieves comparable performance of state-of-the-art methods but with significantly faster processing speeds (approximately 4.2 to 16.7 times), making it well-suited for large-scale and practical applications. See Chapter \ref{chap:enhancemap} for more details.
\end{enumerate}

\chapter{Comprehensive Survey and Benchmark of Model-Building Methods from Cryo-EM Density Maps} \label{chap:cryoem_review}

This chapter is a modified version of a paper by Chenwei Zhang, et al., published in Briefings in Bioinformatics (\url{https://doi.org/10.1093/bib/bbaf322}) \cite{atomicmodelreview_zhang}.

\section{Motivations and Contributions}

\begin{figure}[!ht]
    \centering
    \includegraphics[width=\linewidth, trim={0cm 0cm 0cm 0cm}, clip]{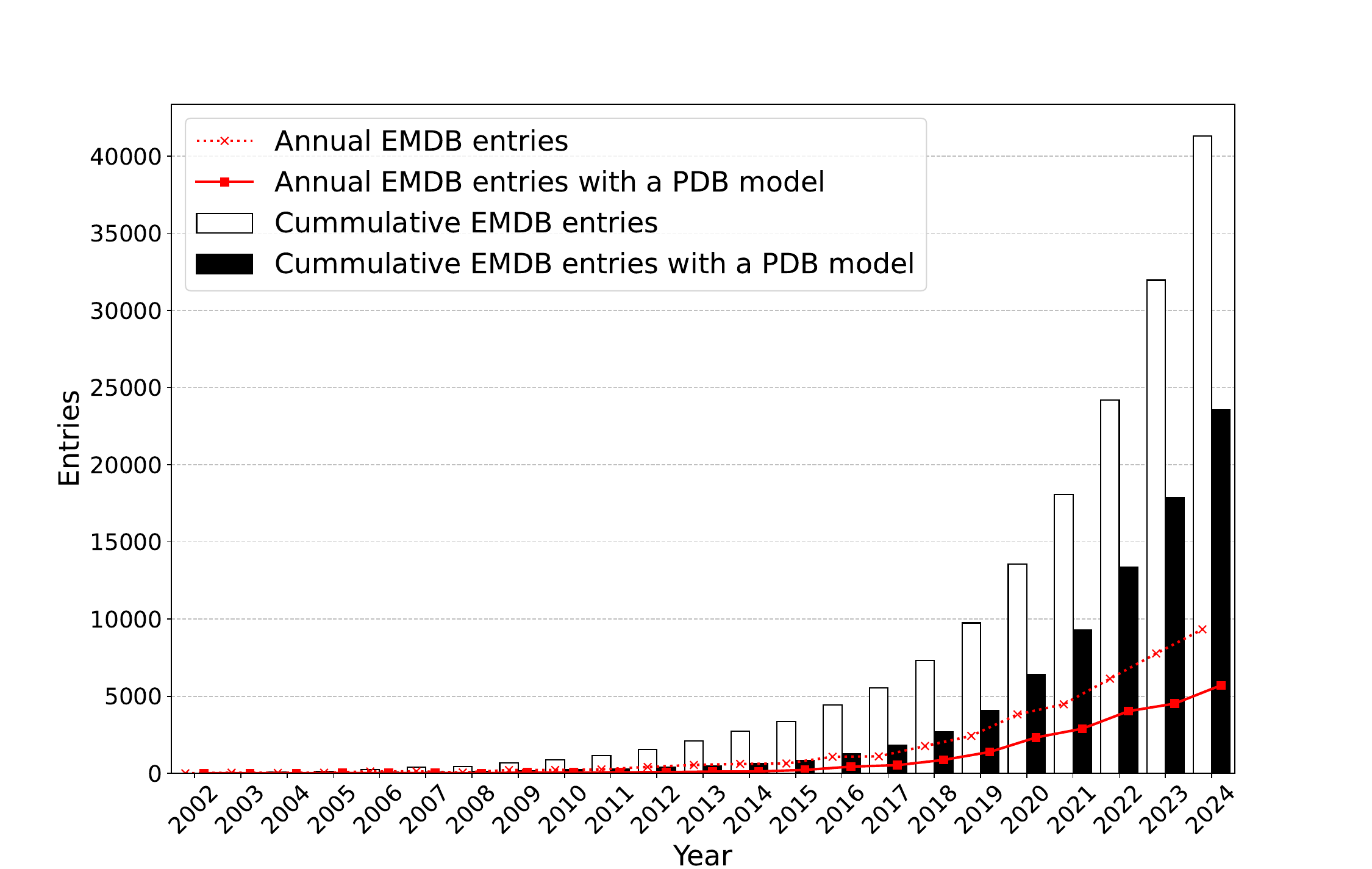}
    \caption{\small Bar chart: cumulative number of all EMDB entries and of those with an associated atomic structure in the PDB at the end of each year. Line plot: number of annually released EMDB and PDB entries. Data is shown until the end of 2024. The statistics were collected from EMDB on 2025-01-09 \cite{EMDB}.}
    \label{fig:emdb_stats}
\end{figure}

As of the end of 2024, more than 40000 electron microscopy density maps have been deposited to the Electron Microscopy Data Bank (EMDB) \cite{EMDB}, with the number of released maps showing an exponential growth. Yet, only approximately 57 \% of the associated atomic coordinates are resolved in the Research Collaboratory for Structural Bioinformatics Protein Data Bank (RCSB PDB) \cite{PDB}, as shown in Figure \ref{fig:emdb_stats}.  
This gap highlights the need for improved tools to accelerate the process of atomic model building from cryo-EM density maps. To address this challenge, a variety of deep learning (DL)-based approaches have been introduced in recent years, demonstrating significant advancements.

Previous reviews have introduced physics-based, ML-based, and some DL-based methods for atomic model building from cryo-EM density maps \cite{diiorio2023novel,giri2023deep,si2022artificial}. However,  they did not quantitatively compare and benchmark DL-based methods, leaving practitioners wondering how recent approaches perform when building atomic models from cryo-EM density maps, and how AlphaFold integration might enhance these methods. To answer these questions, we provide a comprehensive survey of recently proposed DL methods to better understand their common and distinctive features. Furthermore, we present a benchmark comparing several state-of-the-art methods, providing valuable insights and guidance for computer scientists, computational biologists and cryo-EM practitioners in the field.

Our main contributions are:
\begin{itemize}
    \item We run a comprehensive survey of recent state-of-the-art DL-based methods for automated model building, distinguishing  methods that only use density maps, and indirect ones that integrate sequence-to-structure predictions from AlphaFold \cite{alphafold2}. Among these two classes, we also cover the specific architectures, and the benefits and limitations of different methods.
    \item To evaluate the performance of model-building methods, we motivate the need for and define new specific metrics to assess the correspondence between both global structural features and local features, that refine the standard precision, recall, F1, and template matching (TM) scores.
    \item Upon selecting four representative cutting-edge methods and using the improved metrics, we evaluate and compare the performance of these methods by considering the alignment quality of predicted models from atomic to intermediate resolution, runtime and potential benefit of including structure prediction methods. 
    \item We provide valuable insights and future guidance for computer scientists, computational biologists and cryo-EM practitioners in the field of automated atomic model building.
\end{itemize}

\section{Classification of Automated Model Building Approaches} \label{sec2}

In this section, we present a comprehensive survey of DL-based model building methods, distinguishing between \textit{direct model building} approaches that only use density maps, and \textit{indirect model building} ones that integrate sequence-to-structure predictions from AlphaFold. These categories are summarized in Table \ref{tab:modelbuildingsummary}, and detailed next.

\subsection{Direct Model Building}

Direct model-building methods \cite{Emap2sec,Emap2sec+,CNN-classifier,AAnchor,Cascaded-CNN,A2-Net,DeepMM,Structure-Generator,Haruspex,EMNUSS,deeptracer,deeptracer2.0,modelangelo,Cryo2Struct} utilize deep neural networks (DNNs) to generate voxel-wise representations directly from density maps, with each voxel labeled by its associated backbone, SSE, amino acid type, C$_\alpha$ atom, and/or side-chain torsion angle. 
Backbone tracing, which connects predicted C$_\alpha$ atoms into chains, can be challenging due to disordered residues and map noise \cite{Cascaded-CNN}. Advanced algorithms such as Traveling Salesman Problem (TSP) solvers \cite{tsp1,tsp2}, Vehicle Routing Problem (VRP) solvers \cite{vrp1,vrp2}, and Monte Carlo Tree Search (MCTS) \cite{MCTS} are employed to address these challenges. Despite advancement, amino acid type prediction remains limited, mainly due to similarities among some amino acids within density maps. Improvements are achieved by aligning predicted amino acid sequences with target sequences and updating amino acid types accordingly  \cite{deeptracer,modelangelo}. We illustrate this workflow in Figure \ref{fig:modelbuildingworkflow}a. Among direct model building methods, we can distinguish different types of DL architectures, as detailed next.

\paragraph{CNN} 
Convolutional neural networks (CNNs) \cite{cnn} excel in 2D image classification and have been extended to 3D density maps to predict protein structures, including SSEs as well as atom and amino acid types and coordinates.
\emph{Emap2sec} \cite{Emap2sec} adopts a two-phase 3D-CNN to identify SSEs from intermediate-resolution maps, improving accuracy by analyzing adjacent voxels, however it does not precisely locate SSEs coordinates. The improved version, \emph{CNN-classifier} \cite{CNN-classifier} refines SSE prediction, while it is limited to training on simulated maps.
\emph{AAnchor} \cite{AAnchor} and \emph{Cascaded-CNN} \cite{Cascaded-CNN} specialize in annotating amino acids and constructing backbones respectively. Both are limited to high-resolution maps, and the latter lacks training on real-world maps. 
\emph{A$^2$-Net} \cite{A2-Net} and \emph{DeepMM} \cite{DeepMM} focus on connecting and assigning predicted amino acids into individual chains to form complete protein structures, whereas they may overlook some structural features due to limitations inherent in their algorithms.  
\emph{Structure Generator} \cite{Structure-Generator} employs a pre-trained bidirectional long short-term memory (LSTM) network \cite{lstm} to connect CNN-predicted amino acids and build protein models, but it is restricted to simulated maps.
Despite these advancements, CNN-based methods struggle with rotational invariance \cite{cnninvariance,cnnbook} which affects their performance on input density maps with varying orientations and in capturing global features and dependencies \cite{cnnglobal,cnnreview} within density maps.

\paragraph{U-Net} 
Originally designed for 2D medical image segmentation \cite{unet}, U-Net architectures comprise down-sampling encoders and up-sampling decoders with skip connections to capture both the feature information and localization within the input map. 
\emph{Haruspex} \cite{Haruspex} leverages 3D U-Nets to identify protein SSEs and nucleic acids but sometimes misidentifies similar structures (e.g., confusing $\beta$-hairpin turns and polyproline helices with $\alpha$-helical SSEs). On the other hand, \emph{EMNUSS} \cite{EMNUSS} adopts an enhanced version of U-Net with extra connections (UNet++ \cite{unet++}) to better identify $\beta$-sheets and coils. Yet, it still faces challenges with low-resolution maps and complex structures, and neither tool can fully build detailed atom models directly.
Evolved from Cascaded-CNN, \emph{DeepTracer} \cite{deeptracer} utilizes multiple 3D U-Nets to identify and locate different protein structures and has shown promise in modeling coronavirus-related structures, despite occasional inaccuracies in C$_\alpha$ tracing and connectivity faults. \emph{DeepTracer-2.0} \cite{deeptracer2.0} extends this approach to protein-DNA/RNA complexes, enhancing its utility in structure modelling.
While U-Nets are good at extracting features, they struggle with low-resolution maps where critical features are less defined, and are prone to overfitting because of their complex architectures and limited high-resolution data available.

\paragraph{GNN} 
Graph neural networks (GNNs) \cite{GNN,GNNreview} are particularly effective in modeling systems where relationships and dependencies are crucial. They work by treating elements like atoms or amino acids as nodes and their connections---such as bonds or interactions---as edges. This architecture makes GNNs ideal for building protein models.
\emph{ModelAngelo} \cite{modelangelonature,modelangelo}, a tool built on GNNs, enhances protein modeling by integrating cryo-EM map features with protein sequence embeddings derived from ESM-1b \cite{esm1}, a pre-trained protein large language model. However, the performance of GNN-based methods such as ModelAngelo is adversely affected when the node-edge relationships in low-resolution maps are ill-defined.

\paragraph{Transformer}
The Transformer architecture \cite{transformer}, initially a breakthrough in natural language processing, has also been effective in analyzing 2D images \cite{vit} and 3D volumes \cite{tranformer3d}. Attributable to its multi-head attention mechanism, Transformers can emphasize important areas in density maps, such as SSEs, improving the accuracy of protein model construction. 
\emph{Cryo2Struct} \cite{Cryo2Struct} employs this technique to identify protein atoms and amino acids by learning the interactions between atoms across long distances within the map.
A major limitation of Transformers is their large number of parameters, making them prone to overfitting, particularly when trained on limited or highly specific datasets like cryo-EM density maps. Additionally, hyperparameter tuning in Transformers is non-trivial and can greatly impact model performance.

\begin{table}[!p]
\caption{Summary of DL-based methods for atomic model building from cryo-EM density maps.}
\resizebox{\textwidth}{!}{%
\begin{tabular}{ccccccc}
\toprule
 & \thead{Method} & \thead{DL Architecture} & \thead{Training Map} & \thead{Resolution (Å)} & \thead{Built Model} & \thead{Open Source} \\
\midrule
\multirow{18}{*}{Direct} & Emap2sec \cite{Emap2sec} & 3D-CNN & Exp- \& Sim-Map & 5 - 10 & SSEs & Yes \\
\cmidrule{2-7}
 & Emap2sec+ \cite{Emap2sec+} & ResNet & Exp- \& Sim-Map & 5 - 10 & Protein SSEs \& nucleic acids & Yes \\
\cmidrule{2-7}
 & CNN-classifier \cite{CNN-classifier} & 3D-CNN & Sim-Map & 5 - 10 & Protein SSEs & No \\
\cmidrule{2-7}
 & AAnchor \cite{AAnchor} & 3D-CNN & Exp- \& Sim-Map & $\leq$ 3.1 & AA types and coordinates & Yes \\
\cmidrule{2-7}
 & Cascaded-CNN \cite{Cascaded-CNN} & 3D-CNN & Sim-Map & 2.6 - 4.4 & Backbone with C$_\alpha$ atoms, SSEs & Yes \\
\cmidrule{2-7}
 & A$^2$-Net \cite{A2-Net} & 3D-CNN & Sim-Map & $\leq$ 5 & Full-atom model & No \\
\cmidrule{2-7}
 & DeepMM \cite{DeepMM} & 3D-CNN & Exp- \& Sim-Map & 2.6 - 4.9 & Backbone (main-chain, AAs, SSEs) & Yes \\
\cmidrule{2-7}
 & StructureGenerator \cite{Structure-Generator} & CNN, GCN, LSTM & Sim-Map & 1.4 - 1.8 & Backbone (with C$_\alpha$ atoms, AAs) & Yes \\
\cmidrule{2-7}
 & Haruspex \cite{Haruspex} & 3D U-Net & Exp-Map & $\leq$ 4 & Protein SSEs \& nucleic acids & Yes \\
\cmidrule{2-7}
 & EMNUSS \cite{EMNUSS} & 3D U-Net & Exp-Map & 5 - 10 & Protein SSEs & Yes \\
\cmidrule{2-7}
 & DeepTracer \cite{deeptracer} & 3D U-Net & Exp-Map & $\leq$ 4 & Full-atom model & No \\
\cmidrule{2-7}
 & DeepTracer-2.0 \cite{deeptracer2.0} & 3D U-Net & Exp-Map & $\leq$ 4 & Full-atom model \& nucleic acids & No \\
\cmidrule{2-7}
 & ModelAngelo \cite{modelangelonature} & GNN & Exp-Map & $\leq$ 4 & Full-atom model \& nucleic acids & Yes \\
\cmidrule{2-7}
 & Cryo2Struct \cite{Cryo2Struct} & Transformer & Exp-Map & 2.1 - 5.6 & Backbone (main-chain, AAs) & Yes \\
\midrule
\multirow{9}{*}{Indirect} & EMBuild \cite{embuild} & 3D U-Net & Exp-Map & 4 - 8 & Full-atom model & Yes \\
\cmidrule{2-7}
 & FFF \cite{fff} & ResNet & Exp-Map & 1 - 4 & Full-atom model & No \\
\cmidrule{2-7}
 & DiffModeler \cite{DiffModeler} & Diffusion model & Exp-Map & 5 - 10 & Full-atom model & Yes \\
\cmidrule{2-7}
 & DeepMainmast \cite{Deepmainmast} & 3D U-Net & Exp-Map & 3 - 5 & Full-atom model & Yes \\
\cmidrule{2-7}
 & DeepTracer-ID \cite{DeepTracer-ID} & 3D U-Net & Exp-Map & $\leq$ 4.2 & Backbone (main-chain, AAs, SSEs) & No \\
\cmidrule{2-7}
 & DEMO-EM \cite{DEMO-EM} & ResNet & N/A & N/A & Full-atom model & Yes \\
\cmidrule{2-7}
 & DEMO-EM2 \cite{DEMO-EM2} & ResNet & N/A & N/A & Full-atom model & Yes \\
\cmidrule{2-7}
 & Iterative-AlphaFold \cite{phenixiter} & Phenix+AlphaFold & N/A & N/A & Full-atom model & Yes \\
\bottomrule
\end{tabular}
}
\begin{tablenotes}%
\item Note: Exp-Map refers to the deposited experimental density map and Sim-Map refers to the simulated density map. Resolution refers to the resolution of density maps for training. N/A refers to the approaches did not use density maps for training purposes. Full-atom model (Built model column) refers to main and side-chain atoms being built. SSEs refers to Protein secondary structure elements; AA refers to Amino Acid.
\end{tablenotes}
\label{tab:modelbuildingsummary}
\end{table}

\subsection{Indirect Model Building}

Some indirect model-building methods \cite{embuild,fff,DiffModeler} utilize DNNs to enhance the structural features of original maps, yielding representations like backbone probability maps \cite{fff}, main chain probability maps \cite{embuild}, or traced backbone maps \cite{DiffModeler}. 
These methods differ from direct model-building approaches since they rely on sequence-to-structure predictions like AlphaFold to construct models of protein chains, domains, or complexes.
These predicted models are then aligned to the represented maps using various fitting algorithms, including rigid-body \cite{rigid1,rigid2}, partial \cite{empot}, semi-flexible \cite{embuild}, and flexible \cite{flexible1,flexible2,flexible3} fitting, to assemble a complete model. We illustrate this workflow in Figure \ref{fig:modelbuildingworkflow}b.
In addition, some methods \cite{Deepmainmast} integrate structures predicted by AlphaFold into the C$_\alpha$ tracing stage to improve the model construction.
Other methods \cite{phenixiter,DEMO-EM,DEMO-EM2} leverage map features to refine preliminary structures predicted by approaches like AlphaFold. In contrast, some strategies \cite{DeepTracer-ID} utilize AlphaFold to refine the initial models derived directly from density maps.
Similarly to direct model building methods, we can distinguish the following architectures.

\paragraph{U-Net-Based} 
\emph{EMBuild} \cite{embuild} uses a deep U-Net architecture trained on intermediate-resolution density maps to generates main-chain probability (MCP) maps, which extract volumetric information of main-chain atoms. The AlphaFold-predicted chain models are then aligned to MCP maps and assembled to complete protein structures. Nevertheless, the performance of EMBuild may decline if the fitting is sub-optimal.
\emph{DeepMainmast} \cite{Deepmainmast} employs deep U-Nets to identify amino acids, atom types, and C$_\alpha$ positions in density maps. It improves C$_\alpha$ tracing by integrating AlphaFold predictions and aligning them to the map. The final model is selected from the best direct builds or AlphaFold fits. DeepMainmast also has a version called \emph{DeepMainmast-Base}, which builds models directly from maps. Nonetheless, both approaches demand significant computational hours due to intensive optimizations. 
\emph{DeepTracer-ID} \cite{DeepTracer-ID} begins with a preliminary atomic model derived from DeepTracer and iteratively refines it by leveraging AlphaFold predictions. However, this approach is suitable only for high-resolution maps and protein sequences longer than 100 amino acids.

\paragraph{ResNet} 
Residual networks (ResNets) \cite{resnet}, a variant of CNNs, use residual connections that allow the layers to learn from inputs directly, helping to prevent gradient vanishing during training. The residual connection allows for the use of much deeper layers, improving predictions in deeper neural networks designed for complex data.
For instance, \emph{FFF} \cite{fff} adapts a variation of ResNet, known as RetinaNet \cite{retinanet}, to generate backbone probability maps and pseudo-peptide vectors, which facilitate fitting Alphafold predictions into these maps, while this method is sensitive to map resolution.
\emph{DEMO-EM} \cite{DEMO-EM} leverages a deep ResNet to predict distances between protein domains, assisting in the flexible assembly of atomic models. Since these domain structures are produced without map data, however, they may not accurately reflect the actual protein structure. 
Built upon DEMO-EM, \emph{DEMO-EM2} \cite{DEMO-EM2} intertwines both chain-level and domain-level fittings to enhance model accuracy.
Although deeper ResNet models can perform better, they increase the complexity of networks and can be challenging to optimize.

\paragraph{Diffusion Model} 
Diffusion models \cite{diffusion} are used to reduce noise from structured data, such as images or audios. 
Extended to 3D density maps, \emph{DiffModeler} \cite{DiffModeler} takes advantage of denoising diffusion implicit models \cite{ddim} to learn protein backbone features from intermediate-resolution density maps. Using these learned representations, DiffModeler generates traced backbone maps that are used to fit AlphaFold-predicted single-chain models into a full-atom model.
A caveat is that diffusion models require considerable computational resources for both training and inference.  

\paragraph{Iterative-AlphaFold}
Terwilliger \emph{et al.} \cite{phenixiter} have introduced an iterative procedure for Phenix that refines AlphaFold predictions by implicitly incorporating cryo-EM density maps (\emph{Iterative-AlphaFold}). This process involves tweaking an initial AlphaFold model to better fit the map and then using the adjusted model as a custom template for next-cycle AlphaFold predictions. This procedure enhances the model's accuracy but works best with high-resolution maps due to rebuilding constraints inherent in Phenix \cite{phenixdock}.

\section{Evaluation Metrics for Comparison of Predicted and Target Models}

Evaluation metrics for model building depend on the alignment of predicted and target models, where the alignment of C$_\alpha$ atoms accounts for the alignment of residues. Given a predicted structure containing $L_\text{pred}$ residues and a corresponding target structure with $L_\text{target}$ residues, the number of aligned C$_\alpha$ atoms (i.e., paired atoms), denoted as $L_\text{align}$, follows the relationship: $L_\text{align} \leq L_\text{pred} \leq L_\text{target}$. 

\subsection{Limitations of Existing Evaluation Metrics}

\begin{figure}[!p]
    \centering
    \includegraphics[width=0.65\linewidth, trim={1cm 9.6cm 8cm 1cm}, clip]{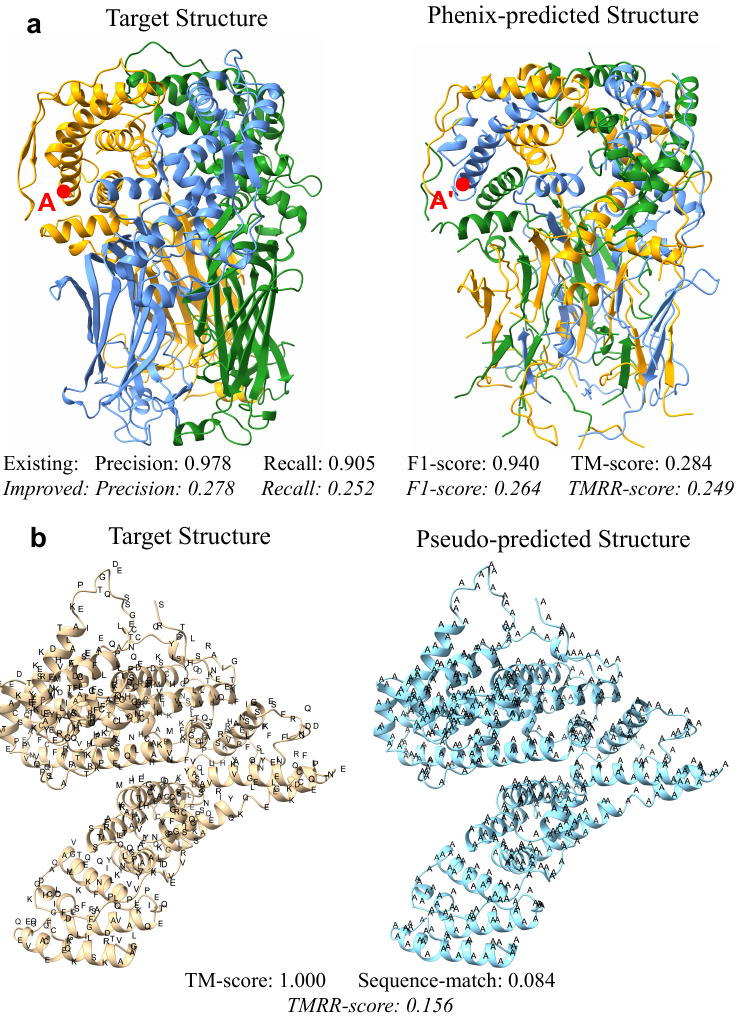}
    \caption{\small Comparison of prediction scores using different evaluation metrics. 
    \textbf{(a)} The overall topology of the Phenix-predicted protein structure closely resembles that of the target one. However, the folding patterns of the predicted protein chains are distinctly different. Existing metrics, such as those outlined by ModelAngelo \cite{modelangelo}, do not account for chain correspondence, thereby resulting in very high precision, recall, and F1-score. Conversely, the TM-score, which considers chain correspondence, is notably low. In our improved metrics, these scores are aligned, showing minimal discrepancies.
    The visualized protein is a structure of rotavirus VP6 (PDB-ID: 3J9S) \cite{3J9S}. Different chains are depicted in different colors. The red dots $A$ and $A^\prime$ refer to C$_\alpha$ atoms in the target and Phenix-predicted structures, respectively. 
    \textbf{(b)} The pseudo-predicted structure shares an identical shape with the target one, achieving a TM-score of 1.000. However, they consist of distinct amino acid types, resulting in a low Sequence-match score of 0.084. In our improved metric, TMRR-score, this value is 0.156. The visualized protein is a structure of human Alpha-fetoprotein (PDB-ID: 7YIM) \cite{7YIM}. The letter represents the one-letter code for amino acids. 
    }
    \label{fig:metriclim}
\end{figure}

Existing metrics such as recall, precision, F1-score, C$_\alpha$ matching score \cite{phenix}, C$_\alpha$ quality score \cite{Cryo2Struct}, and sequence recall \cite{modelangelo} are commonly used to measure the percentage of paired C$_\alpha$ atoms. However, these metrics count paired atoms without considering chain correspondence. Ideally, paired atoms should reside within the same corresponding chains. As illustrated in Figure \ref{fig:metriclim}a, C$_\alpha$ atoms $A$ from the target structure and $A^\prime$ from the Phenix-predicted structure belong to different corresponding chains: $A$ in the yellow chain, while $A^\prime$ in the blue chain. Metrics that disregard chain correspondence incorrectly pair $A$ with $A^\prime$, artificially inflating the precision, recall, and F1-score. In contrast, when chain correspondence is considered, $A$ and $A^\prime$ should not be paired since they belong to different corresponding chains, resulting in significantly lower TM-score \cite{tmscore}. In addition, although the TM-score accounts for chain correspondence, it does not consider the accuracy of residue pairing identity, i.e., whether the paired residues share the same amino acid type, as illustrated in Figure \ref{fig:metriclim}b. We also note that another metric, called Local Distance Difference Test (LDDT) \cite{lddt} has been used to quality of sequence-to-structure predictions. As it requires the same number of residues between two models, its usage for benchmarking model building methods here is limited. Overall, these limitations highlight the need to design better metrics.

\subsection{Improved Evaluation Metrics}

As discussed above, the limitations of existing evaluation metrics include (\romannumeral 1) the ignorance of chain-level correspondence, and (\romannumeral 2) the absence of residue identity matching in TM-score. 
To address (\romannumeral 1), we refine the precision, recall and F1-score, by using a heuristic algorithm, US-align \cite{usalign}, to systematically align the predicted and target structures based on the highest TM-score obtained from all possible combinations of their chains.
To address (\romannumeral 2), we introduce a new metric, named \textbf{\emph{TMRR-score}}, that couples the TM-score with Residue-recall to simultaneously measure both structural and residue-type similarities.

\paragraph{Refined C$_\alpha$ Precision, C$_\alpha$ Recall, and C$_\alpha$ F1-score}
First, we align the two structures using US-algin and collect the aligned C$_\alpha$ atom of each residue from one structure and its counterpart from the other. 
We define a residue as paired, or a C$_\alpha$ atom as correctly predicted, if the distance between the aligned C$_\alpha$ atoms is less than or equal to a specified threshold of 3 {\AA} \cite{phenix}. 
Next, we leverage the defined correctly-predicted C$_\alpha$ atom, termed CP-C$_\alpha$ atom, to calculate the following quantitative measures. \\ 
{\emph{C$_\alpha$ Precision}} is defined as the percentage of predicted C$_\alpha$ atoms that are correct, expressed as:
\begin{equation}
\text{C$_\alpha$ Precision} = \frac{\text{\# of CP-C$_\alpha$ atoms}}{\text{\# of all predicted } C_{\alpha} \text{ atoms}}.
\end{equation} 
{\emph{C$_\alpha$ Recall}} is defined as the fraction of C$_\alpha$ atoms in the target structure that are correctly identified, expressed as:
\begin{equation}
\text{C$_\alpha$ Recall} = \frac{\text{\# of CP-C$_\alpha$ atoms}}{\text{\# of all target } C_{\alpha} \text{ atoms}}.
\end{equation} 
{\emph{C$_\alpha$ F1-score}} is the harmonic mean of precision and recall, expressed as:
\begin{equation}
\text{C$_\alpha$ F1-score} = 2 \times \frac{\text{C$_\alpha$ Precision} \times \text{C$_\alpha$ Recall}}{\text{C$_\alpha$ Precision}+ \text{C$_\alpha$ Recall}}.
\end{equation}

\paragraph{TMRR-Score}
Inspired by the harmonic mean \cite{harmonicmean} that is more sensitive to the lower value, we define the \emph{TMRR-score} by coupling the TM-score and Residue-recall:
\begin{equation}
\text{TMRR-score} = 2 \times \frac{\text{TM-score} \times \text{Residue-recall}}{\text{TM-score} + \text{Residue-recall}},
\end{equation}
\begin{equation} \label{eq-tm}
\text{TM-score} = \frac{1}{L_\text{target}}\left[ \sum_{i=1}^{L_{\text{align}}}{\frac{1}{1+(d_i/d_0)^2}} \right].
\end{equation}
Where $d_0 = 1.24 \times \sqrt[3]{L_{\text{target}}-15} - 1.8$ is a length-dependent scale to normalize the C$_\alpha$ atom distance of the $i$th pair of aligned residues, $d_i$ \cite{tmscore}. \emph{Residue-recall} is defined as the ratio of paired residues that match in amino acid type to the overall count of residues within the target structure. 

The TMRR-score ranges from $(0,1]$ with a higher score indicating a more pronounced similarity in the context of both shape and residue identity. A score of 1 indicates that predicted and target models are identical, i.e., $L_\text{align} = L_\text{pred} = L_\text{target}$. 
Note that we design Residue-recall to measure the accuracy of amino acid type predictions, unlike Sequence-match, which simply measures the percentage of correctly predicted amino acids relative to the total number of predicted residues. Residue-recall accounts for residue coverage. For instance, a predicted structure covering only a fraction of the target structure could still achieve a high Sequence-match score.
We suggest that the sensitivity of the TMRR-score to lower values helps mitigate the risk that an exceptionally high score in one metric could compensate a poor score in the other, thus encouraging a more balanced and unbiased evaluation.

\section{Benchmarking Results} \label{results}

To our knowledge, previous work has not conducted a comprehensive comparison across cutting-edge model-building methods, and the evaluation metrics used vary and are not unified across different methods. Using our new evaluation metrics, we conducted a comprehensive benchmarking of four cutting-edge model-building approaches. Phenix serves as the baseline, and the others are DL-based, namely ModelAngelo \cite{modelangelonature}, EMBuild \cite{embuild}, and DeepMainmast \cite{Deepmainmast}. We selected these three DL-based methods because they are fully automated, open-source, and have consistently demonstrated state-of-the-art performance based on thorough assessments. ModelAngelo is a direct model-building method, whereas EMBuild operates indirectly. DeepMainmast offers two modes: one that integrates AlphaFold predictions (direct) and another does not (indirect). Selecting one physics-based, one deep learning (DL)-direct, one DL-indirect, and one hybrid method incorporating AlphaFold models also ensures a balanced and representative comparison across different methodological categories. 
Specifically, we benchmarked these methods over a diverse set of 50 proteins, including monomers, homomers, and heteromers, with density maps at resolutions ranging from 1.8 {\AA} to 7.8 {\AA} (as detailed in Supplementary Table \ref{tab:metric_emid_data}). Among them, 70\% are multi-chain protein complexes, comprising 18 homomers and 17 heteromers, while the remaining 30\% are single-chain proteins (15 monomers), ensuring biological and structural diversity.
Note that these 50 candidate maps exclude maps from the training datasets of the DL approaches to ensure an unbiased and fair comparison. For more details about our implementation, see Supplementary \ref{metric_implementation}.

\subsection{Performance Comparison Across Methods}

\begin{figure}[!hp]
\centering
\includegraphics[width=\linewidth, trim={0cm 0cm 0cm 0cm}, clip]{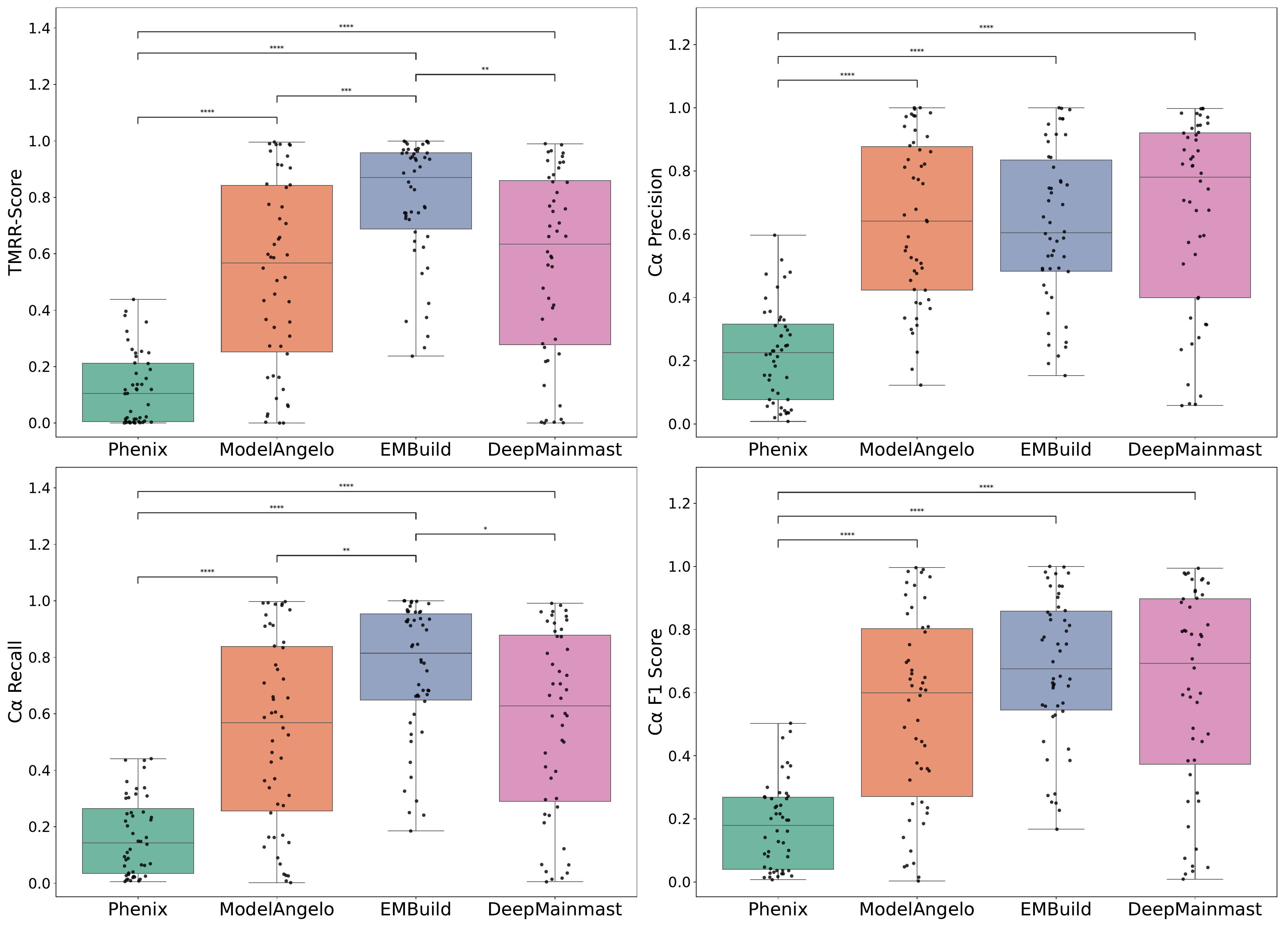}
\caption{ 
Comparison of improved metrics (TMRR-score, C$_\alpha$ Precision, C$_\alpha$ Recall, and C$_\alpha$ F1-score) of 50 protein models generated by Phenix, ModelAngelo, EMBuild, and DeepMainmast. The dots represent the scores for each protein model. Statistical differences between methods were assessed using Kruskal-Wallis tests followed by pairwise Mann-Whitney U tests with Bonferroni correction. Significance levels are indicated as $\text{*}p \leq 0.05$, $\text{**}p \leq 0.01$, $\text{***}p \leq 0.001$, and $\text{****}p \leq 0.0001$.}
\label{fig:metric_boxplot_newmetric}
\end{figure}

In Figure \ref{fig:metric_boxplot_newmetric}, each box-and-whisker plot illustrates the distribution of scores for atomic models constructed from 50 density maps, each assessed using the respective methods\footnote{Phenix failed to construct atomic models for the cryo-EM density maps of EMDB-29290 and EMDB-15560. DeepMainmast failed to construct atomic models for EMDB-29290 and EMDB-8685.}.
Note that for the DeepMainmast and DeepMainmast-Base results, we opted to benchmark the constructed backbone models instead of the full-atom models, since we observed significant distortions in some full-atom models derived from their backbone models using Rosetta-v2021.16, as illustrated in Supplementary Figure \ref{fig:Metric_dmm_problem}. 
Given that our evaluation metrics specifically focus on C$_\alpha$ atoms, they primarily assess the accuracy of backbones, ensuring that the metrics are not biased towards either backbone-only or full-atom models.

\begin{figure}[!ht]
\centering
\includegraphics[width=\linewidth, trim={0cm 0cm 0cm 0cm}, clip]{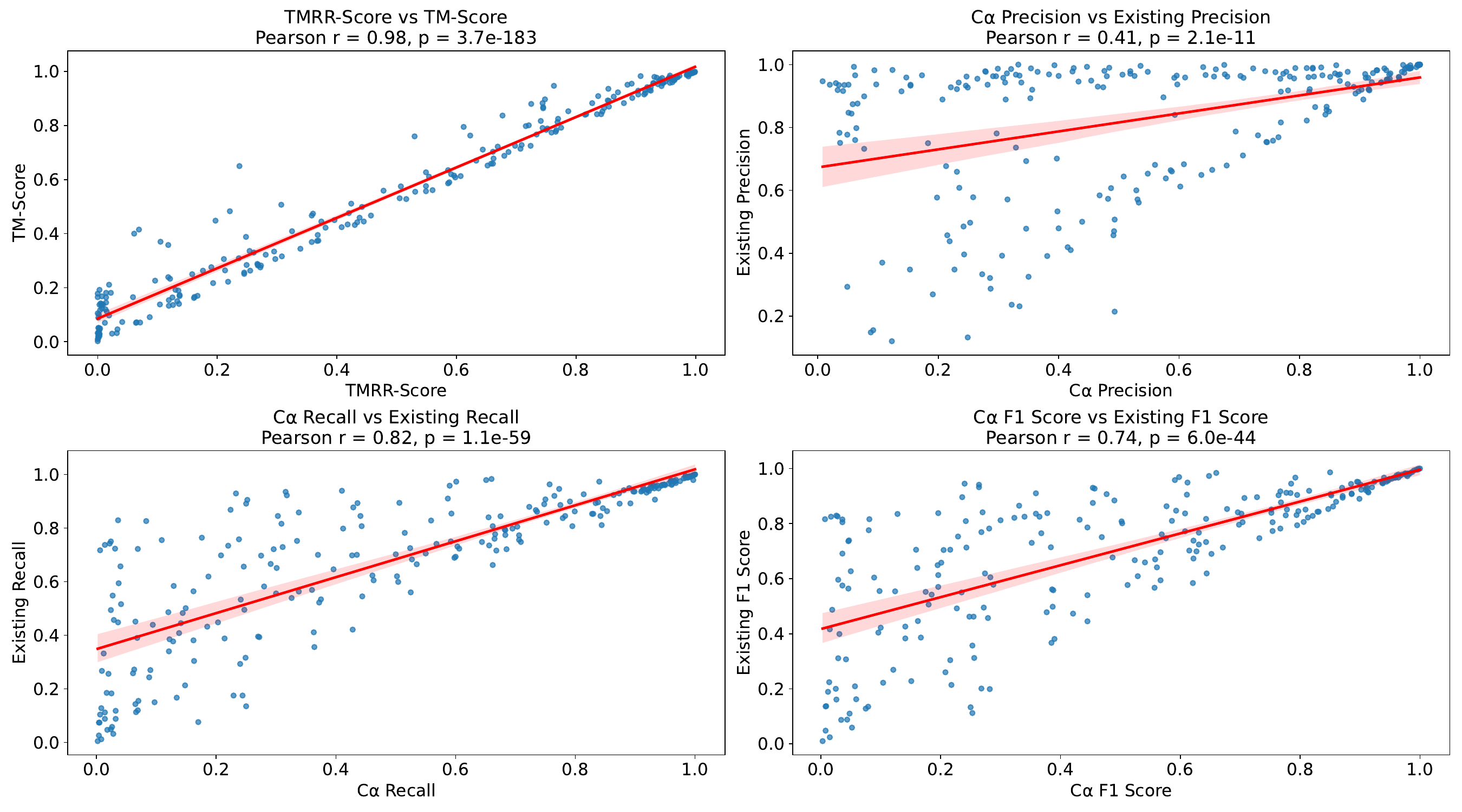}
\caption{
Pairwise Pearson correlation scatter plots between corresponding improved and existing evaluation metrics (TMRR-score vs. TM-score, C$_\alpha$ Precision vs. Existing Precision, C$_\alpha$ Recall vs. Existing Recall, and C$_\alpha$ F1-score vs. Existing F1-score) across all tested samples.
}
\label{fig:Metric_corr_scatter_compare}
\end{figure}

The TMRR-scores revealed that EMBuild outperformed its counterparts, exhibiting exceptional model reliability. Moreover, EMBuild showed the narrowest box width (interquartile range), indicating its more consistent results across the tested density maps. Both ModelAngelo and DeepMainmast achieved comparable mean and median TMRR-scores above 0.5, suggesting that their constructed models closely match the target ones. In contrast, the lower scores of Phenix reflected its underperformance in accurately building protein models and predicting amino acid types. Additionally, DeepMainmast showed the highest mean and median C$_\alpha$ precision, suggesting its effectiveness in producing relevant structures. On the other hand, EMBuild scored the highest mean values in both C$_\alpha$ recall and C$_\alpha$ F1-score, with the narrowest box width emphasizing its high modeling coverage and overall performance, aligning with its TMRR-score.
To assess the statistical significance of differences among these methods, we performed Kruskal-Wallis tests \cite{Kruskal-Wallis} for each evaluation metric, followed by pairwise Mann-Whitney U tests \cite{Mann-Whitney} with Bonferroni correction. The results confirmed that Phenix's performance was significantly poorer than the DL-based methods across all metrics ($\text{****}p < 0.0001$), while EMBuild achieved better performance than both ModelAngelo ($\text{***}p < 0.001$ and $\text{**}p < 0.01$) and DeepMainmast ($\text{**}p < 0.01$ and $\text{*}p < 0.05$) in terms of TMRR-scores and C$_\alpha$ recall, respectively. In contrast, no significant differences were observed between ModelAngelo and DeepMainmast across most metrics.
Table \ref{tab:comparison} summarizes the mean and median scores for each method. Based on these results, we concluded that DL-based methods involving EMBuild, ModelAngelo, and DeepMainmast all showcased good performance across varying evaluation metrics, demonstrating their capabilities to build accurate atomic models from density maps.
We also employed existing metrics including TM-score \cite{tmscore}, precision, recall, and F1-score \cite{modelangelo} to compare each method, with the results presented in Supplementary Figure \ref{fig:Metric_boxplot_oldmetric}. We observed significant discrepancies and inconsistencies between TM-scores and the other three measures. For instance, Phenix exhibited a very low median TM-score below 0.2, while achieving relatively high scores in the other three metrics. 
To quantitatively assess the internal consistency of the improved metrics, we computed Pearson correlations across three settings: between corresponding improved and existing metrics (Figure \ref{fig:Metric_corr_scatter_compare}), among the improved metrics (Supplementary Figure \ref{fig:Metric_corr_scatter_improved}), and among the existing metrics (Supplementary Figure \ref{fig:Metric_corr_scatter_existing}).
Figure \ref{fig:Metric_corr_scatter_compare} shows that some improved metrics, such as TMRR-score and C$_\alpha$ Recall, align well with their existing counterparts, indicating consistency.
Others including C$_\alpha$ Precision and C$_\alpha$ F1-score show moderate correlations, suggesting that they could capture complementary aspects of model quality that are not reflected in existing metrics.
Moreover, all improved metrics demonstrated strong positive correlations with one another, ranging from 0.85 to 0.99, indicating high consistency (see Figure \ref{fig:Metric_corr_scatter_improved}).
In contrast, the existing metrics exhibited much weaker correlations, with TM-scores showing particularly weak correlations with other metrics (See Figure \ref{fig:Metric_corr_scatter_existing}). These results not only validate our measurement framework but underscore the critical importance of implementing these refined metrics.

\begin{table*}[!t]
\caption{Performance comparison of different model-building methods using improved metrics.\label{tab:comparison}}
\tabcolsep=0pt
\begin{tabular*}{\textwidth}{@{\extracolsep{\fill}}lcccccccc@{\extracolsep{\fill}}}
\toprule%
& \multicolumn{2}{@{}c@{}}{TMRR-score $\uparrow$} & \multicolumn{2}{@{}c@{}}{C$_\alpha$ Precision $\uparrow$} & \multicolumn{2}{@{}c@{}}{C$_\alpha$ Recall $\uparrow$} & \multicolumn{2}{@{}c@{}}{C$_\alpha$ F1-score $\uparrow$} \\
\cline{2-3}\cline{4-5}\cline{6-7}\cline{8-9}%
Methods & Mean & Median & Mean & Median & Mean & Median & Mean & Median \\
\midrule
Phenix & 0.119 & 0.104 & 0.222 & 0.226 & 0.166 & 0.143 & 0.179 & 0.179 \\
ModelAngelo & 0.526 & 0.568 & 0.640 & 0.642 & 0.531 & 0.569 & 0.540 & 0.599 \\
EMBuild & \textbf{0.788} & \textbf{0.870} & 0.625 & 0.605 & \textbf{0.751} & \textbf{0.815} & \textbf{0.674} & 0.675 \\
DeepMainmast & 0.563 & 0.634 & \textbf{0.664} & \textbf{0.780} & 0.569 & 0.628 & 0.604 & \textbf{0.693} \\
\bottomrule
\end{tabular*}
\end{table*}

\subsection{Performance Comparison Across Map Resolutions}

\begin{figure}[!p]
\centering
\includegraphics[width=\linewidth, trim={0cm 0cm 3cm 0cm}, clip]{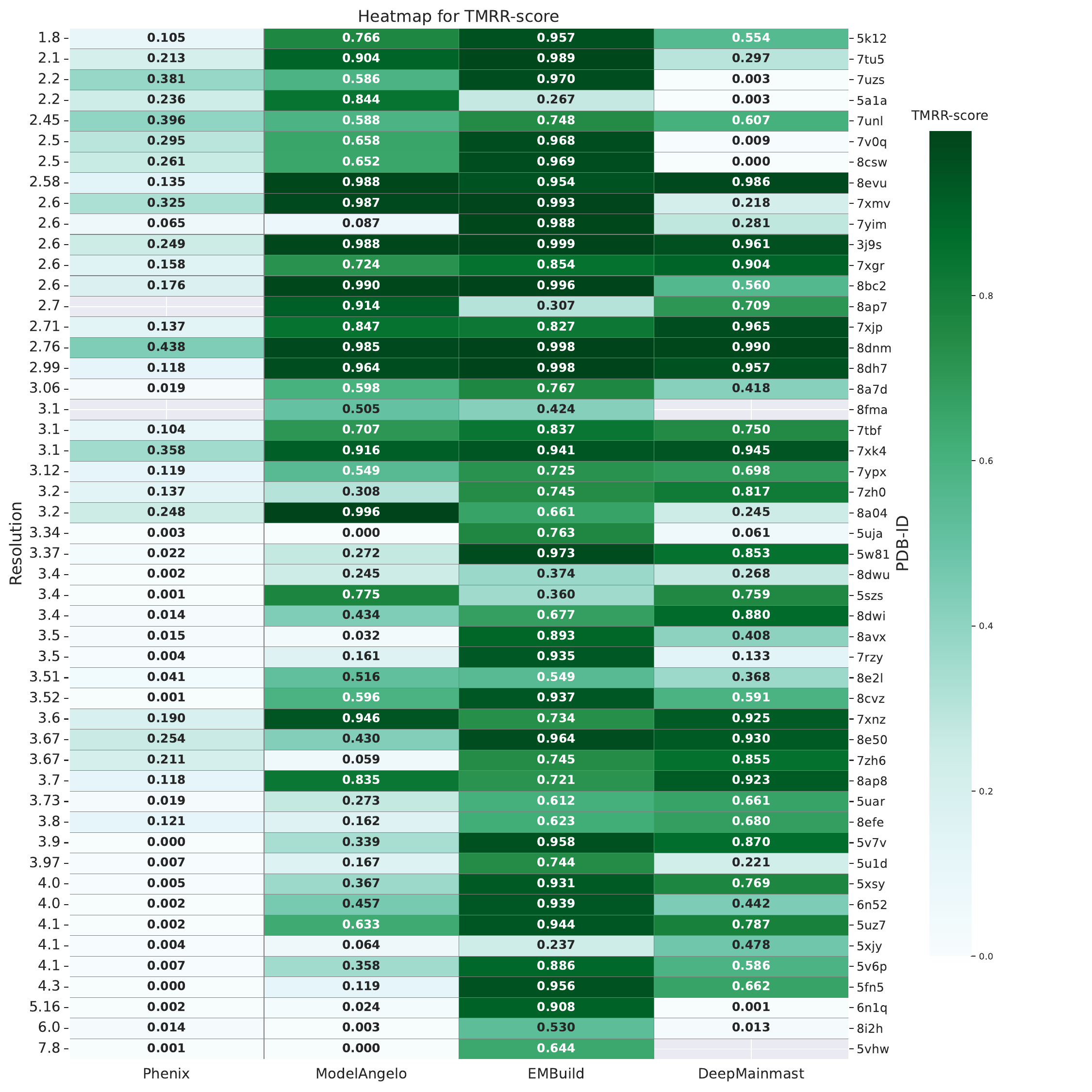}
\caption{\small 
The heat map for the TMRR-score of 50 protein models generated by Phenix, ModelAngelo, EMBuild, and DeepMainmast, sorted by their map resolution from high to low. The darker color refers to the higher score. The value in each cell represents the specific score for the generated model. The left labels show the resolutions and the right labels show the PDB IDs of each generated model.
}
\label{fig:hmap_tmrr}
\end{figure}

We compared different methods against density maps of varying resolutions from 1.8 to 7.8 {\AA}, and their TMRR-scores were depicted as a heat map in Figure \ref{fig:hmap_tmrr}. ModelAngelo demonstrated superior performance at near-atomic resolutions, consistently achieving scores above 0.5, indicating its robustness in constructing meaningful protein models. However, its performance declined at resolutions worse than 3.7 {\AA}. DeepMainmast also performed well at high resolutions, although some models were poorly constructed. In contrast, EMBuild consistently exhibited good performance across a broad range of resolutions, suggesting its independence from map resolution. Phenix, on the other hand, was highly dependent on resolution, with TMRR-scores plummeting to nearly zero at resolutions worse than 4 {\AA}. Furthermore, we observed that all methods except EMBuild struggled to build meaningful models from intermediate-resolution density maps (4 - 8 {\AA}). Interestingly, DeepMainmast was ineffective at very high resolutions (1.8 - 2.5 {\AA}), suggesting that its algorithms failed to adequately capture map features at these finer resolutions.
Additional heat maps of precision, recall, and F1-score are displayed in Supplementary Figures \ref{fig:Metric_hmap_precision}, \ref{fig:Metric_hmap_recall}, and \ref{fig:Metric_hmap_f1score}, respectively, and the results align with their TMRR-scores.

\subsection{Visual Comparison}

\begin{figure}[!p]
    \centering
    \includegraphics[width=0.8\linewidth, trim={0cm 0cm 0cm 0cm}, clip]{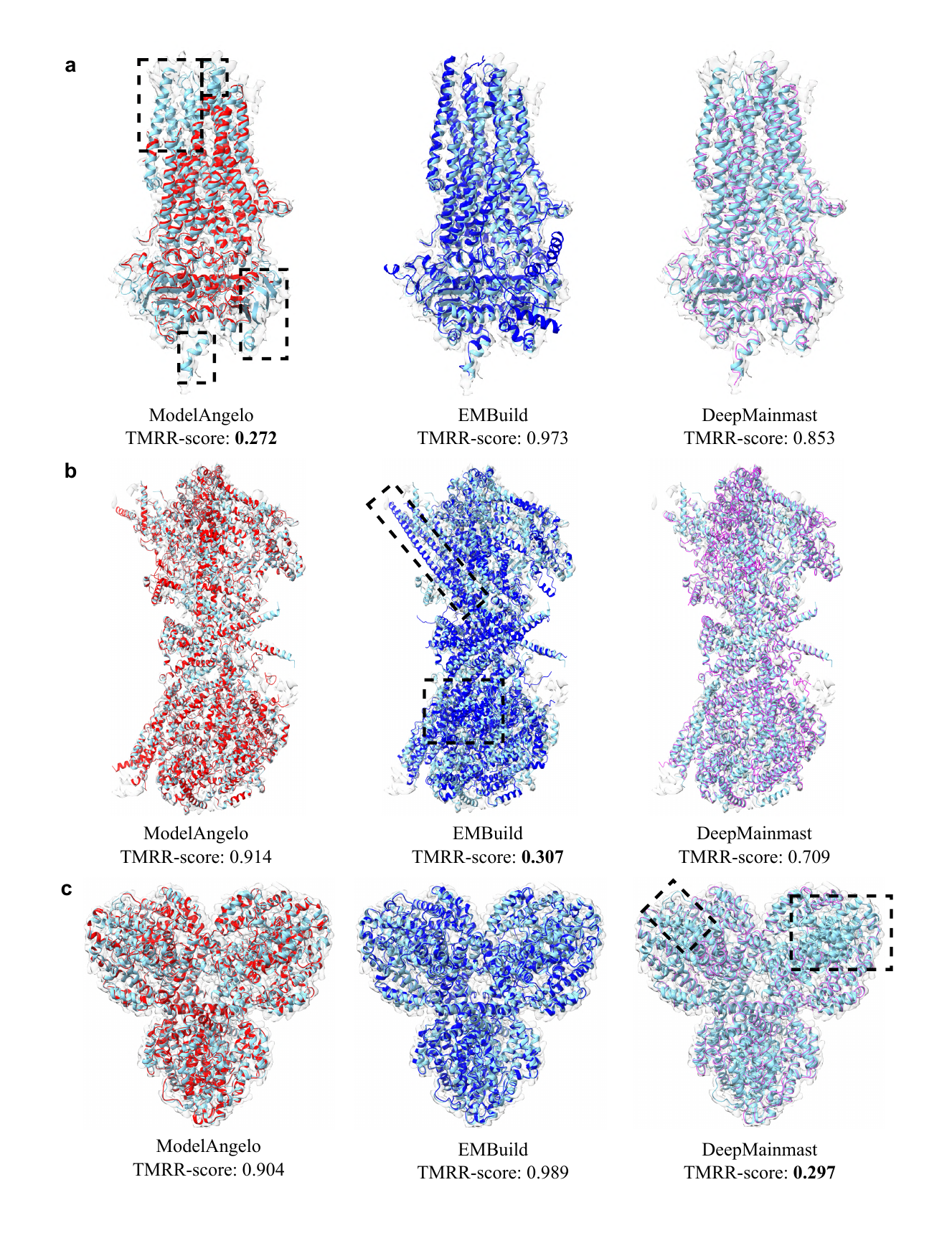}
    \caption{\small \textbf{a-c} Constructed protein atomic models by ModelAngelo (red), EMBuild (blue), and DeepMainmast (purple), respectively, with cryo-EM density maps (transparent gray) and the corresponding reference PDB structures (cyan). Each TMRR-score value is shown at the bottom of the model. Black boxes highlight the poorly modeled regions.
    \textbf{(a)} Phosphorylated, ATP-bound structure of zebrafish cystic fibrosis transmembrane conductance regulator (PDB-ID: 5W81; EMDB-ID: 8782; Resolution: 3.37 {\AA}) \cite{5W81}.
    \textbf{(b)} Membrane region of Trypanosoma brucei mitochondrial ATP synthase dimer (PDB-ID: 8AP7; EMDB-ID: 15560; Resolution: 2.7 {\AA}) \cite{8AP7}.
    \textbf{(c)} Structure of L. blandensis dGTPase in the apo form (PDB-ID: 7TU5; EMDB-ID: 26126; Resolution: 2.1 {\AA}) \cite{7TU5}.}
    \label{fig:review_visual_comparison}
\end{figure}

While the performance comparison in the previous section provides valuable insights from an overall comparison on a large dataset, further insights can be gained by visually examining specific cases where each method performs poorly. To this end, we visualized the constructed atomic models from three representative high-resolution density maps. 
Figure \ref{fig:review_visual_comparison}a shows a poorly ModelAngelo-constructed model with a lower TMRR-score than that of others. We observed that some regions that were enclosed by black boxes were not constructed by ModelAngelo, causing the incompleteness and low TMRR-score, although the rest parts were well constructed. These incompletely modeled areas can be attributed to regions with high noise and low-density values. Figure \ref{fig:review_visual_comparison}b shows a poorly constructed model by EMBuild. We found that some parts of the model were misaligned with the corresponding density volumes, thereby causing an inferior construction. Moreover, we noticed that the completeness was high in the EMBuild-constructed model and almost every density volume was registered due to its explicit usage of AlphaFold predictions. Figure \ref{fig:review_visual_comparison}c reveals a poorly constructed model by DeepMainmast, while perfectly constructed by the other two methods with TMRR-scores above 0.9. 
Although DeepMainmast also integrated AlphaFold in their model-building process, rather than directly aligning the AlphaFold predictions to the map as EMBuild did, it leveraged the predictions to enhance the C$_\alpha$ tracing process. Therefore, DeepMainmast's performance was still strongly dependent on the map quality.

\begin{table}[!ht]
\centering
\caption{Model building time taken by various methods across different sequence lengths. All methods were implemented on one NVIDIA A100 GPU with AMD EPYC 7V12 64-Core Processor. \label{tab:metric_time}} 
\vspace{1em}
\resizebox{\textwidth}{!}{%
\begin{tabular}{ccccccc}
\toprule
Sample & Sequence length & \multicolumn{5}{c}{Model building runtime (hours) $\downarrow$}\\
\cmidrule{3-7}
&  & Phenix & ModelAngelo & EMBuild & DeepMainmast & DeepMainmast-Base \\
\midrule
EMDB-27755  & 495 & \textbf{0.29} & 1.31 & 2.30  & 6.07  & 3.34  \\
EMDB-28081 & 1053 & \textbf{0.65}  & 1.32 & 3.29  & 9.82  & 3.73  \\
EMDB-27574 & 2288 & 1.86 & 2.38 & \textbf{1.80}  & 42.66 & 13.54 \\
EMDB-27022 & 4648 & \textbf{1.43} & 3.99  & 3.82  & 41.31 & 25.52 \\
\bottomrule
\end{tabular}
}
\end{table}

\subsection{Impact of Incorporating AlphaFold}

\begin{figure}[!ht]
\centering
\includegraphics[width=\linewidth, trim={0.5cm 21cm 0.5cm 0cm}, clip]{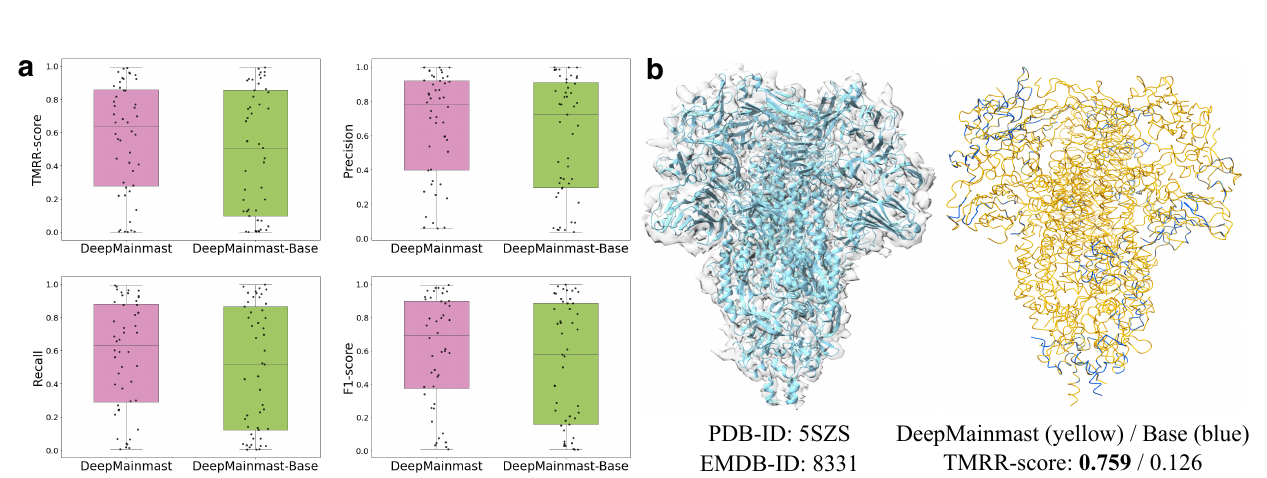}
\caption{ 
\textbf{(a)} Comparison of four metrics (TMRR-score, C$_\alpha$ Precision, C$_\alpha$ Recall, and C$_\alpha$ F1-score) of 50 protein models generated by DeepMainmast and DeepMainmast-Base. The dots represent the scores for each protein model.
\textbf{(b)} Left: the reference PDB structure colored in cyan superimposes onto the corresponding cryo-EM density map colored in transparent gray. Right: the superimposed constructed atomic models by DeepMainmast and DeepMainmast-Base, colored in yellow and blue, respectively. TMRR-score values are shown on the bottom of the models. Glycan shield and epitope masking of a coronavirus spike protein (PDB-ID: 5SZS; EMDB-ID: 8331; Resolution: 3.4 {\AA}) \cite{5SZS}.
}
\label{fig:afimpact_all}
\end{figure}

To investigate how AlphaFold enhances model-building performance, we compared DeepMainmast, which incorporated AlphaFold, against DeepMainmast-Base, which did not. Figure \ref{fig:afimpact_all}a shows that DeepMainmast outperformed DeepMainmast-Base across all four metrics and produced more consistent results, as evidenced by its narrower box width. 
Additionally, we visually compared the models constructed by both methods for the same map, as illustrated in Figure \ref{fig:afimpact_all}b. The results revealed that DeepMainmast can build more complete and less fragmented models than DeepMainmast-Base, because of the integration of AlphaFold. 
Both quantitative and visual results suggest that integrating AlphaFold enhances modeling coverage, achieving a higher degree of completion and reducing fragmentation and disconnections. This achievement stems from AlphaFold's proficiency in constructing a complete topology. However, integrating AlphaFold comes at a cost: increased computational demands and extended runtime. As evidenced in Table \ref{tab:metric_time}, DeepMainmast has a significantly slower runtime compared to DeepMainmast-Base.

\subsection{Runtime Comparison}
The runtime for Phenix included map sharpening and model-building processes; 
for ModelAngelo included the initial model-building step and three-round refinements;
for EMBuild included AlphaFold prediction and model-building process; 
and for DeepMainmast and DeepMainmast-Base included model-building process and AlphaFold prediction for the former. Note that the AlphaFold prediction runtime refers to the time to run AlphaFold locally, as some predictions were not available in the \texttt{AlphaFold Protein Structure Database}. 
Table \ref{tab:metric_time} lists the runtime comparison across various methods for constructing atomic models with different sequence lengths, ranging from 495 to 4648 residues. There is a noticeable variance in computational efficiency, from minutes to days. Phenix demonstrated the fastest runtime for both short and long sequences. ModelAngelo and EMBuild exhibited acceptable runtime, with a slight increase as the sequence length extended. In contrast, DeepMainmast and DeepMainmast-Base showed considerable runtime, with the runtime increasing exponentially as the sequence length extended. Nevertheless, all methods were significantly faster than manual model building, which requires extensive domain expertise.

\section{Conclusion and Future Directions}

In this chapter, we present a comprehensive survey of recent state-of-the-art approaches, categorizing them into direct methods that rely solely on density maps and indirect methods that incorporate sequence-to-structure predictions from tools like AlphaFold. We delve into the specific neural network architectures employed by these methods, examining their design and functionality. Furthermore, we discuss the pros and cons of each approach, providing insights into their effectiveness across various modeling scenarios. 

To generate unbiased comparison and consistent results, we improved the existing evaluation metrics. Upon comparing the performance of four representative methods across 50 density maps spanning a variety of resolutions, our results demonstrate that DL-based methods including ModelAngelo, EMBuild, and DeepMainmast significantly outperform the physics-based approach embodied by Phenix. EMBuild stands out across the entire spectrum of tested resolutions and excels in terms of TMRR-score, recall, and F1-score, showcasing its capability in constructing reliable and accurate atomic models, with the caveat that occasionally EMBuild completely misses the mark. On the other hand, ModelAngelo and DeepMainmast also perform well, particularly at high resolutions, with DeepMainmast achieving the highest precision. 

The integration of AlphaFold into the model-building process has significantly enhanced construction completeness and accuracy. This improvement is particularly evident in the performance of EMBuild and DeepMainmast, where the inclusion of AlphaFold predictions boosts modeling coverage (completeness). However, AlphaFold's dependency on available sequence information limits its applicability. Thus, alternative approaches such as ModelAngelo remain crucial for constructing atomic models in the absence of sequence data.

The field of automated atomic model building from cryo-EM density maps is still in its early stage of development and stands to benefit notably from enhanced collaboration between machine learning scientists and structural biologists. 
The rapid emergence of deep learning architectures and algorithms that achieved significant success in other fields highlights the potential for more sophisticated DL-based approaches to improve atomic model building. Moreover, there is a need for methods that leverage multi-modal data, such as structural templates, amino acid sequences, and 2D cryo-EM projected images, for training purposes. Incorporating physico-chemical constraints, such as rotation angles and bond lengths, should also be considered in future network design. While most methods currently focus on high-resolution density maps, there is also a pressing need to develop methods suitable for intermediate- and low-resolution maps. 

Finally, a critical barrier for the effective development of DL-based methods for automated atomic model building is the  need for high-quality data, specifically, perfect map-model pairs. A large number of maps are poorly aligned with their corresponding models, hindering neural networks' ability to learn effectively. In the future, concerted efforts from research institutions and pharmaceutical companies to share cryo-EM density maps and atomic models with the community would significantly enhance the diversity of current datasets, leading to reduced bias during network training and boost in performance.
\chapter{Struc2mapGAN: Improving Synthetic Cryo-EM Density Maps with Generative Adversarial Networks} \label{chap:gan}

This chapter is a modified version of a paper by Chenwei Zhang, et al., published in Bioinformatics Advances (\url{https://doi.org/10.1093/bioadv/vbaf179}) \cite{struc2mapgan}.

\section{Motivations and Contributions}

For the past few years, DL-based methods have been introduced to automate the construction of macromolecular atomic models from density maps \cite{modelangelonature,deeptracer,embuild} (see Table \ref{tab:modelbuildingsummary}). However, most of these methods require high-resolution experimental maps, given  the lack of reliable experimentally derived map-model pairs at lower resolutions \cite{wriggers2015numerical}. In this context, generating accurate synthetic maps can help to include the vast majority of structures that are limited to be solved at lower resolutions. In addition, these can also be used for ``rigid-body fitting'' in the building process (i.e., aligning a whole or partial atomic model into a density map \cite{empot,eman1}), ``sharpening'' density maps \cite{emready,cryofem} (i.e., enhancing the map to facilitate model building), or providing more ground-truth training targets.

As generating synthetic cryo-EM density maps has become a pivot, various simulation-based methods are available \cite{eman1, eman2,chimerax,tempy2,situs}, essentially using a resolution-lowering point spread function, as detailed in Section \ref{sec:simulation_cryoem}, to convolute atom points extracted from PDB structures (atomic models). Upon treating atoms independently, these methods may fail to characterize more complex features, such as secondary structure elements (SSEs), interatomic interactions, or specific image artifacts inherent to the 3D reconstruction process.  
As a result, notable discrepancies can be observed between simulated maps generated by tools like \emph{molmap} in ChimeraX \cite{chimerax} and actual experimental maps, particularly in the representation of SSEs and local density patterns. To address this gap, a potential solution is to  leverage deep generative model, to learn a mapping from simulated to experimental maps. This led us to develop and present here a new method,  called \textbf{Struc2mapGAN}, that uses a generative adversarial network (GAN) \cite{gan} to enhance the generation of simulations of cryo-EM density maps from PDB structures and current simulation methods.

Our main contributions are:
\begin{itemize}
    \item We introduce the first deep learning-based method, to our knowledge, for generative modeling of synthetic cryo-EM density maps. Our method enhances the learning efficiency by curating high-quality training data and alleviates mode collapse \cite{modecollapse} in GANs through the integration of \texttt{SmoothL1Loss} into the model. 
    \item Our benchmarking shows superior overall performance against simulation-based me\-thods, across various evaluation metrics (e.g., correlation) and over a variety of tested maps. Our experiments also suggest that this performance improvement results from better capture of SSEs.
    \item We benchmark the runtime of \emph{struc2mapGAN} and demonstrate its practical suitability for generating large-scale maps.
    \item Our ablation study demonstrates the importance of integrating an additional \texttt{Smooth\-L1Loss} to guide GAN training and using curated maps as target inputs, both of which enhance model performance.
\end{itemize}

\section{Dataset Preparation}
\paragraph{Training and validation data}
To train and validate the GAN model, our dataset was built with a set of high-resolution cryo-EM density maps ranging from 2.2 {\AA} to 3.9 {\AA} from the EMDB databank \cite{EMDB} and associated PDB structures from the PDB databank \cite{PDB}. To ensure the density maps have proper alignments with their associated PDB structures, we removed maps from the dataset if:
(\romannumeral 1) maps contain extensive regions without corresponding PDB structures;
(\romannumeral 2) maps are misaligned with associated PDB structures;
(\romannumeral 3) maps contain various macromolecules, such as nucleic acids;
(\romannumeral 4) PDB structures only contain backbone atoms and/or include unknown residues.
In addition, pairs of simulated and experimental maps with correlation lower than 0.65 (calculated using ChimeraX) were excluded. 
To eliminate data redundancy, we measured the sequence identity between PDB structures and retained only one when the identity exceeded 30 \%.
After applying these filtering steps, a total of 149 cryo-EM density maps and associated PDB structures remained, as listed in Supplementary Table \ref{tab:gan_trainvaltb}. 134 map-PDB pairs (90 \%) were selected as the training set and 15 pairs (10 \%) as the validation set. 

\paragraph{Test data}
For the test set, we randomly selected 130 PDB structures from the PDB databank and associated cryo-EM density maps from the EMDB databank with resolution ranging from 3 {\AA} to 7.9 {\AA}, as detailed in Supplementary Table \ref{tab:gan_testtb}. Note that these 130 examples do not overlap with the training and validation data sets.

\begin{figure}[!hp]
    \centering
    \includegraphics[width=\linewidth, trim={0cm 1cm 0 0cm}, clip]{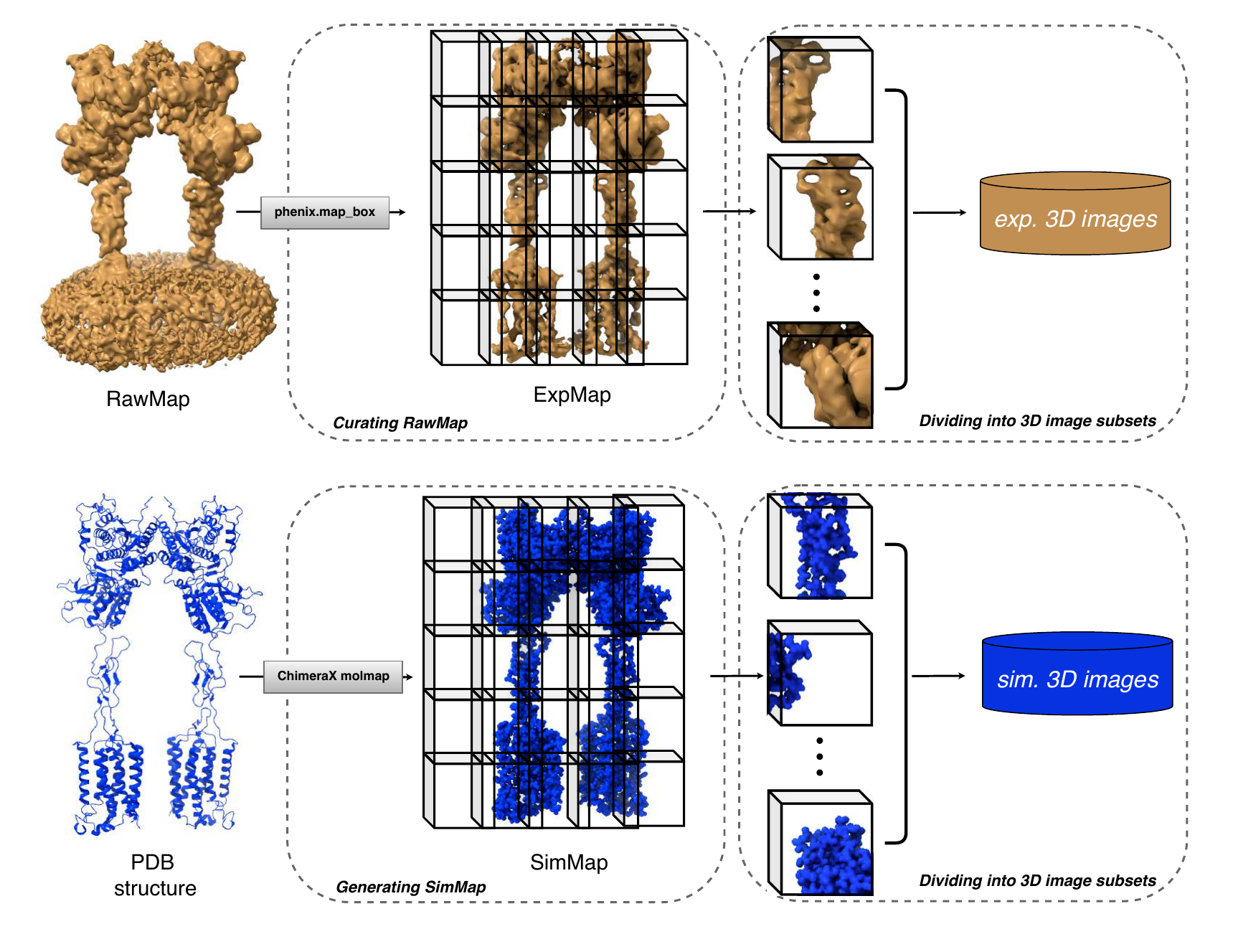}
    \caption{ 
    The data preprocessing workflow. The top panel depicts the process of curating raw experimental maps and dividing the curated maps into 3D image subsets as training targets. The bottom panel depicts the process of generating simulated maps and dividing them into 3D image subsets as training inputs. The illustrated structure is Metabotropic Glutamate Receptor 5 Apo Form (EMDB ID: 0346; PDB ID: 6N52; reported resolution: 4 Å) \cite{6N52}.
    }
    \label{fig:dataprocess}
\end{figure}

\paragraph{Preprocessing}
According to the following preprocessing steps illustrated in Figure \ref{fig:dataprocess}, both experimental maps (denoted as $Raw\-Maps$) and simulated maps from PDB structures (denoted as $Sim\-Maps$) were pre-processed to facilitate the GAN training. In addition, our GAN architecture also takes 3D images of reduced size as input.

\paragraph{Curating Raw\-Maps}
The raw experimental maps consistently contain background noise, such as lipid solvents and artifacts from nanodiscs, which compromise the pairing accuracy between $Raw\-Maps$ and $Sim\-Maps$. To mitigate these effects, we applied a mask to the raw map to isolate the region containing only the protein structure. We first aligned the PDB structure with the corresponding map and employed \emph{phenix.map\_box} to create a rectangular box slightly larger than the targeted region. Subsequently, we resampled the masked map at a grid voxel size of $1 \text{\AA} \times 1 \text{\AA} \times 1 \text{\AA}$ and applied min-max normalization to scale the map's voxel values to the range of $[0, 1]$ to maintain uniformity. The curated map, termed $Exp\-Map$, was then employed for training the network. Note that this curation strategy to make $Exp\-Map$ targets significantly enhanced the performance of the GAN model, as evidenced in our ablation study section.

\paragraph{Generating Sim\-Maps}
The input data $Sim\-Maps$ were simulated using the \emph{molmap} function. This step provides clean and standardized inputs, allowing the network to learn the mapping toward more realistic experimental maps.
We converted the PDB structure into a simulated density map on a grid corresponding to the $Exp\-Map$, with a resolution cutoff at 2 {\AA}. This simulated map was then min-max normalized to the range of $[0, 1]$, aligning with its corresponding $Exp\-Map$. Moreover, in light of enhancing the model's robustness, we utilized \emph{TorchIO} \cite{torchio} to add random Gaussian noise, random anisotropy, and random blur to augment the input data.

\paragraph{Dividing maps into 3D image subsets}
Considering the varying dimensions of each map and the constraints of GPU memory, we zero-padded the $Sim\-Maps$ and $Exp\-Maps$ and divided them into smaller 3D images (denoted as \textit{exp. 3D images} and \textit{sim. 3D images}, respectively) with dimensions of $32\times32\times32$.
To do so, we first created a padded map that exceeded the dimensions of the input map by $2 \times 32$ in each dimension, ensuring no boundary issues. The original map was then centrally placed within this padded map. Following this procedure yielded a total of 403710 3D images for training and 62635 for validation.

\section{The Struc2mapGAN Model}

\subsection{Architecture}

\begin{figure}[!ht]
    \centering
    \includegraphics[width=0.7\linewidth, trim={0cm 10cm 12cm 0cm}, clip]
    {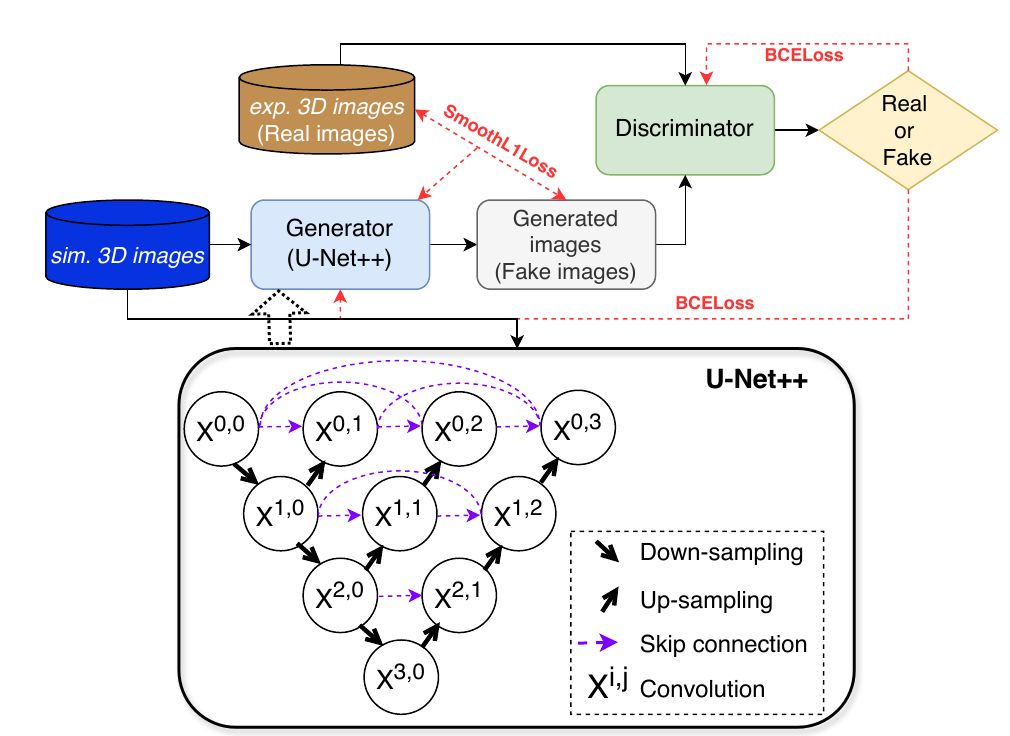}
    \caption{ 
    The \emph{struc2mapGAN} architecture. The bottom panel illustrates the U-Net++ architecture. $X^{i,j}$ refers to the convolution block at depth $i$ and position $j$ of the network.
     }
    \label{fig:GAN-architecture}
\end{figure}

Figure \ref{fig:GAN-architecture} depicts the GAN architecture that comprises a generator and a discriminator. 
We have implemented a nested U-Net architecture (U-Net++) \cite{unet++} as the generator (shown in the bottom panel of Figure \ref{fig:GAN-architecture}). The encoder and decoder blocks of the U-Net++ follow the same design, each consisting of 3D convolution layers with a kernel size of $3\times3\times3$. Unlike standard U-Nets, U-Net++ applies dense skip connections that effectively bridge the encoder and decoder feature maps, for enhanced gradient flow. 3D max-pooling layers, featuring a kernel size of 2 and a stride of 2, are employed for down-sampling, while trilinear interpolation layers with a scale factor of 2 are used for up-sampling.

The discriminator network has been designed for 3D volumetric data classification. It consists of four 3D convolution layers with a kernel size of $3\times3\times3$. An adaptive average-pooling layer is applied after the last convolution layer to reduce the feature map to the size of  $1\times1\times1$, which is then flattened and passed through three fully connected layers. The final layer outputs a single value for binary classification.
For both the generator and discriminator, we employ a parametric rectified linear unit (PReLu) as the activation function between convolution layers, coupled with instance normalization. We opt for instance normalization over the more commonly used batch normalization as the statistics should not be averaged across instances within a batch. This consideration is crucial since local 3D image subsets of cryo-EM density may originate from different maps. The total number of parameters for the generator is 10.3 million and for the discriminator is 1.2 million. 

\subsection{Network Training and Inference}
\paragraph{Training}
During training, \emph{struc2mapGAN}accepts paired \textit{sim. 3D images} and \textit{exp. 3D images} (simulated and experimental maps are divided into smaller 3D images, denoted as \textit{sim. 3D images} and \textit{exp. 3D images}, respectively) as input and output modified images. 
To impose the generator to minimize the difference between the prediction and the target, we incorporated the standard adversarial loss for each input \textit{sim. 3D image} with the smooth L1 loss in the generator, as
\begin{equation} \label{eq:smoothl1loss}
    \text{SmoothL1Loss}(X,Y) =
    \begin{cases}
    0.5(X-Y)^2, & \text{if } |X-Y| < 1, \\
    |X - Y| - 0.5, & \text{otherwise},
    \end{cases}
\end{equation}
where $X$ is the GAN-predicted image, i.e., $G(\textit{sim. 3D image})$, and $Y$ is the paired target image, i.e., \textit{exp. 3D image}. The adversarial loss is calculated as the binary cross entropy loss of the discriminator's predictions, as
\begin{equation}
    \text{Loss}_{\text{adversarial}} = - \log (D(X)).
\end{equation}
Therefore, the generator loss is formulated as a linear combination of adversarial loss and smooth L1 loss:
\begin{equation}
    \text{Loss}_G = \text{SmoothL1Loss} + \alpha \times \text{Loss}_{\text{adversarial}},
\end{equation}
where $\alpha$ is a tuning hyperparameter. 
The discriminator loss adheres to the standard GAN loss formula, aiming to classify $X$ as fake and $Y$ as real:
\begin{equation}
    \text{Loss}_D = - (\log D(Y) + \log (1 - D(X))).
\end{equation}

\paragraph{Inference}
To generate a map from a PDB structure using our trained generator, the input PDB structure was initially converted to a map following the procedure of generating $Sim\-Maps$.
This converted map was subsequently zero-padded and divided into 3D image subsets with dimensions of $32\times32\times32$, following the same strategy employed during training data preprocessing. These image subsets were then refined by the trained generator to yield post-processed images. The post-processed images were then reassembled back to their original map dimensions. To ensure that no spatial information was lost, we exclusively utilized the center $20\times20\times20$ voxels of each 3D image to reconstruct the map, in accordance with the method proposed by Si et al. \cite{Cascaded-CNN}. 

\paragraph{Implementation}
\emph{Struc2mapGAN} was implemented in PyTorch-2.2.2 + cuda-12.1, with all training and validating processes carried out on eight NVIDIA A100 GPUs, each with 80 GB VRAM. This setup supported a batch size of 128. The network was trained over $150$ epochs, with each epoch taking approximately $765$ seconds. The total training duration was approximately 1 day and 8 hours.
For optimization, NAdam \cite{nadam} optimizers were employed with a learning rate set at $0.0001$. 
We tested three different $\alpha$ values: $0.1, 0.01, 0.001$. Ultimately, $\alpha = 0.01$ was selected for yielding the best performance.

\section{Evaluation Metrics}

To measure the accuracy of \emph{struc2mapGAN} in generating maps that are similar to experimental reference ones, we used the following evaluation metrics.

\subsection{Structural Similarity Index Measure} 
The structural similarity index measure (SSIM) evaluates the similarity between two images based on three key features: luminance, contrast, and structure \cite{ssim}. This measure has been extended to evaluate 3D volumetric data and has been adapted as a loss function in cryo-EM density map studies \cite{emready,embuild}. The SSIM score between samples $X$ and $Y$ is calculated as follows:
\begin{equation}
    \text{SSIM}(X,Y) = \frac{(2\mu_{X}\mu_{Y}+c_1)(2\sigma_{XY}+c_2)}{(\mu_{X}^{2}+\mu_{Y}^{2}+c_1)(\sigma_{X}^{2}+\sigma_{Y}^{2}+c_2)},
\end{equation}
where $\mu_X$ and $\mu_Y$, $\sigma_X$ and $\sigma_Y$ are the means and variances of samples $X$ and $Y$, respectively. $\sigma_{XY}$ refers to the covariance of $X$ and $Y$. Constants $c_1$ and $c_2$ are included to stabilize the division. SSIM excels in capturing the local contrast and textures, which are crucial in cryo-EM density maps. In our study, we employed the \emph{scikit-image} package to calculate the SSIM between a \emph{struc2mapGAN}-generated or simulation-based density map and its corresponding experimental map.  

\subsection{ChimeraX Correlation}
UCSF ChimeraX \cite{chimerax} provides a built-in \emph{measure correlation} function to calculate the correlation between two density maps in two ways. One way is to compute the correlation between two maps directly without considering the deviations from their means:
\begin{equation}
    \text{correlation}(X,Y) = \frac{\langle X, Y \rangle}{\|X\|\|Y\|},
\end{equation}
where $\langle X,Y\rangle$ represents the dot product of maps $X$ and $Y$; $\|X\|\|Y\|$ are the norms of the maps. The other way calculates the correlation between deviations of two maps from their means:
\begin{equation}
    \text{correlation about mean}(X,Y) = \frac{\langle X-X_{\text{ave}}, Y-Y_{\text{ave}} \rangle}{\|X-X_{\text{ave}}\|\|Y-Y_{\text{ave}}\|},
\end{equation}
where $X_{\text{ave}}$ and $Y_{\text{ave}}$ are the means of the two maps.
In the context of comparing cryo-EM density maps, we calculated the two correlations to assess the similarity across the entire volume of the two maps.

\subsection{Pearson Correlation Coefficient}
The Pearson correlation coefficient (PCC) is a measure of the linear correlation between two datasets \cite{pcc} and is defined by:
\begin{equation}
    \text{PCC}(X,Y) = \frac{\sum_{i=1}^{n} (X_i - \overline{X})(Y_i - \overline{Y})}{\sqrt{\sum_{i=1}^{n} (X_i - \overline{X})^2 \sum_{i=1}^{n} (Y_i - \overline{Y})^2}},
\end{equation}
where $X_i$ and $Y_i$ represent the individual data points in datasets $X$ and $Y$, respectively. $\overline{X}$ and $\overline{Y}$ are the means of the datasets, and $n$ is the total number of data points.  
In the context of comparing cryo-EM density maps, the PCC provides similarities between the two maps in terms of their density distributions.

\section{Experiments and Results}

We conducted a comprehensive study to evaluate the performance of \emph{struc2mapGAN} using a test set of 130 PDB structures and associated experimental density maps (note that these structures and maps were not used for training the GAN network). In our benchmarking, simulation-based maps were generated using various methods (\emph{molmap} \cite{chimerax}, \emph{e2pbd2mrc} \cite{eman2}, and \emph{StructureBlurrer} \cite{tempy2}), with a resolution cutoff at 2 {\AA}.

\subsection{Learning Curves with Intermediate Maps}
\begin{figure}[!ht]
    \centering
    \includegraphics[width=\linewidth, trim={1cm 11.5cm 2cm 1cm}, clip]{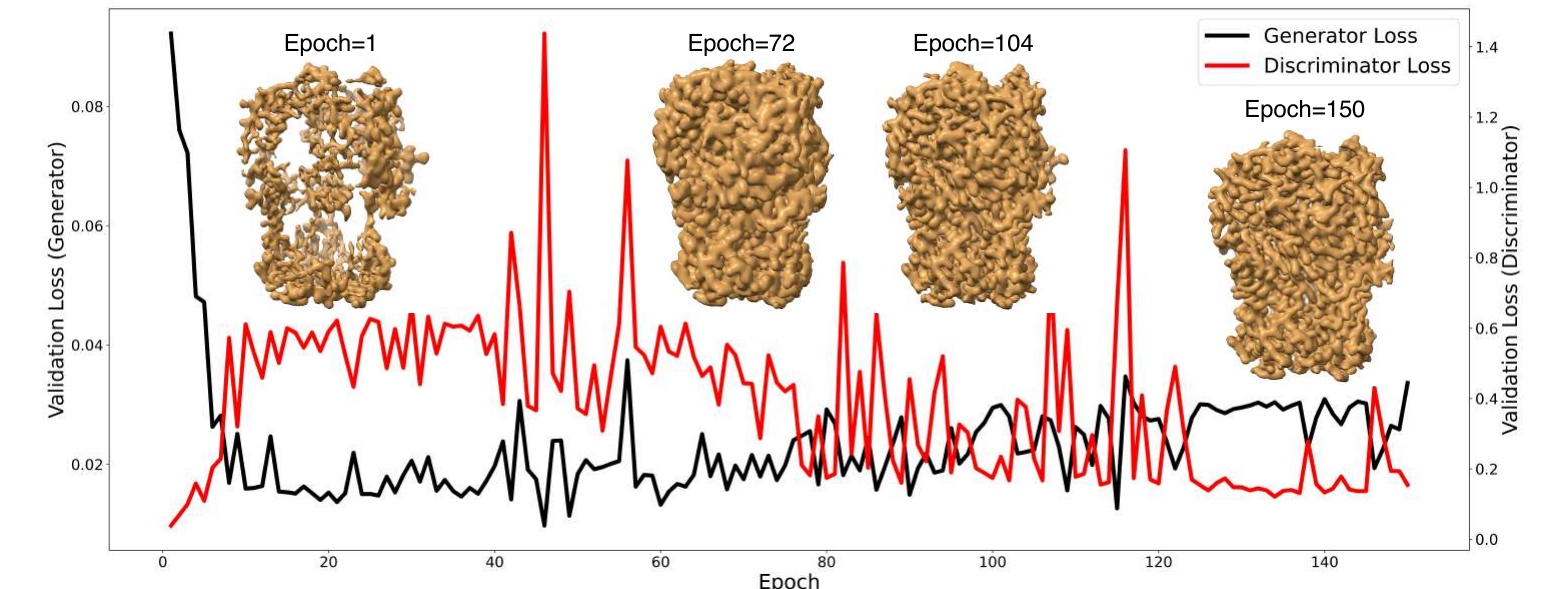}
    \caption{ 
    Validation loss curves of the generator and discriminator in black and red, respectively, with snapshots of generated maps from the models trained at specific epochs. 
    The illustrated structure is Rotavirus VP6 (EMDB ID: 6272; PDB ID: 3J9S; reported resolution: 2.6 {\AA} \cite{3J9S}).
    }
    \label{fig:lossfig}
\end{figure}

To analyze the training process, we visualized the training dynamics of \emph{struc2mapGAN} in Figure \ref{fig:lossfig}. The validation losses of both the generator (black) and discriminator (red) were plotted over 150 epochs. At the beginning of training, the generator loss decreased sharply, reflecting its quick adaptation to produce more realistic maps, while the discriminator loss slightly increased owing to the generator's enhancements. Throughout the training, fluctuations in both losses indicate continuous learning and adaptation. By the end of the training, the losses of both the generator and discriminator tended to converge, suggesting a balanced and stabilized adversarial training outcome. Alongside the loss curves, selected \emph{struc2mapGAN}-generated maps from epochs 1, 72, 104, and 150 were displayed. These intermediate maps visually encapsulated the gradual improvement in the quality and fidelity of the generated maps. For instance, the map from Epoch 1 showed incomplete densities. As training proceeded, the generated maps learned to represent complete densities, with their resolution progressively refined, as illustrated in the maps from Epoch 72 to Epoch 150.

\begin{figure}[!hp]
    \centering
    \includegraphics[width=.88\linewidth, trim={0cm 3.5cm 0cm 0cm}, clip]{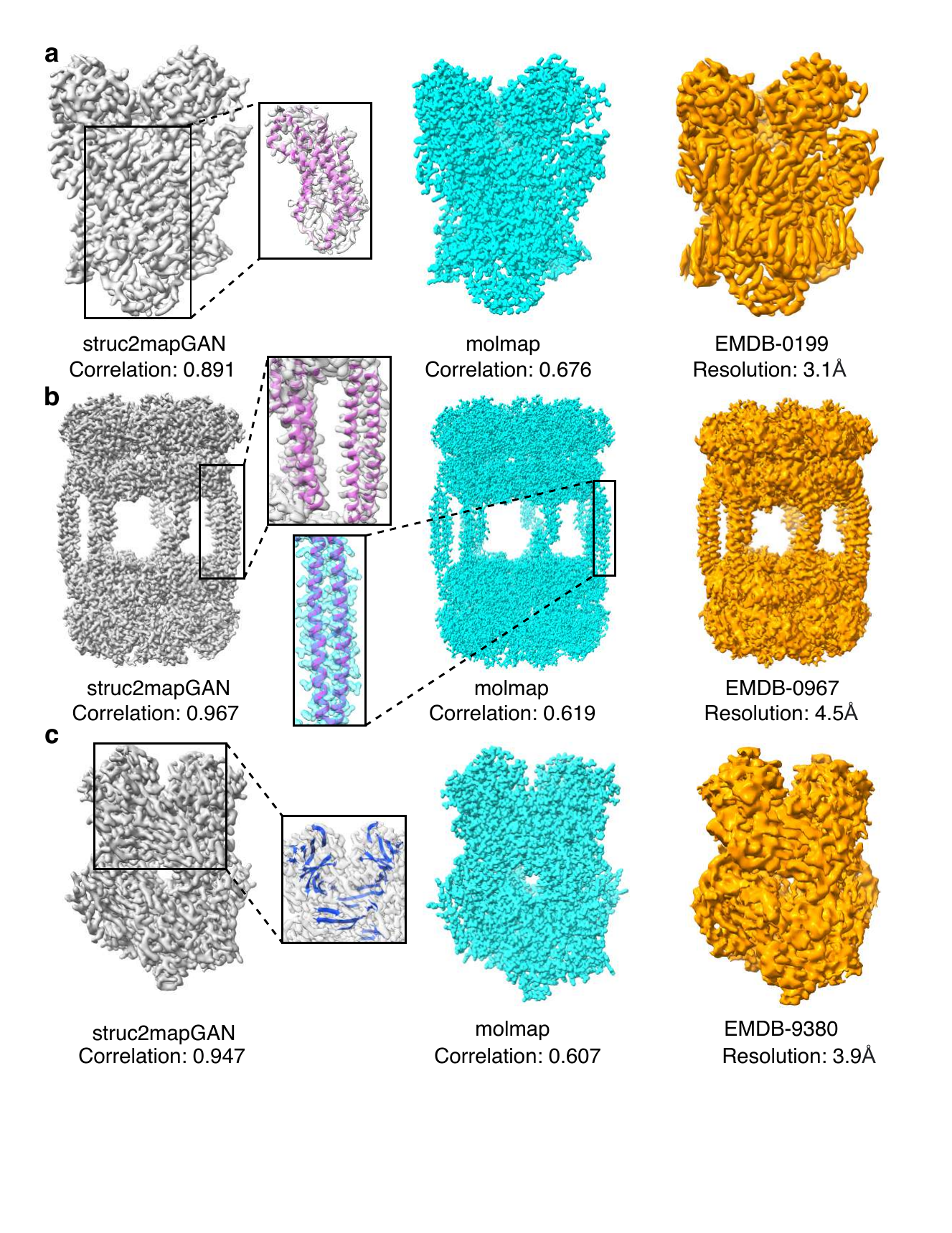}
    \caption{ 
    Examples of \emph{struc2mapGAN} (gray) and \emph{molmap} (cyan) generated maps, and the raw experimental maps (orange). The PDB structures of $\alpha$-helices (pink) and $\beta$-sheets (blue) are superimposed on the maps. 
    \textbf{a.} Human STEAP4 bound to NADP, FAD, heme and Fe(III)-NTA (EMDB ID: 0199; PDB ID: 6HCY; reported resolution: 3.1 {\AA}) \cite{6HCY}.
    \textbf{b.} AAA+ ATPase, ClpL from Streptococcus pneumoniae: ATPrS-bound (EMDB ID: 0967; PDB ID: 6LT4; reported resolution: 4.5 {\AA}) \cite{6LT4}.
    \textbf{c.} Truncated HIV-1 Vif/CBFbeta/A3F complex (EMDB ID: 9380; PDB ID: 6NIL; reported resolution: 3.9 {\AA}) \cite{6NIL}.
    Visualization of cryo-EM density maps and PDB structures was produced by UCSF ChimeraX \cite{chimerax}.
    }
    \label{fig:visual_comp}
\end{figure}

\subsection{Visual Comparison Between Generated Maps by Struc2mapGAN and Molmap}
\begin{figure}[!ht]
    \centering
    \includegraphics[width=\linewidth, trim={0.5cm 17cm 1cm 1cm}, clip]{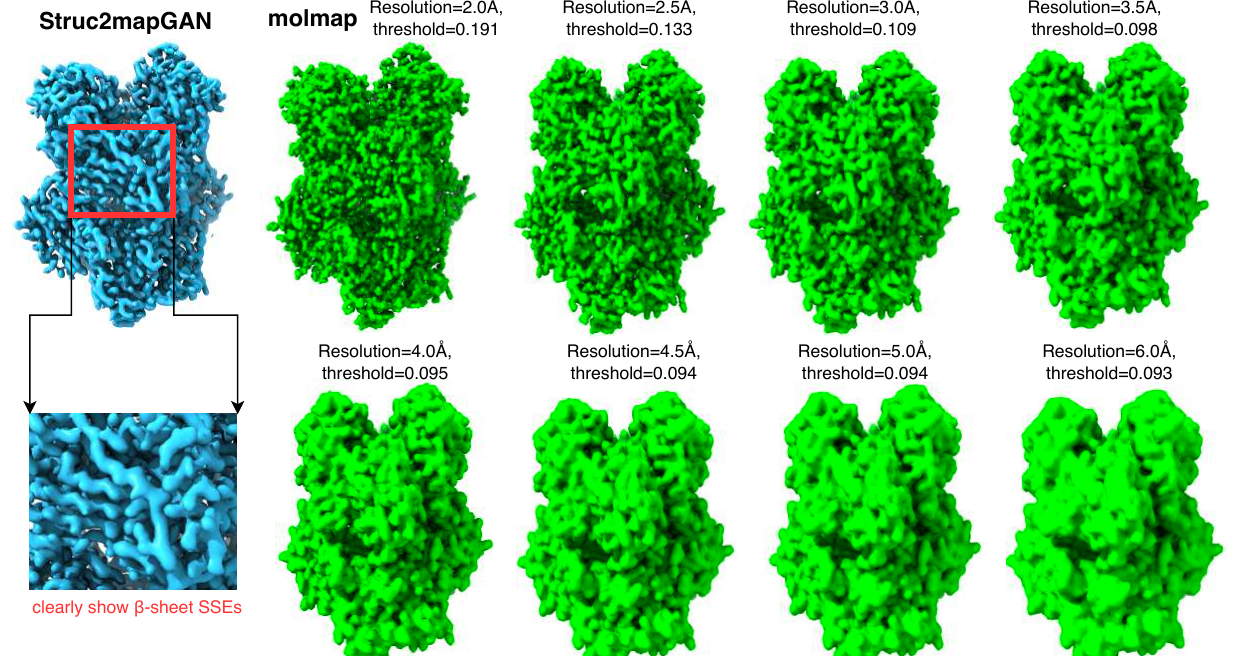}
    \caption{ 
    Struc2mapGAN-generated map is shown in blue, and molmap-simulated maps are shown in green, with varying resolution and threshold.
    The structure visualized here is Truncated HIV-1 Vif/CBFbeta/A3F complex (EMDB ID: 9380; PDB ID: 6NIL; reported resolution: 3.9 {\AA}) \cite{6NIL}.
    }
    \label{fig:rebuttal}
\end{figure}

Upon training, we first visualized three GAN-generated density maps alongside their associated \emph{molmap}-generated maps, with corresponding experimental maps with various resolutions. The visual comparisons are displayed in Figure \ref{fig:visual_comp}. 
We observed that as \emph{struc2mapGAN} is trained to represent high-resolution maps, the GAN-generated maps were produced with consistent detail level across the three examples, with high correlation scores (0.891, 0.967 and 0.947) with experimental maps. 
In contrast, \emph{molmap}-simulated maps displayed fine-grained resolution at the level of individual atoms (0.676, 0.619 and 0.607). In practice, these maps effectively captured the spatial distribution of atomic densities but did not seem to distinctly represent SSEs (see on-set panels in Figure \ref{fig:visual_comp}).
In contrast, \emph{struc2mapGAN} better captures SSEs such as $\alpha$-helices and $\beta$-sheets, as
GAN-generated maps for EMDB-0967 and EMDB-9380  distinctly revealed in Figure \ref{fig:visual_comp}b and c the densities of helical and sheet structures, respectively, indicating that our model effectively learned the complexity of experimental maps.
We further examined whether similar visual characteristics could be replicated by adjusting the resolution and threshold parameters in \emph{molmap}. As shown in Figure \ref{fig:rebuttal}, tuning these parameters primarily smooths the map surface but does not recover SSEs such as $\beta$-sheets, which remain absent in the \emph{molmap}-simulated maps. In contrast, \emph{struc2mapGAN}-generated maps distinctly reveal such features. This highlights that \emph{struc2mapGAN} goes beyond surface-level smoothing by learning to enhance biologically meaningful structural elements that simulation-based methods fail to reproduce.
These visual results demonstrate that the maps synthesized by \emph{struc2mapGAN} achieved visual high resolution quality, with improved representation of SSEs.

\begin{table}[!t]
\caption{Performance comparison of different methods.\label{tab:GAN-comparison}}
\tabcolsep=0pt
\begin{tabular*}{\textwidth}{@{\extracolsep{\fill}}lcccccccccccccccc@{\extracolsep{\fill}}}
\toprule%
& \multicolumn{2}{@{}c@{}}{SSIM} $\uparrow$ & \multicolumn{2}{@{}c@{}}{Correlation} $\uparrow$ & \multicolumn{2}{@{}c@{}}{Correlation about mean} $\uparrow$ & \multicolumn{2}{@{}c@{}}{PCC} $\uparrow$ \\
\cline{2-3}\cline{4-5}\cline{6-7}\cline{8-9}%
Methods & Mean & Median & Mean & Median & Mean & Median & Mean & Median \\
\midrule
\textbf{struc2mapGAN}                 & 0.841 & \textbf{0.896} & \textbf{0.906} & \textbf{0.943} & \textbf{0.466} & \textbf{0.502} & \textbf{0.594} & \textbf{0.621} \\
molmap              & 0.771 & 0.774 & 0.559 & 0.573 & 0.315 & 0.311 & 0.452 & 0.470 \\
StructureBlurrer    & 0.764 & 0.771 & 0.603 & 0.620 & 0.335 & 0.334 & 0.475 & 0.499 \\
e2pdb2mrc           & \textbf{0.848} & \textbf{0.896} & 0.613 & 0.649 & 0.322 & 0.325 & 0.483 & 0.519 \\
\bottomrule
\end{tabular*}
\end{table}

\begin{figure}[!ht]
    \centering
    \includegraphics[width=\linewidth, trim={0cm 0cm 0cm 0cm}, clip]{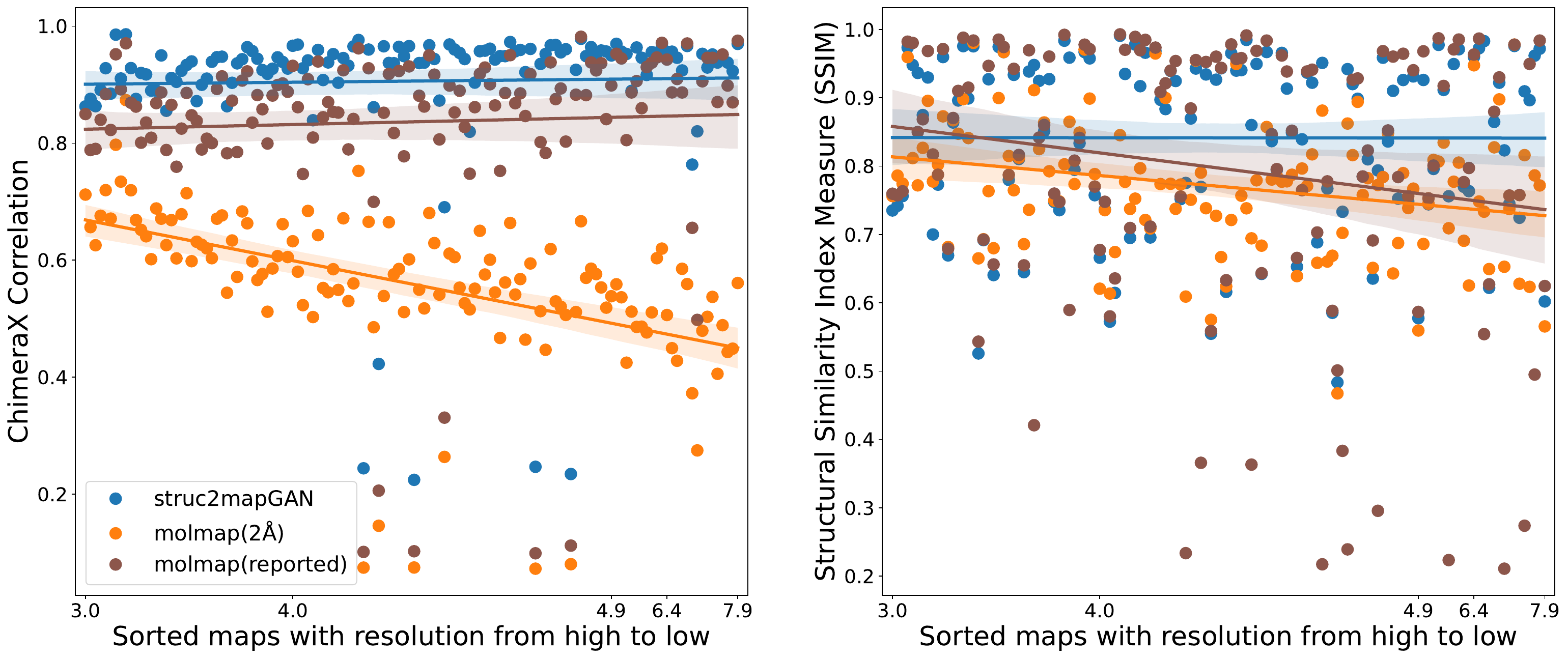}
    \caption{ 
    The scatter plots for comparison of ChimeraX correlation (left) and SSIM (right) for \emph{struc2mapGAN} (blue dots), \emph{molmap} at a resolution cutoff at 2 {\AA} (orange dots), and \emph{molmap} at the reported resolution cutoff (brown dots), across 130 test examples.
    Each point represents one test sample, and the X-axis reflects the sorted order of maps based on their reported resolution, from high (left) to low (right). The x‑tick values correspond to representative resolution values (in {\AA}) sampled from the dataset. Because many maps share identical resolution values, these tick marks are not evenly spaced along the axis.
    The shaded area around each colored regression line represents the confidence interval of the regression estimate.
    }
    \label{fig:scatter_plot}
\end{figure}

\subsection{Quantitative Comparison Over Resolution}
We then studied how the GAN-generated maps from PDB structures compare against the corresponding experimental counterpart, and how a simulation-based method like \emph{molmap} performs in comparison. For this section, we also aimed to consider the resolution of the experimental map (ranging from high (3 {\AA}) to low (7.9 {\AA})), since GAN-generated maps aim to replicate the map features at high resolution.
As depicted in Figure \ref{fig:scatter_plot}, over 95 \% of GAN-generated maps achieved high correlation scores above 0.8, irrespective of their resolution.
Surprisingly, this correlation was also constant on average across resolution levels (as shown by the fitted curve). A similar result was obtained using SSIM (see Methods).
We hypothesize that the information loss in low-resolution maps, that would result in poor placement of atoms in the PDB structure, was compensated by incorporating high-resolution training data, leading to maintaining performance in low-resolution maps.
To support this hypothesis, we performed the same comparison using \emph{molmap}-simulated maps at both high resolution of 2 {\AA}, and the same reported resolution as the experimental one. In Table \ref{tab:GAN-comparison}, we report the mean and median correlation and SSIM scores that indicate better performance of GAN-generated maps. 
As expected, simulated maps at 2 {\AA} showed poorer correlation and similarity as resolution decreases, presumably due to a mismatch between the fine atomic features emphasized in high-resolution simulated maps and the blurred, noisy nature of corresponding experimental maps—particularly at lower resolutions (see Figure \ref{fig:scatter_plot}). This mismatch reduces voxel-wise correlation despite the apparent precision of simulated maps.
Simulated maps at reported resolution would maintain constant correlation (compensating atom misplacement by a more diffuse kernel), but with less value than \emph{struc2mapGAN}, and also a decreasing trend for SSIM, supporting that GAN-generated maps can compensate for the uncertainty of atomic positions in PDB structures generated at low resolution.

\begin{figure}[!hp]
    \centering
    \includegraphics[width=\linewidth, trim={0cm 0cm 0cm 0cm}, clip]{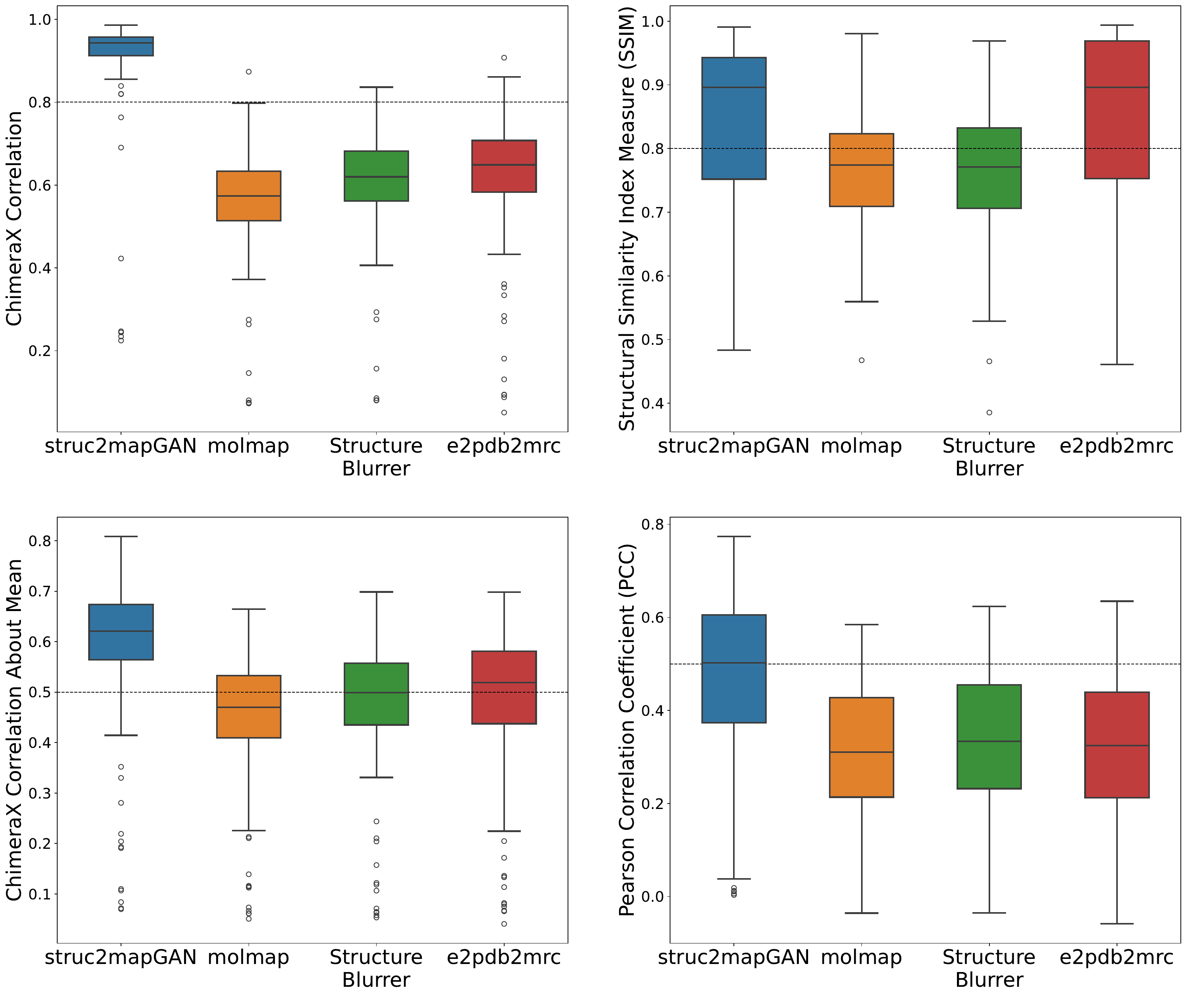}
    \caption{ 
    The box-whisker plots for comparison of different methods (\emph{struc2mapGAN}, \emph{molmap}, \emph{StructureBlurrer}, and \emph{e2pdb2map}) across four evaluation metrics over 130 test examples. For each box-whisker plot, the center line is the median; lower and upper hinges represent the first and third quartiles; the whiskers stretch to 1.5 times the interquartile range from the corresponding hinge; and the outliers are plotted as circles.
    }
    \label{fig:comp_boxplot}
\end{figure}

\subsection{Benchmarking}
We further compared \emph{struc2mapGAN} with other commonly used simulation-based methods, including \emph{StructureBlurrer} and \emph{e2pdb2mrc}, in terms of SSIM, correlation, and PCC scores across all 130 test examples.
The mean and median values of these metrics are listed in Table \ref{tab:GAN-comparison}.
According to the box-and-whisker plots shown in Figure \ref{fig:comp_boxplot}, maps generated by \emph{struc2mapGAN} consistently produced  higher scores in metrics of correlation, correlation about mean, and PCC compared to the other methods. Specifically, \emph{struc2mapGAN} achieved an average correlation of 0.906 and a PCC of 0.594, significantly surpassing the other methods. Although \emph{struc2mapGAN}'s average SSIM score was 0.841, slightly lower than \emph{e2pdb2mrc}'s 0.848, it exhibited a narrower interquartile range, indicating more consistent results. 
The high SSIM scores achieved by \emph{e2pdb2mrc} can be attributed to its effective preservation of local structural details. 
This preservation is facilitated by the method's requirement to select a larger box size ($500\text{\AA}\times500\text{\AA}\times500\text{\AA}$) than the reference for successfully generating a density map. A larger box size encompasses more spatial context around the molecule, thereby retaining more local structural details. Aside from the the SSIM score for \emph{e2pdb2mrc}, we also observe that all three simulation-based methods yield comparable results across all metrics showing their relative similarity.

\begin{figure}[!ht]
    \centering
    \includegraphics[width=\linewidth, trim={0cm 0cm 0cm 0cm}, clip]{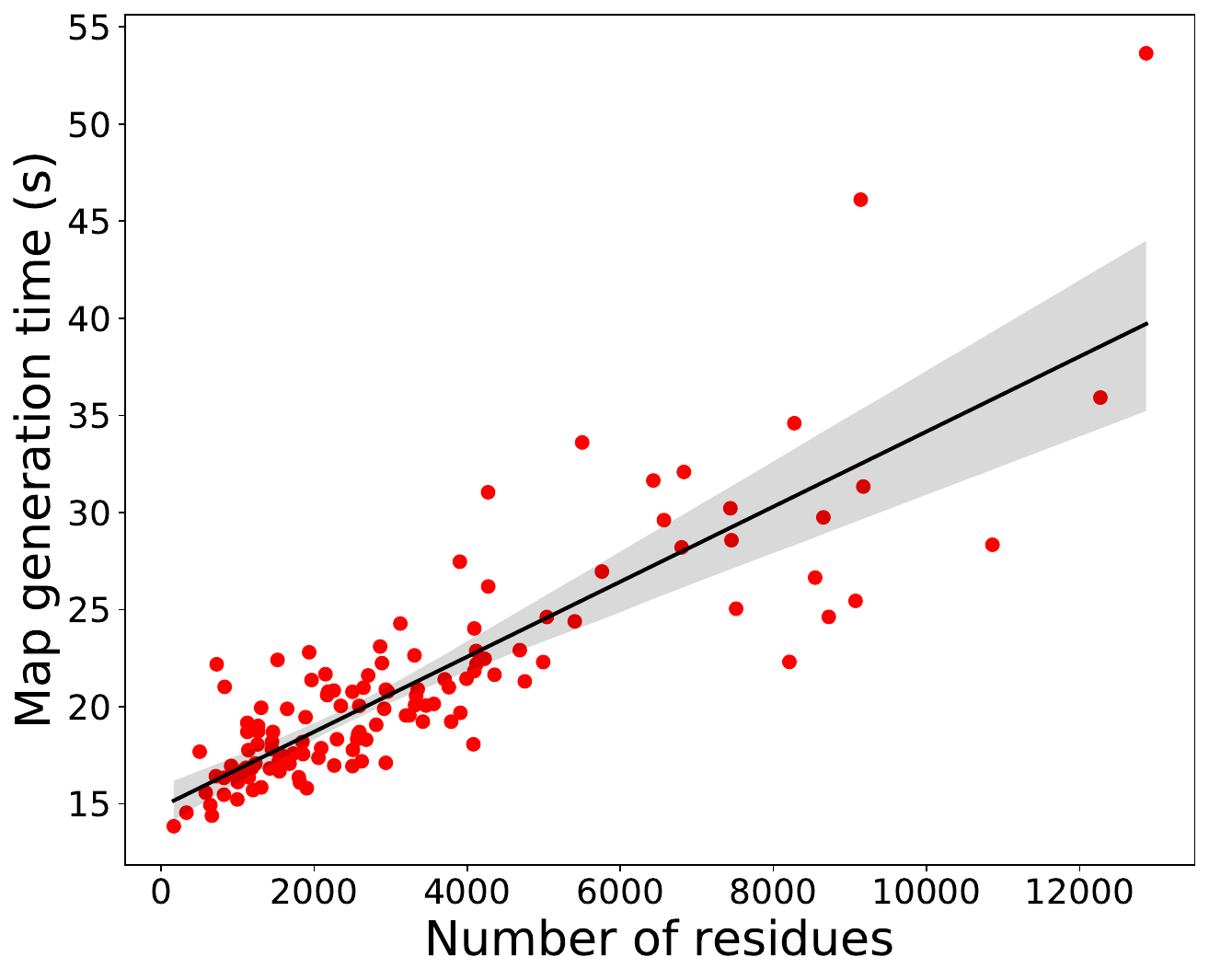}
    \caption{ 
    The scatter plot of map generation time against the number of residues in the molecule. Each red dot represents a molecule's generation based on its residue count. The shaded area around the black regression line represents the confidence interval of the regression estimate. Running times were recorded  for an AMD Ryzen Threadripper 2950X Processor with 32 CPUs.
    }
    \label{fig:inference_time}
\end{figure}

\subsection{Map Generation Time}
To assess the potential use of \emph{struc2mapGAN} in practice, we also recorded the time to generate experimental-like density maps from PDB structures for all 130 test examples. 
Figure \ref{fig:inference_time} shows the wall-clock time plotted against the number of residues of each candidate structure. The scatter plot indicates that the relationship between the map generation time and the number of residues is approximately linear. Upon reviewing several instances, we found that it took around 14 and 20 seconds to generate maps containing 644 and 2501 residues, respectively. For a significantly larger structure with 12868 residues, the map generation time remained within an acceptable range, approximately 53 seconds. These results demonstrate that \emph{struc2mapGAN} scales efficiently with the increasing complexity of protein structures, making it a viable tool for real-time applications.

\subsection{Ablation Study} \label{ablation}

\begin{figure}[!ht]
    \centering
    \includegraphics[width=\linewidth, trim={0cm 0cm 0cm 0cm}, clip]{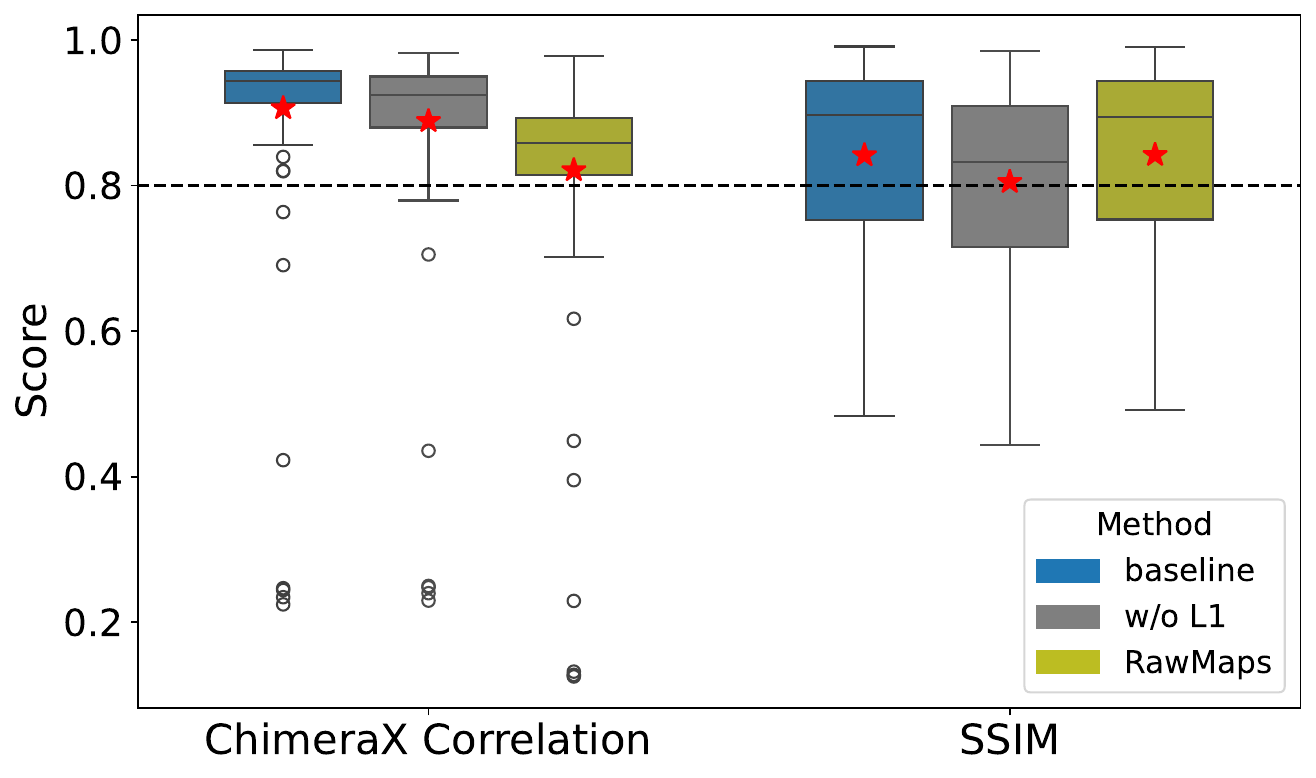}
    \caption{ 
    The box-whisker plots for comparison among \emph{struc2mapGAN} (base), the network trained without \texttt{SmoothL1Loss} (w/o L1), and the network trained with uncurated raw experimental maps (RawMaps). The evaluation was made across two metrics: ChimeraX correlation and SSIM, based on 130 test examples. 
    The red stars represent the mean values of corresponding scores.
    }
    \label{fig:ablation_boxplot}
\end{figure}

Two key factors influence the learning efficiency of \emph{struc2mapGAN}. The first is the incorporation of \texttt{SmoothL1Loss} in the generator as an additional constraint to mitigate mode collapse inherent in GANs and stabilize the training process. The second is the use of curated experimental maps as the learning targets, enabling the model to learn more accurate and reliable mappings between synthetic and experimental maps.
To investigate the impact of these factors on the performance of \emph{struc2mapGAN}, we trained two distinct GAN models: one without \texttt{SmoothL1Loss} (w/o L1) and the other using raw experimental maps without any curation. We then conducted these evaluations on both models using all 130 test examples.
ChimeraX correlation scores are reported in Figure \ref{fig:ablation_boxplot}. The baseline \emph{struc2mapGAN} (i.e., trained with \texttt{SmoothL1Loss} and using curated maps) achieved the highest mean  and median scores, as well as the narrowest interquartile range. These results indicate that incorporating \texttt{SmoothL1Loss} enhances the preservation of overall similarity in the generated maps and leads to more consistent outcomes. 
Similarly, SSIM scores are reported in Figure \ref{fig:ablation_boxplot}, indicating slightly lower mean and median without \texttt{SmoothL1Loss}. 

The model trained with only raw maps achieved SSIM scores with a mean of 0.842 and a median of 0.894, comparable to those of \emph{struc2mapGAN} (mean of 0.841 and  median of 0.896). 
This result was anticipated since the SSIM metric is more sensitive to local structural details. Training with raw maps maintained the local molecular information, resulting in minimal variance in SSIM scores. 
However, most noise, artifacts, and solvents were removed when raw maps were curated, which allows \emph{struc2mapGAN} trained with these curated maps to focus more on overall similarity, thereby yielding higher correlation scores. 
These findings underscore the importance of using an additional \texttt{SmoothL1Loss} to guide network training and employing curated maps as input targets, both of which enhancing the model performance.

\section{Discussion and Future Work}

In this chapter, we present \emph{struc2mapGAN}, a novel method adapted from the GAN architecture to generate cryo-EM density maps from PDB structures, which mimic the unique characteristics of raw maps.  
We enhance \emph{struc2mapGAN}'s training efficiency by excluding the noise and artifacts and selectively extracting the macromolecule regions from the raw maps as training input, as well as incorporating \texttt{SmoothL1Loss} into the generator.
Our benchmarking results show that \emph{struc2mapGAN} outperforms existing simulation-based methods across various evaluation metrics.
Moreover, its rapid synthesis speed makes it suitable for generating large-scale data.

Recent works have made significant progress in simulating realistic 2D cryo-EM images (micrographs) for tasks such as particle picking or pose estimation \cite{cryoGEM, cryoTEM,Topaz-Denoise,abTEM,imageformation,egg}. These methods model the physics of electron scattering and the imaging process to produce authentic 2D micrographs. In contrast, our work focuses on synthesizing 3D cryo-EM density maps from atomic models, serving as volumetric representations of molecular structures that directly used in downstream tasks such as rigid-body fitting and particle picking.
While 2D methods aim to simulate the imaging process, our 3D simulation is designed to mimic the structural characteristics of experimental maps. These two directions are complementary. 2D simulations improve image-level analysis, whereas 3D simulations enhance volumetric representations. In addition, an interesting future direction would be to systematically evaluate how both 2D and 3D simulation strategies contribute to improving particle picking.

Overall, by employing \emph{struc2mapGAN}, structural biologists can synthesize improved exp\-erimental-like cryo-EM density maps in a timely manner, which can then be used for various applications, such as guiding the particle-picking process before reconstruction of experimental maps and assisting rigid-body fitting with use of simulated maps. Furthermore, machine learning scientists can use \emph{struc2mapGAN} to generate numerous experimental-like density maps as targets, enabling the re-training or fine-tuning of the existing models that were originally trained with simulation-based maps, thereby improving the model performance. Although these generated maps can be useful in early-stage analysis or as supplementary inputs, caution must be exercised when using them for validation, particularly when the underlying atomic models are predicted by AlphaFold or derived from low-resolution data.

While our results demonstrate superior performance in generating improved density maps, it is not possible to relate these maps to an exact resolution value. In principle, it would be feasible to precisely modulate resolutions by integrating a resolution-conditioned modulation mechanism into the generator. This could involve conditioning the network on a continuous resolution parameter via learnable embedding layers or feature-wise linear modulation (FiLM) \cite{film}, allowing explicit control over output resolution.
Another promising direction would be to incorporate local resolution information during both training and evaluation. Currently, our model assumes uniform resolution across all voxels. However, in practice, experimental maps often exhibit spatially varying resolution. Future extensions of \emph{struc2mapGAN} could integrate voxel-wise weighting schemes based on local resolution estimates, allowing the model to focus more on high-confidence regions. Similarly, evaluation metrics like correlation and SSIM could be adapted to downweight low-resolution regions, potentially yielding a more biologically meaningful assessment of model performance.
Additionally, developing SSE-specific quantitative evaluation metrics would enable more targeted assessment of structural features, but remains challenging due to the lack of suitable tools. 
Furthermore, with advancements in other generative models \cite{transformer,diffusion}, integrating attention modules into the generator of \emph{struc2mapGAN} could enhance capturing detailed structural information from maps. It would also be interesting to utilize diffusion models for map generation. These will be left for our future work.

Interestingly, our method can also be employed to generate reliable density maps from structures predicted by AlphaFold \cite{alphafold2} and other de novo sequence-to-structure generative methods \cite{rosettafold,esm2}, serving as templates to guide and expedite the particle-picking process.
For instance, given an AlphaFold-predicted structure, practitioners can generate a 3D density map from its atomic model and extract 2D templates by slicing through the volume. These templates can then be used by particle-picking software to more accurately identify protein particles out of micrographs. While Gaussian blobs can also guide particle picking, template-based approaches are generally more effective, as they are theoretically closer to the actual particles \cite{particlepick,recons}. Moreover, these templates have the potential for template matching in cryo-electron tomography \cite{cryoet}, where consistent SSE representation can improve particle identification. Exploring these applications will also be the focus of our future work.

\chapter{CryoSAMU: Enhancing Cryo-EM Density Maps of Protein Structures at Intermediate Resolution with Structure-Aware Multimodal U-Nets} \label{chap:enhancemap}

This chapter is a modified version of a paper by Chenwei Zhang, et al., published in ICML 2025, Generative AI and Biology (GenBio) Workshop (\url{https://openreview.net/forum?id=ga3iu8kdTq}) \cite{cryosamu}.

\section{Motivations and Contributions}

Most existing methods for full-atom model building are tailored for near atomic-resolution density maps, despite the prevalence of lower resolution maps. These methods predominantly employ maps with resolutions finer than 4 {\AA} for both training and evaluation, resulting in a shortfall in their effectiveness at lower resolutions.
To leverage these methods for intermediate-resolution maps, it is crucial to develop techniques that enhance and sharpen these maps, rendering them suitable for model-building approaches originally conceived for high-resolution maps.

Traditional methods rely on B-factor correction, which can be applied globally \cite{phenixautosharpen,relionpostprocess} and locally \cite{LocSpiral,LocalDeblur}. However, these methods struggle with maps exhibiting varying signal-to-noise ratios and lacking prior knowledge (e.g., local resolution) \cite{emready}.
With recent advancements in deep learning (DL), fully data-driven methods have been developed to automatically enhance raw cryo-EM maps for protein structure modeling. Leveraging neural networks such as convolutional neural networks (CNNs) \cite{cnn}, generative adversarial networks (GANs) \cite{gan}, and Transformers \cite{transformer}, these methods \cite{deepemhancer,cryofem,emgan,emready,cryoten} achieved promising results in map enhancement. Yet, they are not optimized for intermediate-resolution maps (i.e., 4-8 {\AA} \cite{emready}) and rely solely on a single modality, i.e., the density map itself, during neural network training, overlooking other relevant modalities such as structural information. This limitation restricts their ability to generalize across diverse protein structures and prevents them from fully leveraging complementary biological information. To address these shortcomings, we thus introduce \textbf{CryoSAMU}, a novel approach that combines 3D map features with structural embeddings derived from the pretrained protein language model ESM-IF1 \cite{esmif} to enhance 3D \textbf{Cryo}-EM density maps with \textbf{S}tructure-\textbf{A}ware \textbf{M}ultimodal \textbf{U}-Nets.

Our main contributions are:
\begin{itemize}\setlength{\itemsep}{0pt}
    \item We propose the first multimodal network that integrates structural information into a 3D U-Net model using cross-attention mechanisms for cryo-EM map enhancement.
    \item We develop a self-attention-based post-processing procedure for ESM-IF1's structural embeddings, effectively preserving both chain and residue relationships while maintaining structural integrity.
    \item We train CryoSAMU on a curated dataset of joint density maps at intermediate resolution and associated protein structures, optimizing it for map enhancement.
    \item We benchmark CryoSAMU against state-of-the-art approaches across various evaluation metrics over diverse tested samples. We achieve competitive level of performance but with significantly faster processing speeds (approximately 4.2 to 16.7 times), making our method well-suited for large-scale and practical applications.
    \item Our ablation study demonstrates significant improvement brought from integrating structural information.
\end{itemize}

\section{Dataset Preparation}

Our dataset was built with a set of cryo-EM density maps at resolutions from 4.0 {\AA} to 7.9 {\AA} from the EMDB databank \cite{EMDB} and their associated protein structures from the PDB databank \cite{PDB}. 
To ensure that density maps are properly aligned with their corresponding PDB structures, we excluded maps and PDBs from the dataset if:
(i) maps contain extensive regions without or misaligned corresponding PDB structures;
(ii) maps contain other macromolecules except proteins;    
(iii) PDB structures contain backbone atoms only and/or unknown residues.
Furthermore, to enhance training efficiency, we measured the correlation between map-PDB pairs using ChimeraX \cite{chimerax}, and removed pairs with correlation score lower than 0.65 to filter out incomplete mappings. To avoid data redundancy, we measured the sequence identity between PDB structures, and retained only one if identity is greater than 30 \%.
As a result, a total of 384 pairs of cryo-EM maps and associated PDB structures remained. Among these data, 247 ($\sim$65 \%), 62 ($\sim$15 \%), and 75 ($\sim$20 \%) map-PDB pairs were selected as training, validation, and test sets, respectively. Details are listed in Supplementary Tables \ref{tab:cryo_train_data}, \ref{tab:cryo_val_data}, and \ref{tab:cryo_test_data}.

\section{Multimodal Representations of Protein Structures}

\paragraph{Generating 3D target maps from protein structures}
For input experimental maps (denoted as \emph{ExpMaps}) in training and validation sets, we simulated the corresponding target maps (denoted as \emph{TgtMaps}) from associated protein structures using the \texttt{Structure\-Blurrer} package in TEMPy2 \cite{tempy2}, in terms of Equation \ref{eq:simulation}. The simulation was performed with a grid interval of 1 {\AA} and a resolution cutoff at 2 {\AA}, based on the convolution of atom points with resolution-lowering point spread functions. 

\begin{figure}[!p]
  \centering
   \includegraphics[width=\linewidth, trim={0.6cm 14.6cm 3.2cm 0.8cm}, clip]{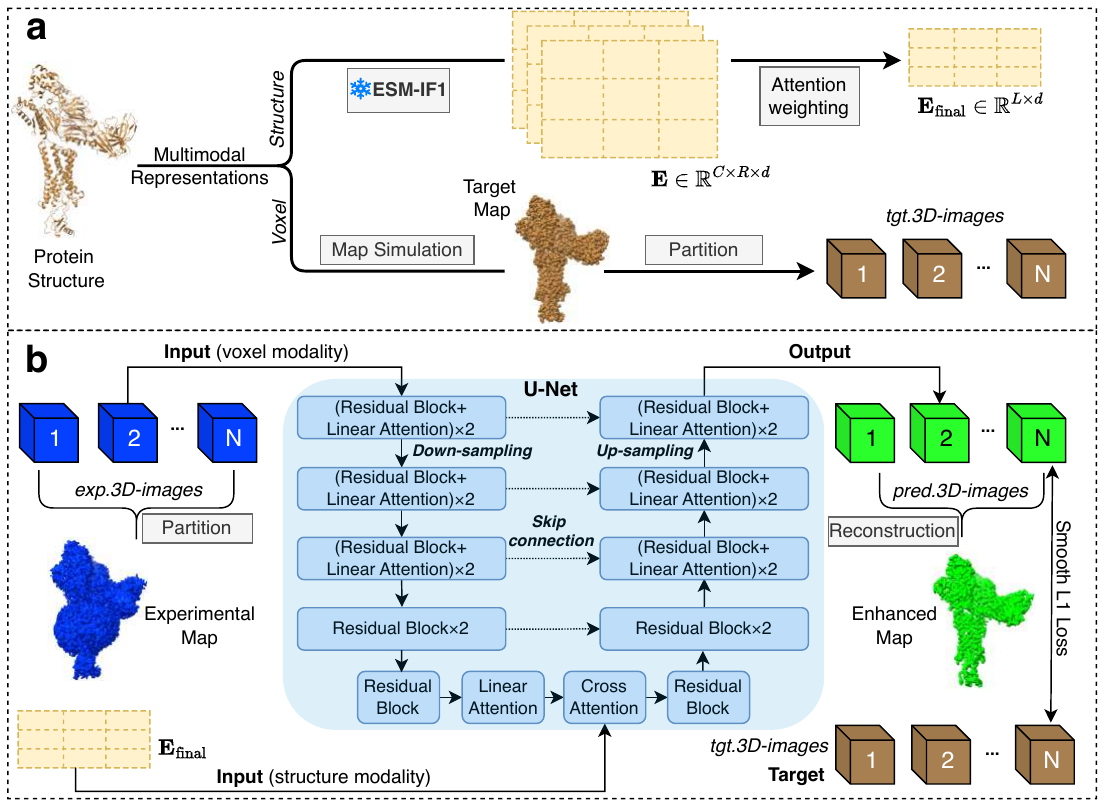}
   \caption{Overview of the CryoSAMU framework.
   \textbf{a} Generating protein multimodal representations: structure features are derived from a frozen pretrained ESM-IF1 model with self-attention weighting for a fixed-size representation; map voxel features are simulated via resolution-lowering point spread function and partitioned into smaller cubes. 
   \textbf{b} The CryoSAMU architecture. The experimental map is partitioned into smaller cubes and processed by a U-Net with residual blocks and linear attention modules. Structural embeddings are integrated into the bottleneck layer with cross-attention mechanism. The output cubes are reconstructed into the full-size enhanced map.
   The illustrated protein complex is a CX3CL1-US28-G11iN18-scFv16 in TL-state (PDB-7RKF, EMDB-24496, reported resolution of 4.00 {\AA}) \cite{7RKF}. 
   }
   \label{fig:cryosamu_model}
\end{figure}

\paragraph{Resampling 3D maps}
We first resampled both \emph{ExpMaps} and \emph{TgtMaps} to 1 {\AA}/voxel since the cryo-EM maps vary in voxel size. Subsequently, we normalized the density values to a range of 0 to 1 using the 99.9th percentile density value of each map. Due to GPU memory constraints, we partitioned \emph{ExpMaps} and \emph{TgtMaps} into smaller 3D subvolume pairs (denoted as \emph{exp.3D-images} and \emph{tgt.3D-images}) with size of $64\times64\times64$, the largest feasible size that allows for a sufficient batch size (See Figure \ref{fig:cryosamu_model}a.). To mitigate boundary artifacts during truncation, we applied zero-padding of 64 voxels on each side along all dimensions. As a result, a total of 29829 \emph{exp-tgt} image pairs were yielded for network training and 4642 for validation.

\paragraph{Generating structural embeddings}
We employed ESM-IF1 \cite{esmif} to generate protein structural embeddings, which will serve as an additional modality for network training. Specifically, we derived the embeddings by first extracting backbone coordinates (N, C$\alpha$, and C atoms) from a PDB file, ensuring that only standard residues with complete backbone information are included. We then fed these coordinates into ESM-IF1 to generate embeddings for each protein chain. Since the lengths of chains varied, we applied zero-padding to standardize the embeddings.

\paragraph{Fixed-size representation with attention weighting}
Following the generation of embeddings, we implemented self-attention weighting to create fixed-size representations while preserving the intrinsic relationships between chains and residues. To this end, we computed attention weights based on embedding similarity to identify the most informative regions. Specifically, given a PDB structure containing $C$ chains and $R$ residues per chain, its structural embedding derived from ESM-IF1 is denoted as $\mathbf{E} \in \mathbb{R}^{C\times R \times d}$, where $d=512$ is the embedding dimension. 
We carried out the refinement process in several steps. First, we computed chain-level embeddings by averaging across residues:
\begin{equation}
    \mathbf{E}_{\text{chain}} = \frac{1}{R} \sum_{j=1}^{R} \mathbf{E}_{:, j, :}, \quad \mathbf{E}_{\text{chain}} \in \mathbb{R}^{C \times d}.
\end{equation}
Next, we computed the similarity matrix to determine the relative importance of each chain:
\begin{equation}
    \mathbf{S} = \mathbf{E}_{\text{chain}} \cdot \mathbf{E}_{\text{chain}}^T, \quad \mathbf{S} \in \mathbb{R}^{C \times C},
\end{equation}
where each element $\mathbf{S}_{ij}$ represents the similarity between chain $i$ and chain $j$: $\mathbf{S}_{ij} = \mathbf{E}_{\text{chain},i} \cdot \mathbf{E}_{\text{chain},j}^T$.
To derive attention weights, we applied a column-wise softmax function to $\mathbf{S}$: 
\begin{equation}
    \mathbf{W}_{ij} = \frac{\exp(\mathbf{S}_{ij})}{\sum_{k=1}^{C} \exp(\mathbf{S}_{ik})}.
\end{equation}
We then aggregated these weights across chains to assign a single importance weight per chain:
\begin{equation}
    \mathbf{w}_i = \frac{1}{C} \sum_{j}^{C} \mathbf{W}_{ij}, \quad i = 1, 2, \dots, C.
\end{equation}
These weights $\mathbf{w} = [\mathbf{w}_1, \mathbf{w}_2, .., \mathbf{w}_C]$ reflect the relative importance of each chain, and we leveraged them to aggregate chain-level embeddings into a unified representation: 
\begin{equation}
    \mathbf{E}_{\text{pooled}} = \sum_{i=1}^{C} w_i \mathbf{E}_{i,:,:}, \quad \mathbf{E}_{\text{pooled}} \in \mathbb{R}^{R \times d}.
\end{equation}
We further measured the importance of each residue in $\mathbf{E}_{\text{pooled}}$ using a residue-level similarity matrix. Following the same procedure as the chain-level weighting, we obtained a scalar weight $\alpha_j$ for each residue $j$, where $j=1, 2, \dots, R$.
Finally, we applied min-max normalization and resampled the embedding $\mathbf{E}_{\text{pooled}}$ based on the attention weights to a fixed-size representation, $\mathbf{E}_{\text{final}} \in \mathbb{R}^{L \times d}$, where $L=800$. When the input length $R > L$, we selected the top-$L$ residues with the highest attention weights. Conversely, when $R < L$, we sorted the residues by their attention weights and repeated them $\lceil L/R \rceil$ times to reach the target length, ensuring each resulting embedding maintains rich representations and consistent dimensions.
Compared to simple averaging or top-L residue selection, our hierarchical attention strategy preserves the distinct contributions of each residue through explicit weighting, as well as emphasize biologically informative patterns across diverse protein complexes.

\section{The CryoSAMU Model}

\subsection{Architecture}

We proposed a structure-aware multimodal 3D U-Net, as depicted in Figure \ref{fig:cryosamu_model}b. The network contains an encoder, bottleneck, and decoder, interconnected by skip connections.

\paragraph{Encoder}
The input to the encoder is a 3D volume with a single channel. The encoder comprises four hierarchical layers. The first three layers each consist of two residual blocks, with each block incorporating a group normalization, a SiLU activation \cite{silu}, and a dropout (p=0.2), followed immediately by a linear self-attention module with 4 heads \cite{linearattention} to capture long-range (global) dependencies across voxels. The channel depth progressively increases as features are abstracted. In the fourth layer, only residual blocks are employed, producing a higher-level feature representation without the addition of attention modules.

\paragraph{Bottleneck}
At the bottleneck layer, the feature representation is first refined by a residue block and then by a linear self-attention module. Subsequently, a cross-attention block is introduced to fuse and align the volumetric features with structural embeddings using multi-head attention with 4 heads, where queries are derived from the volume features and keys/values from structural embeddings. This process enables structural conditioning while preserving spatial relationships. A second residual block is then applied to further fuse the combined features from both modalities. 

\paragraph{Decoder}
The decoder follows a symmetric architecture to the encoder. Feature maps are progressively upsampled using nearest-neighbour interpolation combined with 3D convolutions, and skip connections incorporate corresponding features from the encoder. Finally, a group normalization, a SiLU activation, and a concluding 3D convolution project the processed features to a single output channel.

\subsection{Network Training and Inference}
Protein structural embeddings provide an additional modality containing structure information, serving as key-value pairs in the attention mechanism when training. However, since these embeddings are unavailable during validation and inference, we implemented a specialized mode in which the network bypasses the cross-attention operation. 
In this mode, the network relies exclusively on feedforward transformations with residual connections. 
This design aligns with the well-established principle of conditional training, where auxiliary modalities, despite not available at inference, are used during training to improve the quality and generalization of learned representations from the primary input \cite{auxiliary}.
Moreover, this design also maintains consistency between training and validation/inference phases while preserving the feature representations.

\paragraph{Training}
During training, CryoSAMU accepts an \emph{exp.3D-image} and its corresponding structural embedding as input and generates an enhanced 3D image (denoted as \emph{pred.3D-image}). Previous studies have shown that L1 loss performs well in similar tasks \cite{deepemhancer, cryofem}. However, to improve training stability in the presence of noisy data and outliers that are common in cryo-EM maps, we employed the smooth L1 loss to encourage the generator to minimize the difference between the output \emph{pred.3D-image} and the target \emph{tgt.3D-image}, as introduced in Equation \ref{eq:smoothl1loss}.
Moreover, to enhance the network's robustness, we employed \emph{TorchIO} \cite{torchio} for data augmentation, including random Gaussian noise, anisotropy, and blurring.

\paragraph{Inference}
During inference, the input experimental map was first zero-padded and divided into smaller cubes ($64\times64\times64$), following the same strategy used for training data. Each cube was then individually processed by the trained neural network to generate the enhanced cube. These enhanced cubes were subsequently reassembled to reconstruct the map as its original dimensions. To prevent loss of spatial information and ensure smooth transition between cubes, only the central $50\times50\times50$ voxels from each enhanced cube were used in the final reconstruction, following the method proposed by Si et al. \cite{Cascaded-CNN}.

\paragraph{Implementation}
The network was implemented in PyTorch 2.6.0 with CUDA 12.4, running under Python 3.12.8. Training was conducted using a distributed data parallel (DDP) strategy across two computational nodes connected via NVLink, with each node equipped with four NVIDIA A100 GPUs of 40 GB VRAM. This setup supported a maximum batch size of 18 per GPU. The network was trained over 95 epochs, requiring approximately 63 computational hours. The AdamW \cite{AdamW} optimizer was used with an initial learning rate of 0.0001, along with a cosine annealing learning rate scheduler. To improve training performance while maintaining accuracy, automatic mixed precision training was applied. Additionally, gradient clipping (set to 0.5) was applied to prevent gradient explosion.

\section{Experiments and Results}
We conducted a comprehensive study to assess the performance of CryoSAMU using a test set of 75 intermediate-resolution cryo-EM density maps and associated PDB structures across a wide range of evaluation metrics.

\subsection{Visualization and Quantification of Map Enhancement}
We first visualized a CryoSAMU-enhanced map alongside its associated deposited map using UCSF ChimeraX \cite{chimerax}. For a fair comparison, both sets of maps were illustrated with the same volume, which requires contour level adjustments owing to differences in their volume ranges. Specifically, we first presented the deposited map at its recommended contour level and volume, then adjusted the contour level of the corresponding CryoSAMU-enhanced map to match the recommended volume. In addition, we also visualized both maps at a higher contour level with the same volume. 

\begin{figure}[!ht]
  \centering
   \includegraphics[width=\linewidth, trim={0.5cm 19.2cm 4.5cm 0.5cm}, clip]{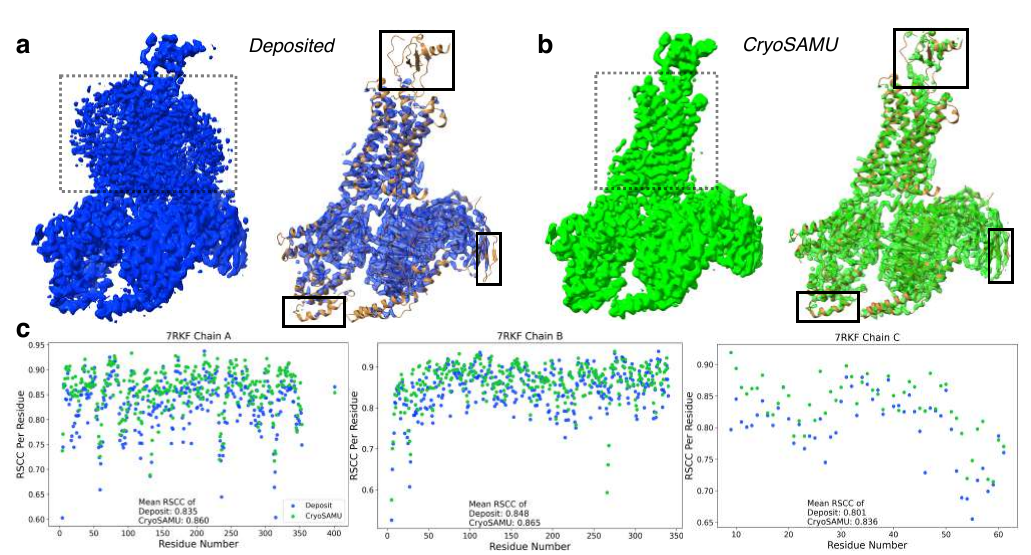}
   \caption{Visual and quantitative comparison of deposited (blue) and CryoSAMU-enhanced (green) maps, with superimposed corresponding PDB structures (brown).
   \textbf{a, b}: Maps are shown at two contour levels. Left: recommended contour level (volume = 85.74e3). Right: higher contour level (volume = 22.57e3).
   \textbf{c}: RSCC comparisons between deposited and CryoSAMU-enhanced maps.
   The example protein is a CX3CL1-US28-G11iN18-scFv16 in TL-state (PDB-7RKF, EMDB-24496, reported resolution of 4.00 {\AA}) \cite{7RKF}. 
   }
   \label{fig:vis_comp}
\end{figure}

As displayed in Figure \ref{fig:vis_comp}b, CryoSAMU significantly suppressed noise in the lip nanodisc regions (highlighted by dashed boxes in Figure \ref{fig:vis_comp}a) of the deposited map for EMDB-24496 (PDB-7RKF). Moreover, the deposited map at a smaller volumes missed certain structural regions corresponding to the protein structures, as highlighted by black boxes in Figure \ref{fig:vis_comp}a. In contrast, the CryoSAMU-enhanced maps exhibited better alignment with the corresponding protein structures, revealing more structural details, as demonstrated by black boxes in Figure \ref{fig:vis_comp}b.
Similar visual results were observed for another protein structure (see Supplementary Figure \ref{fig:CryoSAMU_SI_vis_comp}). 
Furthermore, residue-level real-space correlation coefficient (RSCC) measurements \cite{phenixeval} in Figure \ref{fig:vis_comp}c suggested significant improvements. Specifically, Chains A, B, and C exhibit RSCC increases compared to the deposited map, with correlations rising from 0.835 to 0.860, 0.848 to 0.865, and 0.801 to 0.836, respectively. In addition, 84.9\%, 73.5\%, and 90.6\% of resiudes in Chains A, B, and C, respectively, showcased higher RSCC scores. Consistent RSCC improvements were also observed in other samples (see Supplementary Figure \ref{fig:CryoSAMU_SI_plot_RSCC}).

\subsection{Benchmark I: Improvement of Real and Fourier Space Correlations} \label{bm1}
We then benchmarked CryoSAMU against other state-of-the-art methods, including Autosharpen \cite{phenixautosharpen}, DeepEMhancer \cite{deepemhancer}, and EMReady \cite{emready}, in terms of both real-space and reciprocal-space (i.e., Fourier-space) correlations, across a test set of 75 primary maps. 
For real-space correlation, we computed three correlation metrics using \texttt{phenix.map\_model\_cc} \cite{phenixeval} for each map-model pair (where the model refers to a protein structure): CC\_box, CC\_volume, and CC\_peaks. These metrics differ based on the choice of map regions used in the calculations.
CC\_box considers the entire map. CC\_volume and CC\_peaks focus on regions with the highest density values. However, CC\_volume selects grid points only around atomic centers, while CC\_peaks selects points located anywhere within the volume. For all three metrics, higher values indicate better map performance.
For Fourier-space correlation, we computed Fourier shell correlation (FSC) using \texttt{phenix.mtriage} \cite{phenixeval}, and reported the unmasked map-model FSC05 values. FSC values are typically represented as a function of the inverse map resolution, where lower value indicates better map resolution.

\begin{table}[!t]
\caption{Comparison of different methods across various metrics.}
\label{tab:benchmark}
\resizebox{\textwidth}{!}{
\begin{tabular}{lcccccc}
\toprule
Metric & Deposit & Autosharpen & DeepEMhancer & EMReady &\textbf{CryoSAMU}(ours) & \begin{tabular}[c]{@{}c@{}}CryoSAMU\\(w/o struct.)\end{tabular} \\
\hline
CC\_box $\uparrow$ & 0.731 & 0.679 & 0.618 & \textbf{0.862} & 0.834 & 0.751 \\
CC\_peaks $\uparrow$ & 0.750 & 0.722 & 0.611 & \textbf{0.774} & 0.753 & 0.698 \\
CC\_volume $\uparrow$ & 0.594 & 0.542 & 0.534 & \textbf{0.729} & 0.691 & 0.571 \\
FSC05 $\downarrow$ & 6.124 & 6.147 & 5.283 & \textbf{4.668} & 5.108 & 6.434 \\
\bottomrule
\end{tabular}
}
\end{table}

\begin{figure}[!hp]
  \centering
  \includegraphics[width=\linewidth, trim={0cm 0cm 65.5cm 0cm}, clip]{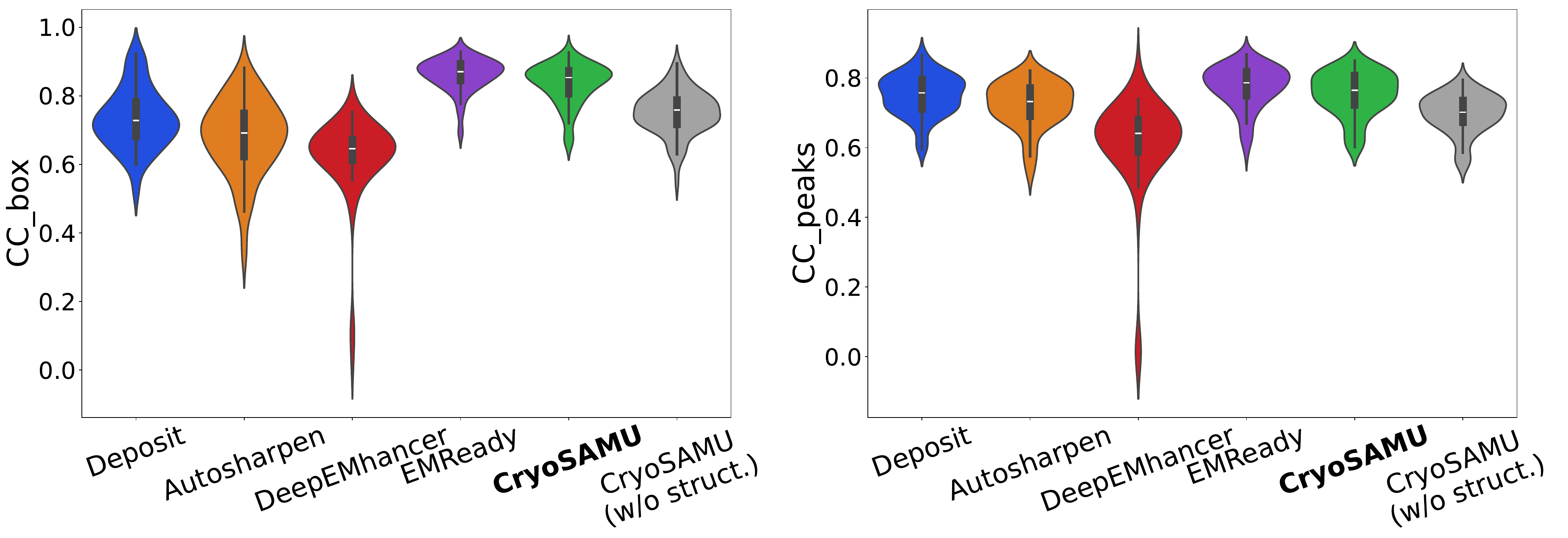}
  \includegraphics[width=\linewidth, trim={65.5cm 0cm 0cm 0cm}, clip]{body_chapters/chap_8/figs/plot_benchmark.pdf}
  \caption{The violin plots for comparison of different methods across four evaluation metrics over 75 test samples.}
   \label{fig:benchmark}
\end{figure}

The average real-space CC and FSC values are listed in Table \ref{tab:benchmark}. 
According to the violin plots shown in Figure \ref{fig:benchmark}, CryoSAMU-enhanced maps demonstrated significant improvements over the deposited maps in terms of CC\_box and CC\_volume, with average values increasing from 0.731 to 0.834 and from 0.594 to 0.691, respectively. The average CC\_peaks score showed a slightly increase from 0.750 to 0.753. These results indicate that CryoSAMU effectively enhances deposited maps in both the entire region and the highest-density regions. In contrast, maps processed by Autosharpen and DeepEMhancer exhibited lower scores across all three metrics. EMReady showed slightly better improvements than CryoSAMU across all three metrics.
For FSC05 scores, CryoSAMU outperformed the deposited map, Autosharpen, and DeepEMhancer, achieving an average value of 5.108 {\AA}. However, it slightly underperformed compared to EMReady, which achieved an average value of 4.668 {\AA}. 
These results demonstrate that both CryoSAMU and EMReady consistently enhance the deposited maps in terms of correlations in both real and Fourier spaces.

\subsection{Benchmark II: Improvement of Protein Structure Modeling} \label{sec:bm2}
As the goal of enhancing cryo-EM density maps is to improve the performance of protein structure modeling from density maps (i.e., map interpretability), we benchmarked protein structures constructed  from CryoSAMU-enhanced maps against those processed by other methods.
Specifically, we used a standard structure modeling tool, known as \texttt{phenix.map\_to\_model} \cite{phenix}, to construct protein structures from 20 maps enhanced by the different tested methods. These maps were randomly selected from the test dataset to ensure that they were not exposed during training, as listed in Supplementary Table \ref{tab:cryo_test_data_m2m}. 
To evaluate these structures, we used \texttt{phenix.chain\_comparison} \cite{phenix} to compare the constructed structures against their corresponding ground-truth PDB protein structures. We reported two metrics: residue coverage and sequence match.
The residue coverage indicates the fraction of residues in the query structure that match the corresponding residues in the target structure within 3.0 {\AA}, regardless of residue type. The sequence match indicates the percentage of matched residues that share identical residue types between the query and target structures.

\begin{table}[!ht]
\caption{Comparison of average residue coverage and sequence match across different methods.}
\label{tab:m2m}
\begin{center}
\begin{tabular}{lcc}
\toprule
Method & \begin{tabular}[c]{@{}c@{}}Residue\\Coverage(\%) $\uparrow$ \end{tabular} & \begin{tabular}[c]{@{}c@{}}Sequence\\Match(\%) $\uparrow$ \end{tabular} \\
\hline
Deposit & 31.71 & 8.42 \\
Autosharpen & 16.00 & 8.13 \\
DeepEMhancer & 24.31 & 10.0 \\
EMReady & 31.61 & \textbf{11.38} \\
\textbf{CryoSAMU}(ours) & \textbf{38.03} & 9.33 \\
CryoSAMU(w/o struct.) & 8.08 & 8.13 \\
\bottomrule
\end{tabular}
\end{center}
\end{table}

\begin{figure}[!hp]
  \centering
   \includegraphics[width=\linewidth, trim={0.5cm 18cm 9cm 0.8cm}, clip]{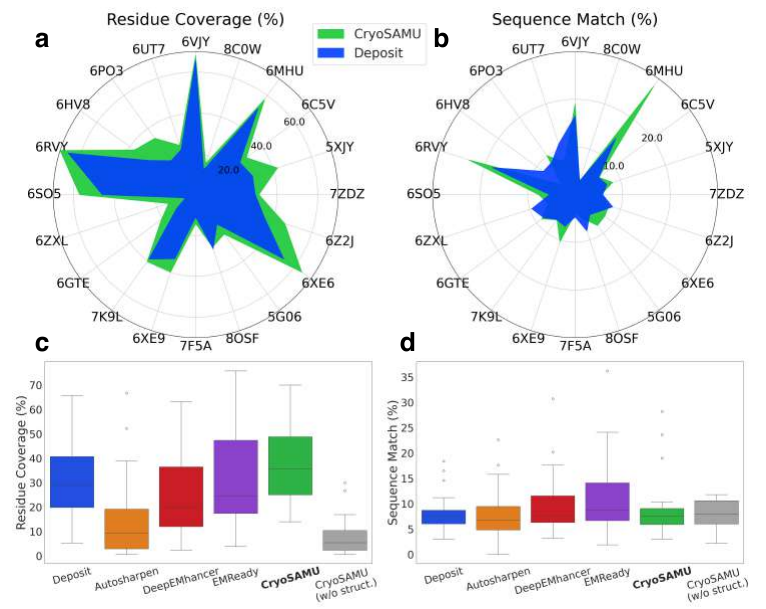}
   \caption{
   \textbf{a-b}: The polar plots for comparison of protein structures constructed from deposited (blue) and CryoSAMU-enhanced (green) maps, using metrics of (\textbf{a}) residue coverage and (\textbf{b}) sequence match.
   \textbf{c-d}: The box-whisker plots for comparison of different methods across two evaluation metrics over 20 test samples.
   }
   \label{fig:fig_m2m}
\end{figure}

The average metric scores from all methods are listed in Table \ref{tab:m2m}.
Figure \ref{fig:fig_m2m}a and b provide a detailed comparison for each individual test examples in terms of residue coverage and sequence match, respectively. The polar plots clearly showcase that after CryoSAMU enhancement, 19 out of 20 samples exhibited an improvement in residue coverage on deposited maps, with the average score increasing from 31.71\% to 38.03\%; and 55\% of samples exhibited an improvement in sequence match, with the average score increasing from  8.42\% to 9.33\%. 
Furthermore, we benchmarked CryoSAMU against other methods, as shown in Figure \ref{fig:fig_m2m}c and d. CryoSAMU achieved the highest residue coverage score among all methods, although its correlation scores were slightly lower than those of EMReady. The sequence match score of CryoSAMU was slightly lower than EMReady and DeepEMHancer, while still better than the deposited maps.
We hypothesize that this reflects a trade-off between local voxel agreement that captured by correlation metrics and model-building capability that captured by residue coverage. The integration of structural embeddings in CryoSAMU may help generate more continuous and interpretable densities, facilitating atomic model reconstruction even if point-to-point correlations are slightly reduced.
These results demonstrate that CryoSAMU enhancement boosts protein structure modeling performance.

\subsection{Benchmark III: Processing Time} \label{bm3}
\begin{figure}[!ht]
  \centering
   \includegraphics[width=\linewidth, trim={0cm 0cm 0cm 0cm}, clip]{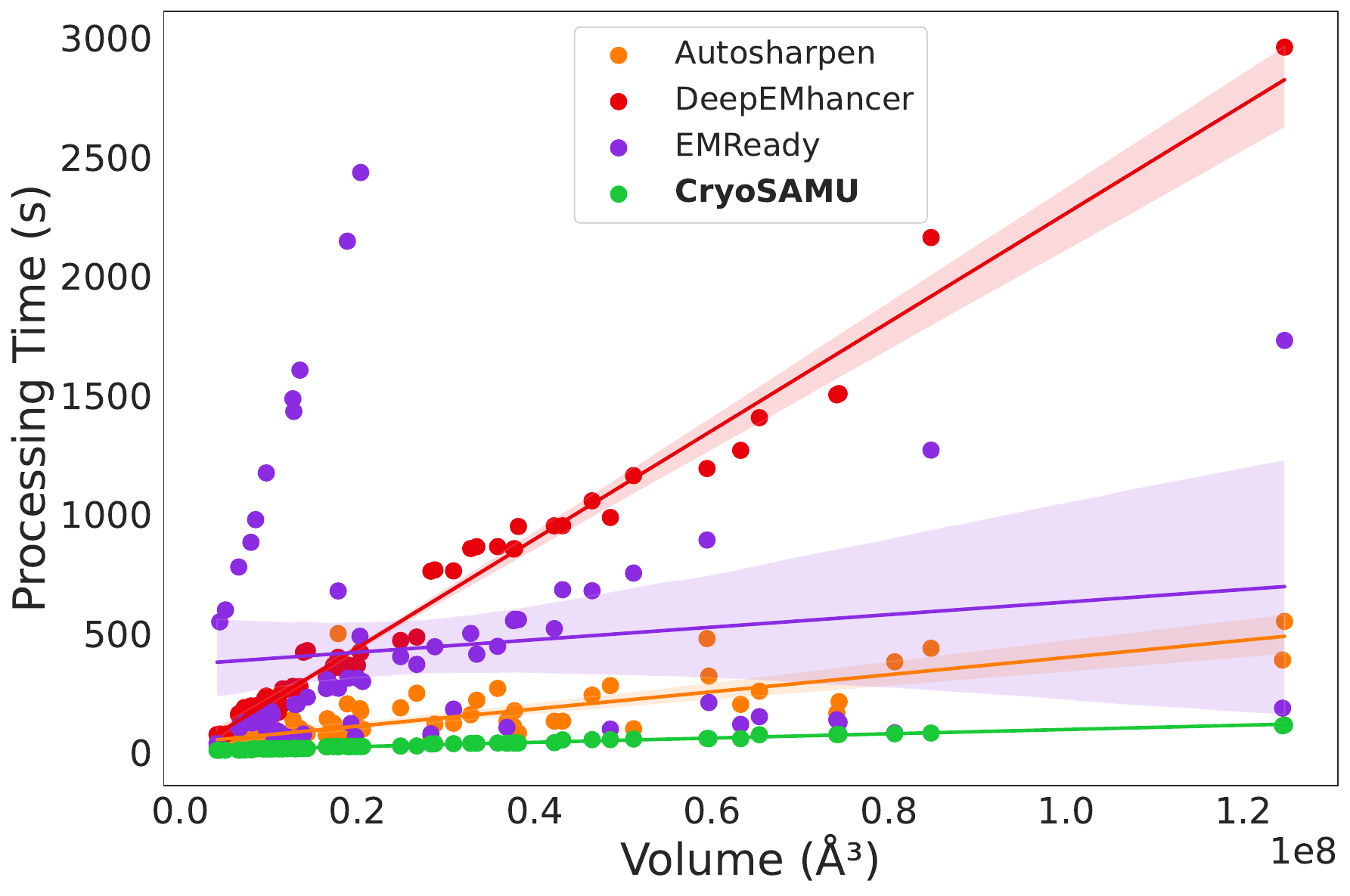}
   \caption{The scatter plot of map processing time against map volume. Each dot represents the processing time for an individual map based on its volume. The shaded area around the regression line denotes the confidence interval of the regression estimate.}
   \label{fig:time}
\end{figure}

To evaluate the scalability of CryoSAMU in practice, we recorded the time required to generate each enhanced map of all 75 test samples and compared it against the processing time of other methods. Figure \ref{fig:time} shows the wall-clock time plotted against the volume size of input experimental maps, ranging from the order of $10^6$ to $10^8$ $\text{Å}^3$.
For a fair comparison, all methods were run on the same workstation equipped with an AMD Ryzen Threadripper 2950X Processor of 32 CPUs and an NVIDIA GeForce RTX 2080 Ti of 12 GB VRAM. Each method was executed with the maximum batch size that our GPU can accommodate: approximately 12 for DeepEMhancer, 64 for EMReady, and 24 for CryoSAMU.

\begin{table}[!ht]
\caption{Average processing time in seconds of different methods.}
\label{tab:cryosamu_time}
\begin{center}
\setlength{\tabcolsep}{4pt} 
\begin{tabular}{l|cccc}
\toprule
Method & Autosharpen & DeepEMhancer & EMReady &\textbf{CryoSAMU}(ours) \\
\cmidrule{1-5}
Avg. Time (s) $\downarrow$ & 138$\pm$118 & 544$\pm$517 & 441$\pm$500 & \textbf{32$\pm$23} \\
\bottomrule
\end{tabular}
\end{center}
\end{table}

CryoSAMU (shown in green) displayed the minimum processing time across maps of varying volumes. Its weak linear dependency on map volume and tight confidence interval around its fit line indicate that CryoSAMU has both optimal scalability and consistent performance. 
In contrast, DeepEMhancer exhibited a strong linear correlation between processing time and map volume, indicating poor scalability as volume size increases.
EMReady showed a wider confidence interval in its linear fit, reflecting high variability in processing time. Notably, several outliers at lower volumes showed significantly longer processing time compared to other methods.
For a significantly large map with a volume size of $1.25 \times 10^8$ $\text{Å}^3$, CryoSAMU took only 116.48 seconds for generating an enhanced map, while Autosharpen, DeepEMHancer, and EMReady took 552.19, 2963.10, and 1731.719 seconds, respectively.
Table \ref{tab:cryosamu_time} lists the average processing time for each method. CryoSAMU achieved an average processing time of 32.49 seconds, approximately 13.6 times faster than EMReady, while generating comparably enhanced maps. These results suggest that CryoSAMU scales efficiently with increasing map volume, making it a promising tool for practical applications.

\subsection{Ablation Study}

We finally conducted an ablation study to evaluate the impact of integrating structural modality. We compared CryoSAMU with (w/) and without (w/o) structural embeddings using 75 test samples for correlation evaluation and 20 test samples for protein structure modeling assessment.

\begin{figure}[!ht]
  \centering
   \includegraphics[width=\linewidth, trim={0cm 0cm 0cm 0cm}, clip]{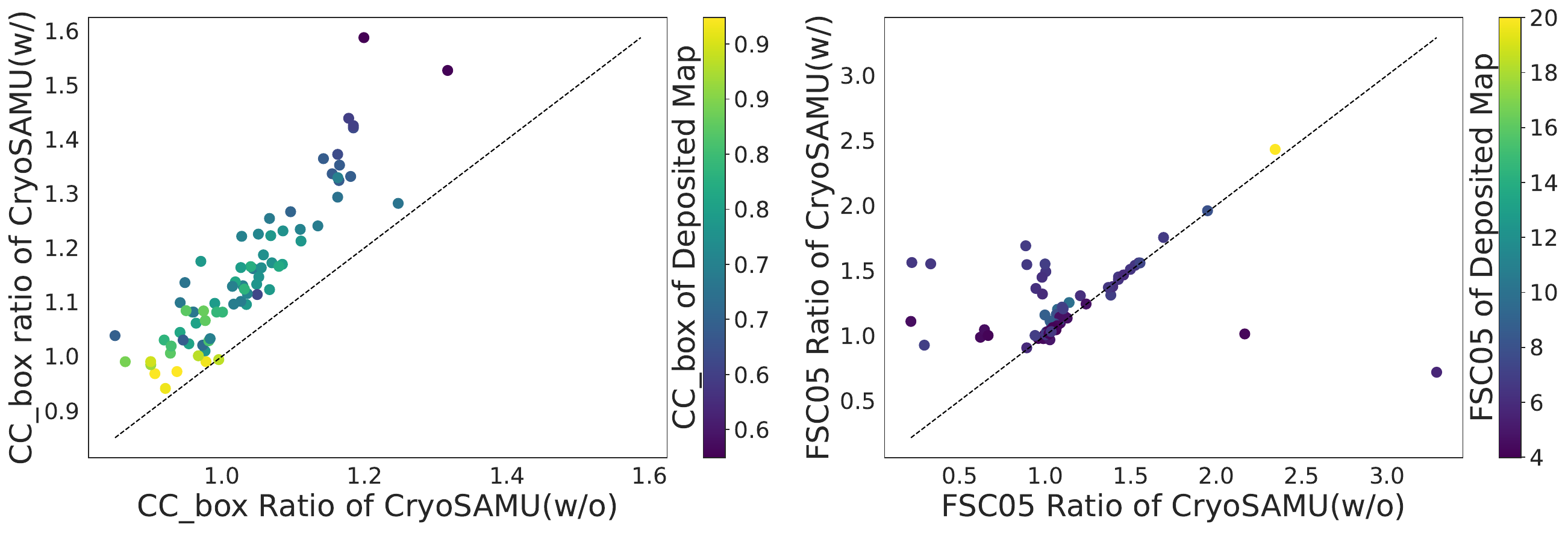}
   \caption{Pairwise comparison of enhanced/deposited ratios for CryoSAMU (w/) and (w/o). Each point represents a single map. Point colors encode the deposited map quality.
   }
   \label{fig:ablation}
\end{figure}

Figure \ref{fig:ablation} shows that CryoSAMU (w/) outperforms CryoSAMU (w/o) in both CC\_box and FSC05, with improvements of 98.7\% and 77.3\%, respectively, as indicated by scatter points above the diagonal line.
Moreover, the most significant gains (points far from 1.0) were observed in poorer‐quality deposited maps, which tend to have lower deposited CC\_box or higher FSC05 values.
Table \ref{tab:benchmark} lists the average real- and Fourier-space metrics, indicating that incorporating structural embeddings derived from ESM-IF1 led to a significant improvement of map enhancement, as also reported in Figure \ref{fig:benchmark}. In terms of protein structure modeling, residue coverage significantly raised from 8.08\% to 38.03\%, while sequence match raised from 8.13\% to 9.33\% with the integration of structural embeddings. This suggests that structural information helps complement map regions with poor resolutions, artifacts, or noise, thereby increasing the completeness (higher residue coverage) and improving accuracy (higher sequence match) during structure modeling. These findings underscore the importance of integrating structural modality to enable the network to develop structural awareness beyond learning solely from 3D density maps.

\section{Discussion and Future Work}

In this chapter, we introduce CryoSAMU, the first structure-aware multimodal network for enhancing cryo-EM density maps at intermediate resolution of protein structures. Our approach combines 3D map features with corresponding structural features through cross-attention mechanisms. 
Notably, structural embeddings are used solely as auxiliary supervision during training to guide the model toward learning biologically-grounded and structure-aware representations from the density maps, which aligns well with the principle of conditional training \cite{auxiliary}.
In addition, we develop a self-attention weighting algorithm to produce fixed-size representations of structural embeddings derived from the pretrained ESM-IF1 model, preserving inter-chain and residue relationships while maintaining structural integrity. 
Our benchmark results demonstrate that CryoSAMU preforms competitively with existing cutting-edge methods, closely approaching the performance of EMReady, the current leading tool for cryo-EM density map enhancement. Notably, CryoSAMU achieves the fastest processing speed among all tested methods, positioning it as a promising solution for large-scale and practical applications. Furthermore, our ablation study reveals that incorporating an additional structural modality significantly boosts CryoSAMU's performance across all evaluation metrics, suggesting a new avenue for future cryo-EM research to explore the effective integration of multimodal data during network training.

Despite CryoSAMU demonstrating superior performance in enhancing cryo-EM maps, its current architecture---based on residual convolutions within a U-Net framework---is primarily designed to capture local information. In practice, capturing global context and long-range dependencies across map voxels could further improve performance. This could be addressed by adopting more hierarchical architectures, such as the Swin Transformer \cite{swintransformer}, which facilitates feature extraction over larger receptive fields. Moreover, incorporating supplementary loss terms, such as the Structural Similarity Index Measure (SSIM) loss, could mitigate overfitting and enhance training efficiency \cite{emready}. These will be explored in our future work.
Furthermore, we plan to explore other cutting-edge pLLMs for structural embeddings, such as ESM3, a multimodal generative language model that offers significantly improved structural representations \cite{esm3}. Unlike structure-based models, ESM3 accepts amino acid sequences as input, which are more readily available.
Additionally, we aim to expand our dataset by including high-resolution maps, which could increase the robustness of the model and further elevate its performance.

\bookmarksetup{startatroot}

\bibliographystyle{unsrt} 
\bibliography{reference} 

\appendix
\clearpage
\newgeometry{bottom=2.5cm}

\phantomsection
\pdfbookmark[-1]{Appendices}{appendix_start}


\chapter{Supplementary Materials for Chapter 1}


\section{Dot-Parenthesis Notation} \label{dp_notation}
Dot-parenthesis (dp) notation is a simple way to represent a pseudoknot-free secondary structure of DNA or RNA. Each character represents a base (except        ``\texttt{+}'', which is a separator for different strands). Dots indicate unpaired bases and matching parentheses indicate paired bases. The number of open and closed parentheses is always equal. For example, in the dp notation  \texttt{\seqsplit{5$'$-{...(((...}-3$'$+5$'$-{...)))...}-3$'$}} for the secondary structure of two DNA strands $P$ (\texttt{\seqsplit{5$'$-TGACGATCA-3$'$}}) and $\bar{P}$ (\texttt{\seqsplit{5$'$-TGATCGTCA-3$'$}}), the left part of the ``\texttt{+}'' sign corresponds to strand $P$ and the right part corresponds to strand $\bar{P}$. Three open parentheses indicate that the bases ``\texttt{CGA}'' in strand $P$ are paired with the bases ``\texttt{TCG}'' in strand $\bar{P}$ which are represented by three closed parentheses.


\chapter{Supplementary Materials for Chapter 3}

\section{Multistrand Simulation Rate Methods and Parameters} \label{multistrand_rates}

\begin{table}[!ht]
\centering
\caption{Kinetic rate models and parameters used in Multistrand simulations for Gao-P4T4, Hata-39, and all Machinek's reactions. The Metropolis rates were adopted from \cite{zolaktafarrhenius} and the Arrhenius rates from \cite{lovrod}.}

\label{tab:multistrand_rates}
\vspace{0.5em}
\begin{tabular}{@{}ll@{}}
\toprule
\multicolumn{2}{l}{\textbf{Gao-P4T4}} \\ 
\midrule
Rate method              & Metropolis        \\
Rate parameters \\
\quad \textit{Unimolecular scaling} & $2.41686715 \times 10^{6} \quad \mathrm{s^{-1}} $ \\
\quad \textit{Bimolecular scaling}  & $8.01171383 \times 10^{5} \quad \mathrm{M^{-1}s^{-1}}$  \\
\addlinespace[0.5em]
\toprule
\multicolumn{2}{l}{\textbf{Hata-39 \& Machinek's Reactions}} \\ 
\midrule
Rate method              & Arrhenius         \\
Rate parameters \\
\quad \textit{Bimolecular scaling}  & $5.40306772408701\times 10^{-2} \quad \mathrm{M^{-1}s^{-1}}$ \\
\quad \textit{lnA\_Stack}           & $7.317929742353791 \quad\mathrm{\ln s^{-1/2}} $ \\
\quad \textit{E\_Stack}              & $1.371171987160233 \quad\mathrm{kcal/mol}$ \\
\quad \textit{lnA\_Loop}            & $9.398865994892086 \quad\mathrm{\ln s^{-1/2}}$ \\
\quad \textit{E\_Loop}               & $6.675295990888666\times 10^{-1} \quad\mathrm{kcal/mol}$ \\
\quad \textit{lnA\_End}             & $1.4182060613367684\times 10^{1} \quad\mathrm{\ln s^{-1/2}}$ \\
\quad \textit{E\_End}                & $3.428091116849789 \quad\mathrm{kcal/mol}$ \\
\quad \textit{lnA\_StackLoop}       & $1.124903532063121\times 10^{1} \quad\mathrm{\ln s^{-1/2}}$ \\
\quad \textit{E\_StackLoop}          & $1.1475124044307237 \quad\mathrm{kcal/mol}$ \\
\quad \textit{lnA\_StackEnd}        & $1.2975095181504653\times 10^{1} \quad\mathrm{\ln s^{-1/2}}$ \\
\quad \textit{E\_StackEnd}           & $-1.2455835738995455 \quad\mathrm{kcal/mol}$ \\
\quad \textit{lnA\_LoopEnd}         & $4.453062436943478\times 10^{-1} \quad\mathrm{\ln s^{-1/2}}$ \\
\quad \textit{E\_LoopEnd}            & $-2.0459543870895307 \quad\mathrm{kcal/mol}$ \\
\quad \textit{lnA\_StackStack}      & $1.15977270694611\times 10^{1} \quad\mathrm{\ln s^{-1/2}}$ \\
\quad \textit{E\_StackStack}         & $-2.4564902855673165 \quad\mathrm{kcal/mol}$ \\
\bottomrule
\end{tabular}
\end{table}

\clearpage

\section{Comparison of Local Structure Preservation} \label{local_preserve}

\begin{table*}[!ht]
\caption{
Comparison of local structure preservation across different embedding methods. The table lists the average difference in energy and graph edit distance (GED) between original state and their K nearest neighbours in the embedding space, for $K=\{10, 50, 100, 200, 500, 1000, 2000, 3000, 4000, 5000\}$. Lower values indicate better preservation. \label{tab:local_preserve}
}
\vspace{0.5em}
\tabcolsep=0pt
\begin{tabular*}{\textwidth}{@{\extracolsep{\fill}}lcccccccccc@{\extracolsep{\fill}}}
\toprule
\multicolumn{11}{c}{\textit{Average difference for K nearest neighbours} $\downarrow$}\\
\addlinespace[0.6em]
& \multicolumn{2}{@{}c@{}}{K@10} & \multicolumn{2}{@{}c@{}}{K@50} & \multicolumn{2}{@{}c@{}}{K@100} & \multicolumn{2}{@{}c@{}}{K@200} & \multicolumn{2}{@{}c@{}}{K@500} \\
\cline{2-11}
Methods & Energy & GED & Energy & GED & Energy & GED & Energy & GED & Energy & GED \\
\midrule
ViDa  & 1.685 & 5.853 & \textbf{1.766} & 6.314 & \textbf{1.810} & 6.565 & \textbf{1.868} & 6.841 & \textbf{1.940} & 7.256 \\
PCA   & 2.916 & 8.367 & 3.038 & 8.839 & 3.102 & 9.101 & 3.180 & 9.407 & 3.306 & 9.886 \\
PHATE & 1.814 & 4.886 & 1.927 & 5.498 & 1.985 & 5.843 & 2.036 & 6.196 & 2.135 & 6.747 \\
UMAP  & 1.785 & 3.870 & 1.892 & 4.407 & 1.980 & 4.853 & 2.094 & 5.476 & 2.264 & 6.590 \\
t-SNE & \textbf{1.629} & \textbf{2.543} & 1.836 & \textbf{3.727} & 1.925 & \textbf{4.390} & 2.037 & \textbf{5.183} & 2.268 & \textbf{6.454} \\
MDS   & 3.936 & 14.572 & 3.922 & 14.559 & 3.923 & 14.560 & 3.924 & 14.560 & 3.925 & 14.564 \\
\addlinespace[0.8em]

\multicolumn{11}{c}{\textit{Average difference for K nearest neighbours} $\downarrow$}\\
\addlinespace[0.6em]
& \multicolumn{2}{@{}c@{}}{K@1000} & \multicolumn{2}{@{}c@{}}{K@2000} & \multicolumn{2}{@{}c@{}}{K@3000} & \multicolumn{2}{@{}c@{}}{K@4000} & \multicolumn{2}{@{}c@{}}{K@5000} \\
\cline{2-11}
Methods & Energy & GED & Energy & GED & Energy & GED & Energy & GED & Energy & GED \\
\midrule
ViDa  & \textbf{2.010} & 7.669 & \textbf{2.143} & 8.220 & \textbf{2.261} & 8.625 & \textbf{2.357} & 8.925 & \textbf{2.450} & 9.195 \\
PCA   & 3.421 & 10.413 & 3.560 & 11.040 & 3.641 & 11.378 & 3.711 & 11.629 & 3.770 & 11.836 \\
PHATE & 2.228 & \textbf{7.267}  & 2.336 & \textbf{7.914}  & 2.438 & \textbf{8.421}  & 2.496 & \textbf{8.834}  & 2.552 & \textbf{9.160} \\
UMAP  & 2.562 & 7.845  & 2.848 & 9.120  & 3.009 & 9.813  & 3.120 & 10.360 & 3.222 & 10.778 \\
t-SNE & 2.511 & 7.592  & 2.849 & 8.827 & 3.064 & 9.529  & 3.214 & 10.039  & 3.317 & 10.448 \\
MDS   & 3.925 & 14.564 & 3.922 & 14.563 & 3.924 & 14.564 & 3.926 & 14.564 & 3.927 & 14.564 \\
\bottomrule
\end{tabular*}
\end{table*}

\clearpage

\begin{table*}[!ht]
\caption{
Comparison of local structure preservation across different ViDa ablations. 
The table lists the average difference in energy and graph edit distance (GED) 
between original state and their K nearest neighbours in the embedding space, 
for $K=\{10, 50, 100, 200, 500, 1000, 2000, 3000, 4000, 5000\}$. 
Lower values indicate better preservation.
\label{tab:vida_ablation_local_preserve}
}
\vspace{0.5em}
\tabcolsep=0pt
\begin{tabular*}{\textwidth}{@{\extracolsep{\fill}}lcccccccccc@{\extracolsep{\fill}}}
\toprule
\multicolumn{11}{c}{\textit{Average difference for K nearest neighbours} $\downarrow$}\\
\addlinespace[0.6em]
& \multicolumn{2}{@{}c@{}}{K@10} & \multicolumn{2}{@{}c@{}}{K@50} & \multicolumn{2}{@{}c@{}}{K@100} & \multicolumn{2}{@{}c@{}}{K@200} & \multicolumn{2}{@{}c@{}}{K@500} \\
\cline{2-11}
Methods & Energy & GED & Energy & GED & Energy & GED & Energy & GED & Energy & GED \\
\midrule
ViDa               & 1.685 & \textbf{5.853} & 1.766 & \textbf{6.314} & 1.810 & \textbf{6.565} & 1.868 & \textbf{6.841} & 1.940 & \textbf{7.256} \\
ViDa\_noGED        & 1.517 & 9.833 & 1.525 & 9.914 & 1.533 & 9.984 & 1.548 & 10.072 & 1.595 & 10.221 \\
ViDa\_noMPT        & 1.681 & 6.631 & 1.744 & 7.022 & 1.777 & 7.257 & 1.816 & 7.529 & 1.880 & 7.922 \\
ViDa\_noMPT\_noGED & \textbf{1.470} & 8.670 & \textbf{1.485} & 8.788 & \textbf{1.495} & 8.876 & \textbf{1.510} & 8.999 & \textbf{1.557} & 9.215 \\
ViDa-\\\_noG\_noMPT\_noGED & 2.400 & 10.356 & 2.401 & 10.382 & 2.402 & 10.408 & 2.406 & 10.459 & 2.430 & 10.589 \\
\addlinespace[0.8em]

\multicolumn{11}{c}{\textit{Average difference for K nearest neighbours} $\downarrow$}\\
\addlinespace[0.6em]
& \multicolumn{2}{@{}c@{}}{K@1000} & \multicolumn{2}{@{}c@{}}{K@2000} & \multicolumn{2}{@{}c@{}}{K@3000} & \multicolumn{2}{@{}c@{}}{K@4000} & \multicolumn{2}{@{}c@{}}{K@5000} \\
\cline{2-11}
Methods & Energy & GED & Energy & GED & Energy & GED & Energy & GED & Energy & GED \\
\midrule
ViDa               & 2.010 & \textbf{7.669} & 2.143 & \textbf{8.220} & 2.261 & \textbf{8.625} & 2.357 & \textbf{8.925} & 2.450 & \textbf{9.195} \\
ViDa\_noGED        & 1.662 & 10.353 & 1.764 & 10.614 & 1.892 & 10.846 & 1.996 & 11.035 & 2.077 & 11.233 \\
ViDa\_noMPT        & 1.950 & 8.267 & 2.087 & 8.741 & 2.198 & 9.115 & 2.291 & 9.418 & 2.379 & 9.683 \\
ViDa\_noMPT\_noGED & \textbf{1.637} & 9.476 & \textbf{1.745} & 9.783 & \textbf{1.879} & 10.098 & \textbf{1.979} & 10.349 & \textbf{2.060} & 10.594 \\
ViDa-\\\_noG\_noMPT\_noGED & 2.483 & 10.763 & 2.612 & 11.056 & 2.742 & 11.326 & 2.821 & 11.533 & 2.874 & 11.702 \\
\bottomrule
\end{tabular*}
\end{table*}

\clearpage

\section{Stack, Mis-Stack, and Hairpin Schematic Representations} \label{hairpin}

\begin{figure}[ht]
    \centering
    \includegraphics[width=\linewidth, trim={0cm 17.5cm 0cm 0cm} ,clip] 
    {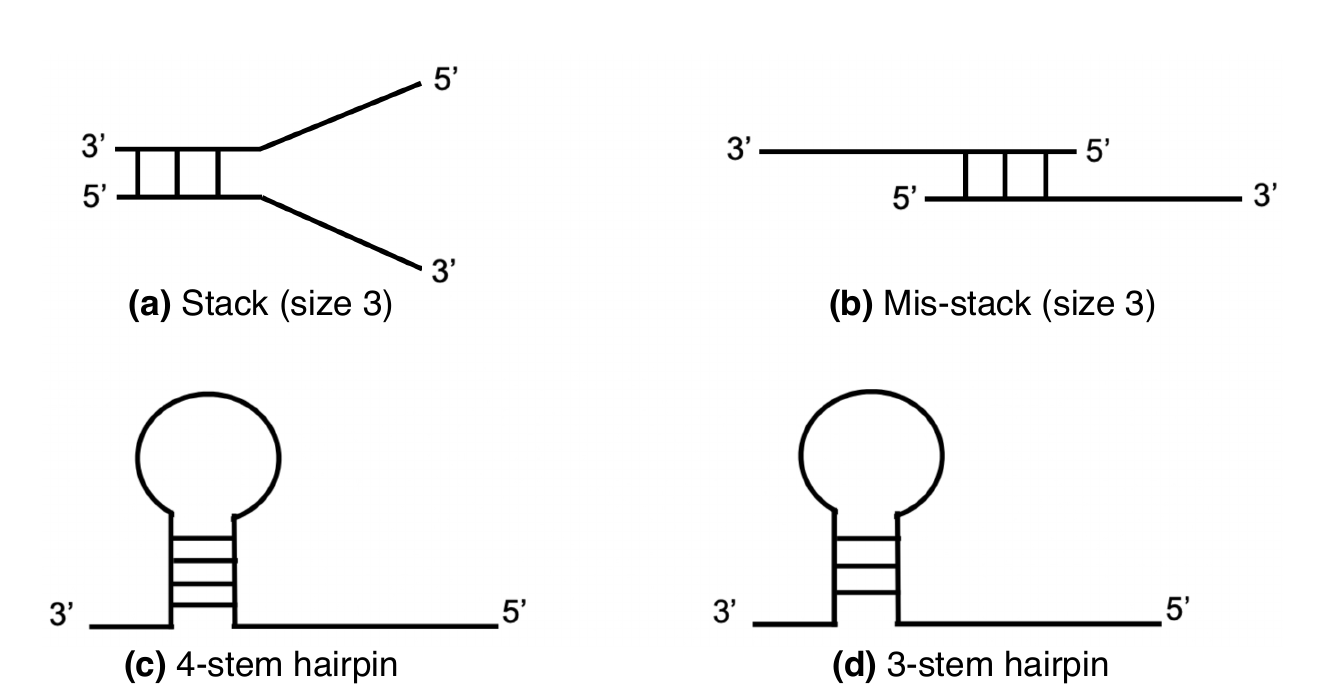}
    \caption{ 
    Schematic representations of stack, mis-stack, and hairpin structures.
    }
    \label{fig:hairpins}
\end{figure}

\clearpage

\section{DBSCAN} \label{DBSCAN}

\begin{figure}[ht!]
  \centering
  \includegraphics[width=\linewidth,trim={0cm 19cm 0cm 0cm}, clip]
  {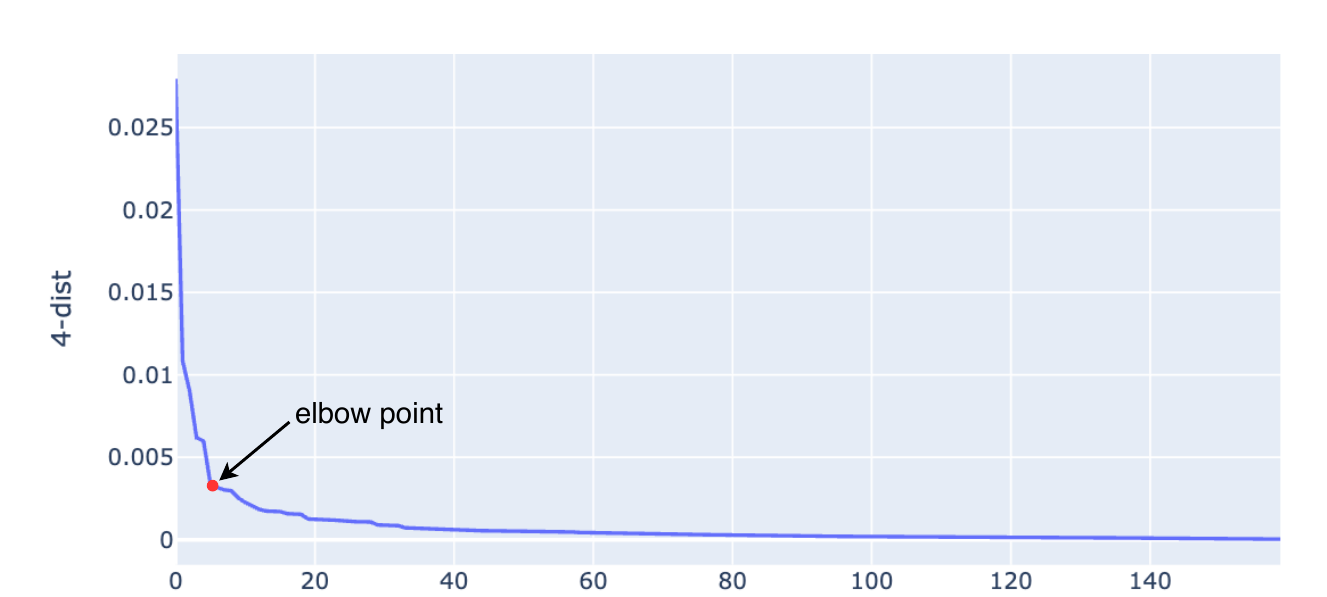}
  \caption{The elbow method plot to determine the epsilon of DBSCAN.}
  \label{fig:dbscan}
\end{figure}

DBSCAN (Density-Based Spatial Clustering of Applications with Noise) is a density-based clustering algorithm widely used in unsupervised learning. It groups data points based on their proximity in a feature space, making it particularly effective for discovering clusters of arbitrary shape. DBSCAN has two hyperparameters: epsilon ($\texttt{eps}$) that defines the maximum distance between two points for one to be considered as in the neighbourhood of the other, and  minimum samples ($\texttt{min\_samples}$) that presents the number of neighbours needed to tell a region is dense. In our work, we chose $\texttt{min\_samples=4}$ as the paper suggests \cite{DBSCAN}. We used the ``elbow'' method to determine $\texttt{eps}$. Specifically, we computed the distance of each point to its 4 nearest neighbours then sorted the points based on the resulting distances. The distances are plotted against sorted points in Figure \ref{fig:dbscan}. Finally, we selected the elbow point represented by a red dot as a reference. Therefore, the value of $\texttt{eps}$ is set from the reference distance, i.e., $\texttt{eps=0.0034}$ in this work.

\clearpage

\section{Kinetic Traps Under Smaller Filtering Threshold} \label{trap_SI}

\begin{figure}[ht!]
  \centering
  \includegraphics[width=\linewidth,trim={0cm 18cm 0cm 0cm}, clip]
  {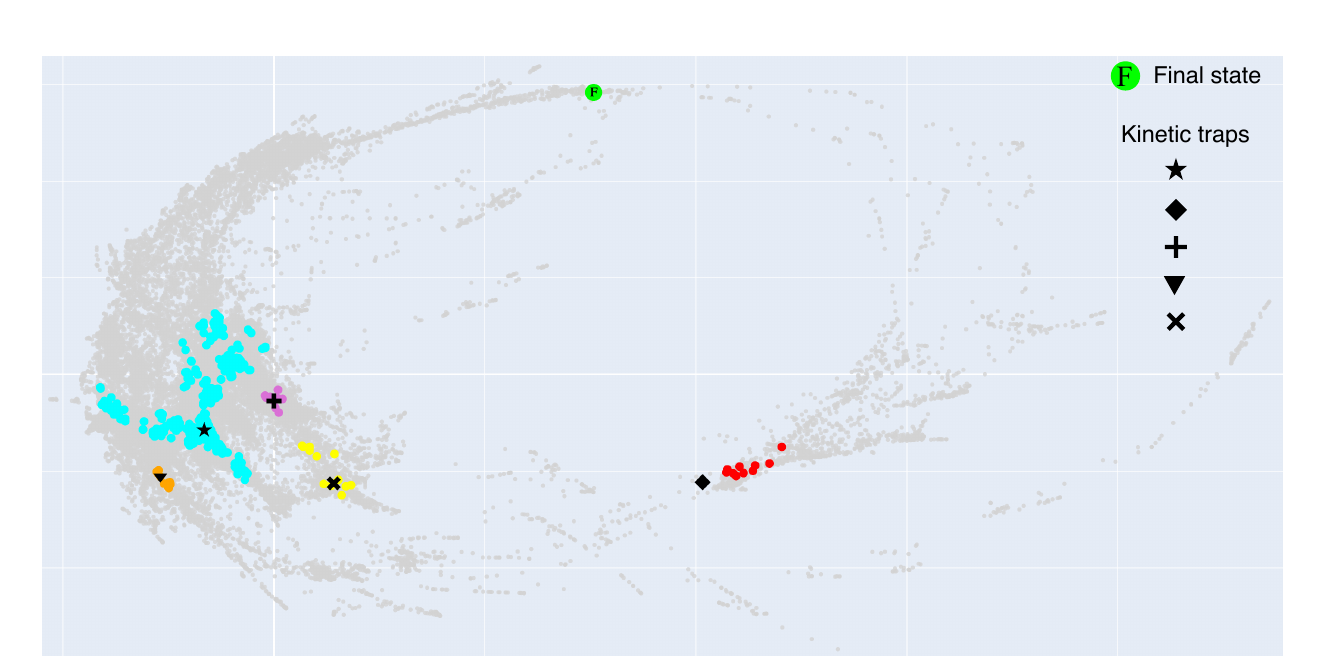}
  \caption{
  DBSCAN results for the Gao-P4T4 reaction on the filtered state space, with the filtering threshold of cumulative time set to $4\times10^{-5} s$. DBSCAN identifies five clusters shown with different colors. Each symbol corresponds to the minimum free energy state (i.e., kinetic trap) within each cluster. States that are clustered as noise or filtered out are shown in light gray. DBSCAN was applied with parameters $\texttt{eps=0.0027}$ and $\texttt{min\_samples=4}$. 
  }
  \label{fig:dbscan_SI}
\end{figure}

We used a smaller cumulative time threshold of $4\times10^{-5} s$ to exclude less significant states and then applied DBSCAN to cluster the post-filtered states for the Gao-P4T4 reaction. The DBSCAN results is shown in Figure \ref{fig:dbscan_SI} and resultant five kinetic traps are listed in Table \ref{tab:gao_5traps}.

\begin{table}[!ht]
\caption{Kinetic traps for the Gao-P4T4 reaction with their associated dot-parenthesis notations, cumulative times, and free energies, filtered using a smaller cumulative time threshold of $4\times10^{-5} s$.}
\vspace{0.5em}
\label{tab:gao_5traps}
\resizebox{\textwidth}{!}{%
\small
\centering
\begin{tabular}{c|ccc}
\toprule
  Kinetic Trap & Secondary Structure with DP-Notation (5' $\rightarrow$ 3')  & Cuml. Time (s) & $\Delta G$ (kcal/mol) \\
\midrule
$\bigstar$ & \texttt{5$'$-.((((.((((..........)))).-3$'$ + 5$'$-.((((..........)))).)))).-3$'$}  &  $1.36\times10^{-1}$  &  $-5.87$  \\
\midrule
$\blacklozenge$  & \texttt{5$'$-....(((.....))).((((((((.-3$'$ + 5$'$-.)))))))).(((.....)))....-3$'$}  & $3.24\times10^{-3}$ &  $-10.6$  \\
\midrule
$\boldsymbol{\times}$  & \texttt{5$'$-((....((((..........)))).-3$'$ + 5$'$-....))....(((.....)))....-3$'$}  & $6.23\times10^{-4}$  &  $-2.73$  \\
\midrule
$\boldsymbol{+}$  & \texttt{5$'$-.((((.((((..........)))).-3$'$ + 5$'$-......(((........))))))).-3$'$}  & $2.32\times10^{-4}$  &  $-2.07$  \\
\midrule
$\blacktriangledown$ & \texttt{5$'$-.(((((((((....)))))......-3$'$ + 5$'$-.((((..........)))).)))).-3$'$}  & $1.80\times10^{-4}$  &  $-2.10$ \\
\bottomrule
\end{tabular}
}
\end{table}

\clearpage

\section{Visualizations by PCA, UMAP, t-SNE, PHATE, and MDS} \label{otherviz}

\begin{figure}[ht!]
\centering
\includegraphics[width=\linewidth,trim={0cm 8cm 0cm 0cm}, clip] {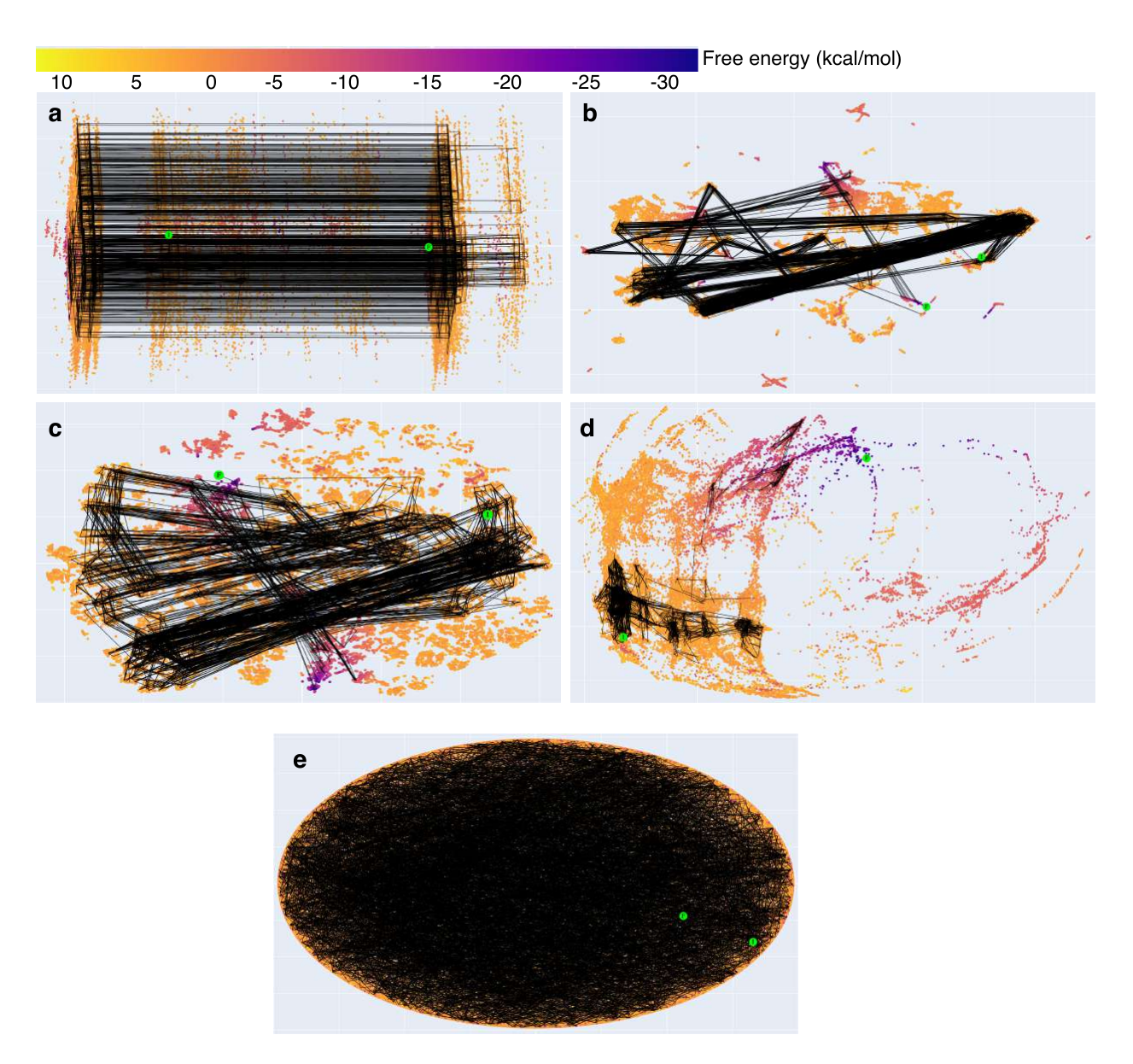}
\caption{ 
Trajectories laid out on the embedding for Gao-P4T4. Each point represents a secondary structure state. The color of each point represents the value of free energy. The black curve represents the same trajectory in each plot, consistent with the trajectory (shown in cyan) in Figure \ref{fig:gao_p4t4_vida}a. The initial and final states are indicated by the green circles marked $I$ and $F$, respectively.
The plot is made by \textbf{(a)} PCA, \textbf{(b)} UMAP, \textbf{(c)} t-SNNE, \textbf{(d)} PHATE, and \textbf{(e)} MDS.
}
\label{fig:novida}
\end{figure}

We assessed the visualizations for Gao-P4T4 by PCA, UMAP, t-SNE, PHATE, and MDS, as shown in Figure \ref{fig:novida}. It can be seen that PCA, UMAP, t-SNE, and MDS's visualizations are quite poor. PHATE's visualization in Figure \ref{fig:novida}d shows that the energy landscape roughly follows the high-to-low trend from the initial to final states. However, its trajectories are not as smooth as ViDa's, owing to the large number of long segments.

\clearpage


\chapter{Supplementary Materials for Chapter 4}

\section{Visualization Plots for Perfect-toehold8} \label{sup:per_t8}

\begin{figure}[!ht]
    \centering
    \includegraphics[width=.8\linewidth, trim={2.8cm 2.8cm 6cm 2cm}, clip]
    {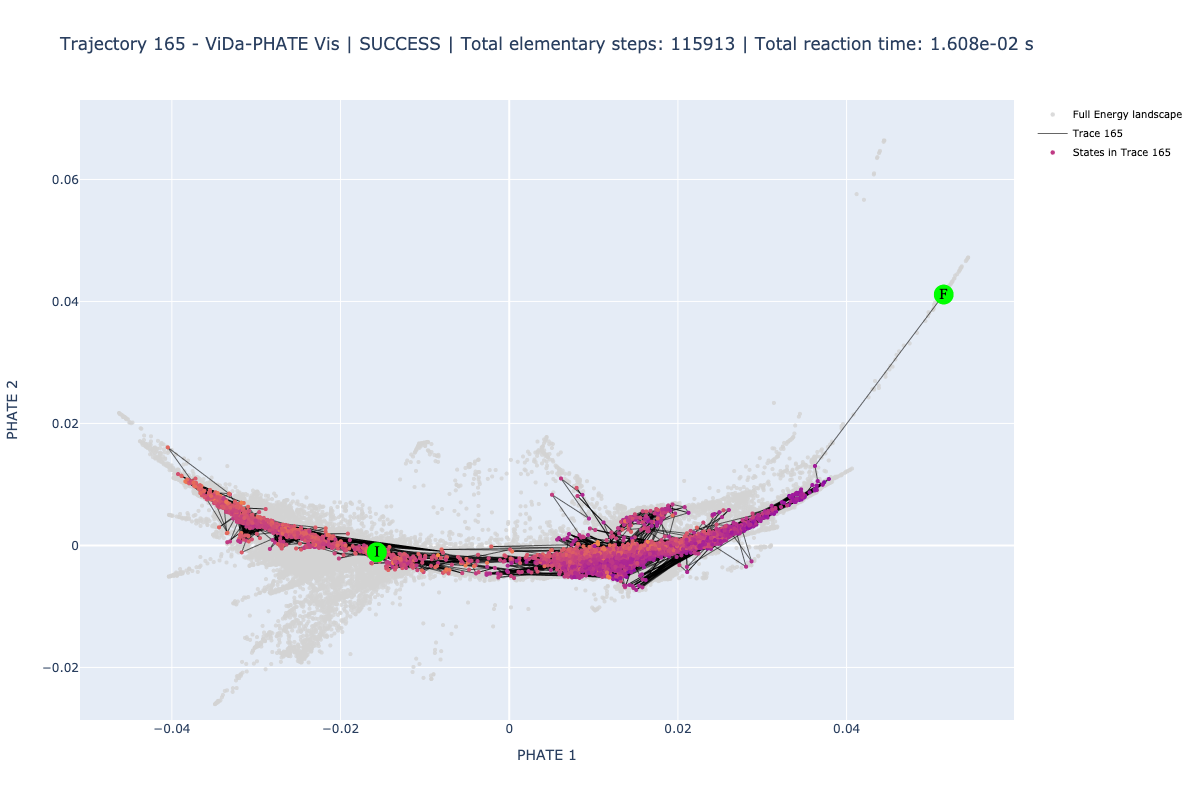}
    \caption{Visualization plot for the perfect-toehold8 reaction. The black trace represents a successful trajectory.}
    \label{fig:per_t8_succ1}
\end{figure}

\begin{figure}[!ht]
    \centering
    \includegraphics[width=.8\linewidth, trim={2.8cm 2.8cm 6cm 2cm}, clip]
    {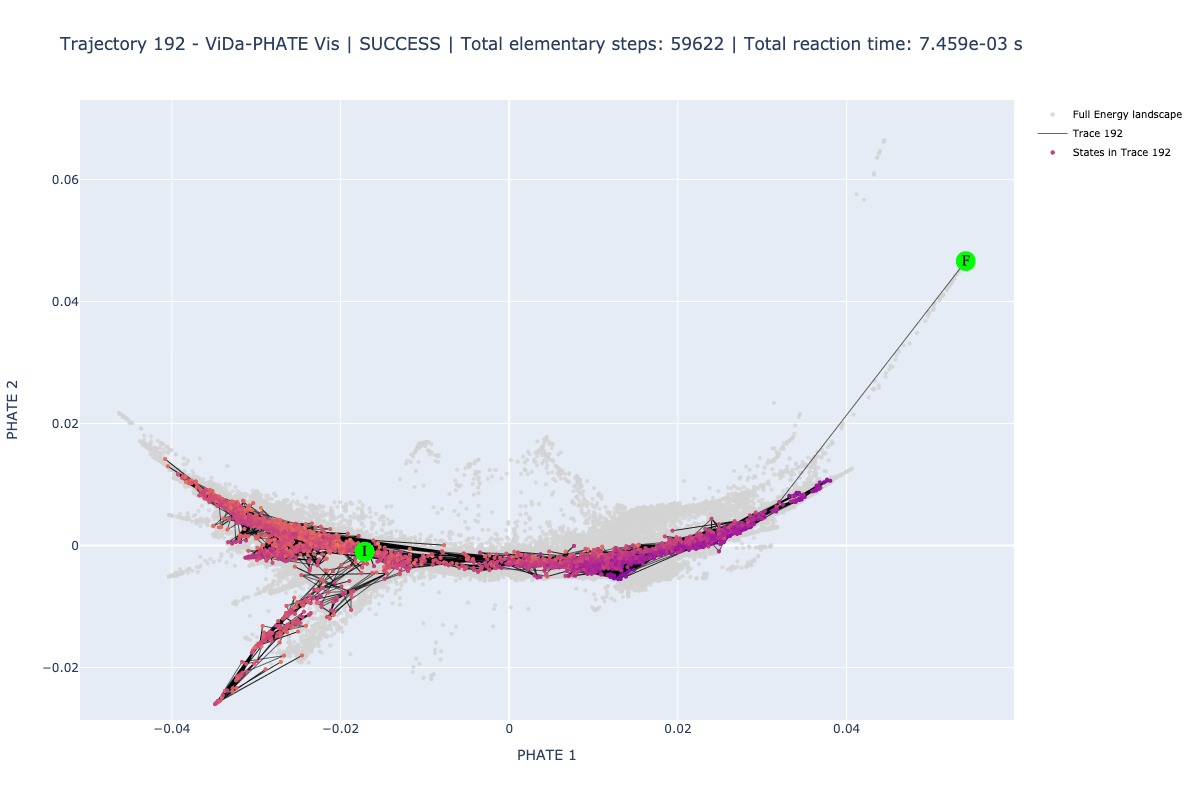}
    \caption{Visualization plot for the perfect-toehold8 reaction. The black trace represents a successful trajectory.}
    \label{fig:per_t8_succ2}
\end{figure}

\begin{figure}[!ht]
    \centering
    \includegraphics[width=.8\linewidth, trim={2.8cm 2.8cm 6cm 2cm}, clip]
    {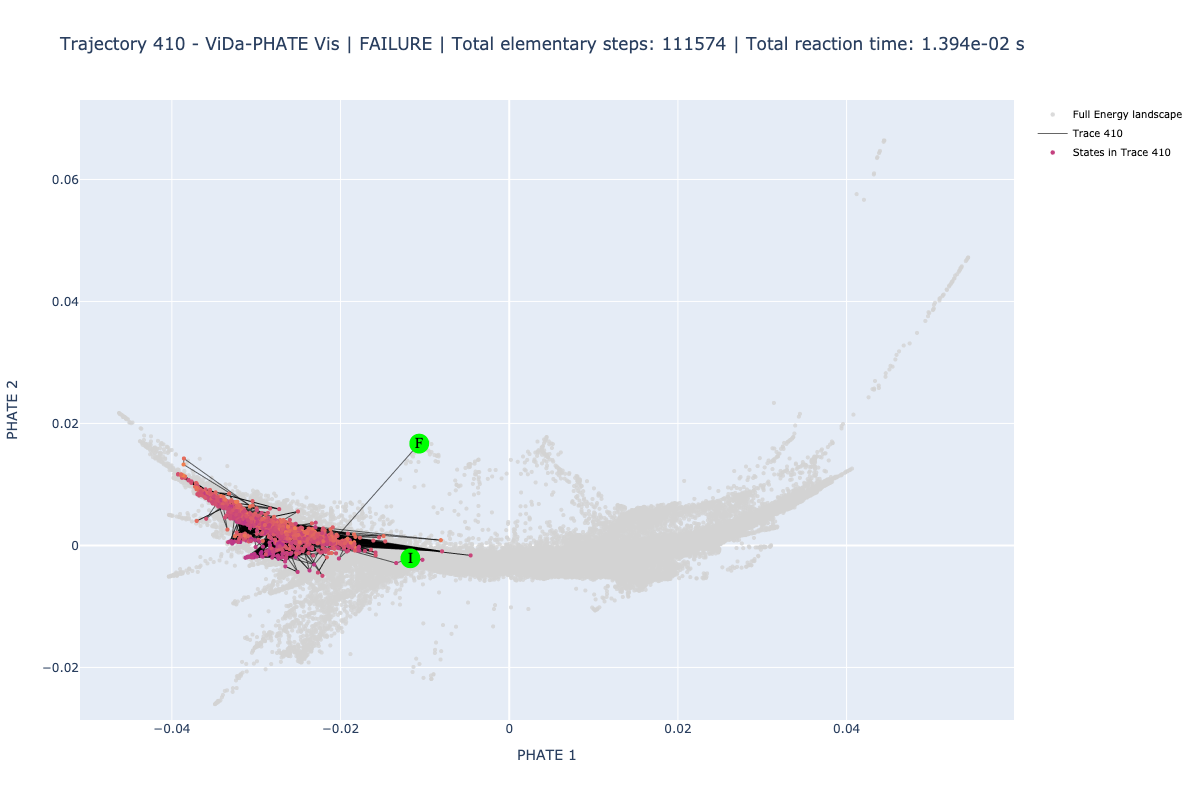}
    \caption{Visualization plot for the perfect-toehold8 reaction. The black trace represents a failed trajectory.}
    \label{fig:per_t8_fail1}
\end{figure}

\begin{figure}[!ht]
    \centering
    \includegraphics[width=.8\linewidth, trim={2.8cm 2.8cm 6cm 2cm}, clip]
    {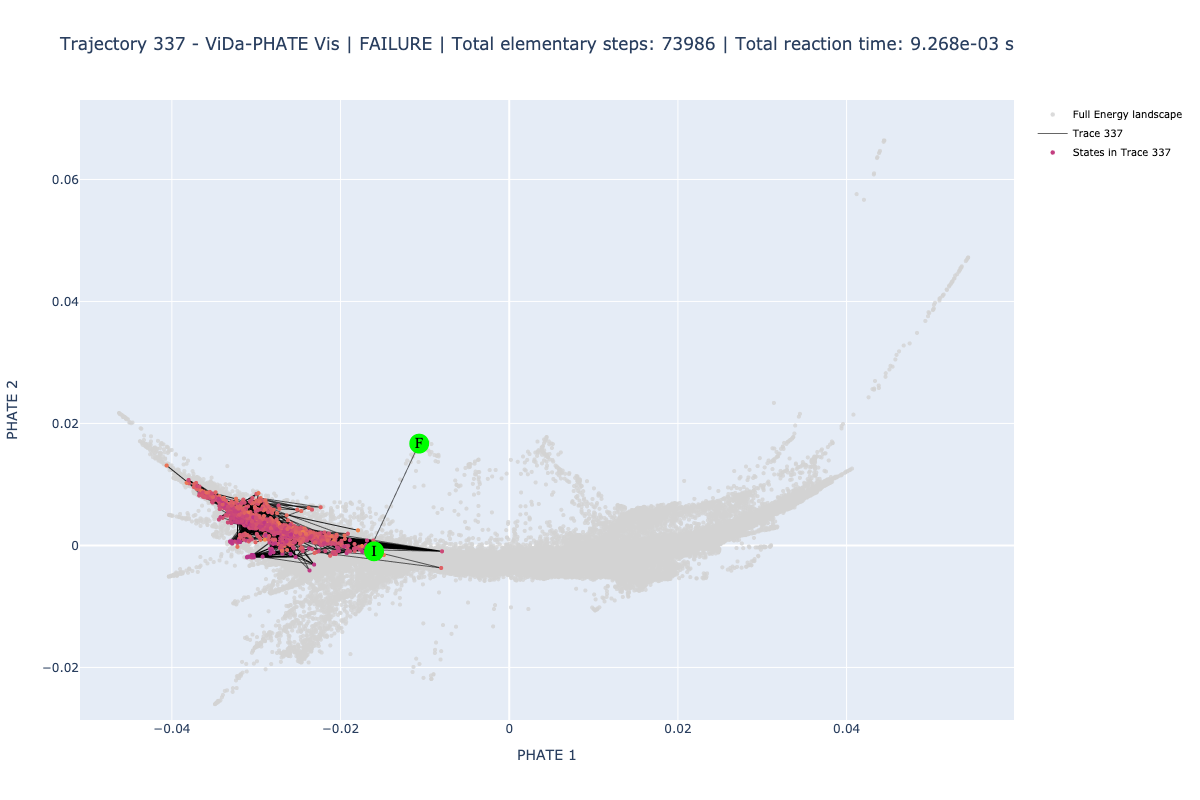}
    \caption{Visualization plot for the perfect-toehold8 reaction. The black trace represents a failed trajectory.}
    \label{fig:per_t8_fail2}
\end{figure}

\clearpage

\section{Visualization Plots for Perfect-toehold7-reporter} \label{sup:per_t7_reporter}

\begin{figure}[!ht]
    \centering
    \includegraphics[width=.8\linewidth, trim={2.8cm 2.8cm 6cm 2cm}, clip]
    {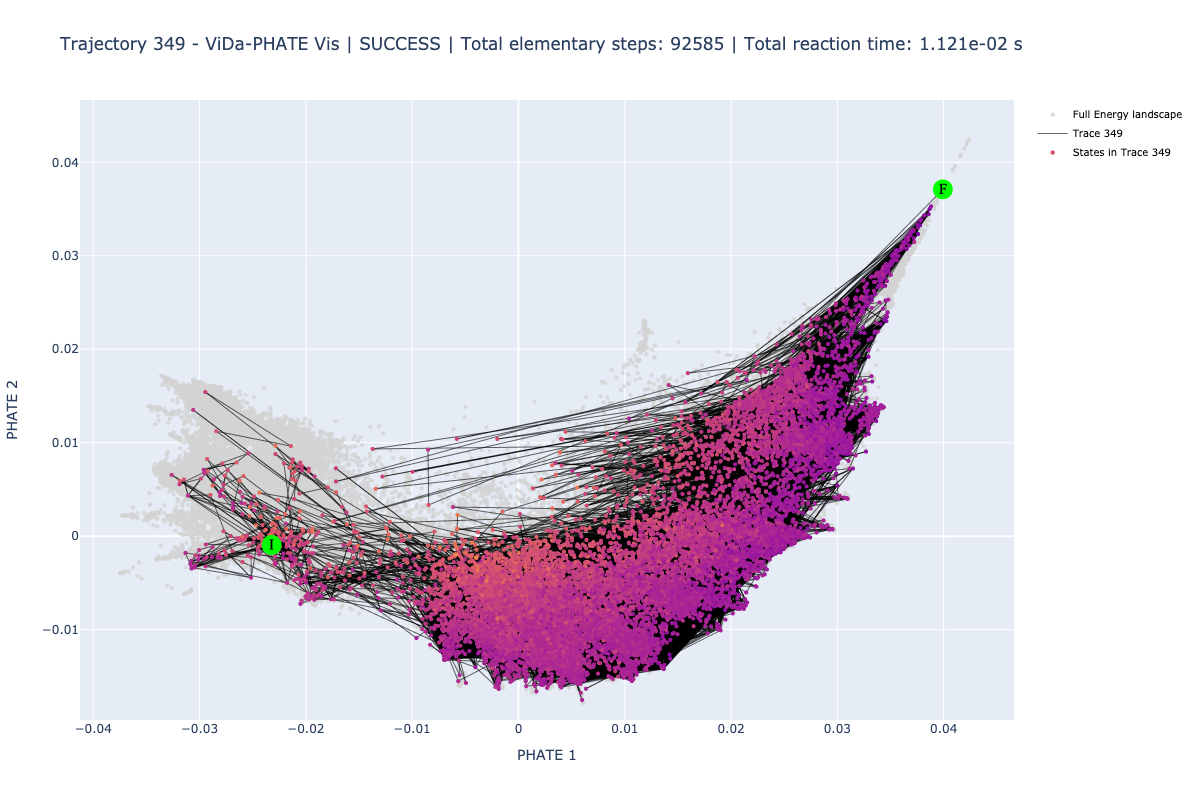}
    \caption{Visualization plot for the perfect-toehold7-reporter reaction. The black trace represents a successful trajectory.}
    \label{fig:per_t7_dangel_succ1}
\end{figure}

\begin{figure}[!ht]
    \centering
    \includegraphics[width=.8\linewidth, trim={2.8cm 2.8cm 6cm 2cm}, clip]
    {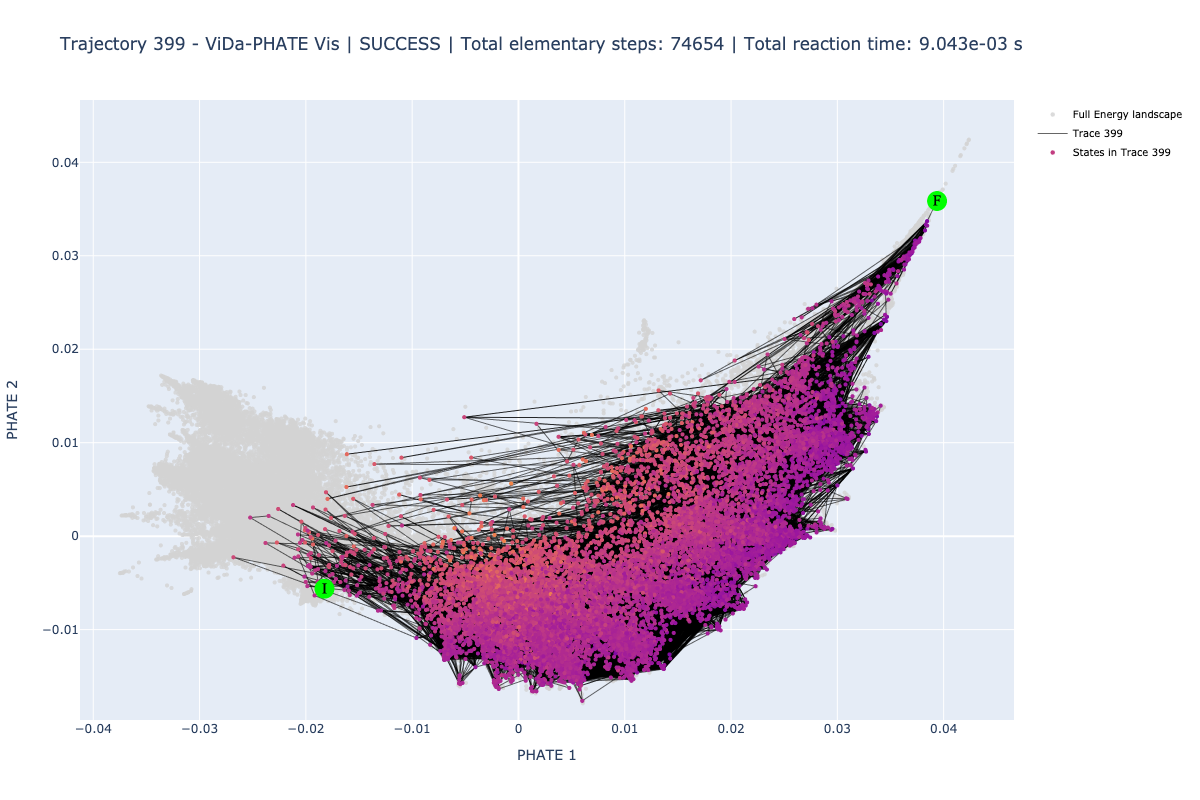}
    \caption{Visualization plot for the perfect-toehold7-reporter reaction. The black trace represents a successful trajectory.}
    \label{fig:per_t7_dangel_succ2}
\end{figure}

\begin{figure}[!ht]
    \centering
    \includegraphics[width=.8\linewidth, trim={2.8cm 2.8cm 6cm 2cm}, clip]
    {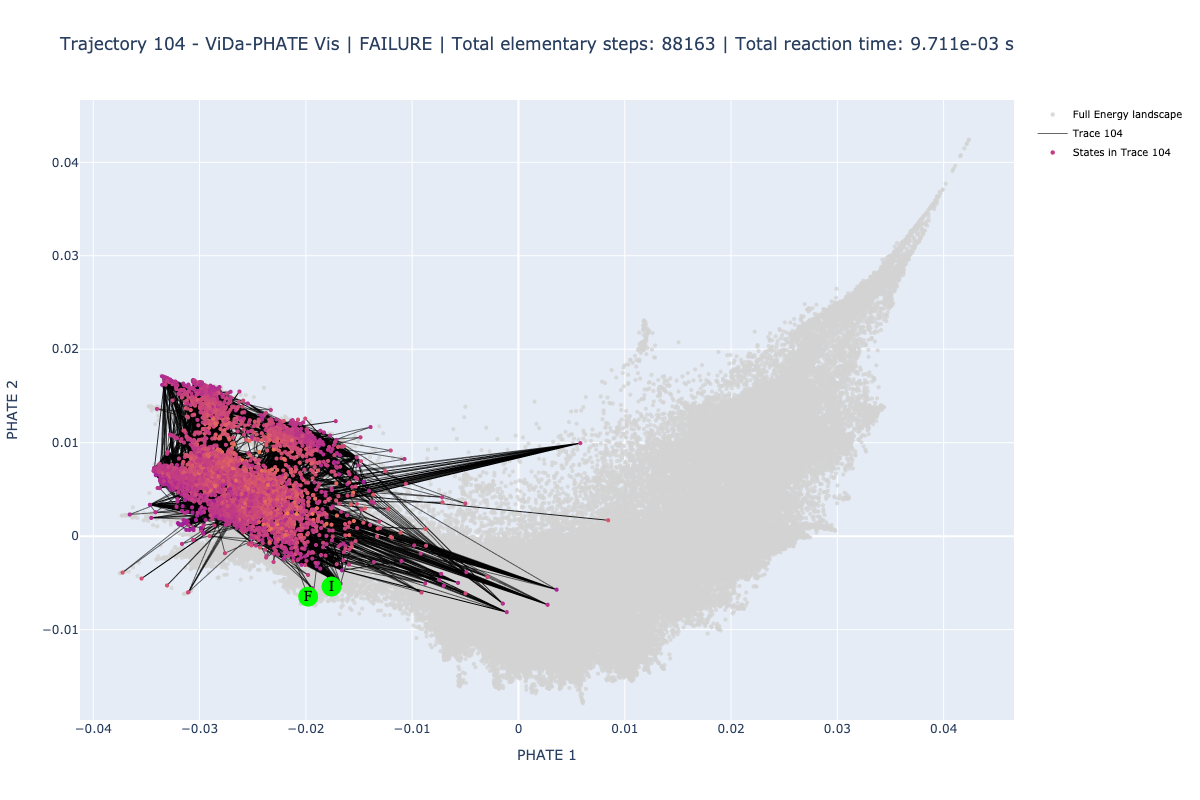}
    \caption{Visualization plot for the perfect-toehold7-reporter reaction. The black trace represents a failed trajectory.}
    \label{fig:per_t7_dangel_fail1}
\end{figure}

\begin{figure}[!ht]
    \centering
    \includegraphics[width=.8\linewidth, trim={2.8cm 2.8cm 6cm 2cm}, clip]
    {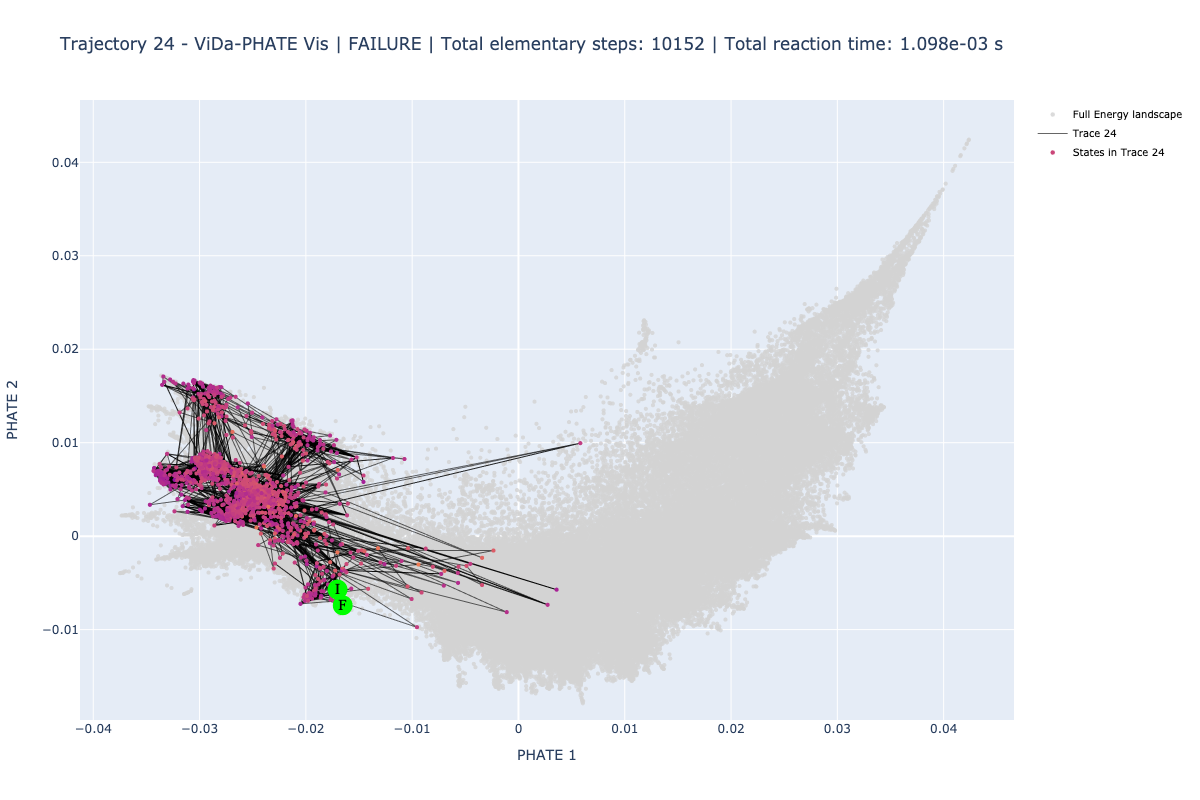}
    \caption{Visualization plot for the perfect-toehold7-reporter reaction. The black trace represents a failed trajectory.}
    \label{fig:per_t7_dangel_fail2}
\end{figure}

\clearpage

\section{Visualization Plots for Proximal-toehold8}  \label{sup:prox_t8}

\begin{figure}[!ht]
    \centering
    \includegraphics[width=.8\linewidth, trim={2.8cm 2.8cm 6cm 2cm}, clip]
    {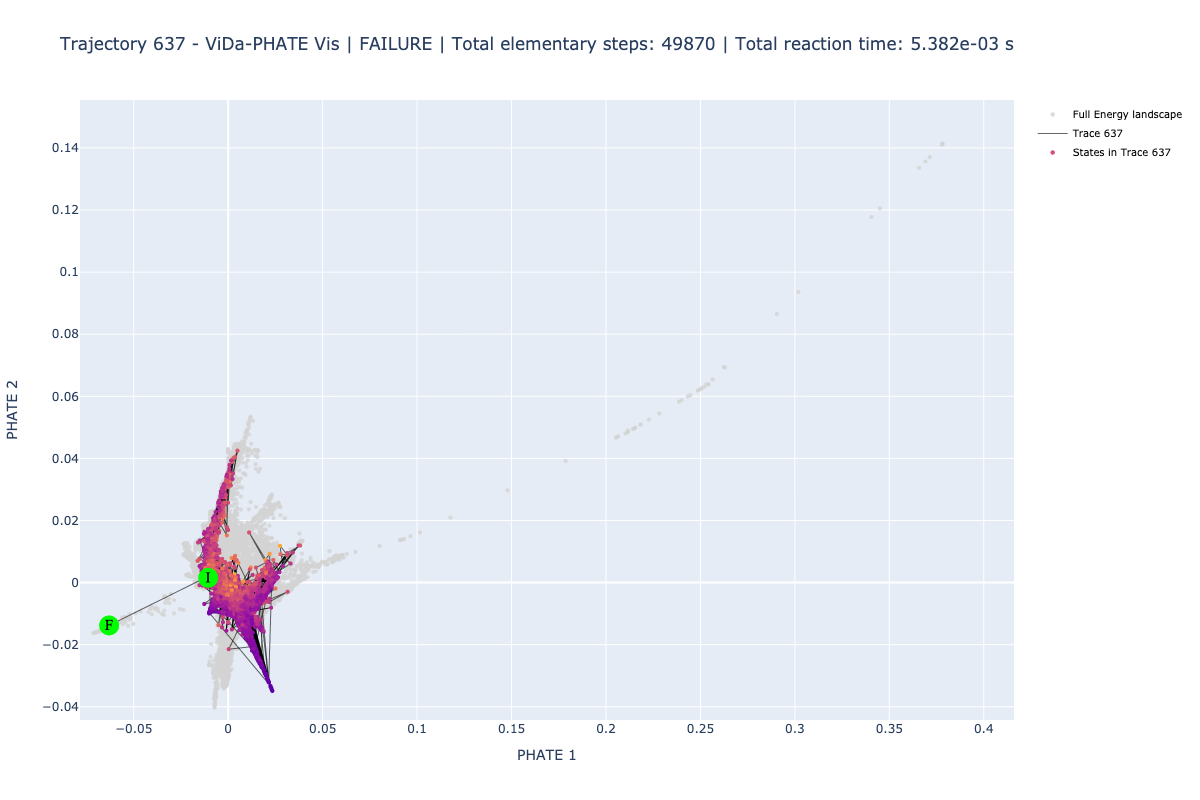}
    \caption{Visualization plot for the proximal-toehold8 reaction. The black trace represents a failed trajectory.}
    \label{fig:prox_t8_fail1}
\end{figure}

\begin{figure}[!ht]
    \centering
    \includegraphics[width=.8\linewidth, trim={2.8cm 2.8cm 6cm 2cm}, clip]
    {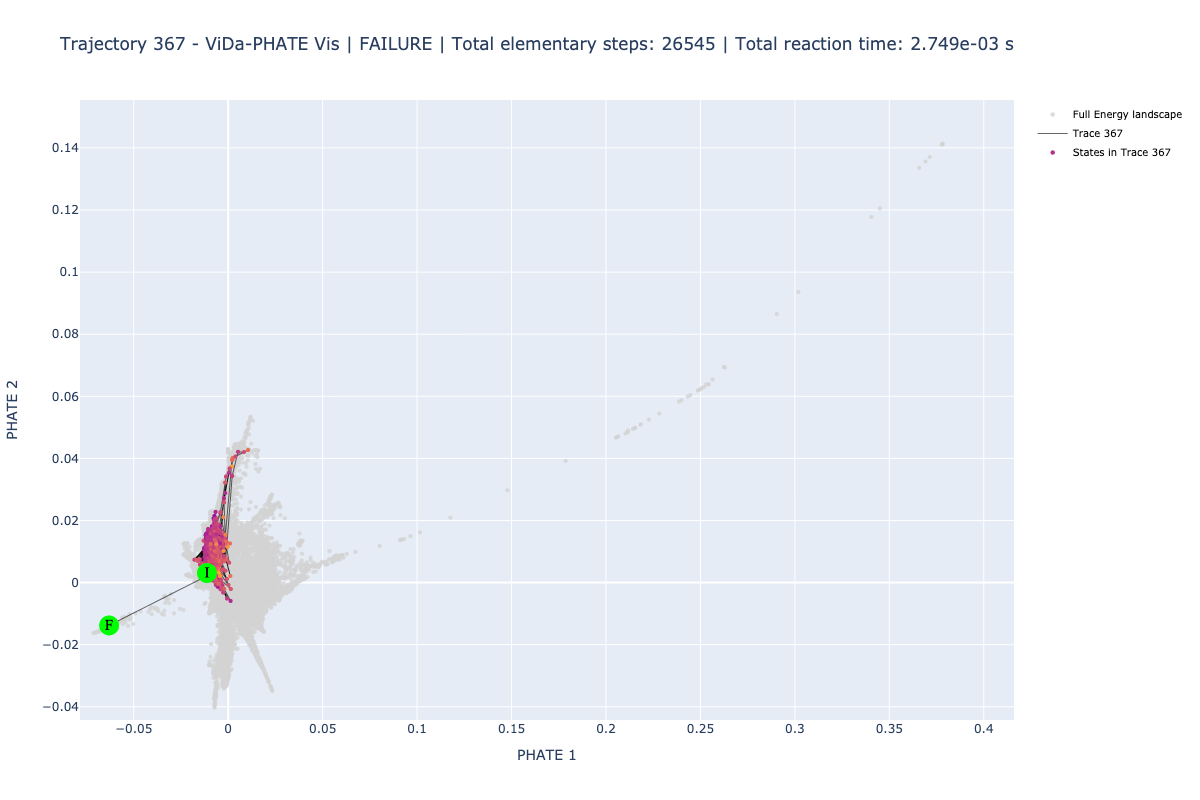}
    \caption{Visualization plot for the proximal-toehold8 reaction. The black trace represents a failed trajectory.}
    \label{fig:prox_t8_fail2}
\end{figure}

\begin{figure}[!ht]
    \centering
    \includegraphics[width=.8\linewidth, trim={2.8cm 2.8cm 6cm 2cm}, clip]
    {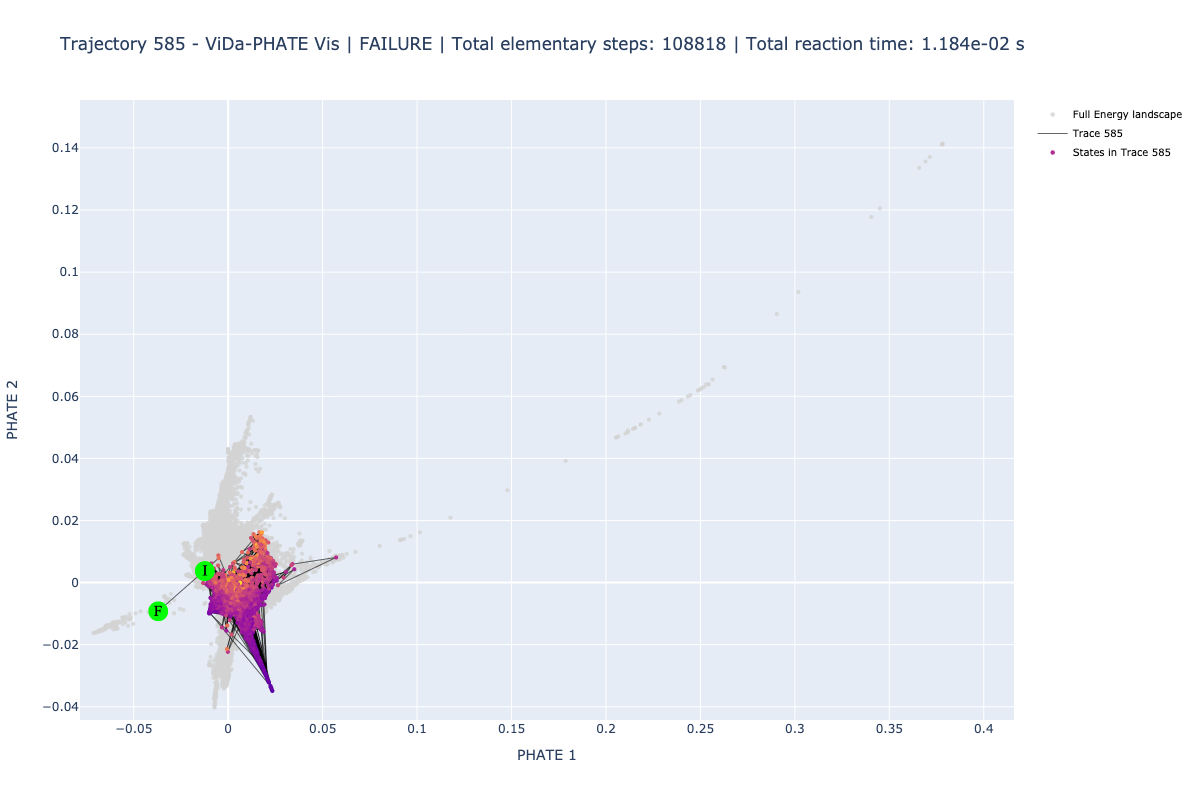}
    \caption{Visualization plot for the proximal-toehold8 reaction. The black trace represents a failed trajectory.}
    \label{fig:prox_t8_fail3}
\end{figure}

\begin{figure}[!ht]
    \centering
    \includegraphics[width=.8\linewidth, trim={2.8cm 2.8cm 6cm 2cm}, clip]
    {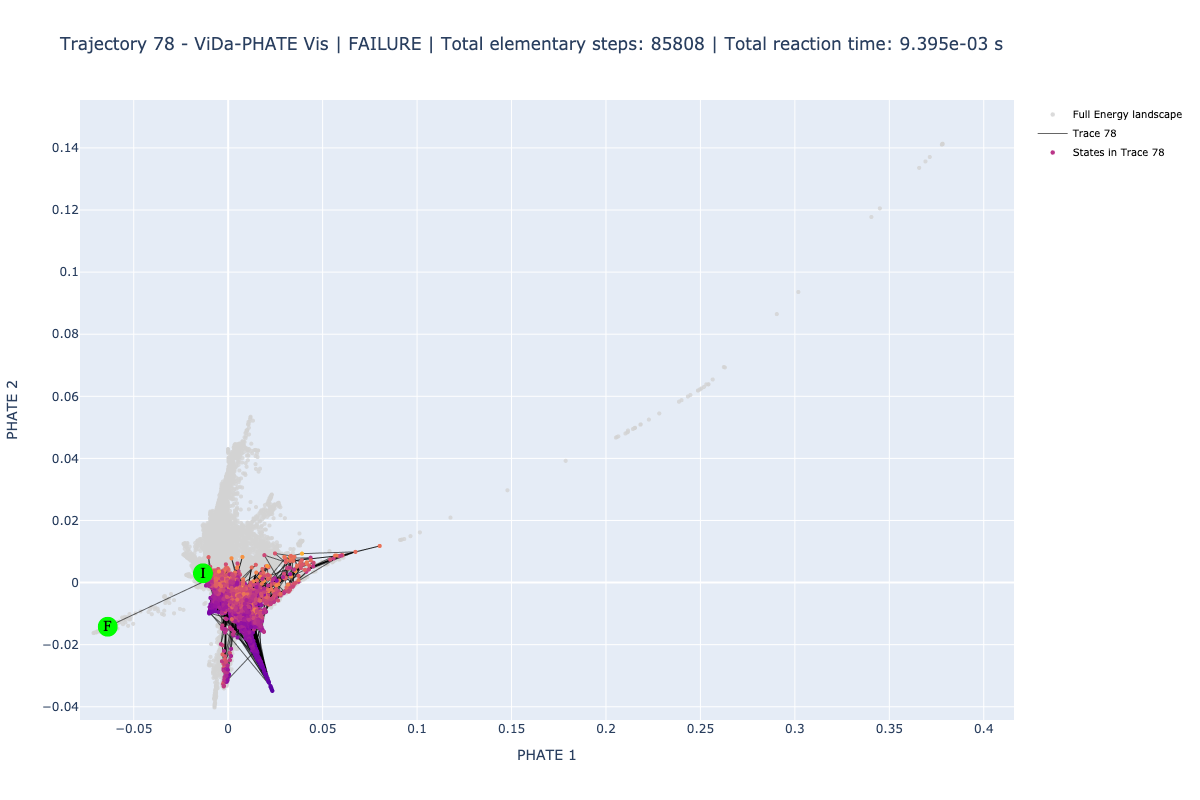}
    \caption{Visualization plot for the proximal-toehold8 reaction. The black trace represents a failed trajectory.}
    \label{fig:prox_t8_fail4}
\end{figure}

\clearpage

\section{Visualization Plots for Proximal-toehold7-reporter}  \label{sup:prox_t7_reporter}

\begin{figure}[!ht]
    \centering
    \includegraphics[width=.8\linewidth, trim={2.8cm 2.8cm 6cm 2cm}, clip]
    {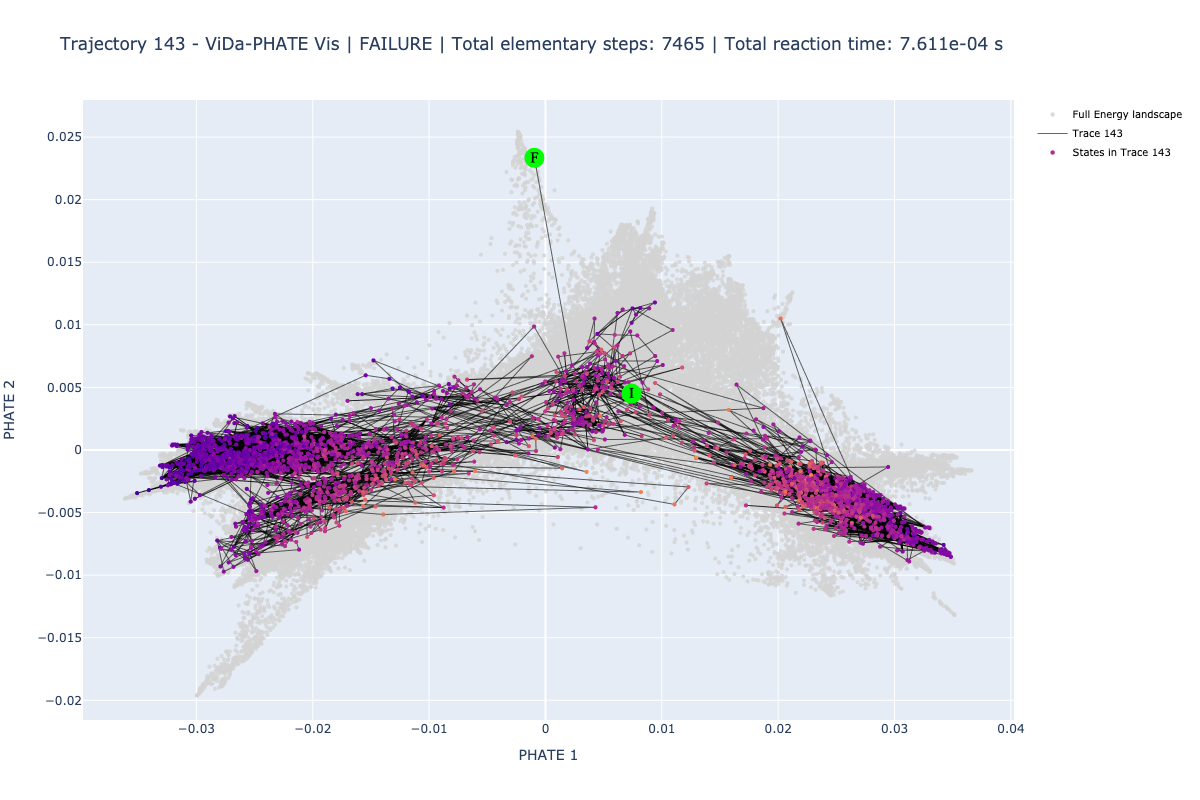}
    \caption{Visualization plot for the proximal-toehold7-reporter reaction. The black trace represents a failed trajectory.}
    \label{fig:prox_t7_dangel_fail1}
\end{figure}

\begin{figure}[!ht]
    \centering
    \includegraphics[width=.8\linewidth, trim={2.8cm 2.8cm 6cm 2cm}, clip]
    {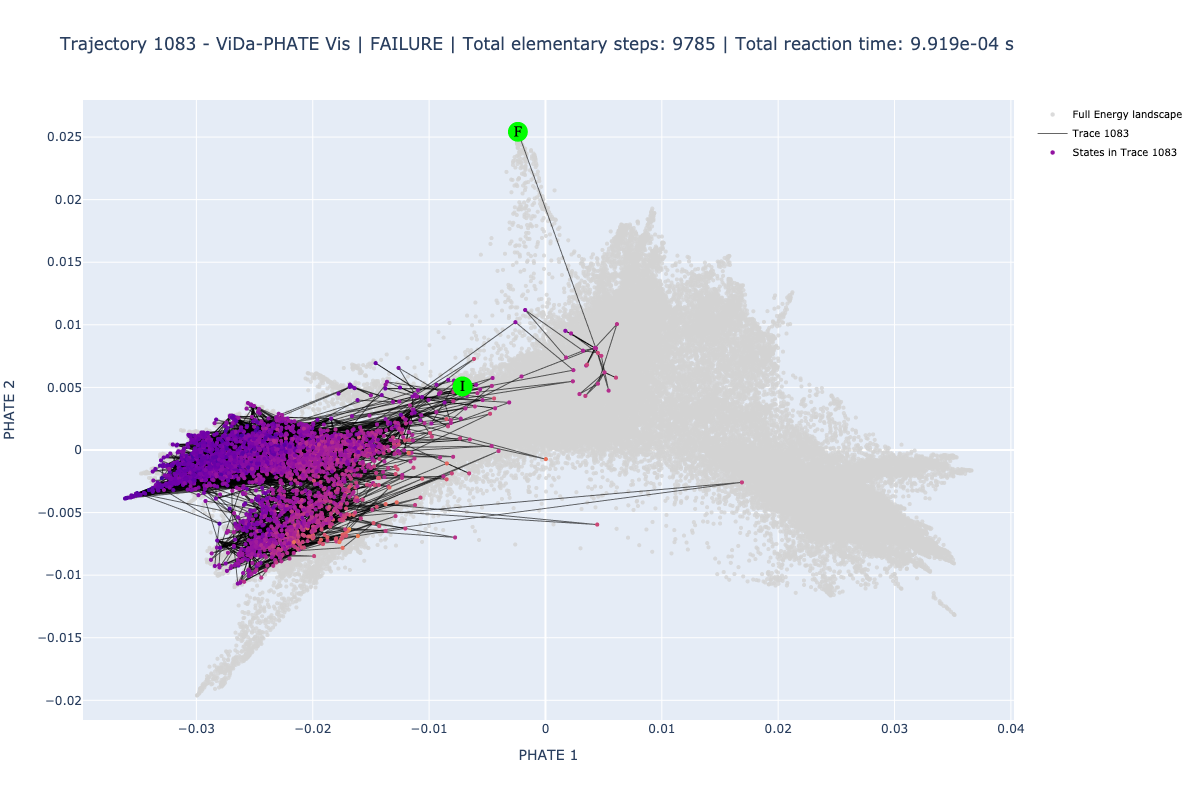}
    \caption{Visualization plot for the proximal-toehold7-reporter reaction. The black trace represents a failed trajectory.}
    \label{fig:prox_t7_dangel_fail2}
\end{figure}

\begin{figure}[!ht]
    \centering
    \includegraphics[width=.8\linewidth, trim={2.8cm 2.8cm 6cm 2cm}, clip]
    {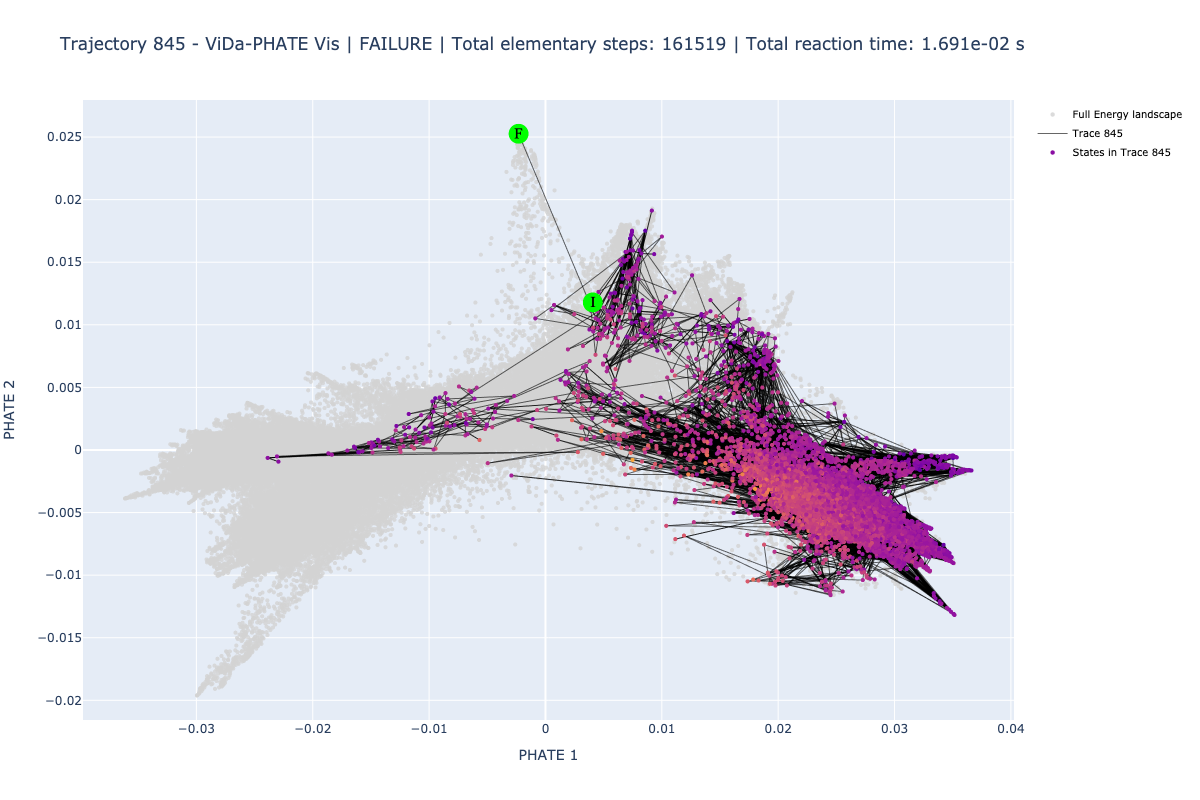}
    \caption{Visualization plot for the proximal-toehold7-reporter reaction. The black trace represents a failed trajectory.}
    \label{fig:prox_t7_dangel_fail3}
\end{figure}

\begin{figure}[!ht]
    \centering
    \includegraphics[width=.8\linewidth, trim={2.8cm 2.8cm 6cm 2cm}, clip]
    {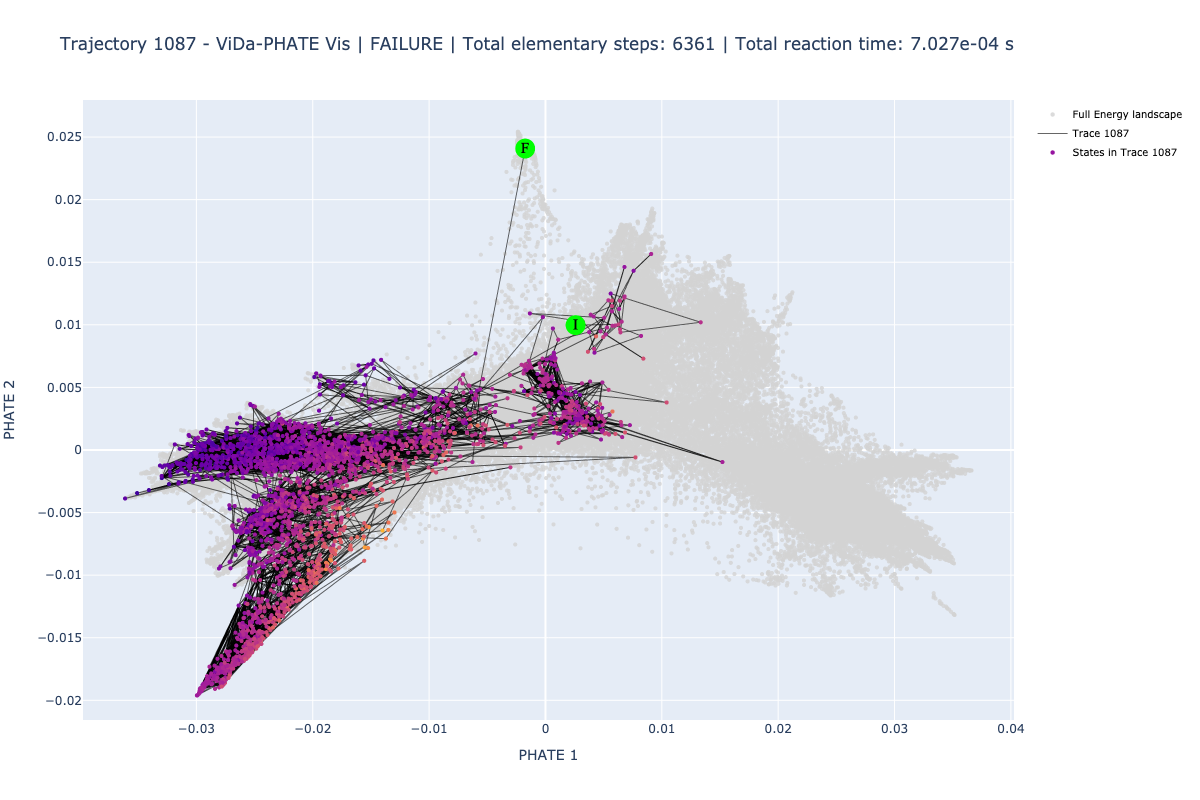}
    \caption{Visualization plot for the proximal-toehold7-reporter reaction. The black trace represents a failed trajectory.}
    \label{fig:prox_t7_dangel_fail4}
\end{figure}


\chapter{Supplementary Materials for Chapter 6}


\section{Benchmarking Implementation} \label{metric_implementation}

We benchmarked the following software versions: Phenix-v1.20, ModelAngelo-v1.0.12, EMBuild (as of February 15, 2024), and DeepMainmast (as of March 13, 2024). 
Rosetta-v2021.16 was used to assist DeepMainmast in producing full-atom models.
All evaluated DL-based approaches used pre-trained networks provided by the developers.
The AlphaFold predictions used for both EMBuild and DeepMainmast were obtained from the \texttt{AlphaFold Protein Structure Database}. For those predictions that were not available in the database, we generated them locally using AlphaFold-v2.3.2. 
All methods were implemented on one NVIDIA A100 GPU with AMD EPYC 7V12 64-Core Processor.
Visualization of cryo-EM density maps and atomic models was produced by UCSF ChimeraX.

\clearpage


\section{Experimental Datasets}

Table \ref{tab:metric_emid_data} lists 50 samples used for benchmarking atomic model building approaches.

\begin{table}[!ht]
\centering
\caption{Summary of benchmarking data.}
\vspace{1em}
\resizebox{\textwidth}{!}{%
\label{tab:metric_emid_data}
\begin{tabular}{@{}ccccc@{}} 
\toprule
EMDB ID  & PDB ID & Resolution (\AA) & \# of Residues & Stoichiometry \\ 
\midrule
26917 & 7UZS & 2.2 & 691 & Monomer - A1 \\
26948 & 7V0Q & 2.5 & 691 & Monomer - A1 \\
33187 & 7XGR & 2.6 & 671 & Monomer - A1 \\
33233 & 7XJP & 2.71 & 1160 & Hetero 2-mer - A1B1 \\
33243 & 7XK4 & 3.1 & 1940 & Hetero 6-mer - A1B1C1D1E1F \\
33306 & 7XMV & 2.6 & 1926 & Homo 6-mer - A6 \\
33331 & 7XNZ & 3.6 & 1912 & Homo 4-mer - A4 \\
33861 & 7YIM & 2.6 & 609 & Monomer - A1 \\
34017 & 7YPX & 3.12 & 1353 & Hetero 6-mer - A3B3 \\
14716 & 7ZH0 & 3.2 & 556 & Monomer - A1 \\
14725 & 7ZH6 & 3.67 & 556 & Monomer - A1 \\
15047 & 8A04 & 3.2 & 717 & Homo 3-mer - A3 \\
15220 & 8A7D & 3.06 & 3239 & Hetero 3-mer - A1B1C1 \\
15560 & 8AP7 & 2.7 & 4822 & Hetero 30-mer - A2B2C2D2E2F2G2H2I2J2K2L2M2N2O2 \\
15561 & 8AP8 & 3.7 & 1165 & Hetero 5-mer - A1B1C1D1E1 \\
15686 & 8AVX & 3.5 & 1876 & Homo 2-mer - A2 \\
15960 & 8BC2 & 2.6 & 2180 & Homo 10-mer - A10 \\
26973 & 8CSW & 2.5 & 691 & Monomer - A1 \\
27022 & 8CVZ & 3.52 & 4648 & Hetero 10-mer - A4B4C2 \\
27431 & 8DH7 & 2.99 & 1052 & Homo 2-mer - A2 \\
27574 & 8DNM & 2.76 & 2288 & Homo 4-mer - A4 \\
27755 & 8DWI & 3.4 & 495 & Monomer - A1 \\
27760 & 8DWU & 3.4 & 3366 & Hetero 9-mer - A5B3C1 \\
27899 & 8E50 & 3.67 & 767 & Monomer - A1 \\
28081 & 8EFE & 3.8 & 1053 & Hetero 2-mer - A1B1 \\
28637 & 8EVU & 2.58 & 1934 & Hetero 6-mer - A2B1C1D1E1 \\
29290 & 8FMA & 3.1 & 22242 & Hetero 22-mer - A11B11 \\
27842 & 8E2L & 3.51 & 3267 & Hetero 6-mer - A4B2 \\
8637  & 5V6P & 4.1 & 814  & Homo 2-mer - A2 \\
26626 & 7UNL & 2.45 & 1008 & Homo 2-mer - A2 \\
6272  & 3J9S & 2.6  & 1194 & Homo 3-mer - A3 \\
8331  & 5SZS & 3.4  & 3975 & Homo 3-mer - A3 \\
2984  & 5A1A & 2.2  & 4088 & Homo 4-mer - A4 \\
8194  & 5K12 & 1.8  & 3348 & Homo 6-mer - A6 \\
26126 & 7TU5 & 2.1  & 2784 & Homo 6-mer - A6 \\
24783 & 7RZY & 3.5  & 2310 & Homo 7-mer - A7 \\
6724  & 5XJY & 4.1  & 2305 & Monomer - A1 \\
0346  & 6N52 & 4    & 1742 & Homo 2-mer - A2 \\
25797 & 7TBF & 3.1  & 873  & Hetero 5-mer - A1B1C1D1E1 \\
6770  & 5XSY & 4    & 2029 & Hetero 2-mer - A1B1 \\
8482  & 5U1D & 3.97 & 1522 & Hetero 3-mer - A1B1C1 \\
3240  & 5FN5 & 4.3  & 1542 & Hetero 4-mer - A1B1C1D1 \\
8623  & 5UZ7 & 4.1  & 1438 & Hetero 5-mer - A1B1C1D1E1 \\
8461  & 5UAR & 3.73 & 1494 & Monomer - A1 \\
8560  & 5UJA & 3.34 & 1460 & Monomer - A1 \\
8642  & 5V7V & 3.9  & 818  & Monomer - A1 \\
8782  & 5W81 & 3.37 & 1494 & Monomer - A1 \\
9317  & 6N1Q & 5.16 & 4088 & Homo 8-mer - A8 \\
35136 & 8I2H & 6    & 682  & Monomer - A1 \\
8685  & 5VHW & 7.8  & 4228 & Homo 4-mer - A4 \\
\bottomrule
\end{tabular}
}
\end{table}

\clearpage


\section{Distortions Using Rosetta}

\begin{figure}[!ht]
\centering
\includegraphics[width=\linewidth, trim={1.cm 13cm 5cm 0cm}, clip]{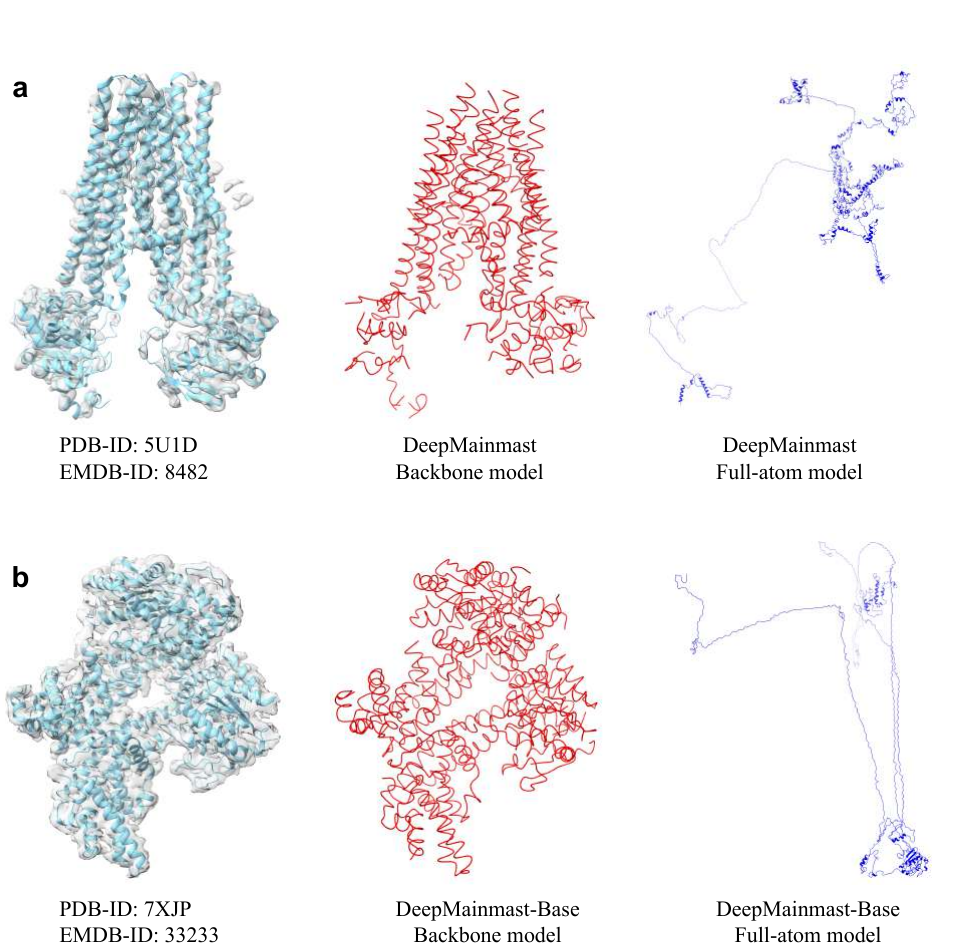}
\caption{ 
Left: the reference PDB structures colored in cyan superimposes onto the corresponding cryo-EM density maps colored in transparent gray. Middle: backbone models constructed by DeepMainmast (at \textbf{a}) and DeepMainmast-Base (at \textbf{b}), colored in red. Right: distorted full-atom models constructed by DeepMainmast (at \textbf{a}) and DeepMainmast-Base (at \textbf{b}), colored in blue.
\textbf{(a)} The human TAP ATP-Binding Cassette Transporter (PDB-ID: 5U1D; EMDB-ID: 8482; Resolution: 3.97 {\AA}) \cite{5U1D}.
\textbf{(b)} EDS1 and SAG101 with ATP-APDR (PDB-ID: 7XJP; EMDB-ID: 33233; Resolution: 2.71 {\AA}) \cite{7XJP}.
}
\label{fig:Metric_dmm_problem}
\end{figure}

Figure \ref{fig:Metric_dmm_problem} shows the distortions
in some full-atom models derived from DeepMainmast's backbone models using Rosetta-v2021.16.

\section{Comparison of Existing Metrics}

\begin{figure}[!ht]
\centering
\includegraphics[width=0.85\linewidth, trim={5.5cm 3.5cm 5.5cm 4.5cm}, clip]{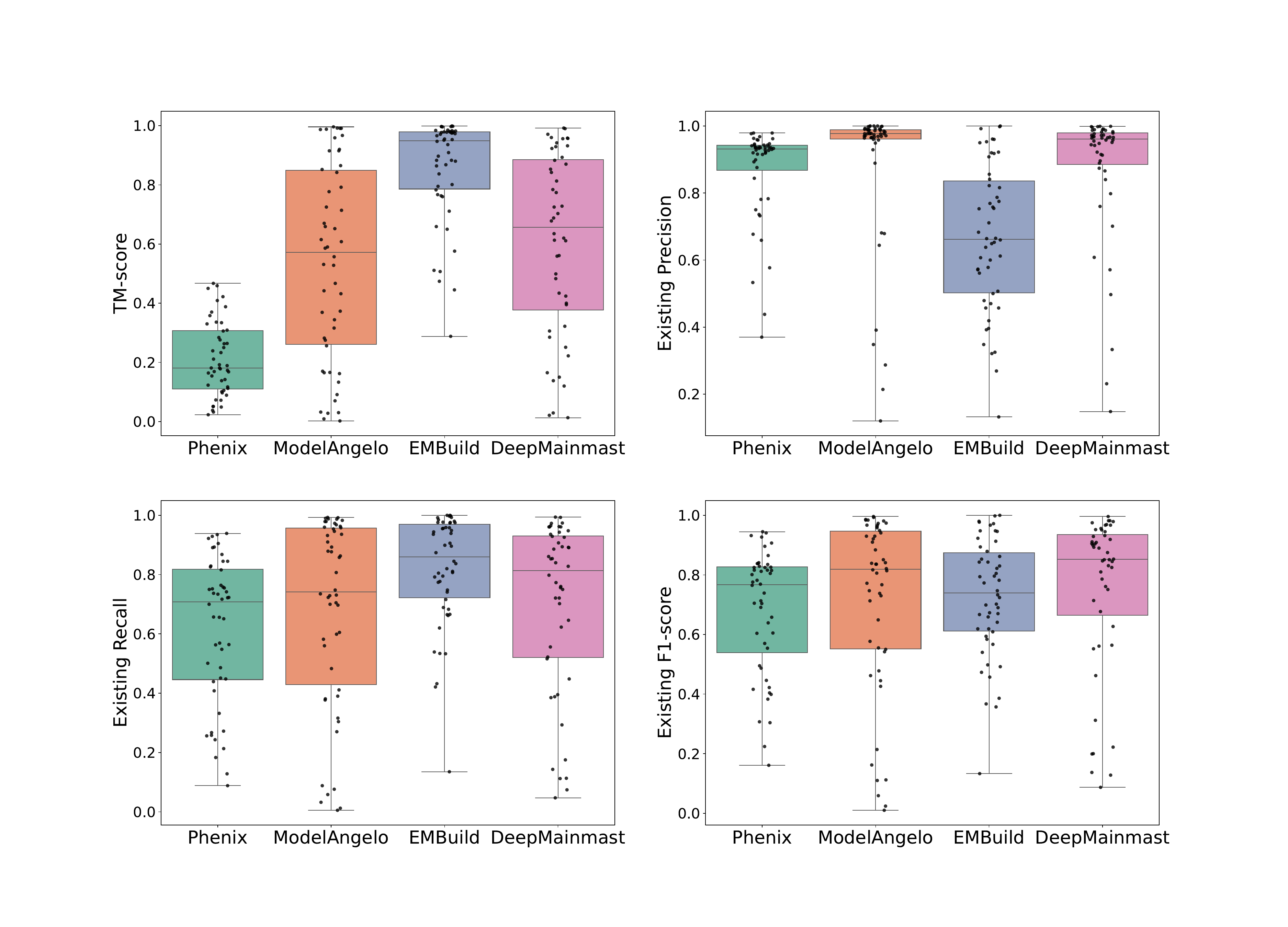}
\caption{ 
Comparison of existing metrics (TM-score, Existing Precision, Existing Recall, and Existing F1-score) of 50 protein models generated by Phenix, ModelAngelo, EMBuild, and DeepMainmast. The dots represent the scores for each protein model.
}
\label{fig:Metric_boxplot_oldmetric}
\end{figure}

Figure \ref{fig:Metric_boxplot_oldmetric} shows the comparison of existing metrics over 50 samples. See the main texts for the explanation of the figure.

\clearpage

\section{Additional Pairwise Pearson Correlation Scatter Plots}

\begin{figure}[!ht]
\centering
\includegraphics[width=\linewidth, trim={0cm 0cm 0cm 0cm}, clip]{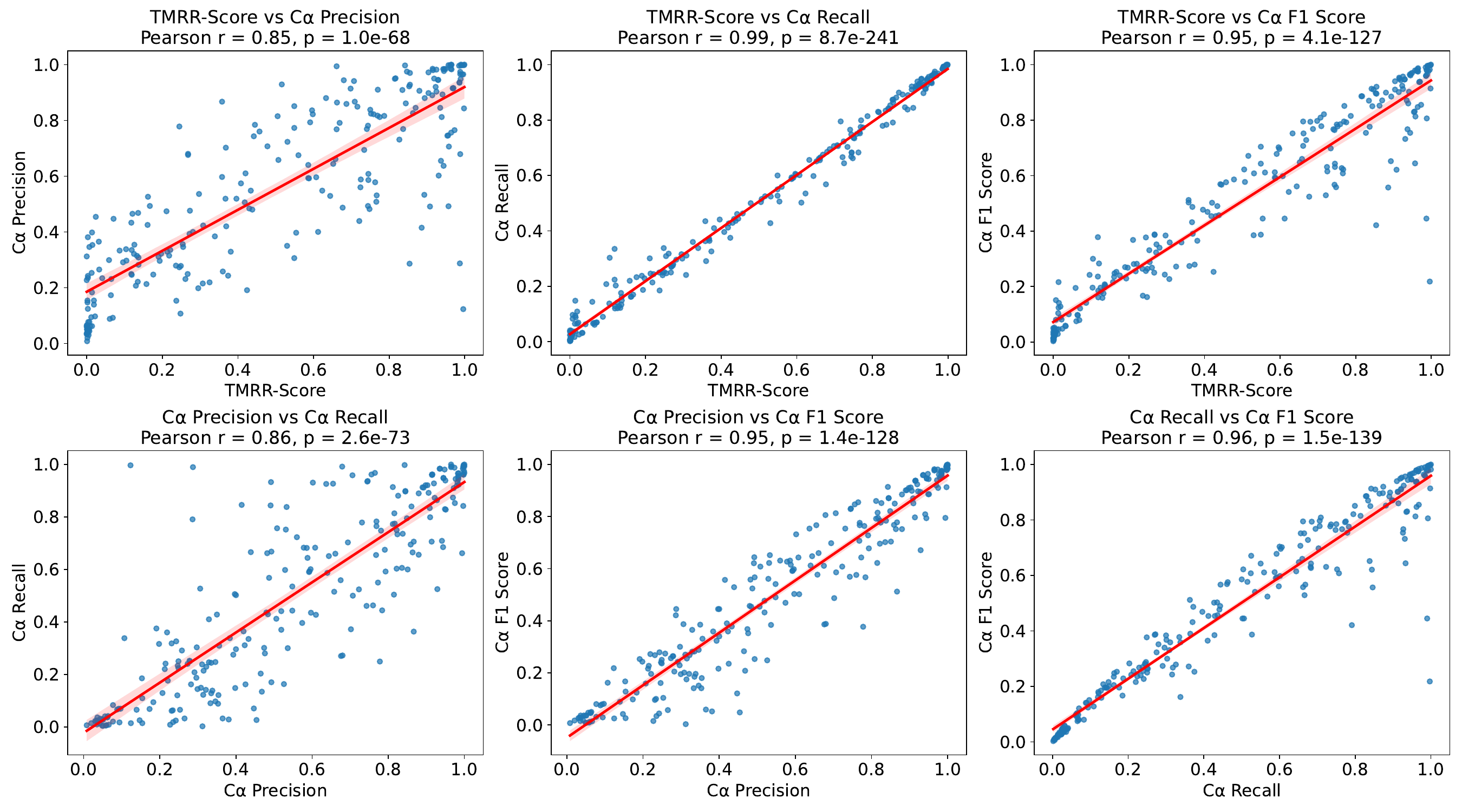}
\caption{
Pairwise Pearson correlation scatter plots between improved evaluation metrics (TMRR-score, C$_\alpha$ Precision, C$_\alpha$ Recall, and C$_\alpha$ F1-score) across all tested samples.
}
\label{fig:Metric_corr_scatter_improved}
\end{figure}

\begin{figure}[!ht]
\centering
\includegraphics[width=\linewidth, trim={0cm 0cm 0cm 0cm}, clip]{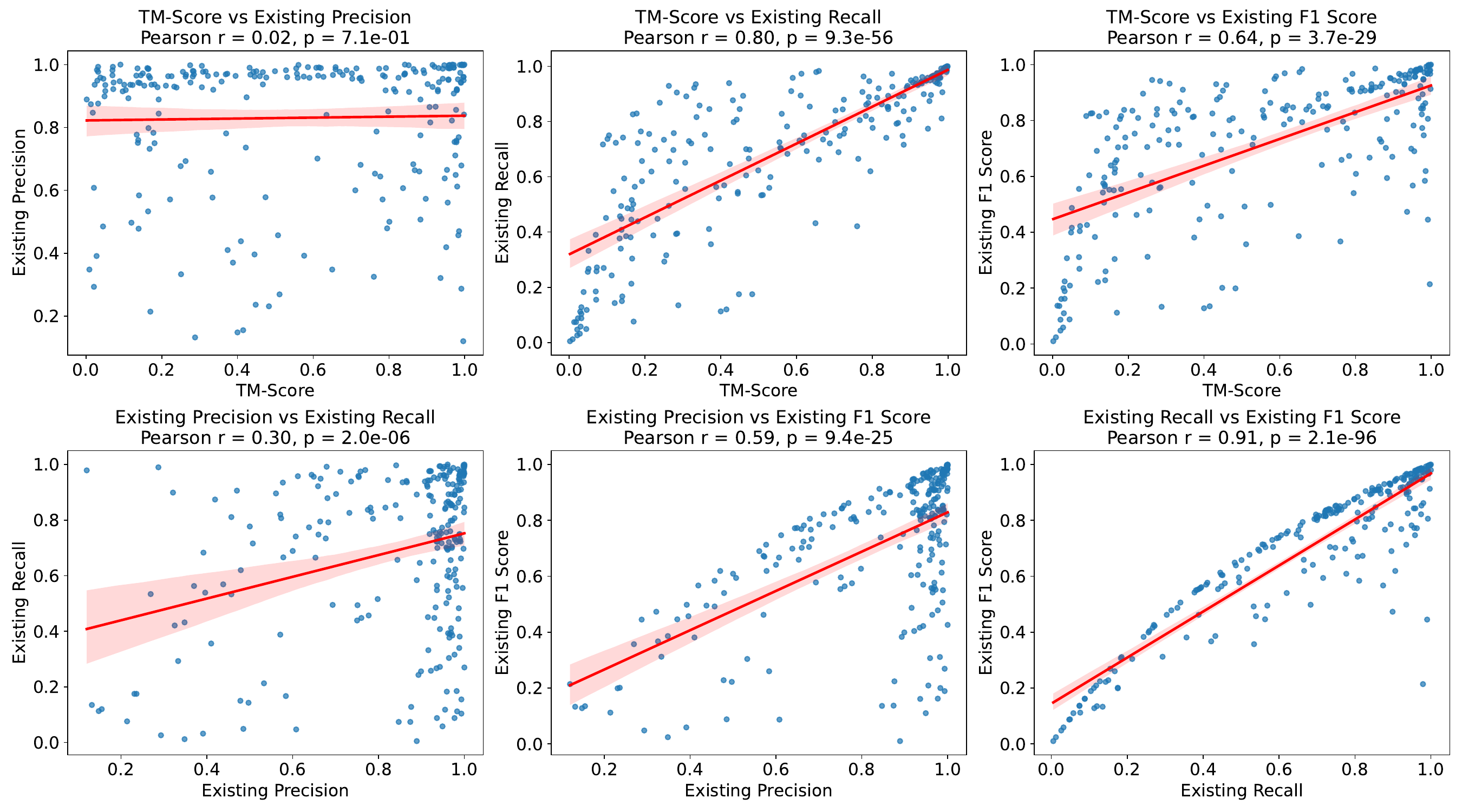}
\caption{
Pairwise Pearson correlation scatter plots between existing evaluation metrics (TM-score, Existing Precision, Existing Recall, and Existing F1-score) across all tested samples.
}
\label{fig:Metric_corr_scatter_existing}
\end{figure}

Figures \ref{fig:Metric_corr_scatter_improved} and \ref{fig:Metric_corr_scatter_existing} show pairwise Pearson correlation scatter plots between improved and existing evaluation metrics, respectively.
See the main texts for the explanation of the figure.

\clearpage

\section{Additional Heat Maps}

\begin{figure}[!ht]
\centering
\includegraphics[width=\linewidth, trim={0cm 0cm 0cm 0cm}, clip]{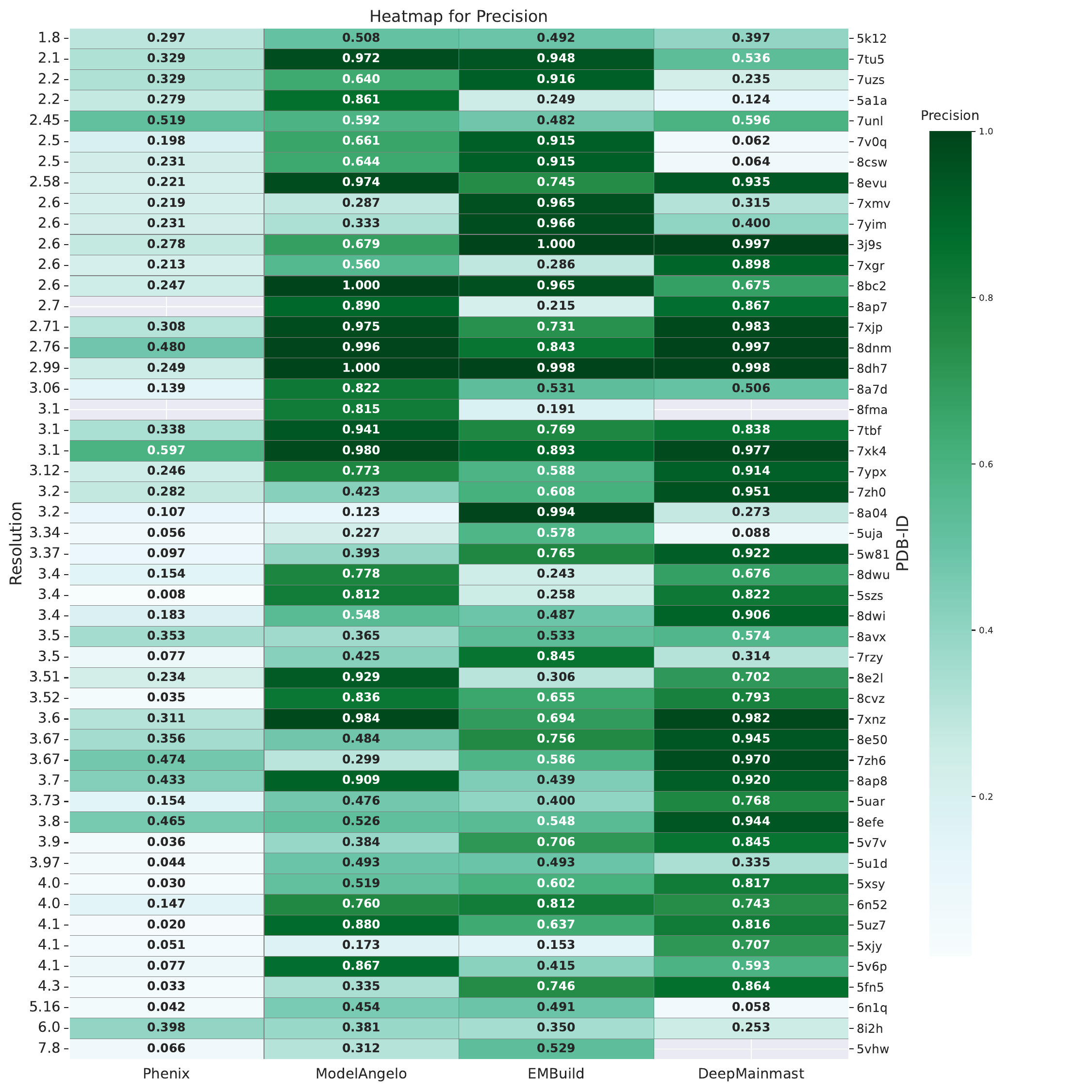}
\caption{ 
The heat map for C$_\alpha$ precision of 50 protein models generated by Phenix, ModelAngelo, EMBuild, and DeepMainmast, sorted by their map resolution from high to low. The darker color refers to the higher score. The value in each cell represents the specific score for the generated model. The left labels shows the resolutions and the right labels shows the PDB IDs of each generated model.
}
\label{fig:Metric_hmap_precision}
\end{figure}

\begin{figure}[!ht]
\centering
\includegraphics[width=\linewidth, trim={0cm 0cm 0cm 0cm}, clip]{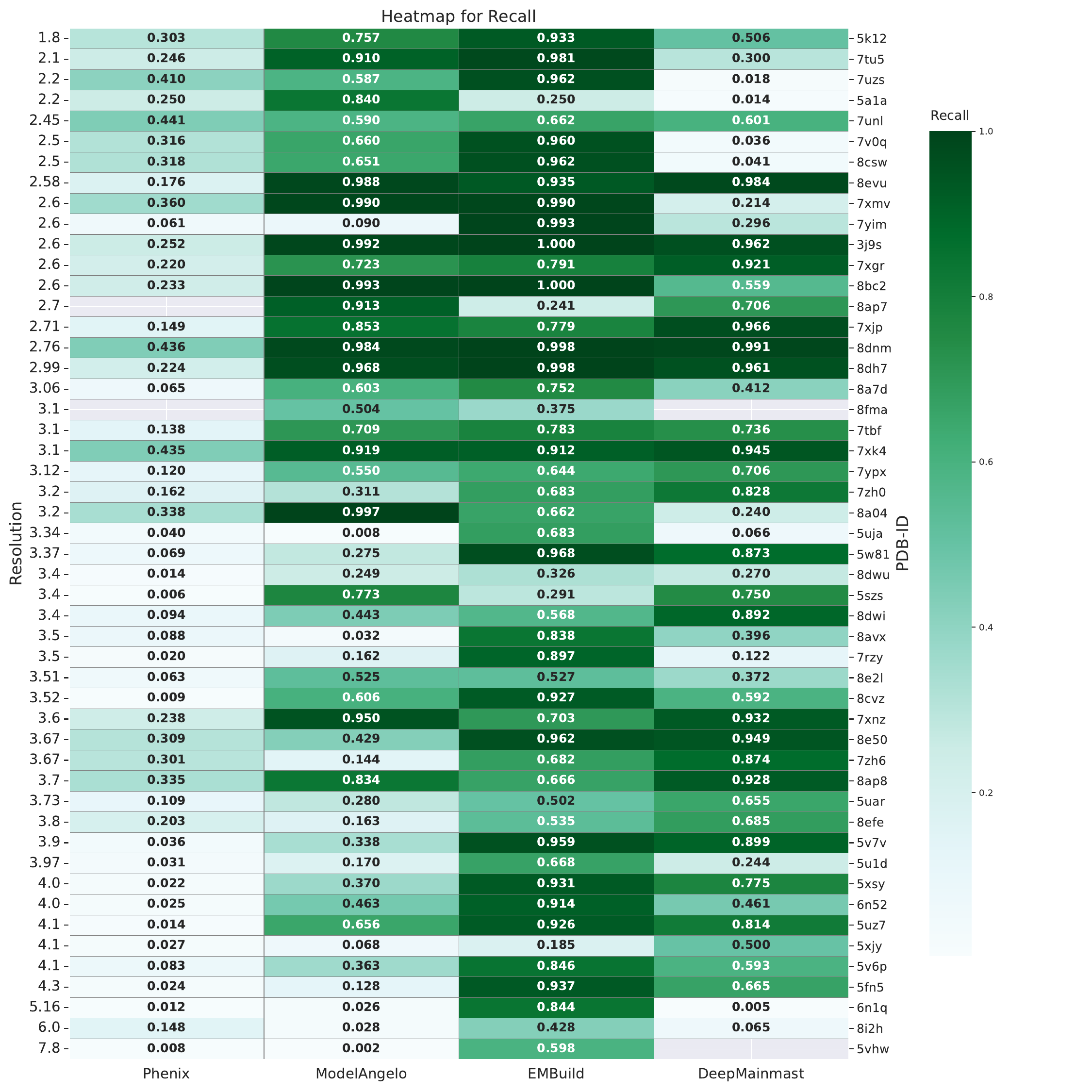}
\caption{ 
The heat map for C$_\alpha$ recall of 50 protein models generated by Phenix, ModelAngelo, EMBuild, and DeepMainmast, sorted by their map resolution from high to low. The darker color refers to the higher score. The value in each cell represents the specific score for the generated model. The left labels shows the resolutions and the right labels shows the PDB IDs of each generated model.
}
\label{fig:Metric_hmap_recall}
\end{figure}

\begin{figure}[!ht]
\centering
\includegraphics[width=\linewidth, trim={0cm 0cm 0cm 0cm}, clip]{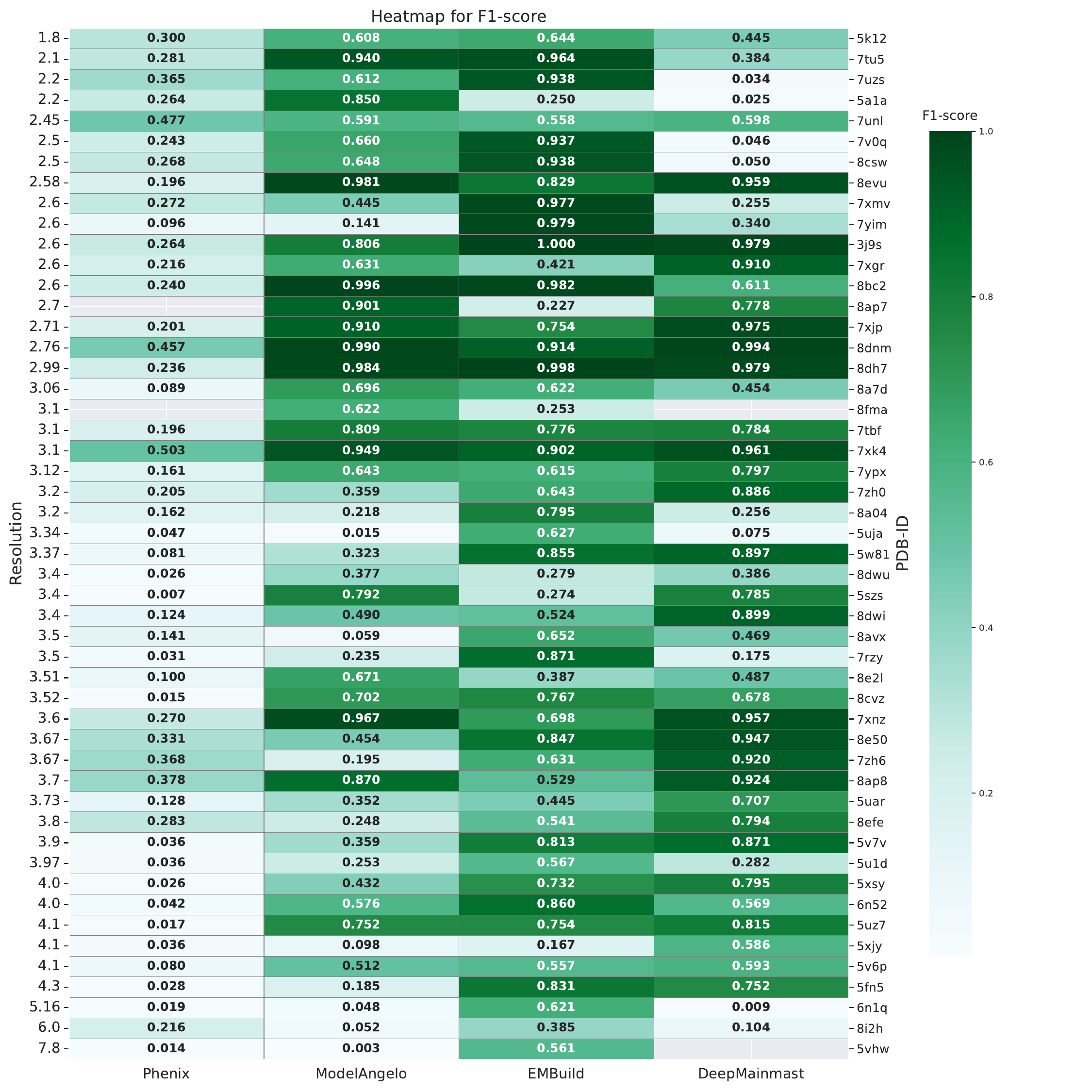}
\caption{ 
The heat map for C$_\alpha$ F1-score of 50 protein models generated by Phenix, ModelAngelo, EMBuild, and DeepMainmast, sorted by their map resolution from high to low. The darker color refers to the higher score. The value in each cell represents the specific score for the generated model. The left labels shows the resolutions and the right labels shows the PDB IDs of each generated model.
}
\label{fig:Metric_hmap_f1score}
\end{figure}

Figures \ref{fig:Metric_hmap_precision}, \ref{fig:Metric_hmap_recall}, and \ref{fig:Metric_hmap_f1score} show heat maps for precision, recall, and F1-score, respectively.
See the main texts for the explanation of the figure.

\chapter{Supplementary Materials for Chapter 7}

\section{Experimental Datasets}

Tables \ref{tab:gan_trainvaltb} and \ref{tab:gan_testtb} list the cryo-EM density maps and associated PDB structures that we used for training and testing our struc2mapGAN model, respectively.


\begin{table}[!ht]
\centering
\caption{List of all EMDB/PDB examples in training and validation sets. Examples in bold are used as the validation set.}
\vspace{1em}
\scriptsize
\begin{tabular}{ccc||ccc||ccc}
\hline
EMDB ID & PDB ID & Res. (\AA) & EMDB ID & PDB ID & Res. (\AA) & EMDB ID & PDB ID & Res. (\AA) \\
\hline
26916 & 7UZQ & 2.17 & 26617 & 7UN9 & 3.3 & 0310 & 6HZ4 & 3.6 \\
26917 & 7UZS & 2.2 & 34963 & 8HR8 & 3.3 & 6324 & 3JA7 & 3.6 \\
26886 & 7UZE & 2.4 & 0089 & 6GYB & 3.3 & 9105 & 6ME0 & 3.6 \\
26974 & 8CSX & 2.4 & 0690 & 6J6K & 3.3 & 9111 & 6MG8 & 3.6 \\
26948 & 7V0Q & 2.5 & 3297 & 5FTL & 3.3 & 30346 & 7CFS & 3.6 \\
26949 & 7V0S & 2.5 & 4571 & 6QK7 & 3.3 & 34738 & 8HGG & 3.64 \\
14774 & 7ZL1 & 2.5 & 6239 & 3J9D & 3.3 & 15635 & 8AT6 & 3.7 \\
15960 & 8BC2 & 2.6 & 6240 & 3J9E & 3.3 & 0609 & 6O2N & 3.7 \\
33233 & 7XJP & 2.71 & 6630 & 3JCZ & 3.3 & 7041 & 6B3Q & 3.7 \\
27937 & 8E78 & 2.77 & 6932 & 5ZJI & 3.3 & 8946 & 6E0F & 3.7 \\
27945 & 8E8O & 2.77 & 6975 & 5ZX5 & 3.3 & 10099 & 6S59 & 3.7 \\
26732 & 7USC & 3 & 9390 & 6NJO & 3.3 & 10399 & 6T8B & 3.7 \\
26733 & 7USD & 3 & 10037 & 6RWB & 3.3 & 20754 & 6UEN & 3.7 \\
26734 & 7USE & 3 & 10836 & 6YLE & 3.3 & 20857 & 6UR8 & 3.7 \\
0785 & 6KZ4 & 3 & 21863 & 6WPK & 3.3 & 0379 & 6N9U & 3.7 \\
9322 & 6N24 & 3 & 7006 & 6AUI & 3.3 & 0688 & 6J6I & 3.7 \\
20442 & 6PPL & 3 & 9898 & 6JZO & 3.3 & 4650 & 6QVE & 3.7 \\
21149 & 6VCD & 3 & 20583 & 6TYI & 3.3 & 20651 & 6U5O & 3.7 \\
21366 & 6VRB & 3 & 20968 & 6V03 & 3.3 & 30238 & 7BXU & 3.7 \\
30305 & 7C8D & 3 & 0327 & 6I1Y & 3.4 & 27645 & 8DQ0 & 3.74 \\
30535 & 7D0I & 3 & 0923 & 6LMX & 3.4 & 28080 & 8EFD & 3.8 \\
6714 & 5XB1 & 3 & 0959 & 6LRR & 3.4 & 28081 & 8EFE & 3.8 \\
21604 & 6WCA & 3 & 9116 & 6MHQ & 3.4 & 0935 & 6LO8 & 3.8 \\
26838 & 7UWQ & 3.05 & 9398 & 6NM9 & 3.4 & 7075 & 6BBJ & 3.8 \\
26978 & 8CT2 & 3.1 & 9887 & 6KUJ & 3.4 & 9971 & 6KFF & 3.8 \\
0590 & 6O1K & 3.1 & 9905 & 6K15 & 3.4 & 10401 & 6T8H & 3.8 \\
0784 & 6KZ3 & 3.1 & 9954 & 6KA4 & 3.4 & 20700 & 6U9H & 3.8 \\
9253 & 6MUR & 3.1 & 10208 & 6SI8 & 3.4 & 20471 & 6PTJ & 3.8 \\
10539 & 6TNY & 3.1 & 10232 & 6SKZ & 3.4 & 0706 & 6KLC & 3.9 \\
22876 & 7KHA & 3.1 & 21913 & 6WUH & 3.4 & 7959 & 6DLZ & 3.9 \\
30016 & 6LYG & 3.1 & 10206 & 6SI7 & 3.4 & 8185 & 5JZH & 3.9 \\
3999 & 6EZJ & 3.1 & 33621 & 7Y5N & 3.45 & 10893 & 6YS8 & 3.9 \\
14716 & 7ZH0 & 3.2 & 26841 & 7UWS & 3.47 & 4749 & 6R81 & 3.9 \\
0936 & 6LOD & 3.2 & 15673 & 8AUR & 3.47 & 4917 & 6RLA & 3.9 \\
4789 & 6RB9 & 3.2 & 34430 & 8H1P & 3.48 & \textbf{33719} & \textbf{7YAT} & \textbf{2.2} \\
4907 & 6RKD & 3.2 & 28666 & 8EY2 & 3.5 & \textbf{26973} & \textbf{8CSW} & \textbf{2.5} \\
9653 & 6IFK & 3.2 & 4581 & 6QLF & 3.5 & \textbf{15361} & \textbf{8ADE} & \textbf{2.78} \\
20650 & 6U5N & 3.2 & 8750 & 5W0S & 3.5 & \textbf{4798} & \textbf{6RBG} & \textbf{3} \\
21146 & 6VBW & 3.2 & 9104 & 6MDR & 3.5 & \textbf{6941} & \textbf{5ZR1} & \textbf{3} \\
21586 & 6WB8 & 3.2 & 9187 & 6MP6 & 3.5 & \textbf{15646} & \textbf{8ATD} & \textbf{3.1} \\
30004 & 6LX3 & 3.2 & 10528 & 6TMV & 3.5 & \textbf{4890} & \textbf{6RIE} & \textbf{3.1} \\
30021 & 6LZ1 & 3.2 & 10573 & 6TT7 & 3.5 & \textbf{15540} & \textbf{8ANE} & \textbf{3.2} \\
30022 & 6LZ3 & 3.2 & 20446 & 6PPR & 3.5 & \textbf{6777} & \textbf{5XWY} & \textbf{3.2} \\
30334 & 7CAL & 3.2 & 20498 & 6PW4 & 3.5 & \textbf{0136} & \textbf{6H3N} & \textbf{3.3} \\
0408 & 6NBD & 3.2 & 20767 & 6UH5 & 3.5 & \textbf{8795} & \textbf{5WC3} & \textbf{3.5} \\
5995 & 3J7H & 3.2 & 22042 & 6X4S & 3.5 & \textbf{0921} & \textbf{6LMV} & \textbf{3.6} \\
20236 & 6P25 & 3.2 & 4746 & 6R7X & 3.5 & \textbf{14725} & \textbf{7ZH6} & \textbf{3.67} \\
20333 & 6PEV & 3.2 & 21307 & 6VP9 & 3.5 & \textbf{4146} & \textbf{5M32} & \textbf{3.8} \\
20334 & 6PEW & 3.2 & 34679 & 8HDS & 3.57 & \textbf{30005} & \textbf{6LXD} & \textbf{3.9} \\
26616 & 7UN8 & 3.3 & 14873 & 7ZQP & 3.6 & & & \\
\hline
\end{tabular}
\label{tab:gan_trainvaltb}
\end{table}

\begin{table}[!ht]
\centering
\caption{List of all EMDB/PDB examples in the test set.}
\vspace{1em}
\scriptsize
\begin{tabular}{ccc||ccc||ccc}
\hline
EMDB ID & PDB ID & Res. (\AA) &
EMDB ID & PDB ID & Res. (\AA) &
EMDB ID & PDB ID & Res. (\AA) \\
\hline
9590 & 6ACF & 3 & 30180 & 7BTO & 4 & 3602 & 5N8Y & 4.7 \\
22414 & 7JPK & 3 & 6425 & 3JD6 & 4.1 & 6862 & 5YZ0 & 4.7 \\
0199 & 6HCY & 3.1 & 3835 & 5ONV & 4.1 & 3439 & 5G5L & 4.8 \\
0927 & 6LN8 & 3.1 & 6746 & 5ZBO & 4.1 & 6875 & 5Z1F & 4.8 \\
9321 & 6N23 & 3.1 & 6987 & 6A69 & 4.1 & 8665 & 5VFR & 4.9 \\
11093 & 6Z6G & 3.1 & 7018 & 6AYE & 4.1 & 8954 & 6E15 & 5.1 \\
22829 & 7KDT & 3.1 & 3866 & 6EGX & 4.1 & 7322 & 6C05 & 5.2 \\
0843 & 6L7E & 3.2 & 4241 & 6FE8 & 4.1 & 3491 & 5MDX & 5.3 \\
9213 & 6MRU & 3.2 & 4286 & 6FO0 & 4.1 & 9537 & 5GRS & 5.4 \\
0636 & 6O6R & 3.2 & 4390 & 6GDG & 4.1 & 4342 & 6G2D & 5.4 \\
20042 & 6OF4 & 3.2 & 9832 & 6JI1 & 4.1 & 8436 & 5TQW & 5.6 \\
10495 & 6TG9 & 3.2 & 0502 & 6NT5 & 4.1 & 10351 & 6SZA & 6 \\
10575 & 6TTF & 3.2 & 5155 & 3IYJ & 4.2 & 3885 & 6EL1 & 6.1 \\
21481 & 6VZ1 & 3.2 & 3237 & 5FN2 & 4.2 & 8470 & 5TWV & 6.3 \\
10847 & 6YMX & 3.2 & 3366 & 5G06 & 4.2 & 0608 & 6O2M & 6.3 \\
11488 & 6ZWM & 3.2 & 4037 & 5LCW & 4.2 & 1874 & 2Y9J & 6.4 \\
22359 & 7JK2 & 3.2 & 4112 & 5LVC & 4.2 & 8230 & 5KBT & 6.4 \\
9361 & 6NF6 & 3.3 & 6734 & 5XMK & 4.2 & 3636 & 5NG5 & 6.5 \\
21458 & 6VYF & 3.3 & 6859 & 5YYS & 4.2 & 7065 & 6B7Y & 6.5 \\
22131 & 6XD3 & 3.3 & 6940 & 5ZQZ & 4.2 & 5245 & 3IZI & 6.7 \\
30165 & 7BSS & 3.3 & 4173 & 6F2D & 4.2 & 5100 & 3IXV & 6.8 \\
9906 & 6K1H & 3.5 & 0311 & 6HZ5 & 4.2 & 3761 & 5O8O & 6.8 \\
10213 & 6SJ7 & 3.5 & 20695 & 6U9E & 4.2 & 6284 & 3J9T & 6.9 \\
20993 & 6V0C & 3.5 & 8398 & 5TCP & 4.3 & 3436 & 5G4F & 7 \\
21436 & 6VXF & 3.5 & 6952 & 5ZSU & 4.3 & 8097 & 5IOU & 7 \\
4339 & 6G1K & 3.6 & 7476 & 6CHS & 4.3 & 8187 & 5JZT & 7.4 \\
0775 & 6KSW & 3.6 & 7793 & 6D3R & 4.3 & 7461 & 6CE7 & 7.4 \\
0836 & 6L53 & 3.6 & 8919 & 6DVW & 4.3 & 8685 & 5VHW & 7.8 \\
30071 & 6M39 & 3.6 & 9214 & 6MRW & 4.3 & 9949 & 6K9K & 7.8 \\
0043 & 6GOV & 3.7 & 20524 & 6PYH & 4.3 & 3186 & 5FJ6 & 7.9 \\
0257 & 6HRA & 3.7 & 10273 & 6SOF & 4.3 &  &  &  \\
0567 & 6O0H & 3.7 & 10549 & 6TQE & 4.3 &  &  &  \\
10092 & 6S3K & 3.7 & 30358 & 7CGN & 4.3 &  &  &  \\
10214 & 6SJB & 3.7 & 2364 & 4BTG & 4.4 &  &  &  \\
21012 & 6V1I & 3.8 & 6668 & 5H64 & 4.4 &  &  &  \\
10617 & 6XT9 & 3.8 & 8751 & 5W1R & 4.4 &  &  &  \\
9626 & 6AHR & 3.9 & 6911 & 5ZBG & 4.4 &  &  &  \\
0071 & 6GVE & 3.9 & 7967 & 6DMW & 4.4 &  &  &  \\
9380 & 6NIL & 3.9 & 0287 & 6HV8 & 4.4 &  &  &  \\
4537 & 6QEL & 3.9 & 9870 & 6JPQ & 4.4 &  &  &  \\
20708 & 6UAN & 3.9 & 9915 & 6K4M & 4.5 &  &  &  \\
7118 & 6BO4 & 4 & 0946 & 6LQI & 4.5 &  &  &  \\
7464 & 6CES & 4 & 0967 & 6LT4 & 4.5 &  &  &  \\
3984 & 6EZ8 & 4 & 30041 & 6M1D & 4.5 &  &  &  \\
0088 & 6GY6 & 4 & 5917 & 4PT2 & 4.6 &  &  &  \\
0258 & 6HRB & 4 & 3776 & 5OFO & 4.6 &  &  &  \\
9883 & 6JT0 & 4 & 9577 & 6KV5 & 4.6 &  &  &  \\
9118 & 6MHU & 4 & 22216 & 6XJX & 4.6 &  &  &  \\
20265 & 6P6W & 4 & 6535 & 3JC5 & 4.7 &  &  &  \\
20501 & 6PW9 & 4 & 2788 & 4V1W & 4.7 &  &  &  \\
\hline
\end{tabular}
\label{tab:gan_testtb}
\end{table}

\chapter{Supplementary Materials for Chapter 8}

\section{Experimental Datasets}

Tables \ref{tab:cryo_train_data}, \ref{tab:cryo_val_data}, and \ref{tab:cryo_test_data} list the cryo-EM density maps and associated PDB structures that we used for training and testing our CryoSAMU model. Table \ref{tab:cryo_test_data_m2m} lists the cryo-EM density maps and associated PDB structures that we used for protein structure modeling.


\begin{table}[!ht]
\centering
\caption{List of all EMDB/PDB examples in training sets.}
\vspace{1em}
\tiny
\begin{tabular}{ccc||ccc||ccc}
\hline
PDB ID & EMDB ID & Res. (\AA) &
PDB ID & EMDB ID & Res. (\AA) &
PDB ID & EMDB ID & Res. (\AA) \\
\hline
6VU8 & 21388 & 4.14 & 6BO4 & 7118  & 4 & 6RWA & 10036 & 4 \\
6ROW & 4975  & 4.5 & 6T6V & 10387 & 4.5 & 3J1P & 5410  & 6.5 \\
6U9F & 20696 & 4.35 & 6AJ2 & 9631  & 4 & 5JZT & 8187  & 7.4 \\
5WSN & 6685  & 4.3 & 7KXY & 23067 & 4.4 & 6D3R & 7793  & 4.3 \\
6VFI & 21173 & 4.5 & 6WYK & 21967 & 4 & 6LXE & 30006 & 4.2 \\
6NY1 & 8994  & 4.2 & 8I4T & 35183 & 5.2 & 6ZPO & 11342 & 4 \\
8BTZ & 16244 & 5.39 & 7JGG & 22326 & 4.9 & 7BG4 & 12170 & 4.2 \\
7KSR & 23024 & 4.1 & 6ZN2 & 11309 & 4.3 & 6VFK & 21185 & 4.3 \\
7YR6 & 34047 & 4.8 & 6WCZ & 21618 & 4 & 6C14 & 7328 & 4.5 \\
6M66 & 30114 & 4.1 & 6JI1 & 9832 & 4.1 & 2Y9J & 1874 & 6.4 \\
6IXH & 9747 & 4 & 6V8P & 21108 & 4.1 & 7CG3 & 30349 & 5.1 \\
7Q3Y & 13797 & 4.34 & 7R7T & 24304 & 4.5 & 6GZV & 0103 & 4 \\
6EL1 & 3885 & 6.1 & 7PTS & 13636 & 5.71 & 6NYB & 0541 & 4.1 \\
7Y1Q & 33570 & 5.03 & 7O24 & 12698 & 4.8 & 5UZ7 & 8623 & 4.1 \\
6Y5K & 10700 & 4.2 & 5KEL & 8240 & 4.3 & 6VEJ & 21363 & 4.3 \\
3J2W & 5577 & 5 & 5NG5 & 3636 & 6.5 & 7PTX & 13642 & 4.03 \\
6BOA & 7122 & 4.2 & 7ELE & 31182 & 4.9 & 5LVC & 4112 & 4.2 \\
5ZQZ & 6940 & 4.2 & 6R3B & 4717 & 4.5 & 6ALF & 8585 & 4.05 \\
6PW9 & 20501 & 4 & 6N1Q & 9317 & 5.2 & 6BVF & 7294 & 4 \\
6SIH & 10210 & 4.7 & 6F2D & 4173 & 4.2 & 6SCT & 0126 & 4.69 \\
6S5T & 10100 & 4.15 & 7O42 & 12716 & 4.1 & 7OZ3 & 13119 & 4.46 \\
7BGJ & 12181 & 6.9 & 6ZYY & 11581 & 4.4 & 6OMA & 20122 & 7.2 \\
6MZC & 9298 & 4.5 & 5LY6 & 4118 & 4.5 & 7KAL & 22774 & 4 \\
6P6F & 20261 & 4.5 & 6UZ2 & 20950 & 4.2 & 6NT8 & 0505 & 6.5 \\
7MO7 & 23919 & 4.8 & 6W1C & 21509 & 5.3 & 7ND2 & 12273 & 4 \\
8EKI & 28204 & 4.5 & 7L30 & 23147 & 4.4 & 7YMX & 33946 & 4.44 \\
5FLC & 3213 & 5.9 & 6TEB & 10479 & 4.1 & 6U9E & 20695 & 4.2 \\
7BE9 & 12154 & 4.2 & 6OFJ & 20047 & 4.5 & 5FN2 & 3237 & 4.2 \\
6WBI & 21590 & 4.4 & 7NTF & 12588 & 5.32 & 7FIF & 31595 & 6.5 \\
6UCV & 20729 & 4.1 & 7K1V & 22630 & 4.6 & 6XPE & 22286 & 4.1 \\
6JSH & 9881 & 5.1 & 6ZPI & 11340 & 4.5 & 6V85 & 21095 & 4.4 \\
5ZAL & 6905 & 4.7 & 6BP7 & 7125 & 4.9 & 7AL3 & 11815 & 4.8 \\
5OF4 & 3802 & 4.4 & 6M1D & 30041 & 4.5 & 6OLM & 20117 & 4.4 \\
3J9P & 6267 & 4.2 & 5H64 & 6668 & 4.4 & 8BAH & 15948 & 4.13 \\
7A1D & 11606 & 4.19 & 5LC5 & 4032 & 4.35 & 6H3L & 0135 & 4.2 \\
6K9K & 9949 & 7.8 & 6RGL & 4876 & 5.4 & 2YEW & 1886 & 6  \\
6PPB & 20432 & 4.3 & 6G2D & 4342 & 5.4 & 7B6E & 12053 & 4.5 \\
8GP3 & 34188 & 4.8 & 6OKR & 20102 & 4.2 & 4UQQ & 2685 & 7.6 \\
6Z0S & 11022 & 5.7 & 6CE7 & 7461 & 7.4 & 6D83 & 7453 & 4.3 \\
6UC2 & 20725 & 4.5 & 7KS3 & 23015 & 5.8 & 6D7L & 7823 & 4 \\
7Q55 & 13826 & 5.7 & 5OAF & 3773 & 4.1 & 6S2E & 10088 & 4.2 \\
6O2M & 0608 & 6.3 & 6JT0 & 9883 & 4 & 6LBA & 0868 & 4.1 \\
6ZQJ & 11366 & 4.2 & 6XYE & 10649 & 4.3 & 6OCE & 20017 & 4.9 \\
7K2V & 22647 & 6.6 & 7KEU & 22233 & 5 & 6HZ5 & 0311 & 4.2 \\
7QJ0 & 14005 & 5.32 & 7B6D & 12052 & 4.3 & 6POF & 20414 & 4.3 \\
6SSM & 10298 & 4.3 & 6VEF & 21156 & 4.08 & 5XMK & 6734 & 4.2 \\
5Y3R & 6803 & 6.6 & 6X0V & 21985 & 4.5 & 6MRW & 9214 & 4.3 \\
6EGX & 3866 & 4.1 & 7ZJ4 & 14740 & 4.43 & 8J5Z & 35996 & 4.75 \\
7JW1 & 22513 & 4.2 & 4BTG & 2364 & 4.4 & 5FKX & 3204 & 6.1 \\
7CA3 & 30323 & 4.5 & 6R5K & 4728 & 4.8 & 6S8F & 10120 & 4 \\
6BBM & 7076 & 4.1 & 6E15 & 8954 & 5.1 & 5N8Y & 3602 & 4.7 \\
5W1R & 8751 & 4.4 & 6W1S & 21514 & 4.02 & 5VFR & 8665 & 4.9 \\
7CTF & 30463 & 4.8 & 7B6H & 12054 & 5.4 & 7MOB & 23923 & 5 \\
6KNB & 0723 & 6.9 & 6KLE & 0709 & 4.5 & 8S91 & 40234 & 4.3 \\
5LJO & 4061 & 4.9 & 6CFZ & 7469 & 4.5 & 5GRS & 9537 & 5.4 \\
6C21 & 7332 & 5.2 & 5TQW & 8436 & 5.6 & 6UWM & 20924 & 5.9 \\
7ZC6 & 14622 & 4.27 & 5KGF & 8246 & 4.54 & 6OGD & 20053 & 4.4 \\
7KTT & 23029 & 4.17 & 5Z1F & 6875 & 4.8 & 6YTV & 10924 & 4.4 \\
6U8Y & 20692 & 4 & 8FTK & 29439 & 4.56 & 7PTQ & 13633 & 4.08 \\
8HMF & 34898 & 4.6 & 8POG & 17791 & 4.15 & 7R0J & 14223 & 4.23 \\
8I9J & 35274 & 6.39 & 5OYG & 3861 & 4.1 & 5OWX & 3856 & 5.2 \\
4PT2 & 5917 & 4.6 & 6B5B & 7055 & 5.2 & 7DN5 & 30781 & 4.11 \\
6TQE & 10549 & 4.3 & 6V69 & 21060 & 4.2 & 6DVW & 8919 & 4.3 \\
8I6Q & 35203 & 4.23 & 6LT4 & 0967 & 4.5 & 7MDI & 23773 & 4.3 \\
8C1C & 16378 & 4.1 & 6D6V & 7821 & 4.8 & 7YL9 & 33215 & 4.7 \\
6I2T & 4400 & 5.7 & 8FNW & 29328 & 6.73 & 7VH1 & 31983 & 4.2 \\
8CA1 & 16515 & 4.3 & 5Y5Y & 6811 & 4.7 & 7SQT & 25394 & 4 \\
6ZGD & 11202 & 4.1 & 5N9Y & 3605 & 4.2 & 5YYS & 6859 & 4.2 \\
5VHW & 8685 & 7.8 & 6SGY & 10188 & 4.6 & 5YZ0 & 6862 & 4.7 \\
6ZLU & 11274 & 4.2 & 8ECI & 28016 & 4 & 6PWP & 20510 & 4.1 \\
8A5Y & 15199 & 4.9 & 5FVM & 3329 & 6.1 & 6V9I & 21121 & 5.2 \\
6U1S & 20613 & 7.6 & 6ZVT & 11470 & 7  & 6VXH & 21437 & 4 \\
6V9H & 21120 & 4.1 & 8HEU & 34692 & 4.6 & 7R9E & 24324 & 4 \\
6C05 & 7322 & 5.15 & 6TGB & 10497 & 5.5 & 6EZ8 & 3984 & 4 \\
7Q5S & 13846 & 4.47 & 3J94 & 6204 & 4.2 & 7DXK & 30912 & 4.1 \\
6ZPG & 11338 & 4.6 & 7NBN & 12260 & 7  & 5TCP & 8398 & 4.3 \\
7D7R & 30610 & 4 & 6SZA & 10351 & 6  & 7DL2 & 30708 & 4.4 \\
6LQI & 0946 & 4.5 & 5WC0 & 8794 & 4.4 & 8GAA & 29849 & 4.24 \\
6PYH & 20524 & 4.3 & 6W4P & 21536 & 6.6 & 6A69 & 6987 & 4.1 \\
6OJ3 & 20086 & 4.5 & 6SHS & 10204 & 4.4 & 6CA0 & 7439 & 5.75 \\
6VOA & 21259 & 4 & 6HCG & 0193 & 4.3 & 6OUA & 20200 & 4.2 \\
6ZY4 & 11549 & 4.1 & 3IZI & 5245 & 6.7 & 6PSN & 20459 & 4.6 \\
5J8V & 8073 & 4.9 &  &  &  &  &  &  \\
\hline
\end{tabular}
\label{tab:cryo_train_data}
\end{table}

\begin{table}[!ht]
\centering
\caption{List of all EMDB/PDB examples in validation sets.}
\vspace{1em}
\begin{tabular}{ccc||ccc}
\hline
PDB ID & EMDB ID & Resolution (\AA) & PDB ID & EMDB ID & Resolution (\AA) \\
\hline
5MDX & 3491  & 5.3  & 7JTH & 22473 & 4  \\
6RKW & 4913  & 6.6  & 6QD0 & 4515  & 4.5  \\
7OZ1 & 13118 & 4  & 6E0H & 8948  & 4.1  \\
3J17 & 5376  & 4.1  & 8C0V & 16372 & 4.1  \\
5ZFU & 6927  & 6.7  & 8JNS & 36450 & 4.2  \\
6B40 & 7046  & 4.3  & 6VFJ & 21174 & 5.35 \\
5ZSU & 6952  & 4.3  & 7CCS & 30341 & 6.2  \\
3J22 & 5465  & 6.3  & 6QVB & 4646  & 4.3  \\
5G4F & 3436  & 7   & 5MKF & 3524  & 4.2  \\
6FSZ & 4301  & 4.6  & 6V3G & 21036 & 4  \\
6QXM & 4669  & 4.1  & 6R22 & 4707  & 5.5  \\
6TSW & 10567 & 4  & 6XJX & 22216 & 4.6  \\
6YRK & 10890 & 4.1  & 6POD & 20412 & 4.05 \\
5FJ9 & 3179  & 4.6  & 6AYE & 7018  & 4.1  \\
6HS7 & 0264  & 4.6  & 3J7V & 6034  & 4.6  \\
5GQH & 9535  & 4.5  & 4V1W & 2788  & 4.7  \\
5KBT & 8230  & 6.4  & 7AHE & 11784 & 4.1  \\
6N52 & 0346  & 4  & 6WCJ & 21611 & 6.3  \\
6CES & 7464  & 4  & 6M5V & 30094 & 4.5  \\
5O8O & 3761  & 6.8  & 6BX3 & 7303  & 4.3  \\
6R4O & 4721  & 4.2  & 5G5L & 3439  & 4.8  \\
7Y59 & 33613 & 4.51 & 6Q0X & 20555 & 4.2  \\
7E8G & 31018 & 4.5  & 3J6X & 5942  & 6.1  \\
3IYJ & 5155  & 4.2  & 7RD8 & 24415 & 5.64 \\
7ZZZ & 15042 & 4.1  & 3JBC & 5888  & 6.5  \\
6M6A & 30117 & 5  & 5FJ6 & 3186  & 7.9  \\
7JPU & 22423 & 5  & 6TY3 & 10615 & 6.3  \\
6IBC & 4447  & 4.4  & 6NI2 & 9375  & 4  \\
6JXA & 9892  & 4.3  & 6OR5 & 9032  & 4  \\
6NT5 & 0502  & 4.1  & 7BST & 30166 & 4.37 \\
6DMW & 7967  & 4.4  & 5ZBO & 6746  & 4.1  \\
\hline
\end{tabular}
\label{tab:cryo_val_data}
\end{table}

\begin{table}[!ht]
\centering
\caption{List of all EMDB/PDB examples in test sets.}
\vspace{1em}
\begin{tabular}{ccc||ccc}
\hline
PDB ID & EMDB ID & Resolution (\AA) & PDB ID & EMDB ID & Resolution (\AA) \\
\hline
6FO0 & 4286  & 4.1  & 5OFO & 3776  & 4.6  \\
7ZDZ & 14678 & 4.3  & 6YTK & 10917 & 4.07 \\
6MHU & 9118  & 4  & 6V0B & 20986 & 4.1  \\
6ZXL & 11524 & 4.2  & 6GDG & 4390  & 4.1  \\
6PO3 & 20408 & 4.28 & 5V5S & 8636  & 6.5  \\
6XE9 & 22145 & 4.3  & 6ZYX & 11579 & 4.3  \\
3J9T & 6284  & 6.9  & 6CHS & 7476  & 4.3  \\
6WAZ & 21582 & 4.1  & 5FWP & 3340  & 7.2  \\
6VJY & 21220 & 4.3  & 6ND1 & 0440  & 4.1  \\
6Z2J & 11041 & 4  & 5TWV & 8470  & 6.3  \\
7U2B & 26311 & 4.1  & 6GY6 & 0088  & 4  \\
8OSF & 17151 & 4  & 5I08 & 8069  & 4  \\
8C0W & 16373 & 4.7  & 8EDG & 28034 & 4.64 \\
6GTE & 0063  & 4.07 & 7RYZ & 24749 & 4.15 \\
5G06 & 3366  & 4.2  & 6S6T & 10105 & 4.1  \\
6UT7 & 20868 & 4.3  & 6YTX & 10925 & 6.23 \\
7F5A & 31463 & 6.4  & 7E2I & 30958 & 4.07 \\
6K4M & 9915  & 4.5  & 6PEK & 20327 & 4.2  \\
7KDV & 22830 & 4.59 & 6VK0 & 21222 & 4.1  \\
6SO5 & 10266 & 4.2  & 5ONV & 3835  & 4.1  \\
6XE6 & 22144 & 4.5  & 3JC5 & 6535  & 4.7  \\
6C5V & 7344  & 4.8  & 7P16 & 13157 & 4.3  \\
6RVY & 10018 & 4.1  & 6PUR & 20479 & 4.4  \\
5XJY & 6724  & 4.1  & 5IOU & 8097  & 7   \\
6HV8 & 0287  & 4.4  & 7KVC & 23046 & 4.7  \\
7K9L & 22754 & 4.9  & 6R7Z & 4748  & 5.14 \\
6JPQ & 9870  & 4.4  & 8POC & 17788 & 4  \\
6R7Y & 4747  & 4.2  & 3JCL & 6526  & 4  \\
6L2T & 0814  & 4.1  & 5ZBG & 6911  & 4.4  \\
6FE8 & 4241  & 4.1  & 8CLS & 16718 & 4  \\
6E10 & 8951  & 4.16 & 8IGG & 35432 & 4.09 \\
7K08 & 22589 & 4.7  & 6M6Z & 30126 & 5.9  \\
6EZO & 4162  & 4.1  & 3J5L & 5771  & 6.6  \\
6R9T & 4772  & 6.2  & 6POS & 20418 & 4.12 \\
8F6R & 28889 & 4  & 6PQX & 20455 & 4.6  \\
6ZPH & 11339 & 6.9  & 6HRB & 0258  & 4  \\
6U1N & 20612 & 4  & 7D0G & 30533 & 5  \\
7RKF & 24496 & 4  &      &       &      \\
\hline
\end{tabular}
\label{tab:cryo_test_data}
\end{table}

\begin{table}[!ht]
\centering
\caption{List of all EMDB/PDB examples for protein structure modeling.}
\vspace{1em}
\begin{tabular}{ccc}
\hline
PDB ID & EMDB ID & Resolution (\AA) \\
\hline
7ZDZ & 14678 & 4.3 \\
5XJY & 6724  & 4.1 \\
6C5V & 7344  & 4.8 \\
6MHU & 9118  & 4 \\
8C0W & 16373 & 4.7 \\
6VJY & 21220 & 4.3 \\
6UT7 & 20868 & 4.3 \\
6PO3 & 20408 & 4.28 \\
6HV8 & 0287  & 4.4 \\
6RVY & 10018 & 4.1 \\
6SO5 & 10266 & 4.2 \\
6ZXL & 11524 & 4.2 \\ 
6GTE & 0063  & 4.07 \\
7K9L & 22754 & 4.9 \\
6XE9 & 22145 & 4.3 \\
7F5A & 31463 & 6.4 \\
8OSF & 17151 & 4 \\
5G06 & 3366  & 4.2 \\
6XE6 & 22144 & 4.5 \\
6Z2J & 11041 & 4 \\
\hline
\end{tabular}
\label{tab:cryo_test_data_m2m}
\end{table}

\clearpage


\section{Additional Visualizations}

\begin{figure}[!ht]
  \centering
   \includegraphics[width=\linewidth, trim={0cm 21cm 4cm 0cm}, clip]{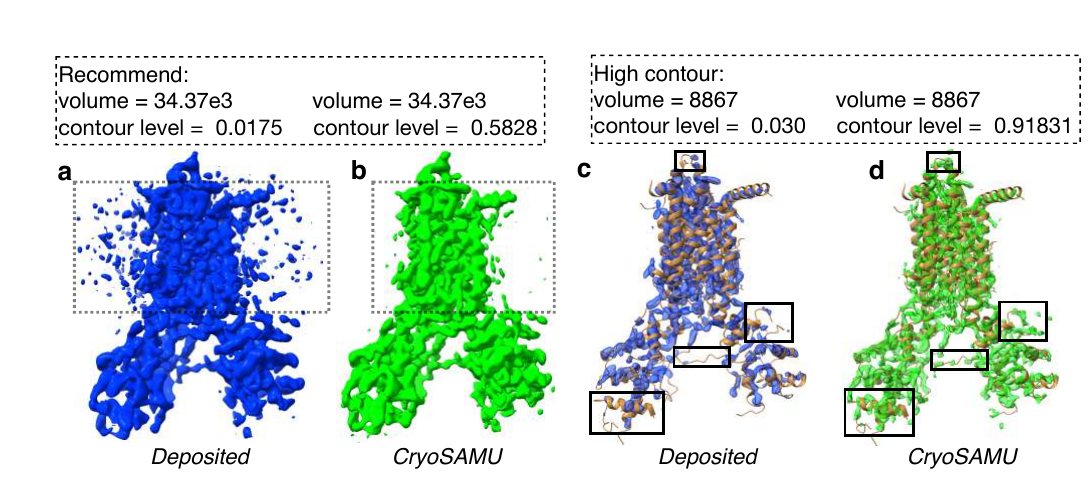}
   \caption{Visualizations of deposited (blue) and CryoSAMU-enhanced (green) maps. The corresponding PDB structures (brown) are superimposed on the maps.
   \textbf{a}: Human Dispatched-1 (PDB-6XE6, EMDB-22144, reported resolution of 4.53 {\AA})~\cite{6XE6}. \textbf{a-b}: Maps displayed at the recommended contour level.
   \textbf{c-d}: Maps displayed at a higher contour level. 
   Visualizations were produced by UCSF ChimeraX \cite{chimerax}.
   The protein structure modeling completeness and accuracy improved after CryoSAMU enhancement. For instance, residue coverage increased from 53.5\% to 64.5\%, as well as sequence match increased from 6.1\% to 7.7\%. 
   }
   \label{fig:CryoSAMU_SI_vis_comp}
\end{figure}

Figure \ref{fig:CryoSAMU_SI_vis_comp} shows visual results of CryoSAMU's enhancements for another protein structures. 
See the main texts for the explanation of the figure.

\clearpage

\section{Additional Real-Space Correlation Coefficient Plots}

\begin{figure}[!ht]
  \centering
   \includegraphics[width=\linewidth, trim={0cm 0cm 0cm 0cm}, clip]{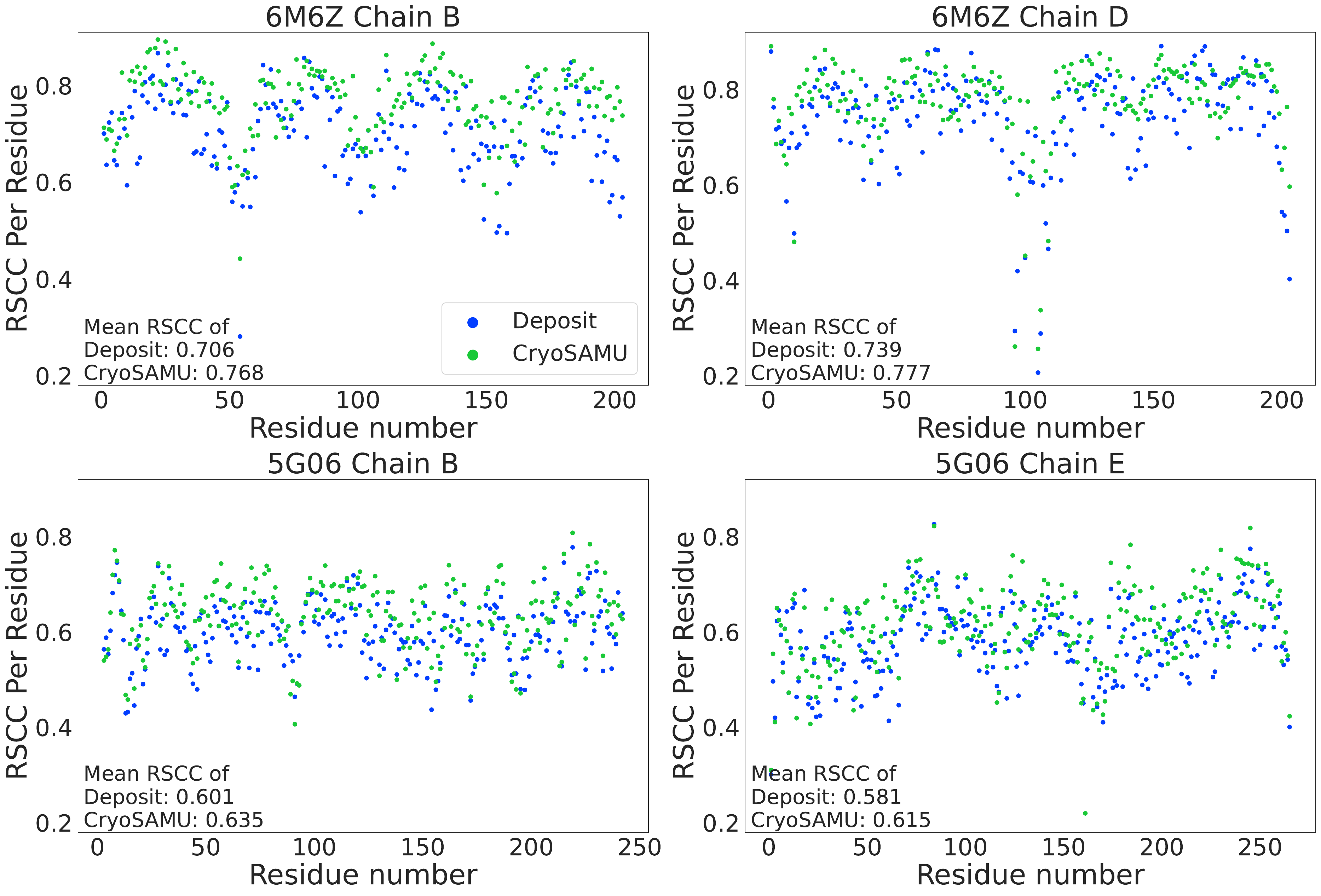}
   \caption{The real-space correlation coefficient (RSCC) comparison between deposited and CryoSAMU-enhanced maps. Top: a transmembrane nanopore TMH4C4 (PDB-6M6Z, EMDB-30126, reported resolution of 5.9 {\AA})~\cite{6M6Z}. Bottom: a yeast cytoplasmic exosome (PDB-5G06, EMDB-3366, reported resolution of 4.2 {\AA})~\cite{5G06}.
   In the first example of PDB-6M6Z, both Chain B and Chain D exhibited significant RSCC improvements compared to the deposited maps, increasing from 0.706 to 0.768 and from 0.739 to 0.777, respectively. In addition, 85.2\% of residues in Chain B and 70.9\% of residues in Chain D showed an increase in their RSCC scores. In the second example of PDB-5G06, the average RSCC scores increased from 0.601 to 0.635 for Chain B, with 82.9\% of 240 residues showing improvement; and increased from 0.581 to 0.615 for Chain E, with 76.2\% of 265 residues perform better.
   }
   \label{fig:CryoSAMU_SI_plot_RSCC}
\end{figure}

Figure \ref{fig:CryoSAMU_SI_plot_RSCC} shows the results of residue-level real-space correlation coefficient (RSCC) for other protein examples.
See the main texts for the explanation of the figure.

\end{document}